\documentclass{article}

\usepackage{PRIMEarxiv}

\usepackage[utf8]{inputenc} 
\usepackage[T1]{fontenc}    
\usepackage{hyperref}       
\usepackage{url}            
\usepackage{booktabs}       
\usepackage{amsfonts}       
\usepackage{nicefrac}       
\usepackage{microtype}      
\usepackage{lipsum}
\usepackage{fancyhdr}       
\usepackage{graphicx}       
\graphicspath{{media/}}     
\usepackage{multirow}
\usepackage{booktabs}
\usepackage{amsmath}
\usepackage{cleveref}
\usepackage{longtable}
\usepackage{xcolor}
\usepackage{setspace}
\usepackage{enumitem}
\usepackage{tikz}
\usetikzlibrary{positioning}  

\usepackage{adjustbox}
\usepackage{array}
\usepackage{tabularx}
\usepackage{xcolor}
\usepackage{tcolorbox}
\usepackage{makecell}
\usepackage[percent]{overpic}
\usetikzlibrary{patterns,backgrounds}
\usepackage{fontspec}
\usepackage{caption}
\usepackage{subcaption}
\usepackage{float}
\usepackage{placeins}

\newfontface\Lora{Lora-Regular.ttf}[Path=figures/fonts/]
\newfontface\CormorantGaramond{CormorantGaramond-Regular.ttf}[Path=figures/fonts/]
\newfontface\Montserrat{Montserrat-Regular.ttf}[Path=figures/fonts/]
\newfontface\BebasNeue{BebasNeue-Regular.ttf}[Path=figures/fonts/]

\title{Graphic-Design-Bench: A Comprehensive Benchmark for Evaluating AI on Graphic Design Tasks}

\date{March 31, 2026}

\author{%
  Adrienne Deganutti \hspace{0.7em} Elad Hirsch\hspace{0.7em} Haonan Zhu \hspace{0.7em} Jaejung Seol \hspace{0.7em} Purvanshi Mehta\thanks{Author names are listed in alphabetical order.} \\
  \texttt{\{adrienne, elad, haonan, jaejung, purvanshi\}@lica.world} \\
}

\begin{document}
\maketitle

\begin{abstract}
We introduce \textsc{GraphicDesignBench} (GDB), the first comprehensive benchmark suite designed specifically to evaluate AI models on the full breadth of professional graphic design tasks. Unlike existing benchmarks that focus on natural-image understanding or generic text-to-image synthesis, \textsc{GDB} targets the unique challenges of professional design work: translating communicative intent into structured layouts, rendering typographically faithful text, manipulating layered compositions, producing valid vector graphics, and reasoning about animation. The suite comprises 49 tasks organized along five axes: \emph{layout}, \emph{typography}, \emph{infographics}, \emph{template \& design semantics} and \emph{animation}, each evaluated under both \emph{understanding} and \emph{generation} settings, and grounded in real-world design templates drawn from the LICA layered-composition dataset. Importantly, for 50\% of these tasks, we provide clear quantitative measures showing that even some of the top-performing models fall far short of usable performance. We evaluate a set of frontier closed-source models using a standardized metric taxonomy covering spatial accuracy, perceptual quality, text fidelity, semantic alignment, and structural validity. Our results reveal that current models fall short on the core challenges of professional design: spatial reasoning over complex layouts, faithful vector code generation, fine-grained typographic perception, and temporal decomposition of animations remain largely unsolved. While high-level semantic understanding is within reach, the gap widens sharply as tasks demand precision, structure, and compositional awareness. \textsc{GDB} provides a rigorous, reproducible testbed for tracking progress toward models that can function as capable design collaborators. The full evaluation framework is publicly available at \url{https://github.com/purvanshi-lica/lica-bench}.
\end{abstract}


\begin{spacing}{0.4} 
    \tableofcontents
\end{spacing}

\section{Introduction}
\label{sec:intro}


 
Design shapes how humans interact with the world, from the layout of a webpage to the typography on a billboard, visual communication is embedded in nearly every product, brand, and piece of media we encounter. At its most structured, this manifests as graphic design: a high-dimensional creative task that integrates spatial layout, typography, color theory, brand identity, and communicative intent into a single visual artifact. Recent advances in vision-language models and diffusion-based image generators have raised the prospect of AI systems that can assist human designers~\cite{inoue2024opencole,cheng2025graphist,lin2024ladeco,jia2024cole,chen2025accordion}, yet the evaluation landscape has not kept pace. Existing benchmarks either measure low-level perceptual quality on natural images~\cite{heusel2017gans,salimans2016improved,hessel2021clipscore} or probe generic visual question-answering, neither of which captures the \emph{structured, intent-driven, and multi-layered} nature of professional design work.
 
This gap matters because design tasks impose constraints that differ qualitatively from those in natural-image benchmarks, along several dimensions:

\begin{itemize}
    \item \textbf{Multi-constraint satisfaction.} A promotional banner is not merely an aesthetically pleasing image; it must render specific text strings legibly, respect a typographic hierarchy, place components within an established grid, and communicate a clear call to action, all simultaneously. Evaluating this requires tasks that test each constraint independently and in combination.
    \item \textbf{Design-specific evaluation.} Measuring whether a model meets these requirements demands task-specific metrics such as OCR-based readability, bounding-box IoU for component placement, and perceptually uniform color distance for style fidelity, none of which are captured by FID or CLIPScore.
    \item \textbf{Continuously evolving standards.} Visual trends shift continuously across cultures, platforms, and time, meaning that what constitutes good design is not a fixed target. A benchmark must account for this moving landscape rather than treating design quality as a static property.
    \item \textbf{Brand and context dependence.} Effective design is often deeply tied to a specific brand identity or creative voice. Unlike object recognition or depth estimation, there is rarely a single objectively correct design solution, making standardized evaluation fundamentally harder than in natural-image settings where ground truth is stable and context-independent.
\end{itemize}

The latter two challenges represent open problems for the field at large. In this work, we focus on the first two: constructing tasks that jointly stress-test multi-constraint satisfaction and measuring performance with design-native metrics. Together, these form the foundation of \textsc{GraphicDesignBench} (GDB) a benchmark suite comprising 49 tasks that collectively evaluate the full spectrum of design-relevant capabilities.
Our contributions are as follows:

\begin{enumerate}
    \item \textbf{The first comprehensive benchmark for graphic design AI.}
    We introduce \textsc{GDB}, covering 49 tasks across layout, typography, SVG \& vector, and animation, spanning both understanding and generation, and representing the broadest evaluation of AI design capabilities to date.

    \item \textbf{A layered-composition evaluation framework.}
    By grounding all tasks in the LICA dataset's ~\cite{hirsch2026lica} full structural metadata (bounding boxes, z-ordering, typography specs, animation properties, SVG source), we enable evaluation settings that are fundamentally impossible with flat raster images, including partial layout completion, layer-aware inpainting, and template variant generation.

    \item \textbf{Design-native metrics.}
    We introduce a multi-faceted metric taxonomy purpose-built for design evaluation, combining spatial accuracy, typographic fidelity, structural validity, and human-aligned preference, moving beyond the perceptual quality scores (FID, CLIPScore) that dominate existing image benchmarks but fail to capture design-specific requirements.

    \item \textbf{A reproducible framework for the community.}
    Rather than a static leaderboard, GDB is designed as an extensible evaluation framework. We evaluate three state-of-the-art frontier model families to establish baselines, reveal persistent failure modes shared across all models, and provide a foundation for the community to benchmark future systems, including open-source models.
\end{enumerate}


The remainder of this paper is organized as follows.
Section~\ref{sec:overview} describes the dataset, evaluated models, and shared metric definitions.
Sections~\ref{sec:layout}--\ref{sec:animation} present tasks grouped by design domain, each covering both understanding and generation settings.
Section~\ref{sec:discussion} synthesizes cross-cutting findings, and Section~\ref{sec:conclusion} concludes with directions for future work.
 
\section{Benchmark Overview}
\label{sec:overview}
\begin{figure*}[t]
    \centering
    \includegraphics[width=\linewidth]{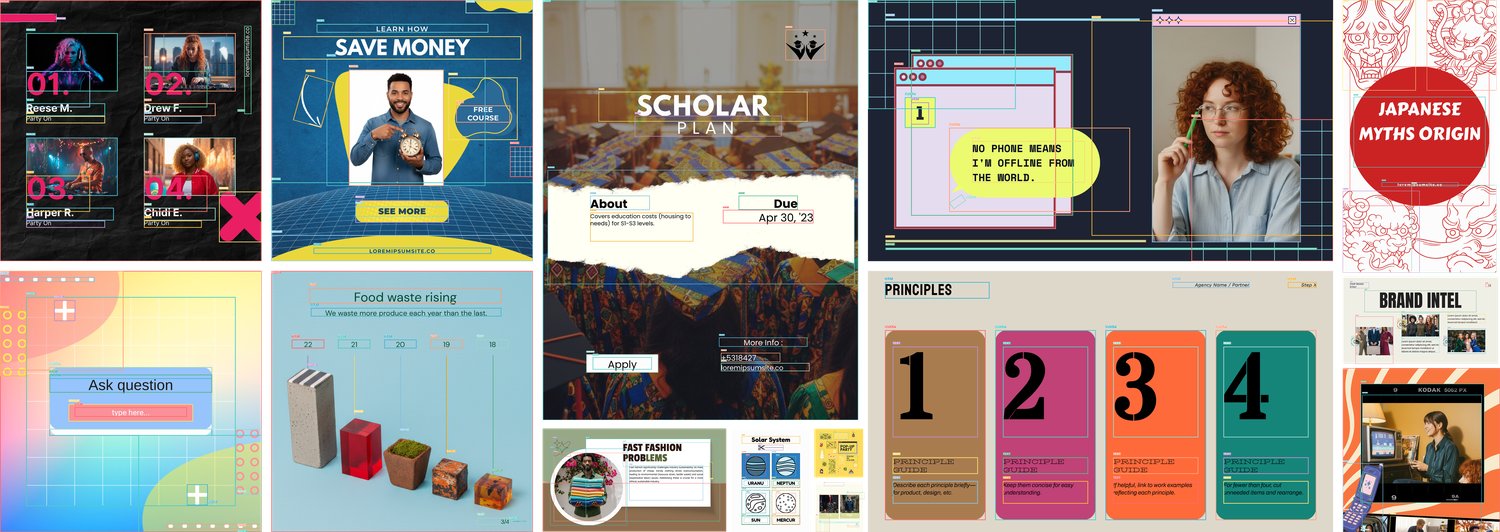}
    \caption{\textbf{LICA samples~\cite{hirsch2026lica}.} Design layouts with structured, component-level annotations capturing full hierarchy and rich metadata beyond coarse bounding boxes, on which we benchmark models in this report.}
    \label{fig:layout_annotations}
\end{figure*}

\textbf{Dataset.}\label{para:dataset}
All tasks in GDB are grounded in the LICA layered-composition dataset~\cite{hirsch2026lica}, a large-scale collection of real-world graphic design templates sourced from a commercial design platform.
Unlike flat raster-image datasets, each LICA template preserves the full layered structure of the original design: individual components are annotated with their type (\texttt{text}, \texttt{image}, \texttt{vector}, \texttt{group}), bounding box, z-order, and styling properties.
Text components carry typographic specifications including font family, size, weight, color, alignment, letter spacing, line height, curvature, rotation, and inline style ranges.
Image and vector components include asset source references, rotation angles, clip paths, and opacity.
Templates are further annotated with global metadata such as canvas aspect ratio, background color, category labels (parent and sub-category), user-intent descriptions, aesthetic tags, and color palettes.
The animated subset additionally provides per-component entrance animation attributes: motion type (from 32 canonical categories), duration, start-time offset, and keyframe sequences.
Each template may have multiple \emph{sibling layouts}, instantiations that share the same structural theme but differ in content, color scheme, or imagery.
This sibling structure enables evaluation settings such as template variant understanding (Section~\ref{sec:semantics}) and template variant generation, which cannot be constructed from isolated designs.
Figure~\ref{fig:layout_annotations} shows examples from the data.

\textbf{Evaluation subsets.}\label{par:eval-subsets}
From the full dataset, we construct task-specific evaluation subsets with standardized filtering criteria.
Table~\ref{tab:dataset_summary} summarizes the primary evaluation sets.
Component-level tasks sample up to three elements per layout to control evaluation cost while maintaining coverage.

The evaluation sizes reflect the full scope of available annotated data at each level of granularity.
Layout-level tasks use all 989 non-video layouts in the dataset; component-level tasks (typography, partial completion) further multiply the effective sample size by sampling up to three elements per layout, yielding $\sim$2{,}500 instances.
Generation tasks use smaller evaluation sets (typically 100 samples, filtered for resolution and aspect-ratio compatibility) due to the cost and complexity of multiple inference calls per sample to closed-source generation APIs.
Animation tasks are evaluated on all 100 compositions with complete temporal annotations; the 50-sample Lottie subset and the 10-brief video generation set reflect the smaller pool of animated assets with full structural metadata.
In all cases, the sample sizes are sufficient to support the conclusions drawn: the inter-model performance gaps reported throughout the paper are typically an order of magnitude larger than the statistical uncertainty at the corresponding sample sizes.

\begin{table}[h]
\centering
\caption{Evaluation subsets used across GDB. Component-level tasks sample up to 3 elements per layout.}
\label{tab:dataset_summary}
\small
\begin{tabular}{llrl}
\toprule
\textbf{Task Group} & \textbf{Section} & \textbf{Samples} & \textbf{Key Annotations Used} \\
\midrule
Layout understanding        & \ref{sec:layout}      & 989 layouts        & Bboxes, z-order, aspect ratio, component types \\
Typography understanding    & \ref{sec:typography}   & 2{,}568 text elems & Font, color, size, weight, alignment, curvature \\
Intent-to-layout generation & \ref{sec:layout}      & \textsuperscript{\dag} & User intent, image description, style cues \\
Partial layout completion   & \ref{sec:layout}      & 989 layouts & Layered composites, component bboxes \\
Layer-aware inpainting      & \ref{sec:layout}      & ---\textsuperscript{\dag} & Layered composites, object masks, asset refs \\
Multi-aspect ratio adaptation      & \ref{sec:layout}   & 13 layouts & Multi aspect image pairs \\
Styled text generation      & \ref{sec:typography}   & ---\textsuperscript{\dag} & Typography specs, rendered ground truth \\
Text removal                & \ref{sec:typography}   & ---\textsuperscript{\dag}    & Text masks, clean backgrounds \\
SVG understanding \& editing & \ref{sec:svg}         & 300 SVGs       & SVG code, code complexity \\
SVG generation              & \ref{sec:svg}          & 300 SVGs       & SVG code, Text descriptions, rendered PNGs \\
SVG generation (by type)    & \ref{sec:svg}          & 150 SVGs           & \shortstack{50 per type: SVG code, element types, Text \\ descriptions, renderred PNGs} \\
Lottie generation           & \ref{sec:svg}          & 50 animations      & Lottie JSON, keyframe PNGs, descriptions \\
Category classification     & \ref{sec:semantics}    & 989 layouts        & Parent + sub-category labels \\
User intent prediction      & \ref{sec:semantics}    & 989 layouts        & Natural-language intent descriptions \\
Template variant understanding & \ref{sec:semantics} & 1250 problems & Template IDs, sibling groups \\
Template variant generation & \ref{sec:semantics}    & 340 problems       & Sibling layouts, JSON representations \\
Keyframe ordering           & \ref{sec:animation}    & 100 animations     & 4 keyframes per animation \\
Motion type classification  & \ref{sec:animation}    & 100 animations     & 32 motion types, component-level labels \\
Animation property extraction & \ref{sec:animation}  & 100 animations     & Durations, start-time offsets \\
Short-form video layout generation & \ref{sec:animation} & 10 briefs & Natural-language marketing brief \\
\bottomrule
\multicolumn{4}{l}{\textsuperscript{\dag}\,Exact sample counts will be reported in the task-specific sections.}
\end{tabular}
\end{table}

\textbf{Evaluated tasks.}\label{par:tasks}
\textsc{GDB} comprises 49 tasks across five domains: layout, typography, SVG \& vector, template semantics, and animation, each evaluated under understanding and generation modes (Table~\ref{tab:tasks_highlevel}). These domains were chosen to reflect the core skill axes a professional designer exercises: arranging space, specifying type, working with vector code, interpreting design intent, and reasoning about motion.

Each domain is evaluated under both \emph{understanding} and \emph{generation} settings.
Understanding tasks ask whether a model can perceive and reason about an existing design, e.g., identifying fonts, counting components, parsing animation timing, while generation tasks ask whether it can produce or modify design artifacts that satisfy structured constraints. 
As the two settings may involve different model families (e.g.\ a vision-language model for understanding vs.\ a dedicated image-generation model from the same provider), we aim to explore what current AI systems can perceive about a design with what they can actually produce.

\begin{table}[h]
\centering
\small
\setlength{\tabcolsep}{5pt}
\caption{GDB organizes its 49 tasks along two orthogonal axes: \emph{design domain} (layout, typography, SVG \& vector, template semantics, and animation) and \emph{capability mode} (understanding vs.\ generation). Understanding tasks probe a model's ability to perceive and reason about existing designs, while generation tasks evaluate whether a model can produce or edit design artifacts that satisfy structured constraints.}
\begin{tabular}{>{\raggedright\arraybackslash}p{3cm}>{\raggedright\arraybackslash}p{2.5cm}>{\raggedright\arraybackslash}p{1cm}>{\raggedright\arraybackslash}p{6cm}}
\toprule
\textbf{Domain} & \textbf{Mode} & \textbf{\#} & \textbf{Description} \\
\midrule
\multirow{2}{*}{Layout} 
    & Understanding & 8 & Spatial reasoning over design canvases: aspect ratio, element counting, component type and detection, layer order, rotation, crop shape, and frame detection. \\[6pt]
    & Generation & 4 & Producing and completing layouts: intent-to-layout generation, partial layout completion, and layer-aware inpainting, multi-aspect ratio adaptation. \\
\midrule
\multirow{2}{*}{Typography} 
    & Understanding & 10 & Perceiving fine-grained text properties: font family, color, size, weight, alignment, spacing, curvature, style ranges, and rotation. \\[6pt]
    & Generation & 2 & Rendering and editing text: styled text generation and text removal with background reconstruction. \\
\midrule
\multirow{2}{*}{Infographics} 
    & Understanding & 5 & SVG code reasoning and editing: perceptual and semantic Q\&A, bug fixing, code optimization, and style editing. \\[6pt]
    & Generation & 5 & Generating vector graphics and animations: text-to-SVG, image-to-SVG, combined image and text to SVG, text to Lottie file generation,  combined image and text to Lottie file generation. \\
\midrule
\multirow{2}{*}{Template \& Semantics} 
    & Understanding & 5 & Interpreting design intent and structure: category classification, user intent prediction and template matching, ranking, and clustering. \\[6pt]
    & Generation & 2 & Producing template-faithful layouts: style completion and color scheme variation \\
\midrule
\multirow{2}{*}{Animation} 
    & Understanding & 5 & Perceiving temporal design properties: keyframe ordering, motion type classification, and duration (video and component) and start-time prediction. \\[6pt]
    & Generation & 3 & Generating animated design content: animation parameter generation, motion trajectory synthesis, and short-form video generation. \\
\midrule
\textbf{Total} & & \textbf{49} & \\
\bottomrule
\end{tabular}
\label{tab:tasks_highlevel}
\end{table}

\textbf{Evaluated models.}\label{par:models}
\label{sec:models}
We evaluate models spanning three frontier model families, covering proprietary API-served systems. These families were selected because they represent the current state of the art across the broadest range of creative tasks, spanning text understanding, image generation, and video synthesis, making them the most natural candidates for assessing design capability in AI systems. Our goal is not simply to rank these models, but to use them as strong baselines to expose where the field as a whole falls short on design-specific tasks. Table~\ref{tab:models} summarizes the models, their access methods, and the task groups in which each participates. Not every model is evaluated on every task: some tasks require image-generation capabilities (available only in GPT-Image-1.5, Gemini~3.1 Pro, and Gemini~3.1 Flash Image), while others require structured code output or video understanding. Although we compare model performance, the goal of the paper is to provide a benchmark and insights into current capabilities in this field, rather than highlight specific failure modes of each model.

\textbf{Decoding and prompting.}\label{par:decoding}
All models are evaluated with greedy decoding (temperature~$0$) to ensure reproducibility.
Each task uses a single fixed prompt template across all models; prompt templates are provided in Appendix~\ref{app:prompts} for reproducibility.
Where a task defines multiple prompting conditions, such as open-vocabulary versus label-constrained (Sections~\ref{sec:semantics}), all models receive the same prompt variant.

\begin{table}[htbp]
\centering
\caption{Models evaluated in GDB. ``Modalities'' indicates the input types accepted for the benchmark tasks.}
\label{tab:models}
\small
\begin{tabular}{lllll}
\toprule
\textbf{Model} & \textbf{Provider / Access} & \textbf{Input Modalities} & \textbf{Output Modality} \\
\midrule
Gemini-3.1-Pro          & Google API        & Text, image, video  & Text \\
Gemini-3.1-Flash-Image  & Google API        & Text, image  & Image \\
GPT-5.4                 & OpenAI API        & Text, image               & Text \\
GPT-Image-1.5           & OpenAI API        & Text, image & Image \\
Claude-Opus-4.6         & Anthropic API     & Text, image               & Text \\
Sora-2    & OpenAI API      & Text, image               & Video \\
Veo-3.1    & Google API      & Text, image               & Video \\
\bottomrule
\end{tabular}
\end{table}

\textbf{Modality conditions.}\label{par:modality}
Several tasks are evaluated under multiple input modality conditions to disentangle the contribution of visual and structural signals: \emph{text-only} (the model receives layout JSON or text metadata), \emph{image-only} (the model receives rendered PNG), and \emph{both} (JSON and PNG together).
Not all modality conditions are available for every model; missing entries are marked in the per-task results tables.
For animation tasks (Section~\ref{sec:animation}), Gemini~3.1 Pro natively accepts video as a first-class input modality and processes the full rendered animation.
GPT-5.4 and Claude Opus~4.6 do not support native video ingestion; for these models, we extract uniformly sampled keyframes from each animation and supply them as an ordered sequence of still images.


\subsection{Evaluation Metrics}
\label{sec:metrics}

GDB tasks report metrics drawn from a shared taxonomy.
We define all recurring metrics here; task-specific metrics are introduced in their respective sections.
Unless otherwise noted, $\uparrow$ denotes higher-is-better and $\downarrow$ denotes lower-is-better.

Throughout the paper we assign each task one of three solvability labels.
\textbf{Mostly Solved}: best-model performance exceeds 95\% (or an equivalent metric-specific threshold), with limited room for improvement at the current evaluation granularity.
\textbf{Partially Solved}: best-model performance falls between 80--95\%, or the task exhibits a clear split where one sub-condition is tractable while another exposes a fundamental gap (e.g.\ single- vs.\ multi-element, coarse vs.\ fine-grained, open-vocab vs.\ constrained).
\textbf{Unsolved}: best-model performance is below 80\%, or the task remains structurally beyond current capabilities, e.g.\ an orders-of-magnitude gap relative to established baselines in other domains.
We set this threshold in consultation with design experts to reflect practical usability, as even small errors (e.g., a few pixels) typically require manual correction.

\paragraph{Spatial accuracy.}
\textbf{mIoU} (mean Intersection over Union) measures average overlap between predicted and ground-truth bounding boxes.
\textbf{bbox F1} is used as a complementary overlap-quality metric. When explicit target boxes are unavailable, we estimate boxes with an LLM-based detector; implementation details and detector selection are provided in Appendix~\ref{app:bbox-detector-selection}.
\textbf{mAP@0.5} and \textbf{mAP@0.5:0.95} follow the COCO detection protocol~\cite{lin2014microsoft}: mAP@0.5 counts a detection as correct if its IoU with the ground-truth box exceeds 50\%; mAP@0.5:0.95 averages across IoU thresholds from 0.5 to 0.95 in steps of 0.05, rewarding tighter localization.
\textbf{MAE} (mean absolute error) and \textbf{MSE} (mean squared error) are used for continuous regression targets such as element counts, font sizes, and animation durations.

\paragraph{Perceptual quality.}
\textbf{LPIPS}~\cite{zhang2018unreasonable} $(\downarrow)$ computes learned perceptual distance between two images using deep features; lower values indicate greater perceptual similarity.
\textbf{SSIM}~\cite{wang2004image} $(\uparrow)$ measures structural similarity based on luminance, contrast, and structure.
\textbf{DreamSim}~\cite{fu2023dreamsim} $(\downarrow)$ captures higher-level, human-aligned visual similarity beyond low-level pixel statistics.
\textbf{FID} (Fr\'{e}chet Inception Distance)~\cite{heusel2017gans} $(\downarrow)$ measures distributional distance between generated and reference image sets in Inception feature space.
\textbf{MSE} (pixel-level) $(\downarrow)$ is the mean squared pixel error between rendered outputs and references, used primarily in SVG evaluation.
\textbf{PSNR} (peak signal-to-noise ratio) $(\uparrow)$ measures reconstruction fidelity on a logarithmic scale.

\paragraph{Semantic alignment.}
\textbf{CLIP Score}~\cite{hessel2021clipscore} $(\uparrow)$ computes cosine similarity between a text prompt and a generated image in CLIP embedding space, measuring text-image alignment.
\textbf{PickScore}~\cite{kirstain2023pick} $(\uparrow)$ is a learned human-preference metric trained on pick-a-pic data, providing a scalar alignment score.
\textbf{BERTScore}~\cite{zhang2019bertscore} $(\uparrow)$ performs soft token-level alignment between generated and reference texts using contextual embeddings (we use the RoBERTa-large variant).
\textbf{Embedding cosine similarity} $(\uparrow)$ computes cosine similarity between mean-pooled last hidden states of generated and reference texts using Llama-3.2-1B.

\paragraph{Human-aligned preference.}
\textbf{NIMA}~\cite{talebi2018nima} $(\uparrow)$ predicts mean opinion scores for aesthetic quality.
\textbf{ImageReward}~\cite{xu2023imagereward} $(\uparrow)$ is a reward model trained on human preference data for text-to-image generation.
\textbf{HPSv3}~\cite{ma2025hpsv3} (Human Preference Score v3) $(\uparrow)$ provides a scalar human-preference prediction calibrated on recent generation models.

\paragraph{Text fidelity.}
\textbf{OCR Accuracy} $(\uparrow)$ measures whether generated text is recognized as the intended target string by an off-the-shelf OCR system.
\textbf{Font Family Accuracy} $(\uparrow)$ and \textbf{Text Align Accuracy} $(\uparrow)$ are discrete match rates for font-family and alignment predictions.
\textbf{Font Size MAE} $(\downarrow)$ is the mean absolute error on predicted font sizes (in pixels).
\textbf{Color distance} is reported as \textbf{$\Delta E$} (CIEDE2000) $(\downarrow)$, a perceptually uniform color-difference metric, alongside \textbf{RGB~$\ell_2$} distance and \textbf{hue-bucket accuracy} (8-bucket quantisation).
A prediction is considered perceptually acceptable when $\Delta E < 5$.
\textbf{Letter Spacing MAE} $(\downarrow)$ measures error in estimated letter spacing.
When generating images or components, text-fidelity metrics rely on a custom Text-Params-Predictor, a lightweight model that recovers typographic parameters from rendered text patches for comparison against ground-truth specifications.

\paragraph{Structural validity.}
\textbf{JSON Valid} $(\uparrow)$ is the fraction of model outputs that parse as valid JSON with the expected schema (used in template variant generation).
\textbf{SVG Valid} $(\uparrow)$ is the fraction of generated SVGs that parse and render without error.
\textbf{Compression Ratio} $(\downarrow)$ measures the byte-level ratio between an optimized SVG and the original, with lower values indicating more aggressive size reduction.

\paragraph{Rank correlation.}
\textbf{Kendall's~$\tau$} $(\uparrow)$ measures the fraction of correctly ordered element pairs (robust to local perturbations).
\textbf{Spearman's~$\rho$} $(\uparrow)$ is a rank correlation coefficient that penalizes large rank displacements.
Both are used for layer-order and ranking tasks, where a score of~0.5 corresponds roughly to random ordering.

\paragraph{Clustering.}
\textbf{ARI} (Adjusted Rand Index)  $(\uparrow)$ and \textbf{AMI} (Adjusted Mutual Information) $(\uparrow)$ are chance-corrected measures of cluster agreement.
\textbf{V-measure} $(\uparrow)$ is the harmonic mean of homogeneity and completeness.
\textbf{FMI} (Fowlkes--Mallows Index) $(\uparrow)$ is the geometric mean of precision and recall at the pair level.

\paragraph{Set-based.}
For multi-label tasks such as aesthetic tag prediction, we report \textbf{precision}, \textbf{recall}, and \textbf{F1} computed per sample and macro-averaged.
We evaluate under two matching criteria: \emph{exact matching} (string identity after normalization) and \emph{fuzzy matching} (substring containment with greedy one-to-one assignment).

\paragraph{Judge-based.}
\textbf{M-Judge}\cite{Patnaik_2025_CVPR} $(\uparrow)$ is an MLLM-based pairwise win-rate metric that compares a model output against the ground-truth layout under the sample-specific intent. The judge selects the better layout based on aesthetics, clarity, usability, creativity, and consistency. In our benchmark, we use M-Judge for layout generation (Section~\ref{sec:layout}), where holistic design quality cannot be fully captured by coordinate- or pixel-level metrics alone. The prompt template is provided in Appendix~\ref{tab:prompt-m-judge}.          
 
\section{Composition Tasks}
\label{sec:layout}

\subsection{Layout Understanding}
\label{sec:layout-understanding}

\begin{figure*}[t]
\centering

\begin{minipage}[t]{0.27\textwidth}
\raggedright\small
\centering
\textbf{(a)} Layouts spanning multiple aspect ratios (e.g.\ $1\!:\!1$, $16\!:\!9$, $3\!:\!4$, $2\!:\!5$).
\end{minipage}
\hfill
\begin{minipage}[t]{0.27\textwidth}
\raggedright\small
\centering
\textbf{(b)} A layout containing a rotated image component.
\end{minipage}
\hfill
\begin{minipage}[t]{0.27\textwidth}
\raggedright\small
\centering
\textbf{(c)} An image placed inside a decorative frame with a circular crop.
\end{minipage}

\vspace{2pt}

\begin{minipage}[c]{0.27\textwidth}
\centering
\includegraphics[width=\linewidth]{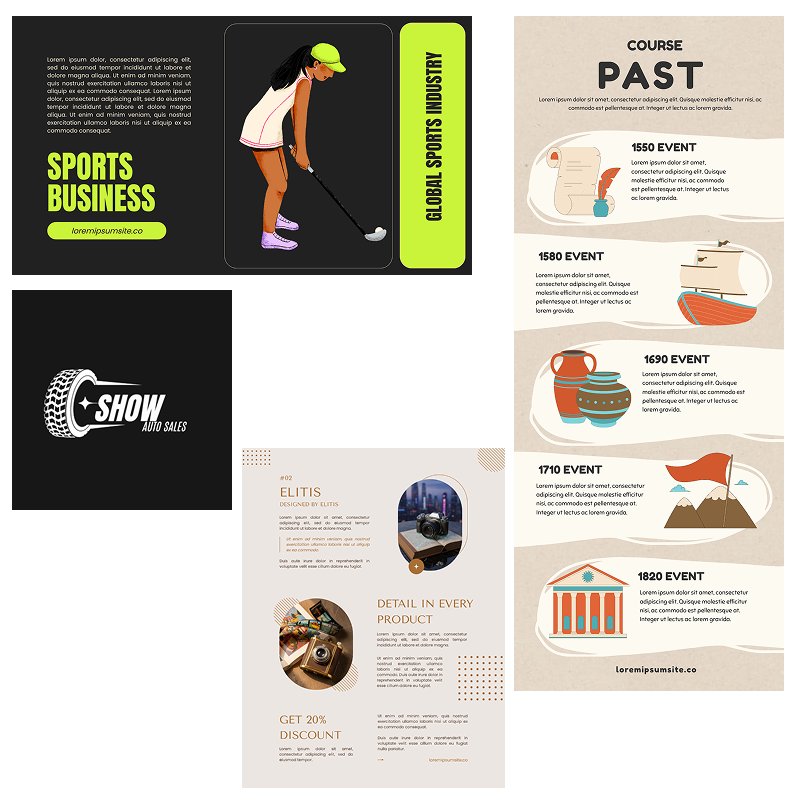}
\end{minipage}
\hfill
\begin{minipage}[c]{0.27\textwidth}
\centering
\includegraphics[width=\linewidth]{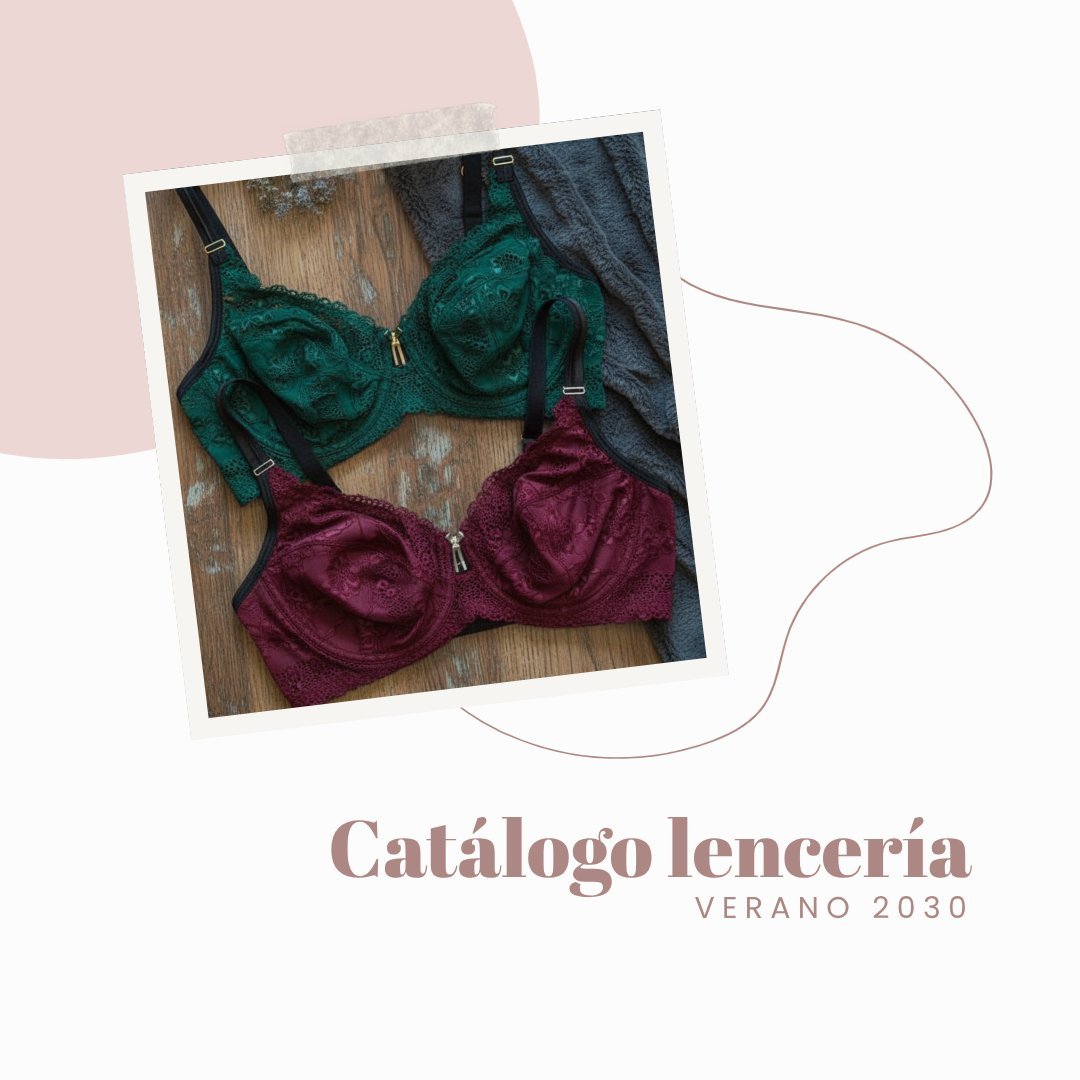}
\end{minipage}
\hfill
\begin{minipage}[c]{0.27\textwidth}
\centering
\includegraphics[width=\linewidth]{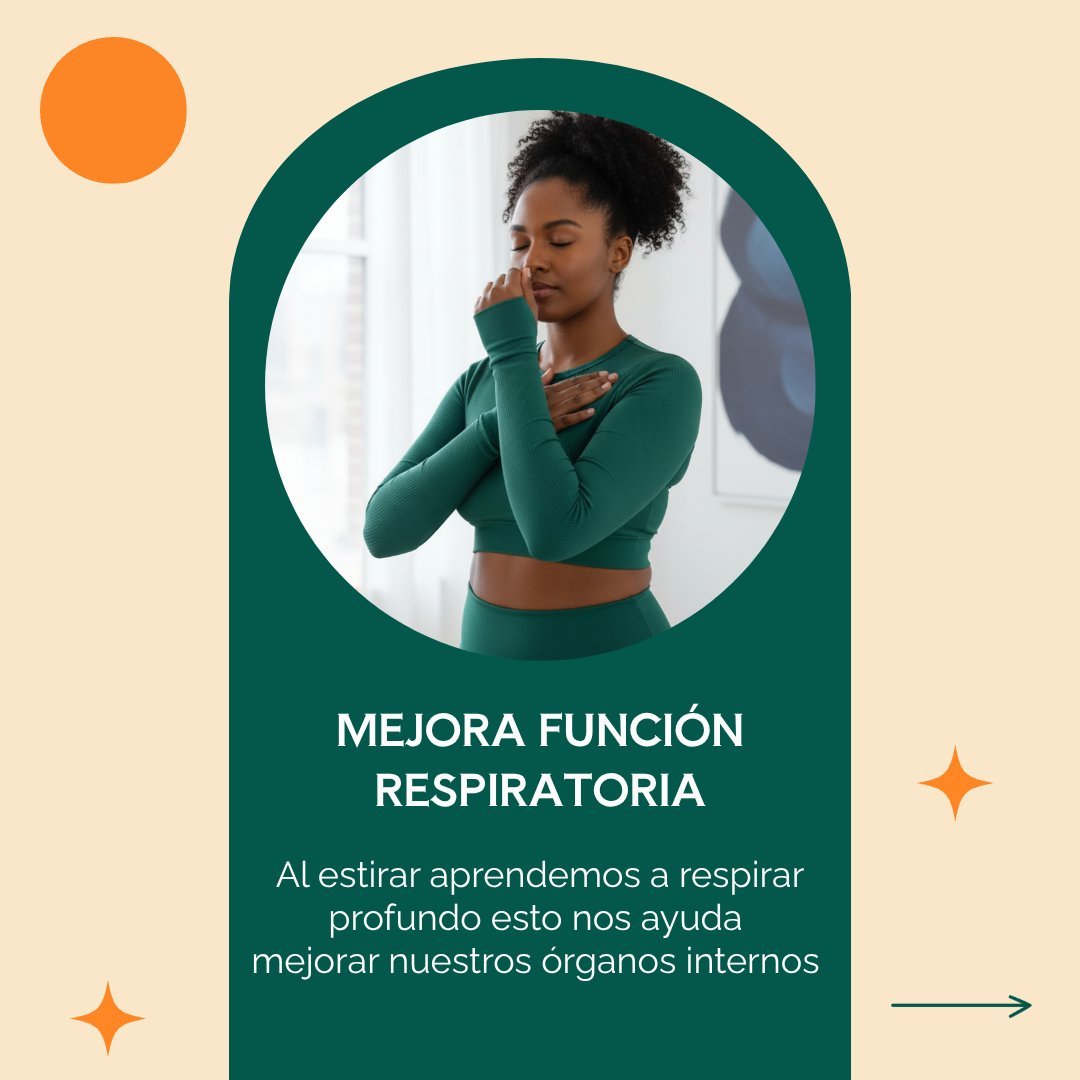}
\end{minipage}

\caption{Example design templates illustrating the variance of layout properties evaluated in this section. The layouts differ in aspect ratio, number and type of components, spatial composition, and visual complexity~(a), and may contain rotated image elements~(b) or images placed inside decorative frames with non-rectangular crops~(c).}
\label{fig:layout-examples}
\end{figure*}

A layout defines the spatial arrangement of visual components on a canvas, encompassing their positions, dimensions, z-order stacking, and type. It forms the structural backbone of any graphic design, and understanding it is the most fundamental capability a design-aware model must possess. A model that cannot reliably identify how many components a layout contains, what types they are, or how they are stacked cannot be expected to modify or extend that layout coherently. We evaluate eight progressively harder tasks (Table~\ref{tab:layout_tasks}) on 989 designs from the LICA dataset (\ref{para:dataset}). The evaluation set spans a realistic distribution of canvas formats — $16\!:\!9$ (34.8\%), $1\!:\!1$ (26.2\%), and $3\!:\!4$ (20.4\%) together cover 80\% of layouts, with rarer formats such as $5\!:\!4$ and $2\!:\!3$ appearing fewer than 10 times each — and a wide range of layout complexity, with layouts containing on average 15.7 top-level components (median\,=\,12, $\sigma$\,=\,14.0). This class imbalance is reflected in the gap between accuracy and macro-F1 reported below. Tasks range from global geometric inference to fine-grained component localization and depth reasoning, and results show that this capability remains largely unsolved by current frontier systems. Figure~\ref{fig:layout-examples} shows example layouts illustrating the variance in layout properties.



\begin{table}[h]
\centering
\small
\setlength{\tabcolsep}{4pt}
\caption{Spatial composition understanding task definitions and dataset statistics. All tasks draw from the same 989 layouts. Sample counts and distribution might vary across tasks because component-level tasks apply validity filters (e.g., minimum element size) and per-layout sampling caps.}
\begin{tabular}{>{\raggedright\arraybackslash}p{2.2cm}>{\raggedright\arraybackslash}p{4.2cm}>{\raggedright\arraybackslash}p{1cm}>{\raggedright\arraybackslash}p{1.8cm}>{\raggedright\arraybackslash}p{2.5cm}>{\raggedright\arraybackslash}p{3cm}}
\toprule
\textbf{Task} & \textbf{Description} & \textbf{N} & \textbf{Classes} & \textbf{Distribution} & \textbf{Metrics} \\
\midrule
Aspect Ratio Classification & Predict the canvas aspect ratio from 9 categories. Tests whether the model can infer global geometric proportions from visual content alone. & 989 & 9 ratios & 16:9 34.8\%, 1:1 26.2\%, 3:4 20.4\%, 4:3 8.7\%, 4:5 5.0\%, other 4.8\% & Acc, Macro-F1 \\
\midrule
Element Counting & Predict the total number of visible components in a rendered design, probing compositional parsing without explicit localization. & 989 & — & Median 12, P25--P75: 7--18, range [1, 104]; 80\% of layouts have 5--34 elements & MAE, MSE \\
\midrule
Component Type Classification & Classify a component type within a bounding-box region as text, image, vector, or group. Up to 3 components sampled per layout. & 2,943 & 4 types & text 41.5\%, image 29.5\%, vector 17.1\%, group 11.8\% & Acc, Macro-F1 \\
\midrule
Component Detection & Localize and classify all components via bounding boxes simultaneously. The most demanding layout task, requiring recognition across variable-length component sets. & 989 & 4 types & 15.1 boxes/layout (median 11); text 45.5\%, image 27.3\%, vector 19.5\%, group 7.7\% & mAP@.5, mAP@.5:.95 \\
\midrule
Layer Order Prediction & Predict the z-order stacking of all visible components back-to-front. Requires occlusion and depth reasoning only partially recoverable from a 2D render. & 989 & — & Median 12 elements, P25--P75: 7--18, range [2, 104] & Kendall $\tau$, Spearman $\rho$ \\
\midrule
Image Rotation Prediction & Predict the rotation angle in degrees of a target image component identified by its alt-text. Only non-cropped components included to prevent ambiguity. & 2,565 & $[-180, 180]°$ & 84.4\% unrotated; of 15.6\% rotated: $|{\theta}|$ median $87°$, 39\% in $[45,90]°$, 14\% in $[135,180]°$ & Rot.\ Acc, Angle MAE \\
\midrule
Image Crop Shape Prediction & Predict whether an image is cropped non-rectangularly and classify into one of six shapes: none, rectangle, rounded rectangle, circle, polygon, or organic. & 2,565 & 6 categories & none 91.8\%, rectangle 4.1\%, rounded rect.\ 2.6\%, circle 0.7\%, organic 0.6\%, polygon 0.2\% & Crop Acc, Shape Acc \\
\midrule
Frame Detection & Predict whether an image resides inside a decorative frame such as a shaped mask or ornamental cutout. Plain rectangular bounding boxes do not qualify. & 1,863 & Binary & Not framed 85.0\%, framed 15.0\% & Acc, Precision, Recall, F1 \\
\bottomrule
\end{tabular}
\label{tab:layout_tasks}
\end{table}

\begin{figure}[t]
\centering

\begin{minipage}[t]{0.3\columnwidth}
\raggedright\small
\centering
\textbf{(a)} Original layout with correct layer ordering. All components are visible and correctly stacked.
\end{minipage}
\hspace{0.05\columnwidth}
\begin{minipage}[t]{0.3\columnwidth}
\raggedright\small
\centering
\textbf{(b)} Layout re-rendered using the z-order predicted by GPT-5.4.
\end{minipage}

\vspace{2pt}

\begin{minipage}[c]{0.3\columnwidth}
\centering
\includegraphics[width=\linewidth]{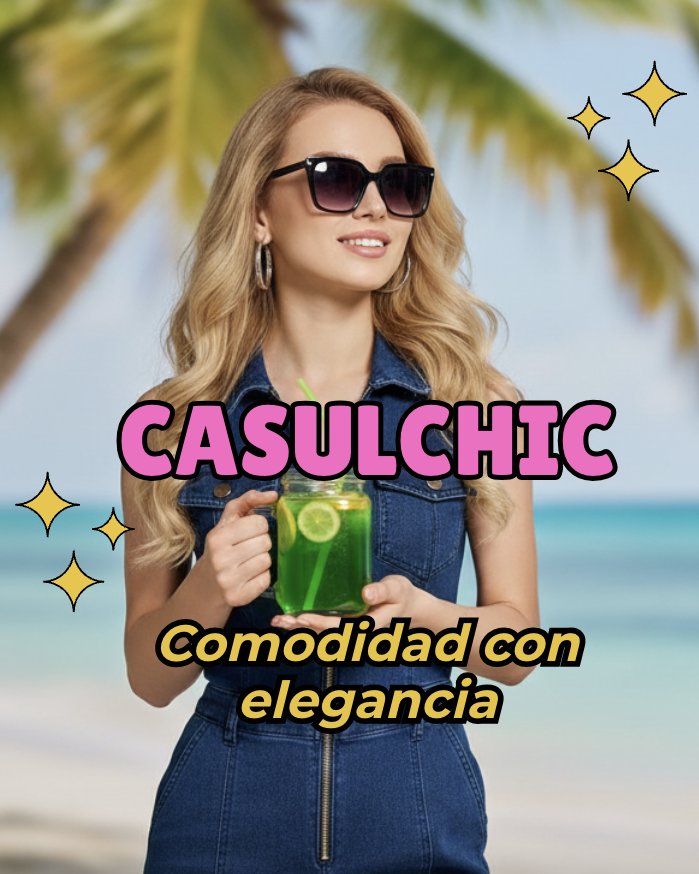}
\end{minipage}
\hspace{0.05\columnwidth}
\begin{minipage}[c]{0.3\columnwidth}
\centering
\includegraphics[width=\linewidth]{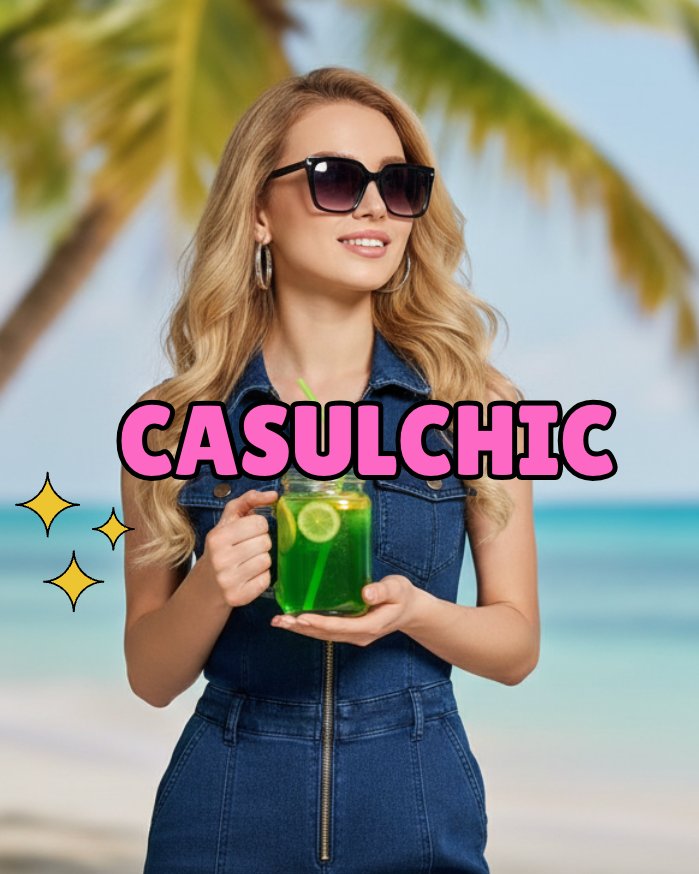}
\end{minipage}

\caption{Layer order prediction failure case (GPT-5.4). The model predicts an incorrect z-order that buries foreground elements (text and decorative shape) behind the background, illustrating that even small rank errors can render a layout unusable.}
\label{fig:layer-order-failure}
\end{figure}

\paragraph{Results.}
Tables~\ref{tab:layout-results} and~\ref{tab:image-rotation-results} present the full results and all metrics are defined in Section~\ref{sec:metrics}.
Figures~\ref{fig:layer-order-failure} shows qualitative examples. The results reveal four distinct patterns across the spatial composition tasks:
\begin{itemize}
    \item \textbf{Spatial reasoning is highly uneven across models.} Aspect ratio classification (93.9\% best accuracy), element counting (MAE\,=\,5.81 best), and component type classification (46.1\% best accuracy) show large performance gaps, with models trailing by up to $2\times$ on counting tasks.
    \item \textbf{Component detection remains effectively unsolved.} The best model achieves only 6.4\% mAP@0.5, orders of magnitude below natural-image benchmarks, indicating a fundamental gap in design-domain spatial grounding across all evaluated systems.
    \item \textbf{Z-order inference is a distinct capability.} Layer order prediction does not correlate with performance on other layout tasks, suggesting that depth and occlusion reasoning requires different visual capabilities than bounding-box prediction.
    \item \textbf{Visual container reasoning is a separate skill.} Crop shape detection and frame detection reveal a capability dimension orthogonal to other spatial tasks, with the best models achieving 76.9\% and F1\,=\,0.504 respectively.
\end{itemize}
\FloatBarrier 

\subsection{Layout Generation}
\label{sec:layout-generation}

Layout generation evaluates whether a model can synthesize or modify a graphic design under structured constraints, rather than only interpret an existing layout. Across the four tasks in this subsection, the model must generate a full design from intent, complete missing components, reinsert masked assets, or adapt a layout to a new canvas ratio. For intent-to-layout generation and layer-aware inpainting, we evaluate on 100 randomly sampled examples after filtering out images with extreme resolutions or aspect ratios.

For the multi-aspect-ratio adaptation task, we additionally report three human-evaluated binary metrics that capture structural preservation more directly than generic preference models. \textbf{TextAcc} is 1 if the required text is preserved correctly and legibly in the adapted output, and 0 otherwise. \textbf{Recall} is 1 if all core assets from the source layout are retained appropriately after adaptation, and 0 otherwise. \textbf{Hallucination} is 1 if any unsupported text or visual element is introduced by the model, and 0 otherwise. All three metrics are averaged over evaluation samples.

\begin{table}[H]
\centering
\caption{Layout understanding results.
Best result per metric in \textbf{bold}.
$\uparrow$\,=\,higher is better; $\downarrow$\,=\,lower is better.}
\label{tab:layout-results}
\small
\begin{tabular}{@{}l cc cc cc@{}}
\toprule
& \multicolumn{2}{c}{\textbf{Aspect Ratio}}
& \multicolumn{2}{c}{\textbf{Elem.\ Counting}}
& \multicolumn{2}{c}{\textbf{Comp.\ Type Clf.}} \\
\cmidrule(lr){2-3} \cmidrule(lr){4-5} \cmidrule(lr){6-7}
\textbf{Model}
  & Acc$\uparrow$ & F1$\uparrow$
  & MAE$\downarrow$ & MSE$\downarrow$
  & Acc$\uparrow$ & F1$\uparrow$ \\
\midrule
Gemini-3.1-Flash-Lite & 0.236 & 0.105 & 12.50 & 375.7 & 0.252 & 0.224 \\
Gemini-3.1-Pro        & 0.245 & 0.085 & 12.35 & 342.9 & 0.281 & 0.236 \\
GPT-5.4               & \textbf{0.939} & \textbf{0.679} & \textbf{5.81} & \textbf{134.0} & \textbf{0.461} & \textbf{0.359} \\
Claude-Opus-4.6       & 0.093 & 0.179 & 6.46 & 150.7 & 0.072 & 0.101 \\
\midrule\midrule
& \multicolumn{4}{c}{\textbf{Component Detection}}
& \multicolumn{2}{c}{\textbf{Layer Order}} \\
\cmidrule(lr){2-5} \cmidrule(lr){6-7}
\textbf{Model}
  & mAP@.5$\uparrow$ & mAP@.5:.95$\uparrow$
  & AP\textsubscript{txt}$\uparrow$ & AP\textsubscript{img}$\uparrow$
  & $\tau$$\uparrow$ & $\rho$$\uparrow$ \\
\midrule
Gemini-3.1-Flash-Lite & 0.006 & 0.002 & 0.003 & 0.008 & \textbf{0.567} & 0.566 \\
Gemini-3.1-Pro        & 0.010 & 0.004 & 0.002 & 0.004 & 0.495 & 0.519 \\
GPT-5.4               & \textbf{0.064} & \textbf{0.036} & \textbf{0.073} & 0.020 & 0.492 & 0.537 \\
Claude-Opus-4.6       & 0.018 & 0.008 & 0.020 & \textbf{0.028} & 0.542 & \textbf{0.573} \\
\bottomrule
\end{tabular}
\end{table}

\begin{table}[H]
\centering
\caption{Image rotation prediction results ($n$\,=\,2{,}585).
Best result per metric in \textbf{bold}.}
\label{tab:image-rotation-results}
\small
\begin{tabular}{@{}l ccc@{}}
\toprule
\textbf{Model}
  & Rot.\ Acc$\uparrow$ & Angle MAE$\downarrow$ & Angle MAE\textsubscript{rot}$\downarrow$ \\
\midrule
Gemini-3.1-Flash-Lite & 0.716 & 17.07 & 74.93 \\
Gemini-3.1-Pro        & 0.750 & 16.29 & 73.48 \\
GPT-5.4               & \textbf{0.800} & \textbf{13.76} & \textbf{69.43} \\
Claude-Opus-4.6       & 0.766 & 15.81 & 75.75 \\
\bottomrule
\end{tabular}
\end{table}

\begin{table}[H]
\centering
\caption{Image crop shape prediction ($n$\,=\,2{,}585) and frame detection ($n$\,=\,1{,}863) results.
Best per metric in \textbf{bold}.}
\label{tab:crop-frame-results}
\small
\begin{tabular}{@{}l ccc cccc@{}}
\toprule
& \multicolumn{3}{c}{\textbf{Image Crop Shape}}
& \multicolumn{4}{c}{\textbf{Frame Detection}} \\
\cmidrule(lr){2-4} \cmidrule(lr){5-8}
\textbf{Model}
  & is-crop$\uparrow$ & shape$\uparrow$ & shape\textsubscript{crop}$\uparrow$
  & Acc$\uparrow$ & Prec$\uparrow$ & Rec$\uparrow$ & F1$\uparrow$ \\
\midrule
Gemini-3.1-Flash-Lite & 0.557 & 0.531 & 0.066 & 0.725 & 0.181 & 0.237 & 0.205 \\
Gemini-3.1-Pro        & 0.704 & 0.688 & 0.043 & 0.734 & 0.149 & 0.165 & 0.156 \\
GPT-5.4               & 0.605 & 0.596 & \textbf{0.431} & 0.823 & 0.402 & 0.373 & 0.387 \\
Claude Opus 4.6       & \textbf{0.769} & \textbf{0.757} & 0.346 & \textbf{0.846} & \textbf{0.487} & \textbf{0.523} & \textbf{0.504} \\
\bottomrule
\end{tabular}
\end{table}

\begin{figure*}[H]
\centering

\begin{minipage}[c]{0.31\textwidth}
\centering
\includegraphics[width=0.8\linewidth]{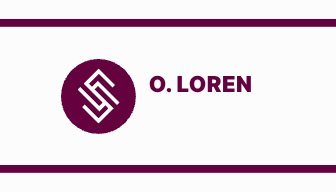}
\end{minipage}
\hfill
\begin{minipage}[c]{0.31\textwidth}
\centering
\includegraphics[width=0.8\linewidth]{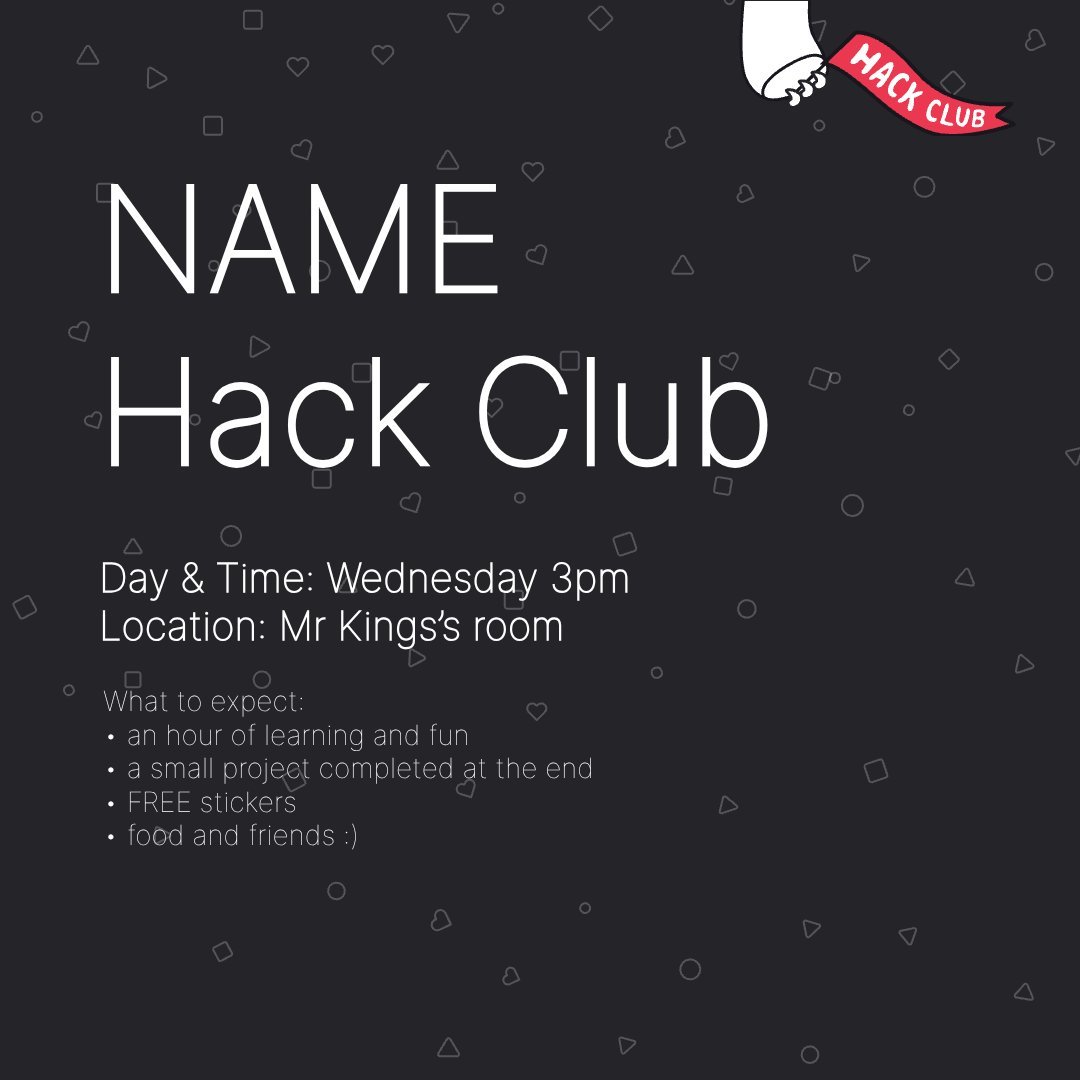}
\end{minipage}
\hfill
\begin{minipage}[c]{0.31\textwidth}
\centering
\includegraphics[width=0.8\linewidth]{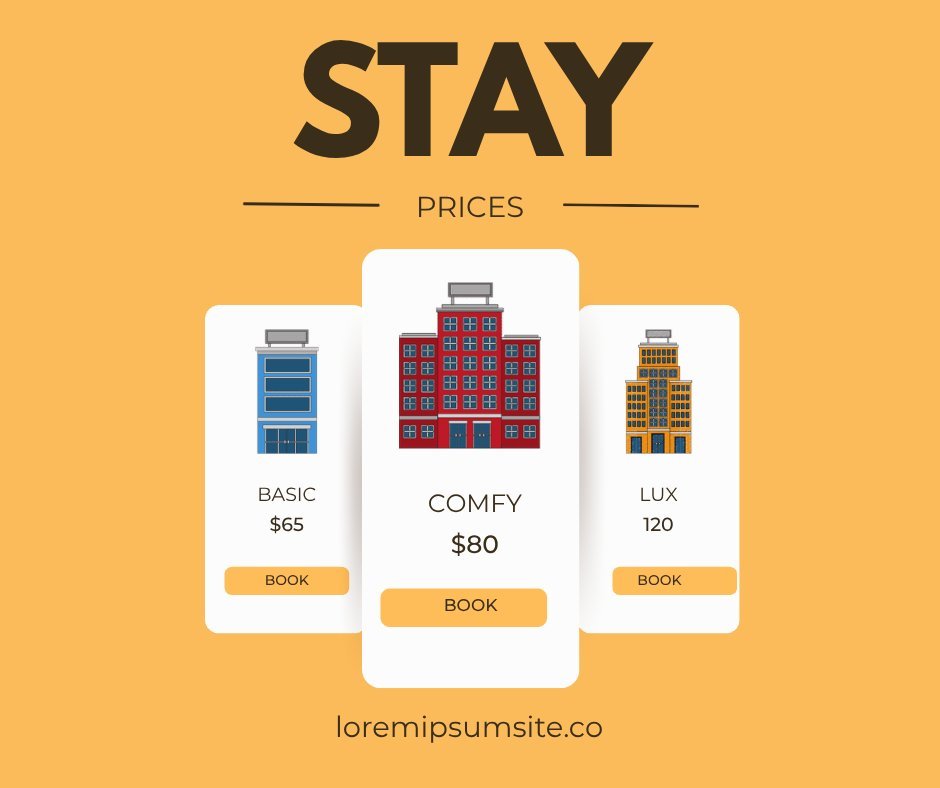}
\end{minipage}

\vspace{2pt}

\begin{minipage}[t]{0.31\textwidth}
\raggedright\small
\textbf{(a) Aspect Ratio Classification}\\[2pt]
\textbf{GT:} 16:9\quad \textbf{GPT-5.4:} 2:3\\
A wide-format banner is misclassified as a portrait ratio. GPT-5.4 confuses the dominant vertical text column for a portrait canvas, ignoring the actual image dimensions.
\end{minipage}
\hfill
\begin{minipage}[t]{0.31\textwidth}
\raggedright\small
\textbf{(b) Element Counting}\\[2pt]
\textbf{GT:} 5\quad \textbf{GPT-5.4:} 95\\
A simple layout with 5 top-level components is predicted to contain 95 elements. The model appears to count individual characters or decorative sub-pixels rather than functional design components.
\end{minipage}
\hfill
\begin{minipage}[t]{0.31\textwidth}
\raggedright\small
\textbf{(c) Component Detection}\\[2pt]
\textbf{GT:} 22 components (text, image, vector)\\
\textbf{Gemini Pro:} 14 boxes, all labelled ``text''\\
The model collapses images and vectors into text labels and produces bounding boxes in a rotated coordinate system unrelated to the actual layout.
\end{minipage}

\caption{Representative failure cases for layout understanding tasks.
Models exhibit systematic errors including ratio confusion~(a), order-of-magnitude overcounting~(b), and type collapse~(c).}
\label{fig:layout-failures}
\end{figure*}

\begin{table}[h]
\centering
\small
\setlength{\tabcolsep}{5pt}
\caption{Layout generation task definitions.}
\label{tab:layout-generation-tasks}
\begin{tabular}{>{\raggedright\arraybackslash}p{3.1cm}>{\raggedright\arraybackslash}p{5.4cm}>{\raggedright\arraybackslash}p{1.6cm}>{\raggedright\arraybackslash}p{3.2cm}}
\toprule
\textbf{Task} & \textbf{Description} & \textbf{Samples} & \textbf{Metrics} \\
\midrule
Intent-to-Layout Generation & Generate a complete layout image from communicative intent, layout description, style cues, required text, and target aspect ratio. & 100 layouts & Pick, CLIP, NIMA, ImgRwd, HPS, OCR \\
\midrule
Partial Layout Completion & Predict placements of missing components from an incomplete layout, isolated assets, and structured metadata. & 989 layouts & mIoU, LPIPS, DreamSim, M-Judge \\
\midrule
Layer-Aware Inpainting & Generate and integrate a missing asset within a masked layout while preserving its identity and overall compositional harmony. 
& 100 layouts & CLIP, DINOv2, DreamSim, LPIPS, PSNR, SSIM, ImgRwd, HPS \\
\midrule
Multi-Aspect Ratio Adaptation & Recompose a layout for a new canvas ratio while preserving text, core assets, and visual balance; structural preservation is scored with human-evaluated binary metrics. & 13 layouts & Human TextAcc, Recall, Halluc., M-Judge, PickScore, HPSv3, ImgRwd, DreamSim \\
\bottomrule
\end{tabular}
\end{table}

Table~\ref{tab:layout-generation-tasks} lists the tasks used to evaluate this subsection along with their primary settings and metrics.

\begin{figure*}[htbp]
    \centering
    \includegraphics[width=\textwidth]{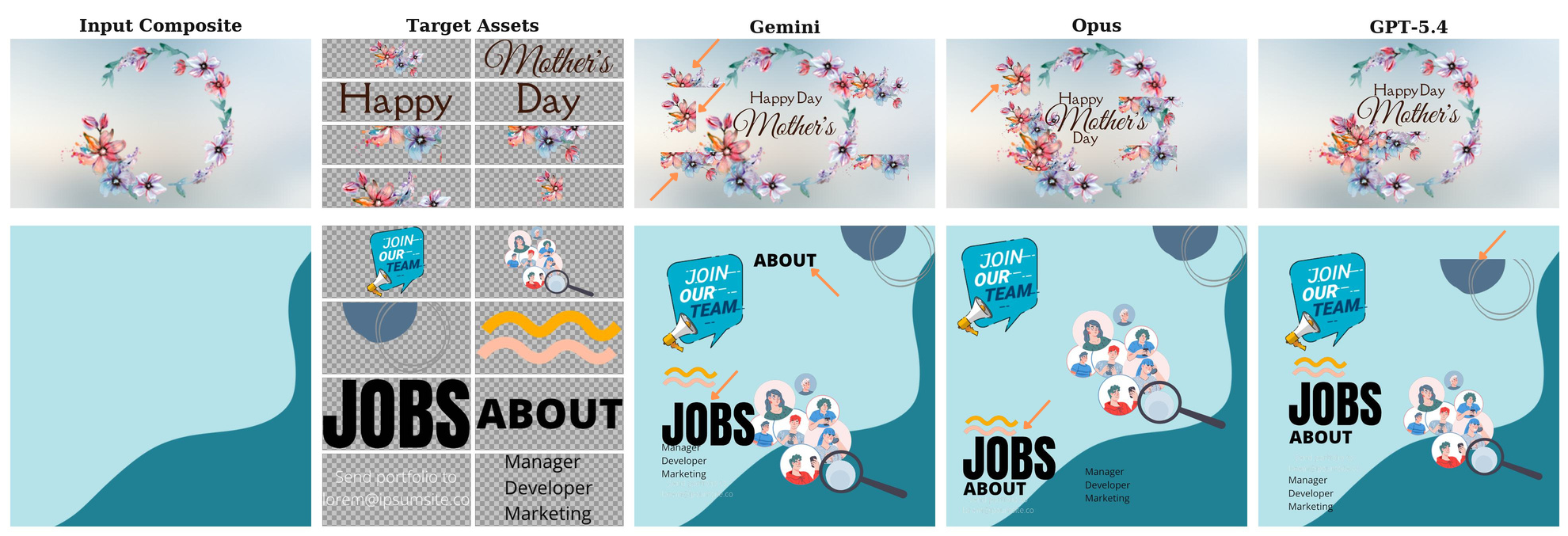}
    \caption{\textbf{Partial layout completion visualization.}
From left to right, we show the composite input image, the multiple target assets with a checkerboard background, and the predictions from three models: Gemini, Opus, and GPT.
Both rows illustrate the \textbf{multiple}-placement setting.
In the first row, the models fail to account for the cropped nature of the object asset, leading to visually unnatural placements of truncated content, as indicated by the \textcolor[HTML]{FF914D}{orange arrows}.
In the second row, the models fail to preserve the shared visual concept of related text elements such as “Jobs” and “About,” placing them too far apart or in the incorrect order, and potentially cropping or misplacing other shapes (GPT-5.4).}
    \label{fig:partial_layout_completion_vis}
\end{figure*}

\paragraph{Results.}
Tables~\ref{tab:generation_eval_metrics}, \ref{tab:partial-completion}, \ref{tab:inpainting}, and \ref{tab:multi-aspect-adaptation} present the full results for the four generation settings, with qualitative examples in Figures~\ref{fig:partial_layout_completion_vis}, \ref{fig:layer_aware_inpainting_combined}, and \ref{fig:multi_aspect_ratio_adaptation}. The main takeaways are:
\begin{enumerate}
    \item \textbf{Intent-to-layout generation remains bottlenecked by text fidelity.} The two evaluated models are closely matched on alignment and aesthetic metrics, but OCR remains in the 73--75\% range, indicating that a substantial fraction of required strings are still rendered incorrectly or illegibly.
    \item \textbf{Partial layout completion degrades sharply in the multi-element setting.} Single-element completion is comparatively tractable, but jointly reasoning about several missing placements remains a substantially harder compositional problem.
    \item \textbf{Layer-aware inpainting exposes a trade-off between identity preservation and global harmony.} Models that stay closer to the reference asset do not necessarily produce the most coherent final composition, and the gap between image-conditioned and text-conditioned inputs remains relatively modest.
    \item \textbf{Multi-aspect-ratio adaptation is not well captured by a single scalar metric.} GPT-Image-1.5 better preserves text and achieves stronger preference-model scores, while Gemini preserves core assets more reliably and scores better on M-Judge and DreamSim, indicating complementary strengths.
\end{enumerate}

\phantomsection\label{sec:intent-to-layout}

\begin{table}[H]
    \centering
    \caption{Intent-to-layout generation results.
    Metrics are grouped by text--image alignment, aesthetic quality, and text accuracy.}
    \label{tab:generation_eval_metrics}
    \small
    \setlength{\tabcolsep}{4pt}
    \begin{tabular}{@{}l cc ccc c@{}}
        \toprule
        & \multicolumn{2}{c}{\textbf{Alignment}}
        & \multicolumn{3}{c}{\textbf{Aesthetic Quality}}
        & \textbf{Text} \\
        \cmidrule(lr){2-3} \cmidrule(lr){4-6} \cmidrule(lr){7-7}
        \textbf{Model}
          & Pick$\uparrow$ & CLIP$\uparrow$
          & NIMA$\uparrow$ & ImgRwd$\uparrow$ & HPS$\uparrow$
          & OCR$\uparrow$ \\
        \midrule
        Gemini-3.1-Flash-Img & 19.81 & 0.310 & 5.14 & 0.250 & 10.28 & 0.726 \\
        GPT-Image-1.5        & 20.21 & 0.308 & 5.18 & 0.278 & 10.45 & 0.754 \\
        \bottomrule
    \end{tabular}
\end{table}

\phantomsection\label{sec:partial-completion}

\begin{table}[H]
    \centering
    \caption{Partial layout completion results for single-element and multiple-element settings.
    Best result per metric in \textbf{bold}.}
    \label{tab:partial-completion}
    \small
    \setlength{\tabcolsep}{5pt}
    \begin{tabular}{@{}ll cccc@{}}
        \toprule
        \textbf{Setting} & \textbf{Model}
          & mIoU$\uparrow$ & LPIPS$\downarrow$ & DreamSim$\downarrow$ & M-Judge$\uparrow$ \\
        \midrule
        \multirow{3}{*}{Single}
          & GPT-5.4         & 0.2299 & 0.0582 & \textbf{0.0131} & 0.29 \\
          & Claude-Opus-4.6 & 0.1878 & 0.0677 & 0.0164 & 0.26 \\
          & Gemini-3.1-Pro  & \textbf{0.2413} & \textbf{0.0512} & 0.0137 & \textbf{0.31} \\
        \midrule
        \multirow{3}{*}{Multiple}
          & GPT-5.4         & 0.1620 & 0.3349 & 0.0829 & 0.17 \\
          & Claude-Opus-4.6 & 0.1286 & 0.3586 & 0.1016 & 0.17 \\
          & Gemini-3.1-Pro  & \textbf{0.2007} & \textbf{0.3127} & \textbf{0.0669} & \textbf{0.33} \\
        \bottomrule
    \end{tabular}
\end{table}

\phantomsection\label{sec:inpainting}

\begin{table}[h]
    \centering
    \caption{Layer-aware inpainting results for image-conditioned insertion and text-conditioned synthesis.
    Identity preservation metrics and reconstruction quality are computed on the inserted object crop; ImageReward and HPSv3 are computed on the final composite.}
    \label{tab:inpainting}
    \small
    \setlength{\tabcolsep}{4pt}
    \begin{tabular}{@{}ll cccc cc cc@{}}
        \toprule
        & & \multicolumn{4}{c}{\textbf{Identity Preservation}}
        & \multicolumn{2}{c}{\textbf{Reconstruction}}
        & \multicolumn{2}{c}{\textbf{Design Quality}} \\
        \cmidrule(lr){3-6} \cmidrule(lr){7-8} \cmidrule(lr){9-10}
        \textbf{Setting} & \textbf{Model}
          & CLIP$\uparrow$ & DINOv2$\uparrow$ & DreamSim$\downarrow$ & LPIPS$\downarrow$
          & PSNR$\uparrow$ & SSIM$\uparrow$
          & ImgRwd$\uparrow$ & HPS$\uparrow$ \\
        \midrule
        \multirow{2}{*}{Image-cond.}
          & GPT-Image-1.5          & 0.818 & 0.479 & 0.530 & 0.327 & 13.61 & 0.6911 & \textbf{0.098} & \textbf{9.585} \\
          & Gemini-3.1-Flash-Image & \textbf{0.829} & \textbf{0.478} & \textbf{0.502} & \textbf{0.078} & 22.074 & 0.9173 & $-$0.083 & 8.760 \\
        \midrule
        \multirow{2}{*}{Text-cond.}
          & GPT-Image-1.5          & 0.821 & 0.461 & 0.550 & 0.314 & 13.697 & 0.7041 & \textbf{0.061} & \textbf{9.439} \\
          & Gemini-3.1-Flash-Image & \textbf{0.827} & \textbf{0.470} & \textbf{0.522} & \textbf{0.107} & 20.518 & 0.9022 & $-$0.033 & 8.734 \\
        \bottomrule
    \end{tabular}
\end{table}

\FloatBarrier

\begin{figure*}[htbp]
\centering

\newcommand{\laiimg}[1]{%
    \includegraphics[width=\linewidth]{#1}%
}

\newcommand{\laicheckerasset}[1]{%
    \begingroup
    \sbox0{\includegraphics[width=0.85\linewidth]{#1}}%
    \begin{tikzpicture}[baseline=(current bounding box.center), inner sep=0pt, outer sep=0pt]
        \fill[white] (0,0) rectangle (\wd0,\ht0);
        \fill[pattern=checkerboard, pattern color=black!22] (0,0) rectangle (\wd0,\ht0);
        \node[anchor=south west, inner sep=0pt] at (0,0) {\usebox0};
    \end{tikzpicture}%
    \endgroup
}

\newcommand{\laigtfrom}[1]{%
    \IfFileExists{#1/gt.png}{%
        \laiimg{#1/gt.png}%
    }{%
        \IfFileExists{#1/full_layout.png}{%
            \laiimg{#1/full_layout.png}%
        }{%
            \laiimg{#1/original_layout.png}%
        }%
    }%
}

\newcommand{\laipromptbox}[1]{%
\begin{tcolorbox}[
    nobeforeafter,
    width=\linewidth,
    colback=gray!5,
    colframe=gray!45,
    boxrule=0.25pt,
    arc=1.2pt,
    left=2.5pt,right=2.5pt,top=2.5pt,bottom=2.5pt
]
{\fontsize{5.8}{6.6}\selectfont #1}
\end{tcolorbox}%
}

\newcommand{\laipromptitem}[2]{%
\textbf{#1}~#2\par\vspace{0.8pt}
}

\newcommand{\ColImg}{0.175\textwidth}
\newcommand{\ColText}{0.245\textwidth}

{\normalsize\textbf{(a) Image-conditioned}}\par\vspace{5pt}

\setlength{\tabcolsep}{3.0pt}
\begin{tabular}{@{} c c c c c @{}}
\parbox[c]{\ColImg}{\centering\textbf{Masked Image}} &
\parbox[c]{\ColText}{\centering\makecell[c]{\textbf{Asset}}} &
\parbox[c]{\ColImg}{\centering\textbf{GT}} &
\parbox[c]{\ColImg}{\centering\makecell[c]{\textbf{Gemini-3.1}\\\textbf{Flash}}} &
\parbox[c]{\ColImg}{\centering\textbf{GPT-Image 1.5}} \\[12pt]

\parbox[c]{\ColImg}{\centering\laiimg{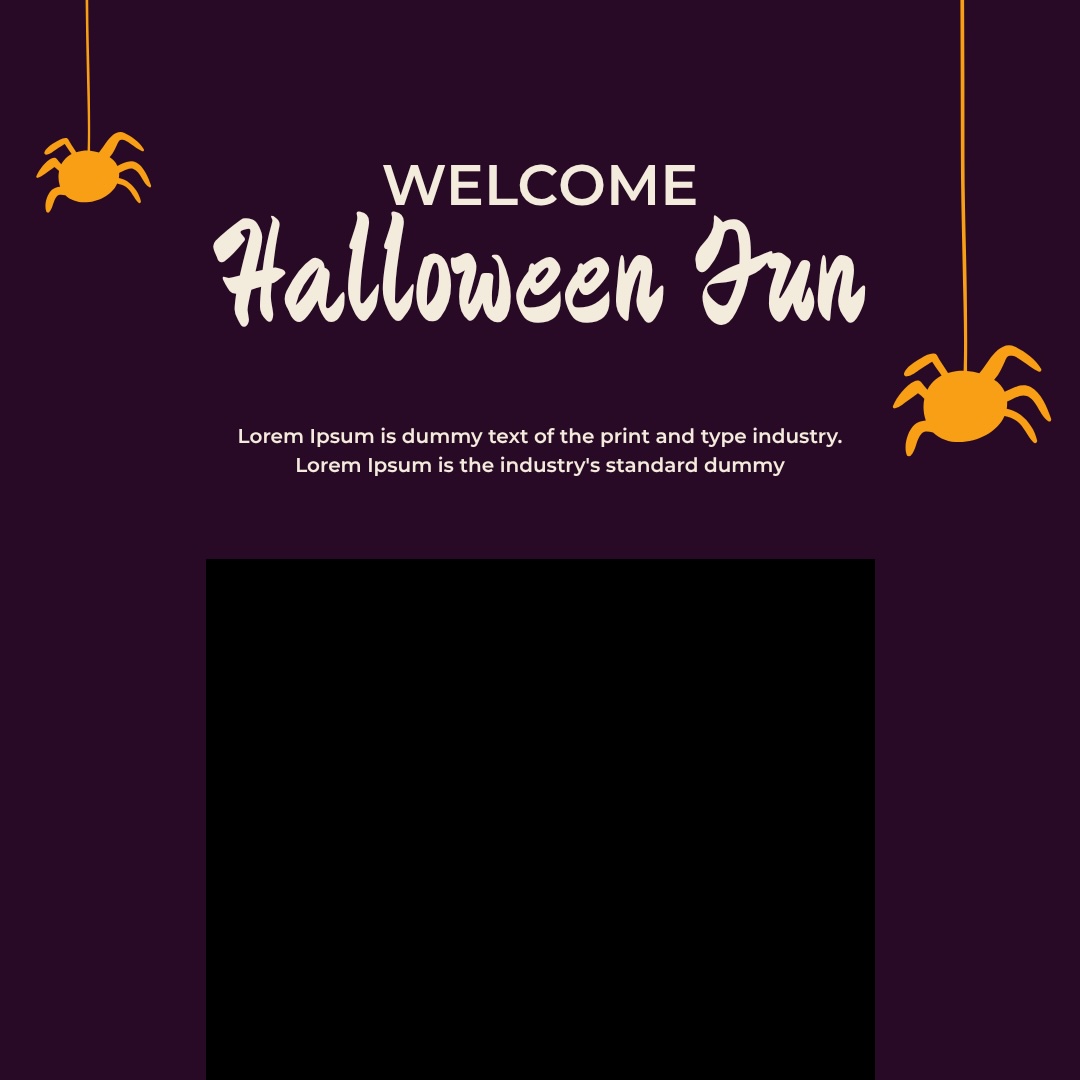}} &
\parbox[c]{\ColText}{\centering\laicheckerasset{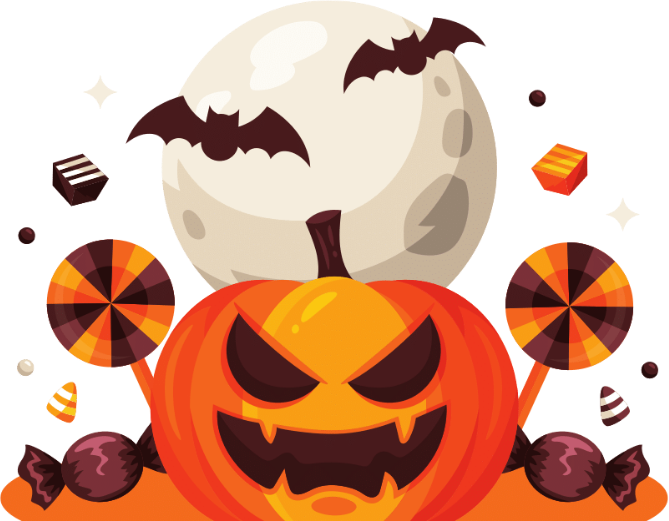}} &
\parbox[c]{\ColImg}{\centering\laiimg{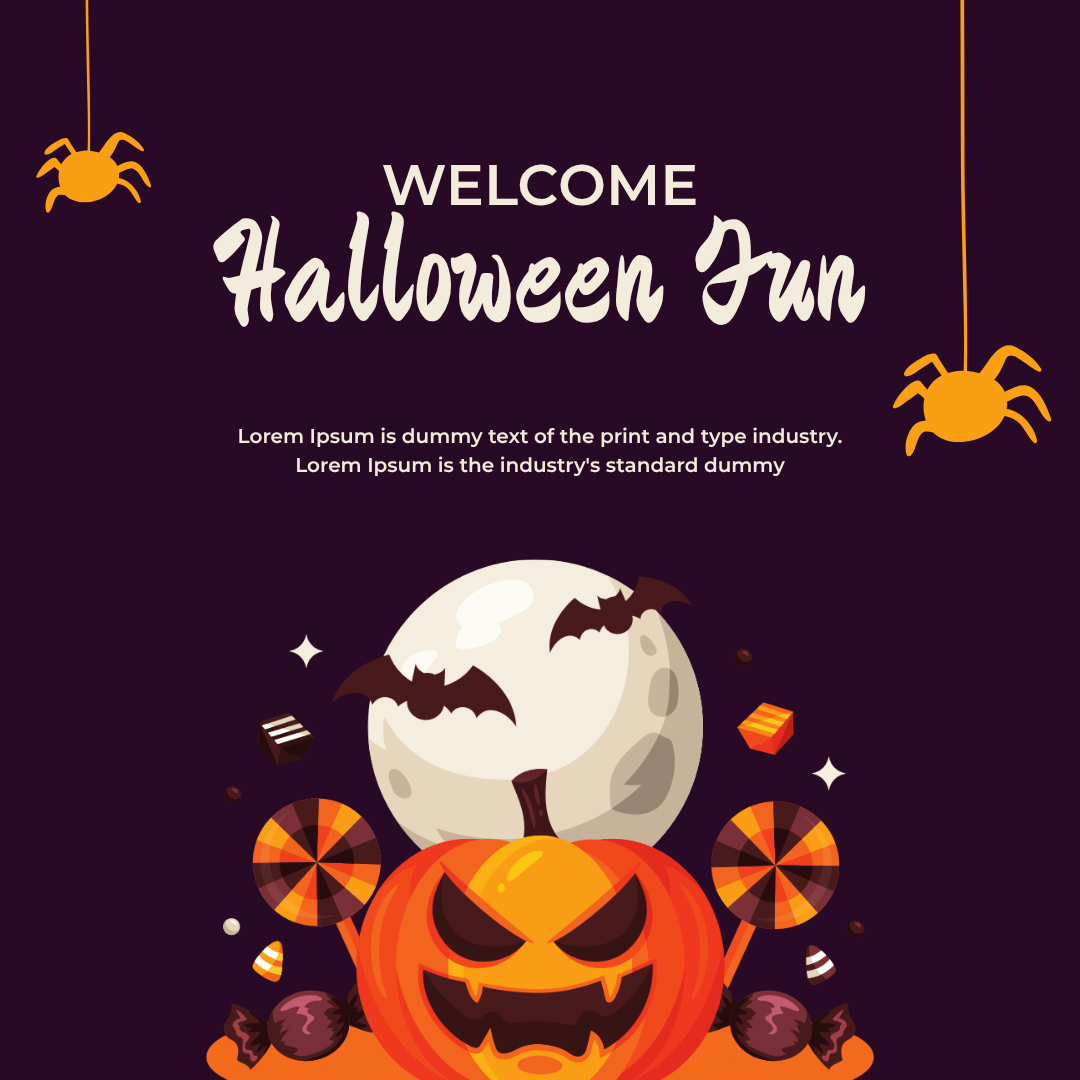}} &
\parbox[c]{\ColImg}{\centering\laiimg{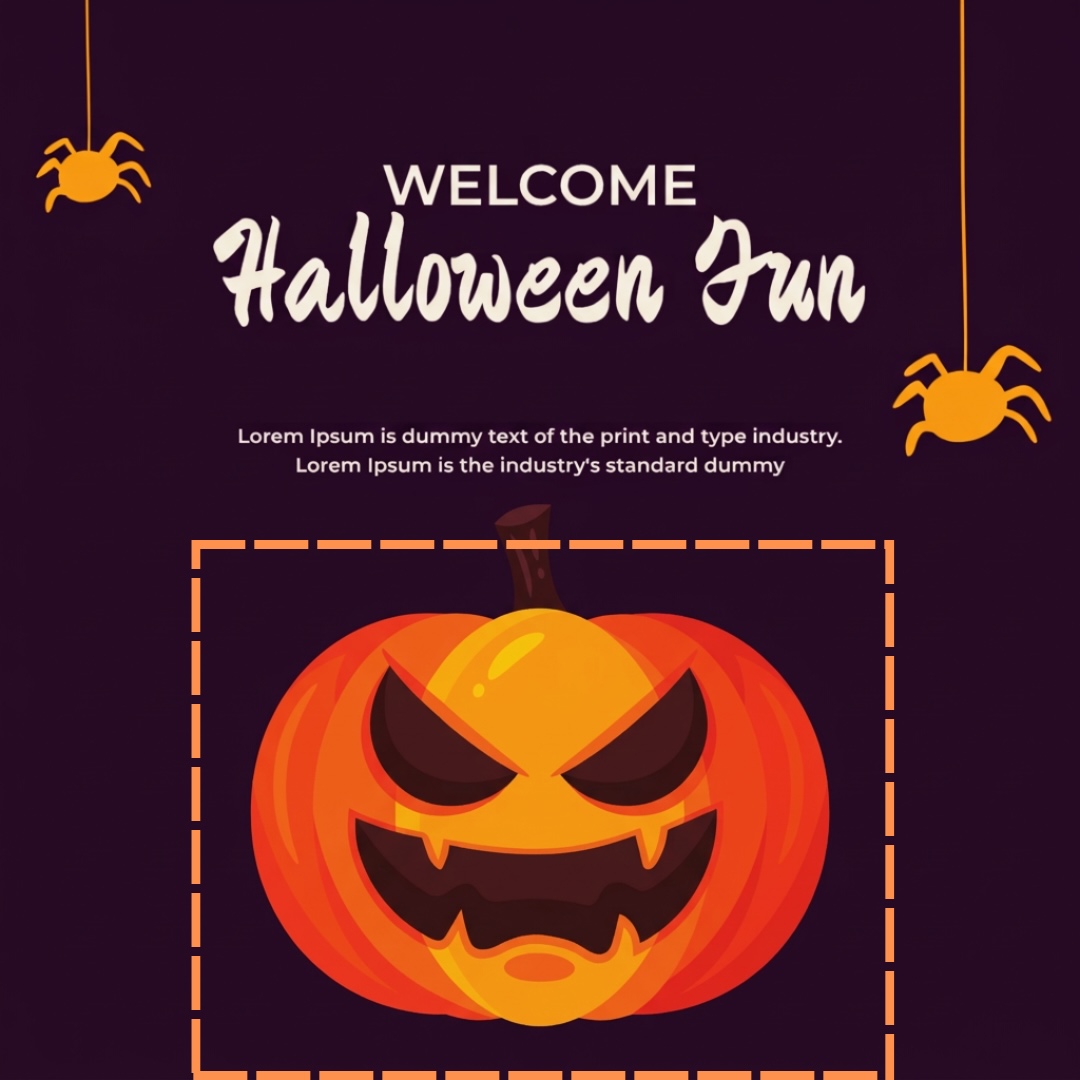}} &
\parbox[c]{\ColImg}{\centering\laiimg{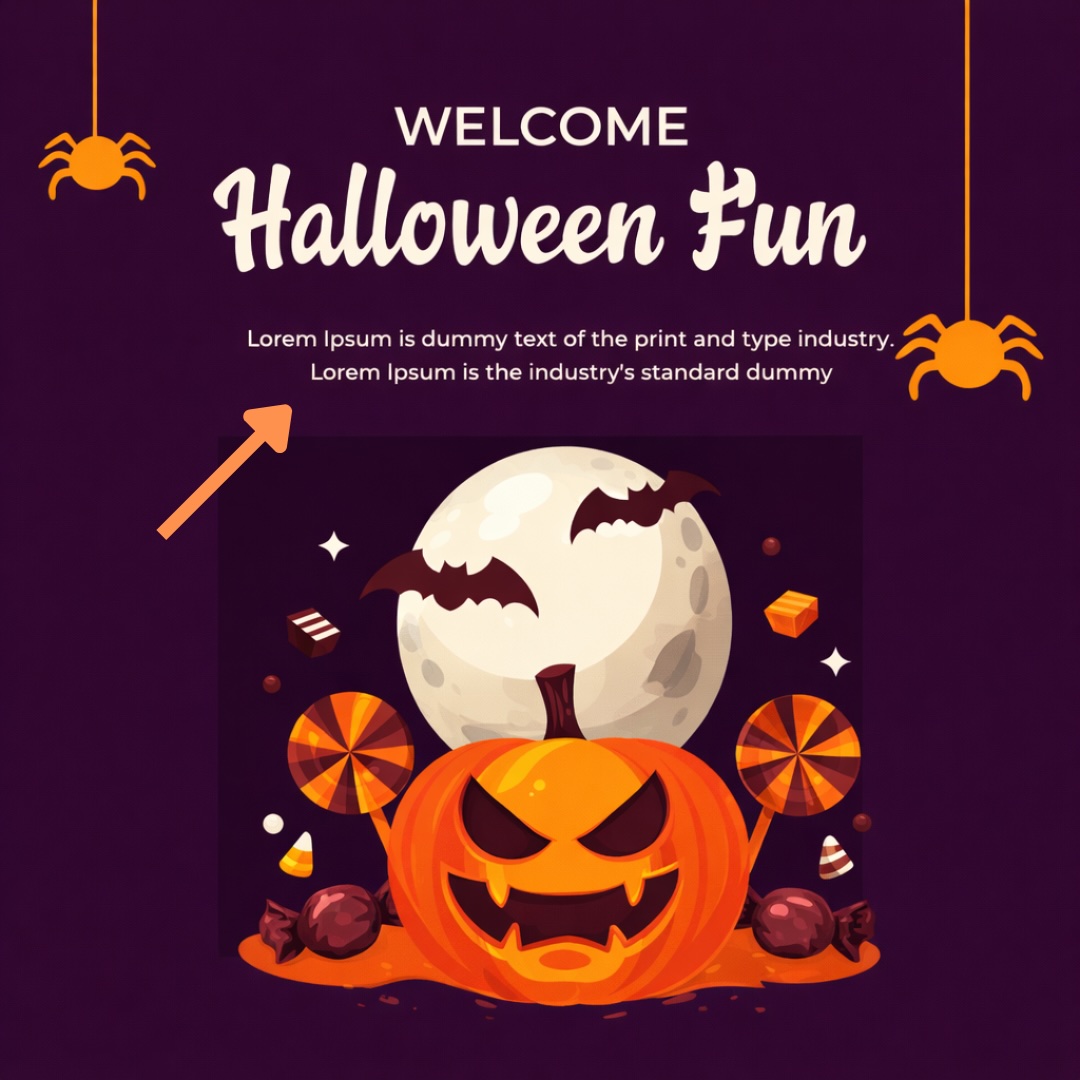}} \\
\end{tabular}

\vspace{6pt}

{\normalsize\textbf{(b) Text-conditioned}}\par\vspace{5pt}

\setlength{\tabcolsep}{3.0pt}
\begin{tabular}{@{} c c c c c @{}}
\parbox[c]{\ColImg}{\centering\laiimg{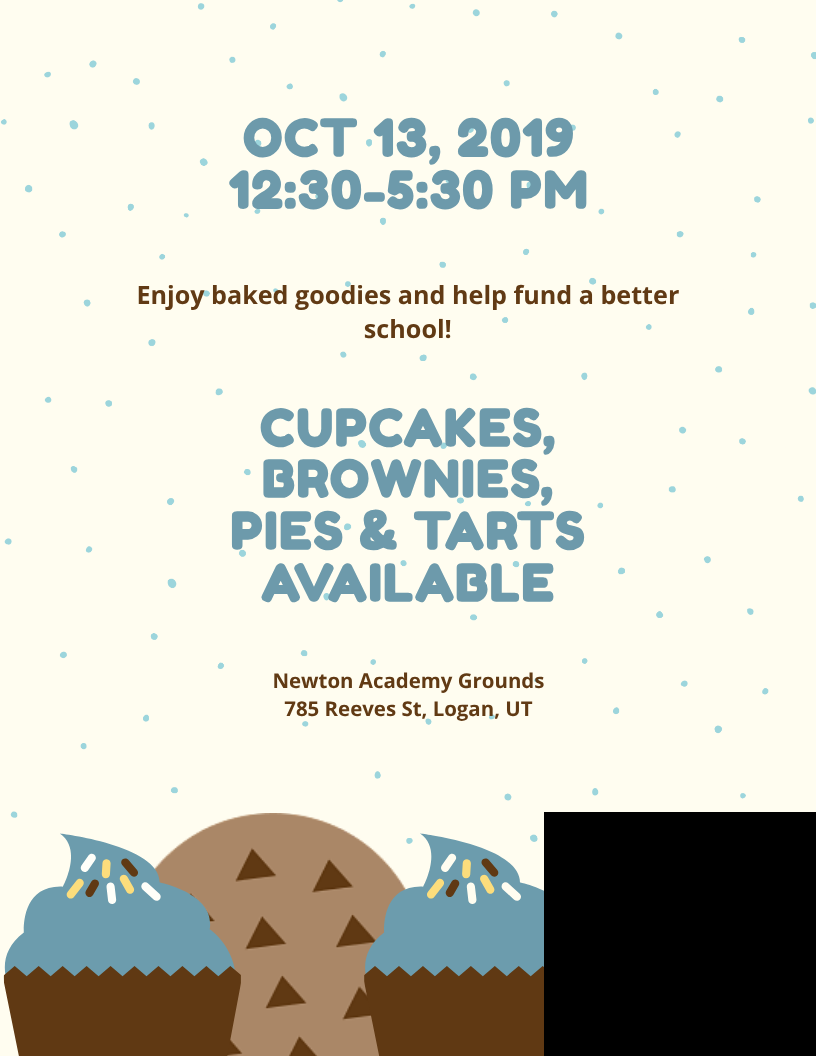}} &
\parbox[c]{\ColText}{\laipromptbox{%
\laipromptitem{Description:}{This image features a flat, minimalist illustration of a chocolate chip cookie on a transparent background. The main subject is a perfectly circular, medium brown cookie base, \textcolor{red}{adorned with eleven scattered}, darker brown triangular shapes representing chocolate chips. The overall style is simple and graphic, using solid colors and basic geometric forms.}
}} &
\parbox[c]{\ColImg}{\centering\laiimg{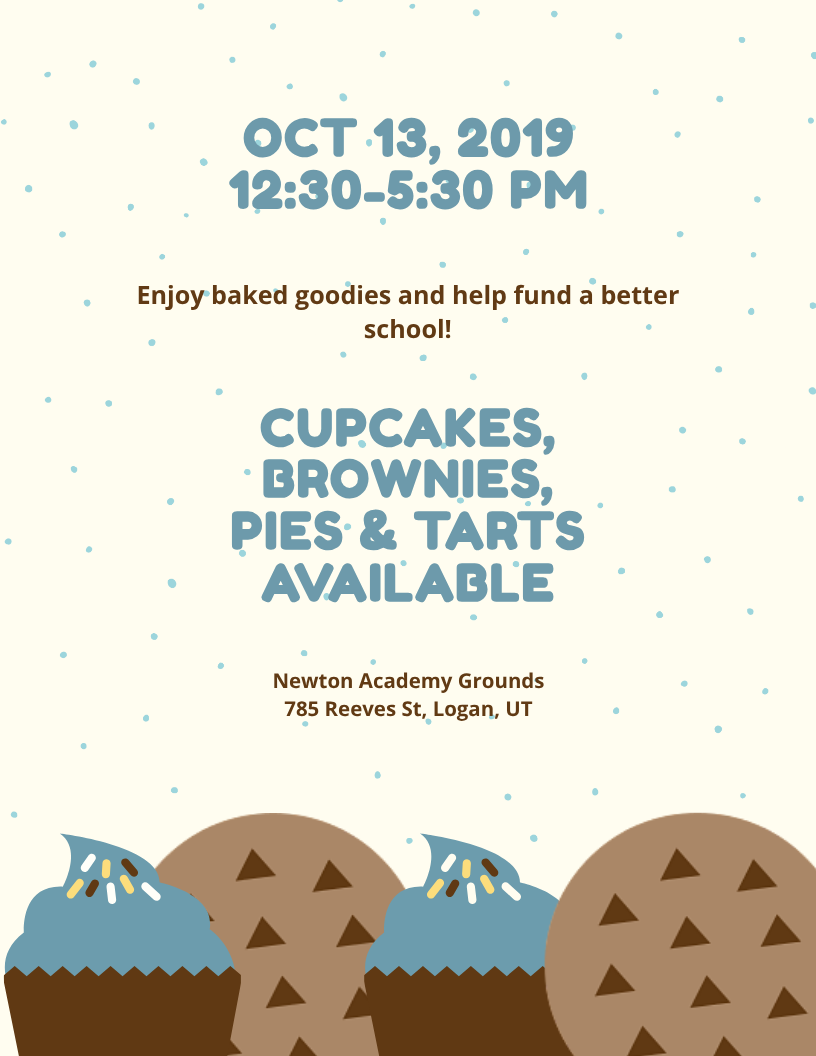}} &
\parbox[c]{\ColImg}{\centering\laiimg{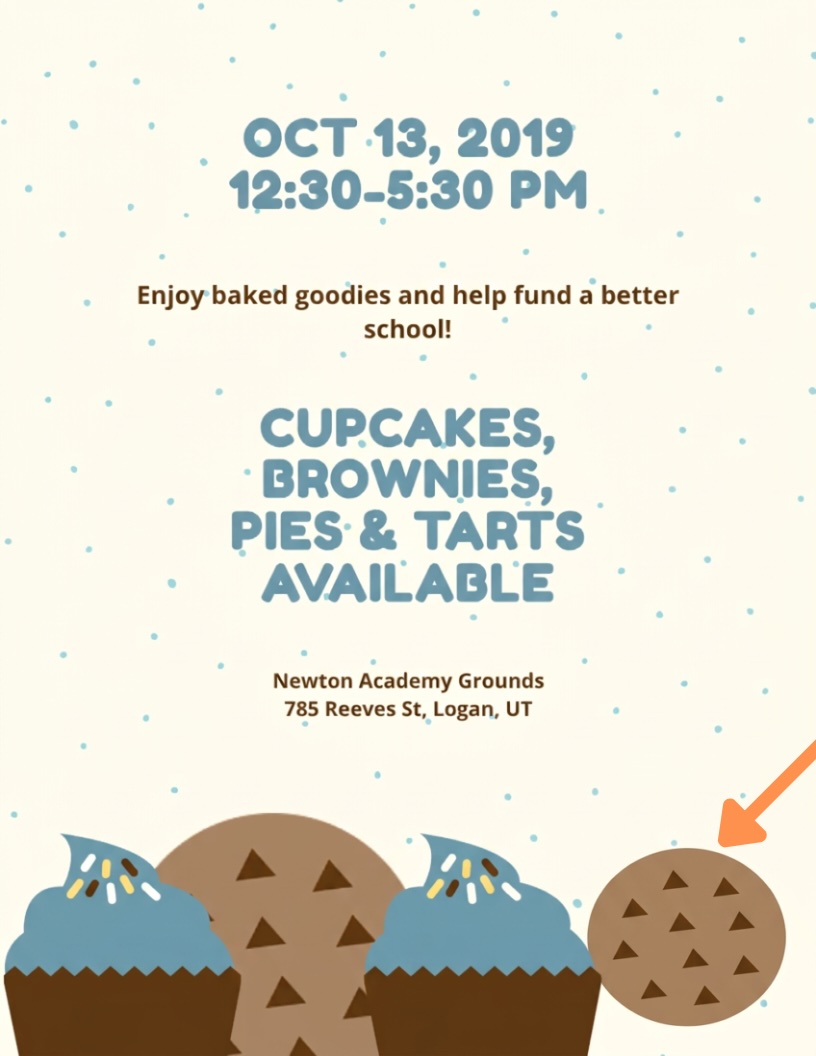}} &
\parbox[c]{\ColImg}{\centering\laiimg{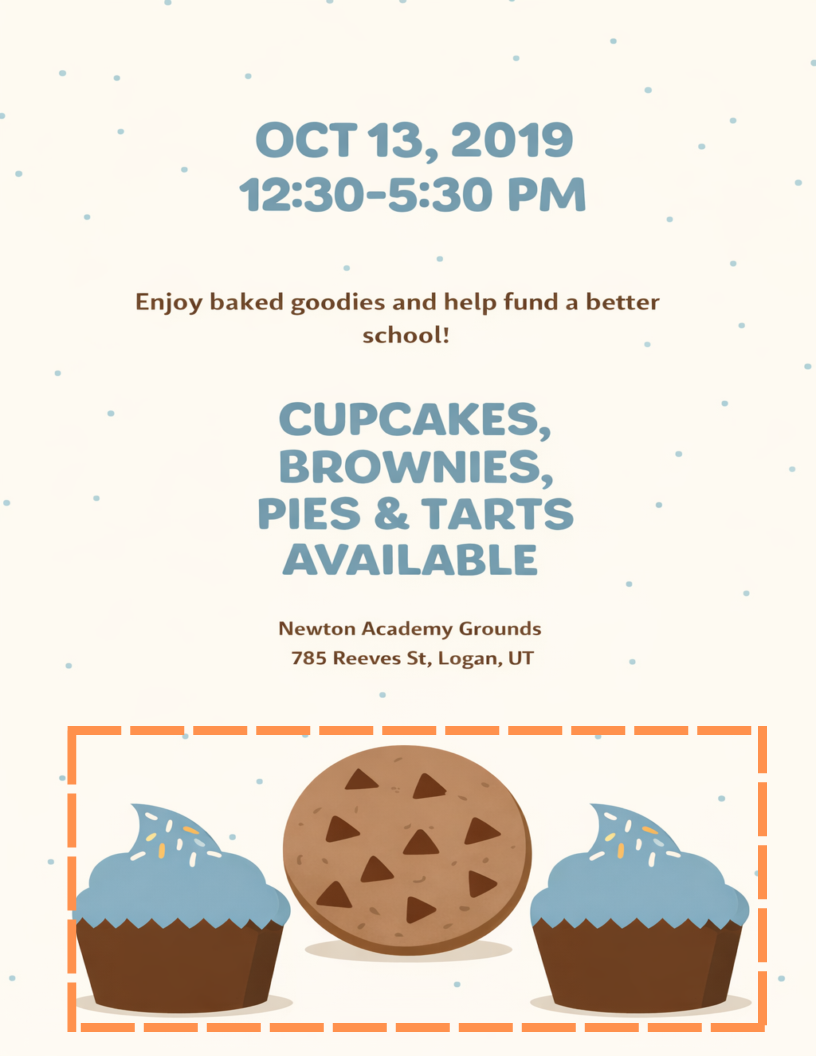}} \\
\end{tabular}

\caption{\textbf{Layer-aware inpainting.}
(a) Image-conditioned inpainting uses the original RGBA asset as reference, yet both models fail to preserve it: GPT-Image-1.5 introduces unintended changes to background color and content, while Gemini-3.1-Flash-Image severely distorts the input asset.
(b) Text-conditioned inpainting fails to satisfy the instruction, producing incorrect content and altering the surrounding layout.
\textcolor[HTML]{FF914D}{Orange arrows and boxes} highlight these failures.
Overall, both models fail to preserve background consistency and adhere to input constraints, resulting in incoherent designs.}
\label{fig:layer_aware_inpainting_combined}
\end{figure*}
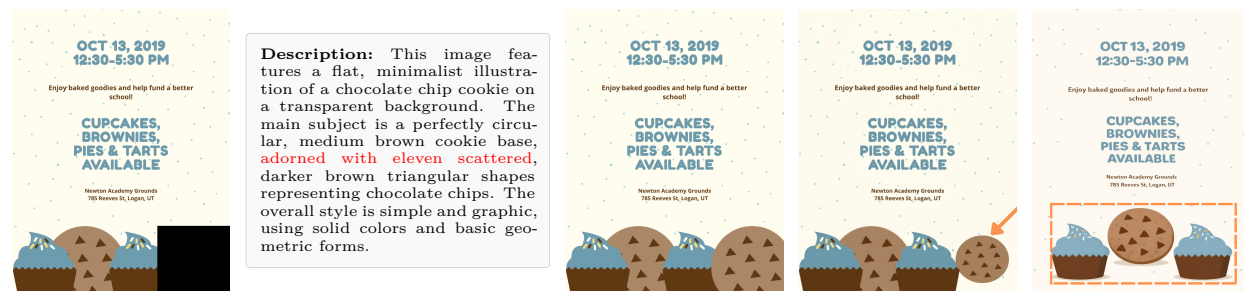

\FloatBarrier

\phantomsection\label{sec:aspect}

\begin{figure*}[htbp]
\centering

\newlength{\maraCellH}
\setlength{\maraCellH}{0.17\textheight}

\newcommand{\maraimg}[1]{%
\begin{minipage}[c][\maraCellH][c]{\linewidth}
    \centering
    \includegraphics[
        width=\linewidth,
        height=\maraCellH,
        keepaspectratio
    ]{#1}%
\end{minipage}%
}

\setlength{\tabcolsep}{2pt}
\begin{tabular}{
    >{\centering\arraybackslash}p{0.245\textwidth}
    >{\centering\arraybackslash}p{0.245\textwidth}
    >{\centering\arraybackslash}p{0.245\textwidth}
    >{\centering\arraybackslash}p{0.245\textwidth}
}
\makecell[c]{\textbf{Source Layout}\\\textbf{(9:16)}} &
\makecell[c]{\textbf{Target Layout}\\\textbf{(1:1)}} &
\makecell[c]{\textbf{Gemini-3.1}\\\textbf{Flash Image}} &
\makecell[c]{\textbf{GPT-Image 1.5}} \\[3pt]

\maraimg{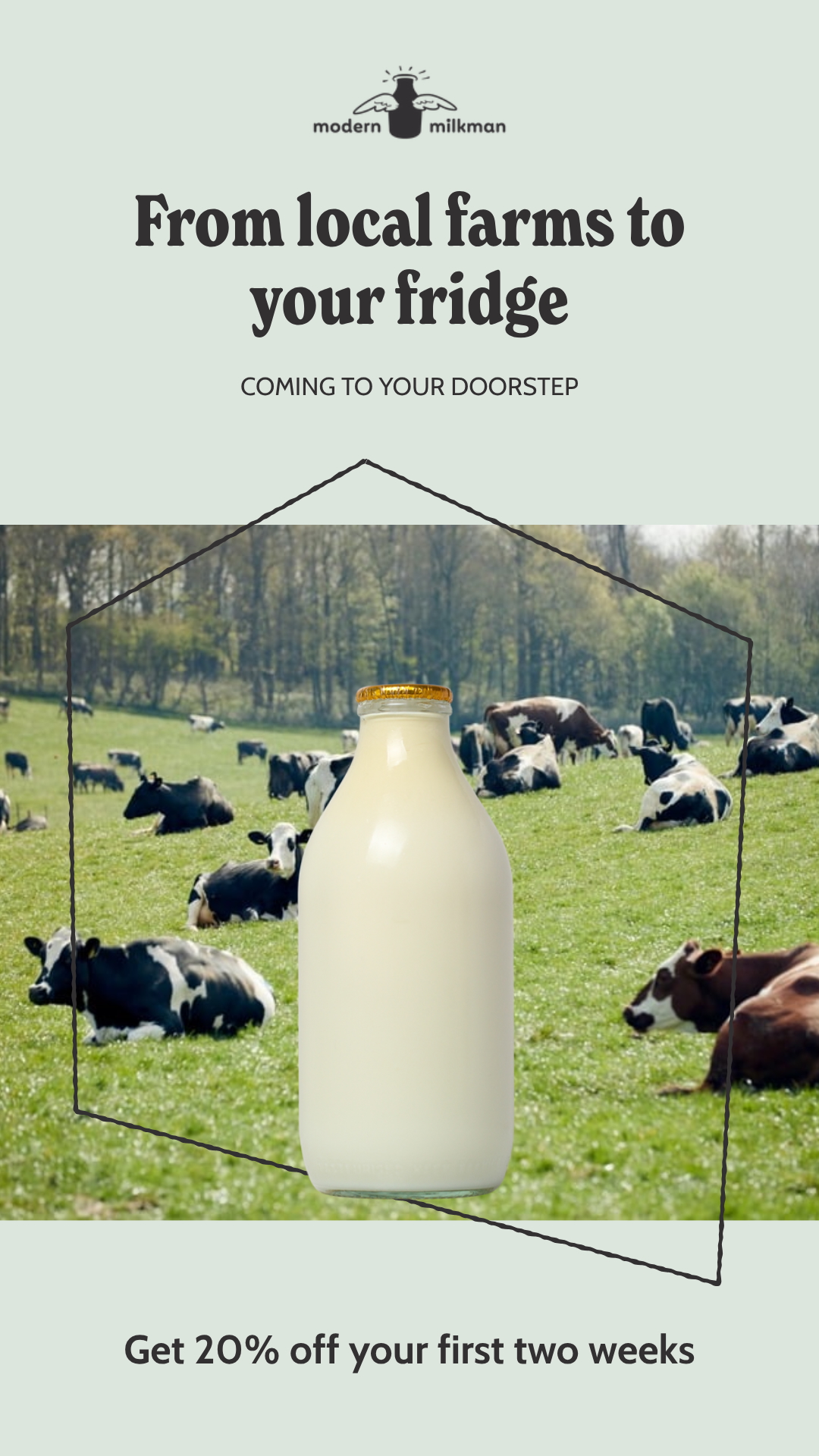}
&
\maraimg{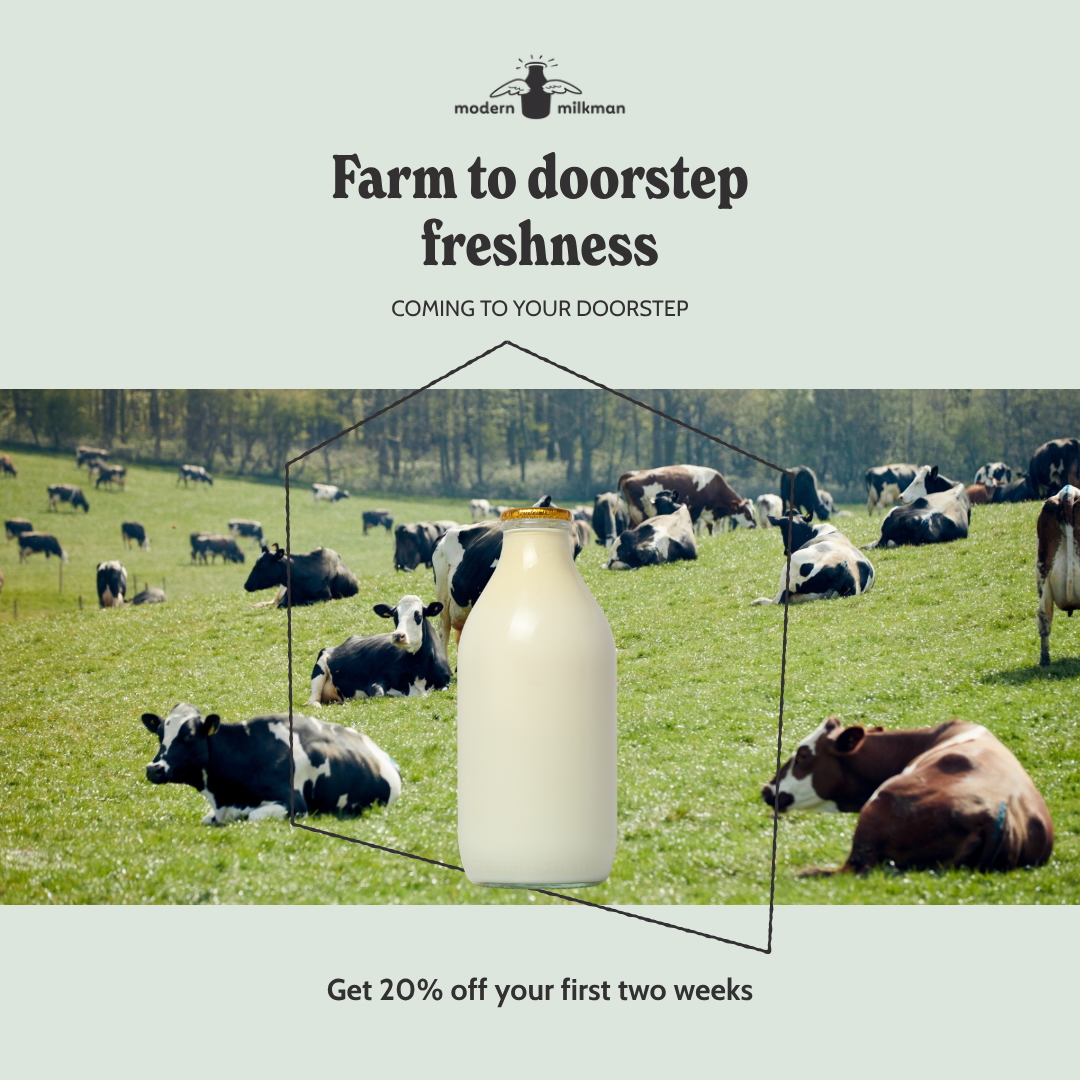}
&
\maraimg{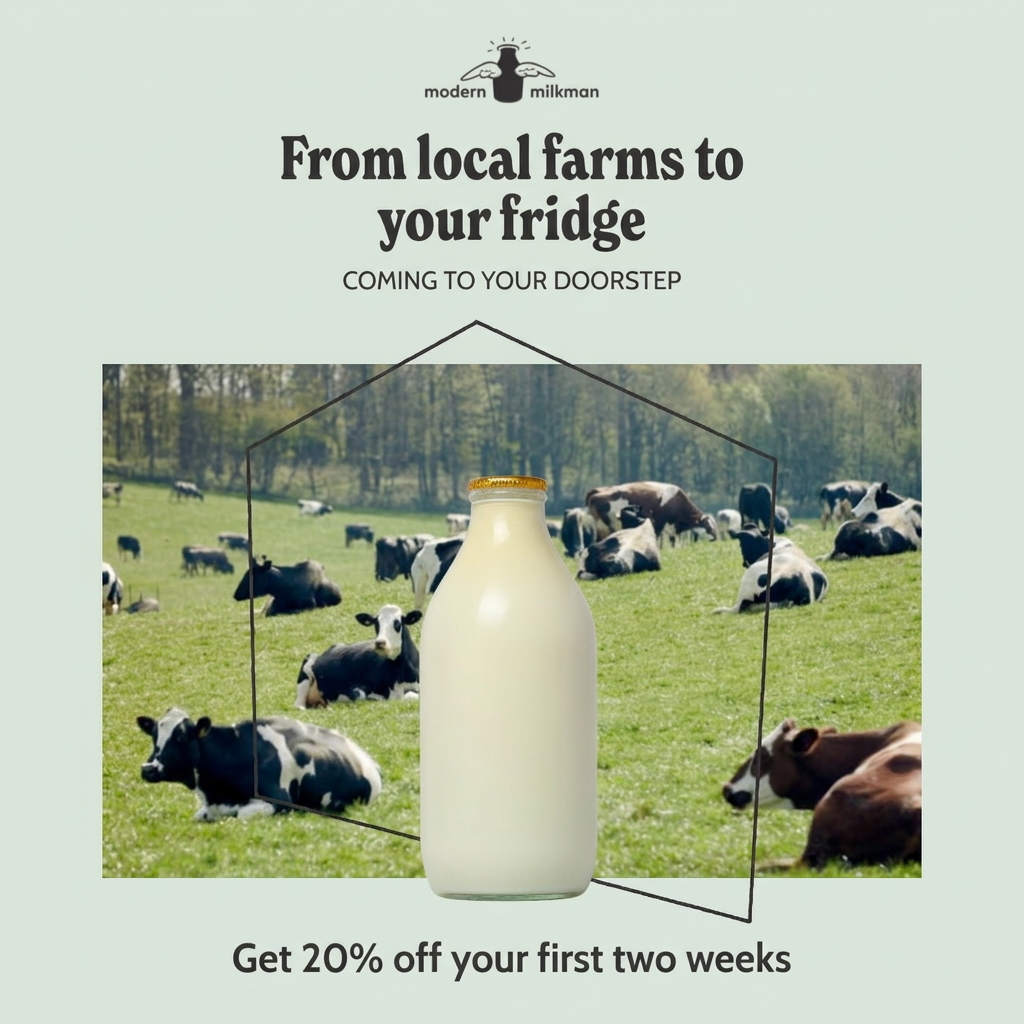}
&
\maraimg{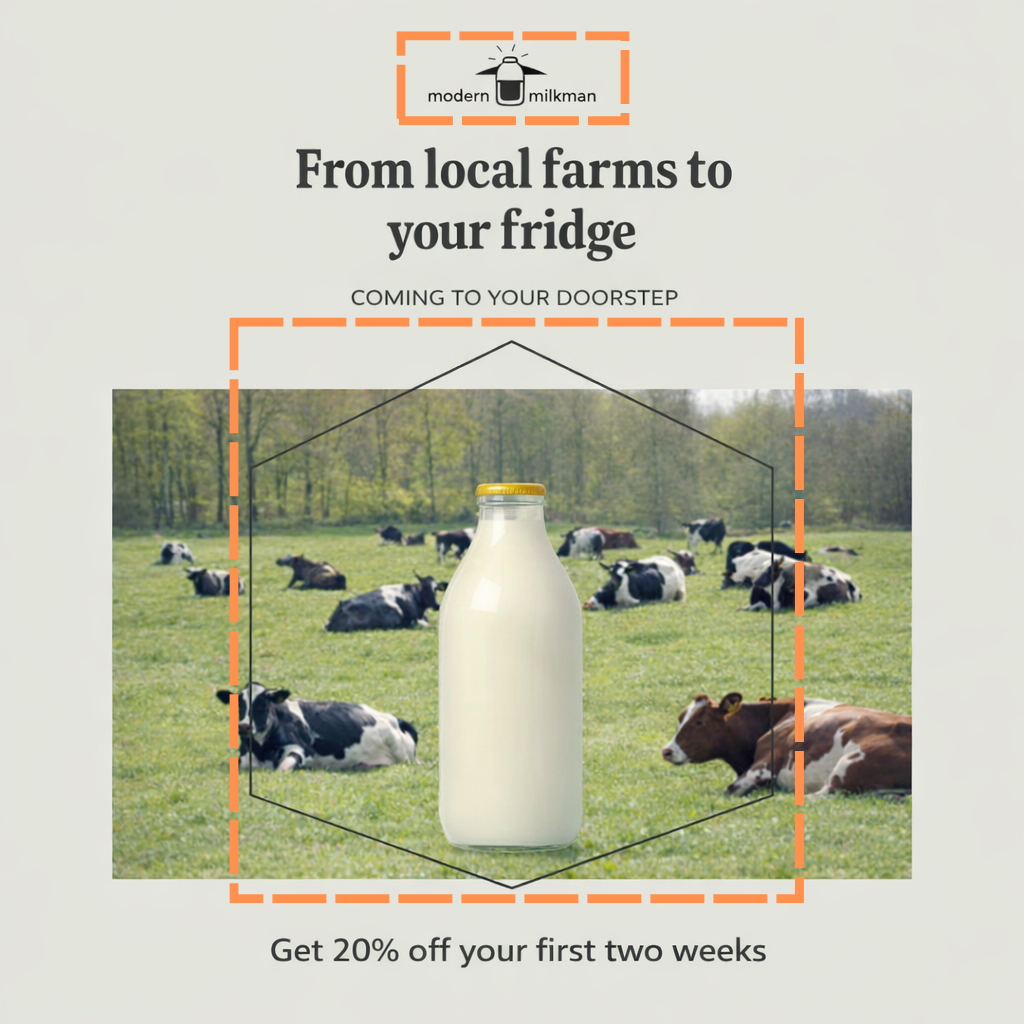}
\\[5pt]

\maraimg{figures/multi-aspect-ratio/G4_pair_001_M6REXUoS68ixhpN4fXmj_to_9sMhRCJ2XcTdwqTG5skx_input_image.jpg}
&
\maraimg{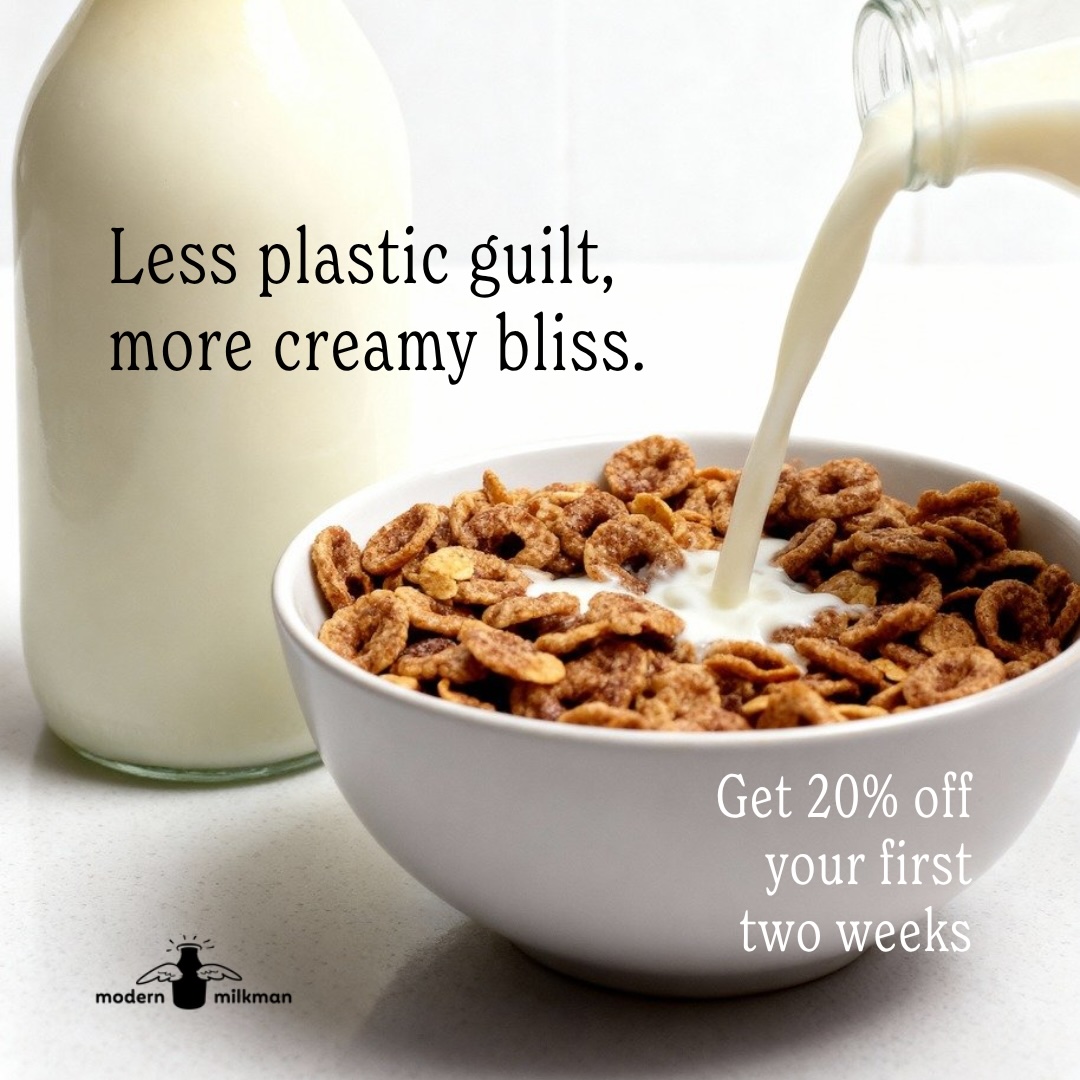}
&
\maraimg{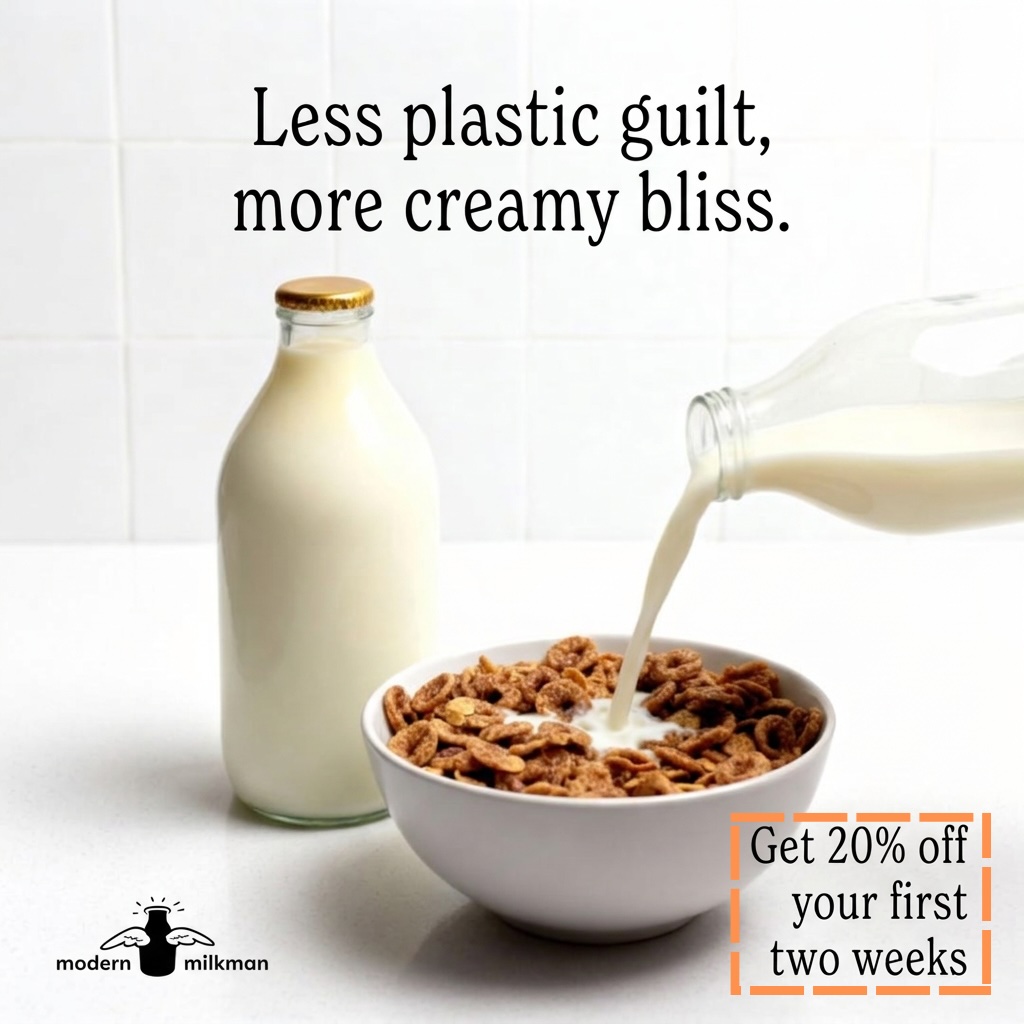}
&
\maraimg{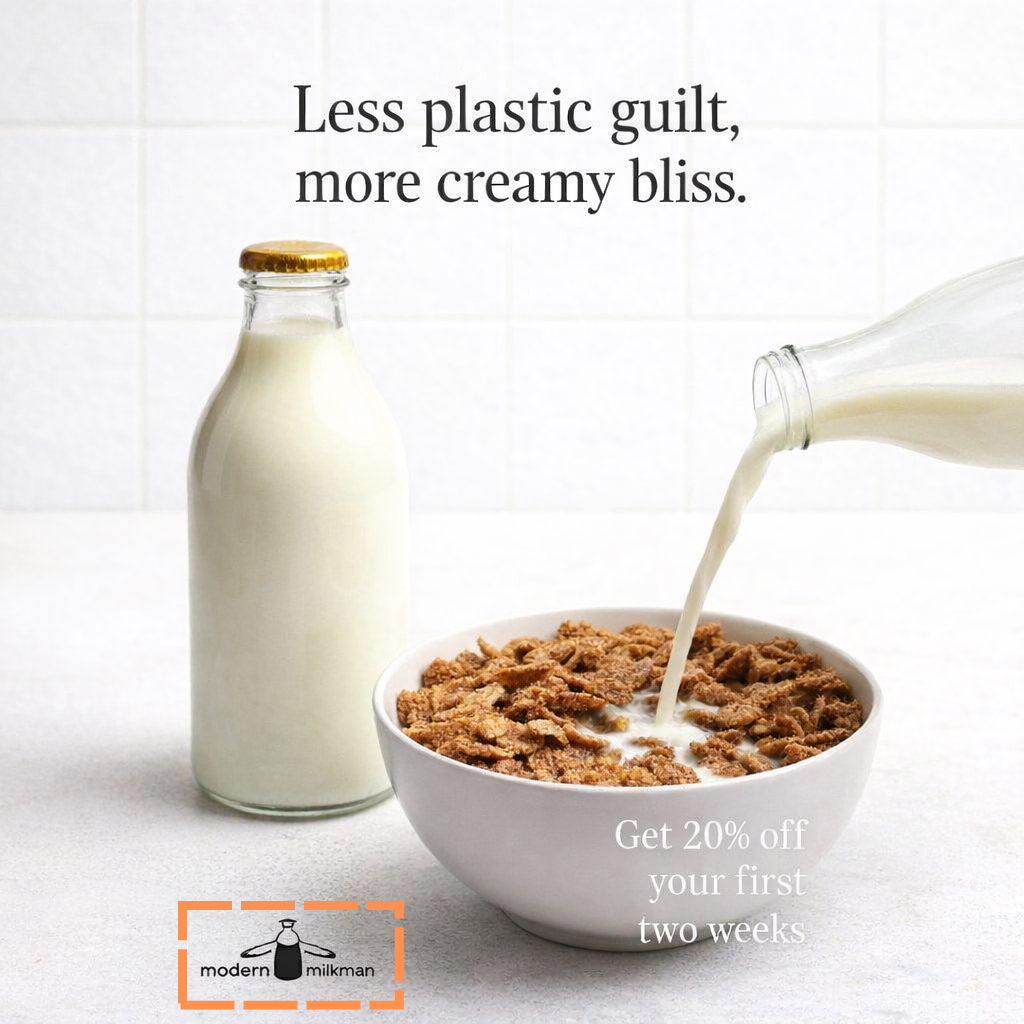}
\\

\end{tabular}

\caption{\textbf{Qualitative results for multi-aspect-ratio adaptation (9:16 $\rightarrow$ 1:1).}
For each example, we show the original source layout, the ground-truth target layout, and the adapted outputs from Gemini-3.1 Flash Image and GPT-Image 1.5.
The figure highlights whether each model preserves reading order, core assets, and text content while recomposing the design for a substantially different canvas ratio.}
\label{fig:multi_aspect_ratio_adaptation}
\end{figure*}

\vspace{1in}

\begin{table*}[h]
\centering
\caption{Multi-aspect-ratio adaptation results (long-to-short) using direct image generation. Metrics are grouped into human-evaluated structural-preservation measures and automated aesthetic and preference scores. The best result for each metric is shown in bold.}
\label{tab:multi-aspect-adaptation}
\small
\setlength{\tabcolsep}{4pt}
\begin{tabular}{@{}l ccc ccccc@{}}
\toprule
& \multicolumn{3}{c}{\textbf{Structural Preservation}} & \multicolumn{5}{c}{\textbf{Aesthetic \& Preference}} \\
\cmidrule(lr){2-4} \cmidrule(l){5-9}
\textbf{Model} & TextAcc$\uparrow$ & Recall$\uparrow$ & Halluc.$\downarrow$ & M-Judge$\uparrow$ & PickScore$\uparrow$ & HPSv3$\uparrow$ & ImgRwd$\uparrow$ & DreamSim$\downarrow$ \\
\midrule
GPT-Image-1.5          & 1.0 & 0.23 & 0.0 & 0.6153 & 21.583 & 9.2665 & 0.9913 & 0.1192 \\
Gemini-3.1-Flash-Image & 0.92 & 1.0 & 0.0 & 0.6667      & 18.79 & 8.5528 & -0.0204 & 0.0910 \\
\bottomrule
\end{tabular}
\end{table*}

\clearpage         

\begin{table}[H]
\centering
\small
\setlength{\tabcolsep}{5pt}
\caption{Summary of key findings across layout tasks.}
\begin{tabular}{>{\raggedright\arraybackslash}p{2.8cm}>{\raggedright\arraybackslash}p{3.5cm}>{\raggedright\arraybackslash}p{2.5cm}>{\raggedright\arraybackslash}p{2cm}>{\raggedright\arraybackslash}p{2cm}}
\toprule
\textbf{Task Group} & \textbf{Key Finding} & \textbf{Best Performance} & \textbf{Best Model} & \textbf{Status} \\
\midrule
Aspect Ratio \& Counting & Large performance gaps; counting errors are heavy-tailed on complex layouts & 93.9\% acc, MAE 5.81 & GPT-5.4 & Partially solved \\
\midrule
Component Type \& Detection & Models identify text reliably but fail on images and vectors; detection is orders of magnitude below natural-image benchmarks & 46.1\% type acc, 6.4\% mAP@.5 & GPT-5.4 & Unsolved \\
\midrule
Layer Order Prediction & Z-order inference is a distinct capability uncorrelated with other spatial tasks; Gemini leads despite trailing on all other layout tasks & $\tau$ = 0.567 & Gemini Flash Lite & Unsolved \\
\midrule
Image Rotation & All models struggle when images are rotated; rotated-only MAE exceeds 70° across the board & 80.0\% binary acc, 13.76° MAE & GPT-5.4 & Partially solved \\
\midrule
Crop Shape \& Frame Detection & Visual container reasoning is orthogonal to other spatial tasks; Claude leads despite trailing elsewhere & 76.9\% crop acc, F1 = 0.504 & Claude-Opus-4.6 & Unsolved \\
\midrule
Intent-to-Layout Generation & Models converge on similar aesthetic quality; text fidelity is a critical bottleneck with 1 in 4 strings rendered illegibly & OCR 75.4\%, Pick 20.21 & GPT-Image-1.5 & Unsolved \\
\midrule
Partial Layout Completion & Single-element placement is tractable; multi-element completion exposes a clear gap in spatial interdependency reasoning & mIoU 0.2413 (single), 0.2007 (multi) & Gemini-3.1-Pro & Unsolved \\
\midrule
Layer-Aware Inpainting & Trade-off between identity preservation and compositional harmony; input modality has little effect on output quality & LPIPS 0.078, HPSv3 9.585 & Gemini-3.1-Pro (identity), GPT-5.4 (quality) & Partially solved \\
\midrule
Multi-Aspect Ratio Adaptation{\footnotesize\,(Layout Generation)} & Text preservation and structural preservation diverge under large canvas changes; no single system consistently preserves all design constraints & (Human Eval) TextAcc 1.0, Recall 1.0, M-Judge 0.6667 & Metric-dependent & Partially solved \\

\bottomrule
\end{tabular}
\label{tab:layout_findings}
\end{table}

\clearpage

\section{Typography Tasks}
\label{sec:typography}

\subsection{Typography Understanding}
\label{sec:typography-understanding}
\begin{figure}[t]
\centering
\includegraphics[width=\linewidth]{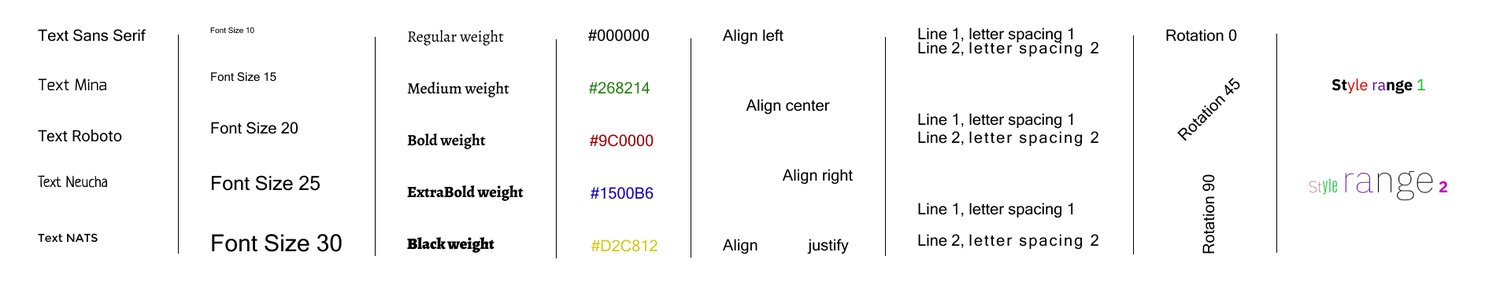}
\caption{Typography Properties. From left to right: font family, font size, font weight, text color, alignment, line spacing \& letter spacing, rotation and style ranges. An example for curved text appears in Figure~\ref{fig:typo-failures}.}
\label{fig:typography_properties}
\end{figure}

In graphic design, text is not merely readable content but a visual element that conveys hierarchy, tone, and brand identity through its typographic specification. We evaluate whether models can perceive fine-grained text properties across ten tasks on 2,568 text elements extracted from 989 layouts (Section~\ref{para:dataset}), spanning font identification, color estimation, property prediction, curved-text detection, inline style-span recovery, and rotation estimation. Table~\ref{tab:typography_tasks} lists the task definitions and dataset statistics for this domain.
Visual examples of the different typographic properties appear in Figure~\ref{fig:typography_properties}.

\paragraph{Results.}

Tables~\ref{tab:typo-results-font-color}--\ref{tab:typo-results-style-range} present the full results. The results reveal five key patterns across the typography understanding tasks:

\begin{table}[h]
\centering
\small
\setlength{\tabcolsep}{5pt}
\caption{Typography understanding task definitions and dataset statistics. All tasks draw from text components across 989 layouts, sampling up to 3 text elements per layout. For color prediction, 58 examples are excluded due to missing (optional) values.}
\begin{tabular}{>{\raggedright\arraybackslash}p{2.5cm}>{\raggedright\arraybackslash}p{5.5cm}>{\raggedright\arraybackslash}p{1.2cm}>{\raggedright\arraybackslash}p{2cm}>{\raggedright\arraybackslash}p{2.5cm}}
\toprule
\textbf{Task} & \textbf{Description} & \textbf{Samples} & \textbf{Classes} & \textbf{Metrics} \\
\midrule
Font Family Prediction & Identify the font family used for a target text string. Evaluation set contains 167 distinct typefaces with a long-tail distribution skewed toward Poppins, Montserrat, and Roboto. & 2,556 & 167 families & Top-1 Acc, Macro-F1 \\
\midrule
Text Color Prediction & Predict the foreground color of a text element as a hex value. A prediction is perceptually acceptable when $\Delta$E $<$ 5 (CIEDE2000). & 2,498 & Continuous & $\Delta$E, $\Delta$E$<$5, Hue Acc \\
\midrule
Font Size Prediction & Estimate the font size in pixels of a target text element from the rendered design. & 2,556 & Continuous (px) & MAE \\
\midrule
Font Weight Prediction & Predict the CSS font weight on the 100--900 scale. Weight 400 (regular) accounts for roughly half the dataset and weight 700 (bold) for about 30\%. & 2,556 & 100--900 & Accuracy \\
\midrule
Text Alignment Prediction & Predict whether a text element is aligned left, center, right, or justify. & 2,556 & 4 classes & Accuracy \\
\midrule
Line Height Prediction & Estimate the line height in pixels of a target text element. & 2,556 & Continuous (px) & MAE \\
\midrule
Letter Spacing Prediction & Estimate the letter spacing in em units of a target text element. & 2,556 & Continuous (em) & MAE \\
\midrule
Curved Text Detection & Determine whether a text element follows a curved path and estimate its curvature value in $[-100, +100]$. Only 1.0\% of elements are curved ($n=26$), making the curved-only metrics the informative ones. & 2,556 & $[-100, 100]$ & is-curved Acc, Curv. MAE \\
\midrule
Style Range Detection & Identify which character spans carry a distinct inline style and return their character-index intervals with associated properties. Of 2,568 samples, only 19 have two or more style ranges. & 2,556 & 1--6+ ranges & Span IoU, Exact Match \\
\midrule
Text Rotation Prediction & Predict the rotation angle in degrees of a target text element. Most text is axis-aligned, so the rotated-only MAE isolates angle estimation quality. & 2,556 & $[-180, 180]°$ & is-rot. Acc, Angle MAE \\
\bottomrule
\end{tabular}
\label{tab:typography_tasks}
\end{table}

\FloatBarrier

\begin{table}[H]
\centering
\caption{Font Family and Text Color results.
Best result per metric in \textbf{bold}.}
\label{tab:typo-results-font-color}
\small
\begin{tabular}{@{}l cc cccc@{}}
\toprule
& \multicolumn{2}{c}{\textbf{Font Family}}
& \multicolumn{4}{c}{\textbf{Text Color}} \\
\cmidrule(lr){2-3} \cmidrule(lr){4-7}
\textbf{Model}
  & Acc$\uparrow$ & F1$\uparrow$
  & $\Delta$E$\downarrow$ & RGB$\ell_2$$\downarrow$ & $\Delta$E${<}5$$\uparrow$ & HueAcc$\uparrow$ \\
\midrule
Gemini-3.1-Flash-Lite & 0.065 & 0.002 & 52.41 & 192.5 & 0.118 & 0.219 \\
Gemini-3.1-Pro        & 0.047 & 0.004 & 54.97 & 201.3 & 0.114 & 0.194 \\
GPT-5.4               & \textbf{0.237} & \textbf{0.055} & \textbf{5.57} & \textbf{20.0} & \textbf{0.633} & \textbf{0.881} \\
Claude-Opus-4.6       & 0.006 & 0.003 & 14.05 & 47.1 & 0.382 & 0.696 \\
\bottomrule
\end{tabular}
\end{table}

\begin{table}[H]
\centering
\caption{Typographic property results.
Best result per metric in \textbf{bold}.}
\label{tab:typo-results-properties}
\small
\begin{tabular}{@{}l c c c cc@{}}
\toprule
& \textbf{Font Size}
& \textbf{Weight}
& \textbf{Align}
& \textbf{Line Ht.}
& \textbf{Letter Sp.} \\
\cmidrule(lr){2-2} \cmidrule(lr){3-3} \cmidrule(lr){4-4} \cmidrule(lr){5-5} \cmidrule(lr){6-6}
\textbf{Model}
  & MAE$\downarrow$ & Acc$\uparrow$ & Acc$\uparrow$ & MAE$\downarrow$ & MAE$\downarrow$ \\
\midrule
Gemini-3.1-Flash-Lite & 38.42 & 0.390 & 0.428 & 43.88 & 0.045 \\
Gemini-3.1-Pro        & 35.53 & 0.407 & 0.441 & 42.16 & 0.038 \\
GPT-5.4               & \textbf{8.97} & \textbf{0.552} & 0.851 & \textbf{15.66} & \textbf{0.016} \\
Claude-Opus-4.6       & 10.84 & 0.491 & \textbf{0.861} & 17.17 & 0.034 \\
\bottomrule
\end{tabular}
\end{table}

\begin{table}[H]
\centering
\caption{Curved text detection and text rotation results.
Best per metric in \textbf{bold}.
``Curv-only'' columns restricted to the 26 curved elements.}
\label{tab:typo-results-curvature-rotation}
\small
\begin{tabular}{@{}l cccc ccc@{}}
\toprule
& \multicolumn{4}{c}{\textbf{Curved Text ($n$=2{,}568)}}
& \multicolumn{3}{c}{\textbf{Text Rotation ($n$=2{,}568)}} \\
\cmidrule(lr){2-5} \cmidrule(lr){6-8}
\textbf{Model}
  & Acc$\uparrow$ & MAE$\downarrow$ & Det\textsubscript{curv}$\uparrow$ & MAE\textsubscript{curv}$\downarrow$
  & Acc$\uparrow$ & MAE$\downarrow$ & MAE\textsubscript{rot}$\downarrow$ \\
\midrule
Gemini-3.1-Flash-Lite & 0.972 & 1.26 & 0.000 & 59.50 & 0.927 & 4.18 & 53.99 \\
Gemini-3.1-Pro        & 0.979 & 1.15 & 0.000 & 59.50 & 0.928 & 4.08 & 55.23 \\
GPT-5.4               & \textbf{0.995} & \textbf{0.58} & 0.885 & \textbf{43.77} & \textbf{0.980} & \textbf{0.93} & \textbf{18.67} \\
Claude-Opus-4.6       & 0.975 & 1.23 & \textbf{0.923} & 79.92 & 0.972 & 1.41 & 30.79 \\
\bottomrule
\end{tabular}
\end{table}

\begin{table}[H]
\centering
\caption{Style range detection: span IoU and exact match, stratified by number of ground-truth style ranges.}
\label{tab:typo-results-style-range}
\small
\begin{tabular}{@{}l cc cc cc@{}}
\toprule
& \multicolumn{2}{c}{\textbf{All ($n$=2{,}568)}}
& \multicolumn{2}{c}{\textbf{1-range ($n$=2{,}549)}}
& \multicolumn{2}{c}{\textbf{$\geq$\,2-range ($n$=19)}} \\
\cmidrule(lr){2-3} \cmidrule(lr){4-5} \cmidrule(lr){6-7}
\textbf{Model}
  & IoU$\uparrow$ & EM$\uparrow$
  & IoU$\uparrow$ & EM$\uparrow$
  & IoU$\uparrow$ & EM$\uparrow$ \\
\midrule
Gemini-3.1-Flash-Lite & 0.376 & 0.000 & 0.377 & 0.000 & 0.244 & 0.000 \\
Gemini-3.1-Pro        & 0.375 & 0.000 & 0.376 & 0.000 & 0.215 & 0.000 \\
GPT-5.4               & 0.975 & 0.000 & 0.977 & 0.000 & \textbf{0.824} & 0.000 \\
Claude-Opus-4.6       & \textbf{0.982} & 0.000 & \textbf{0.984} & 0.000 & 0.775 & 0.000 \\
\bottomrule
\end{tabular}
\end{table}

\FloatBarrier

\vspace{0.5in}

\begin{figure*}[t]
\centering

\begin{minipage}[c]{0.31\textwidth}
\centering
\includegraphics[width=\linewidth]{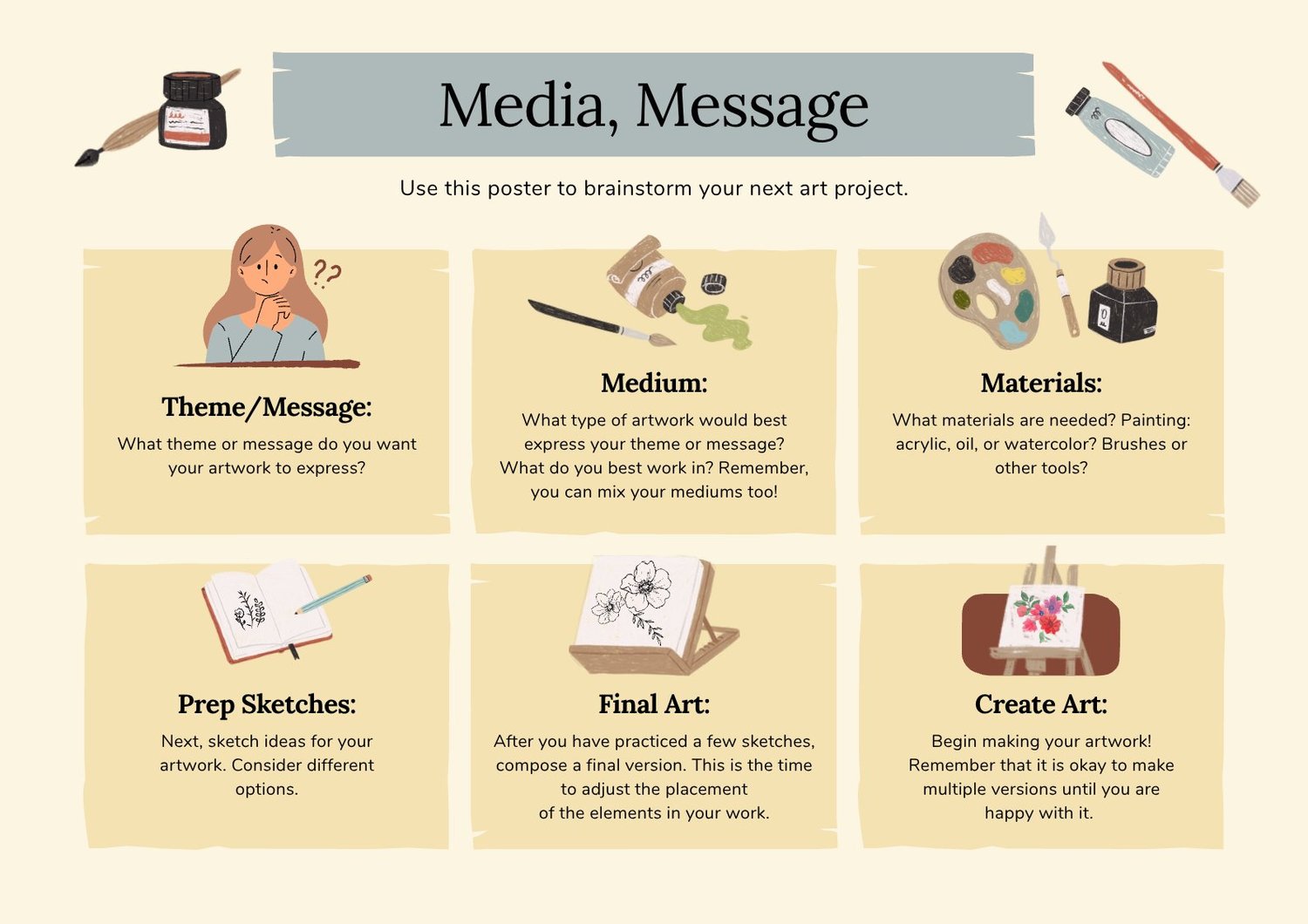}
\end{minipage}
\hfill
\begin{minipage}[c]{0.31\textwidth}
\centering
\includegraphics[width=\linewidth]{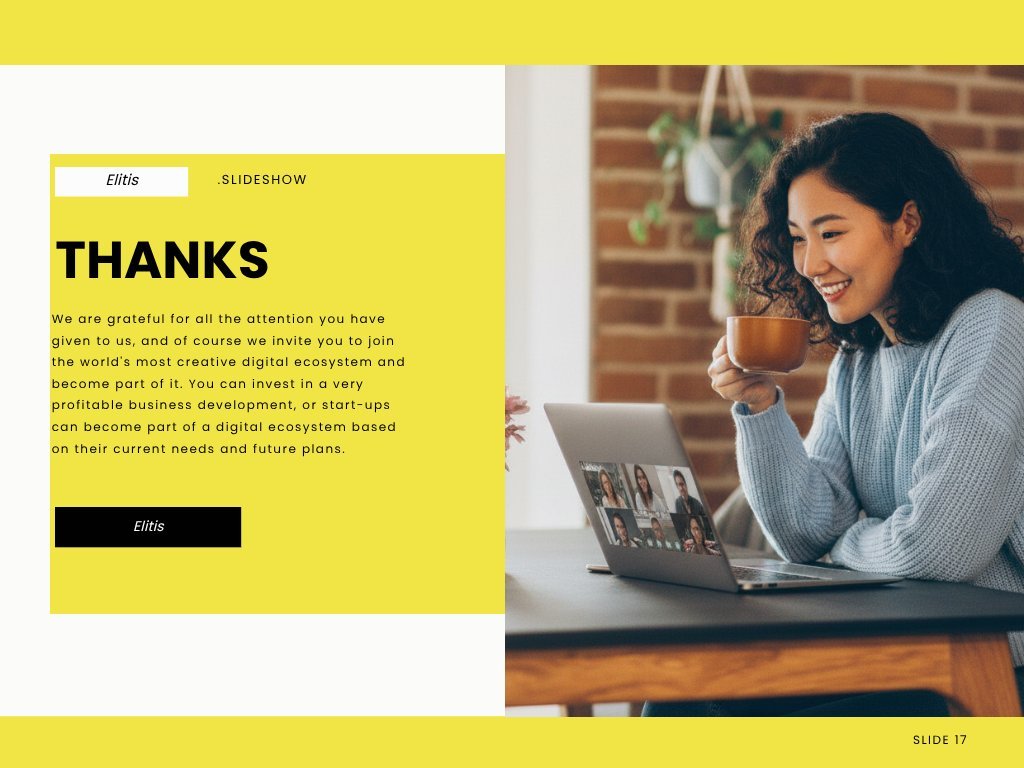}
\end{minipage}
\hfill
\begin{minipage}[c]{0.31\textwidth}
\centering
\includegraphics[width=\linewidth]{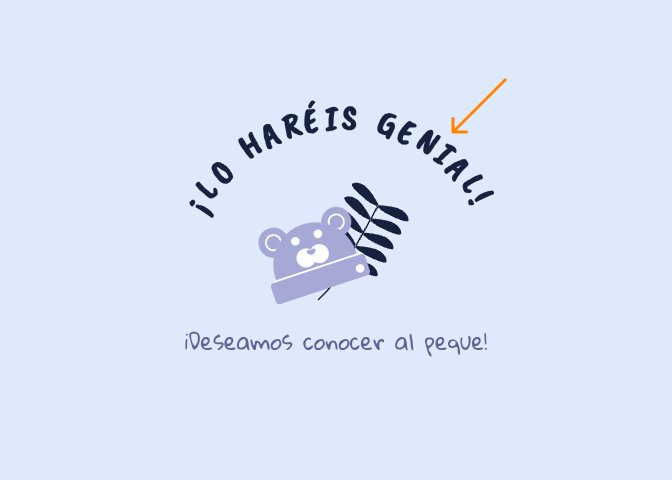}
\end{minipage}

\vspace{2pt}

\begin{minipage}[t]{0.31\textwidth}
\raggedright\small
\textbf{(a) Font Family Prediction}\\[2pt]
\textbf{GT:} \\ {\Lora "Message"} -- Lora \\
\textbf{GPT-5.4:} \\ {\CormorantGaramond "Message"} -- Cormorant Garamond\\
\textbf{Gemini Lite:} \\ {\Montserrat "Message"} -- Montserrat \\
\textbf{Gemini Pro:} \\ {\BebasNeue "Message"} -- Bebas Neue \\
GPT recognizes the serif style but picks the wrong family. Both Gemini variants predict fonts from entirely different categories.
\end{minipage}
\hfill
\begin{minipage}[t]{0.31\textwidth}
\raggedright\small
\textbf{(b) Text Color Prediction}\\[2pt]
\textbf{GT:} \texttt{rgb(1,1,1)} \,\fcolorbox{black}{black}{\rule{0pt}{5pt}\rule{5pt}{0pt}}\, \\
\textbf{GPT-5.4:} \texttt{\#FFFFFF} \,\fcolorbox{black}{white}{\rule{0pt}{5pt}\rule{5pt}{0pt}}\\
The model predicts the exact opposite color. Gemini Pro returns empty strings on multiple color queries, while Gemini Lite defaults to \texttt{\#000000} regardless of the actual color.
\end{minipage}
\hfill
\begin{minipage}[t]{0.31\textwidth}
\raggedright\small
\textbf{(c) Curved Text Detection}\\[2pt]
\textbf{GT:} \\ curved, curvature = 55\\
\textbf{Gemini Lite:} \\ straight, curvature = 0\\
A clearly arced text element is classified as straight. Both Gemini models achieve zero detection rate on all 26 curved elements, predicting straight for every sample.
\end{minipage}

\caption{Representative failure cases for typography understanding tasks.
Models confuse font categories~(a), predict inverted colors~(b), and miss curved text entirely~(c).
See task prompts in Table~\ref{tab:prompts-typo}.
}
\label{fig:typo-failures}
\end{figure*}

\begin{itemize}
    \item \textbf{Font recognition is weak across the board.} The best model achieves only 23.7\% top-1 accuracy ($3.6\times$ over the next best), with Macro-F1 an order of magnitude lower, confirming that correct predictions concentrate on high-frequency typefaces while the vast majority of the 167 families score zero F1.
    \item \textbf{Color perception varies dramatically across models.} The best model achieves $\Delta$E\,=\,5.57 with 63.3\% of predictions within the perceptually acceptable $\Delta$E\,$<$\,5 threshold, while the weakest models produce $\Delta$E\,$>$\,52, colors that are perceptually unrelated to the ground truth. Claude occupies an intermediate position ($\Delta$E\,=\,14.05, hue accuracy 69.6\%).
    \item \textbf{Typographic property estimation is unevenly solved.} Font size and line height show the largest gaps (up to $4\times$ difference; GPT-5.4 MAE\,=\,8.97\,px vs.\ Gemini ${\sim}$36\,px), while letter spacing is the easiest property for all models (MAE\,$<$\,0.045\,em). Text alignment is the most consistently predicted property, with the best model reaching 86.1\%.
    \item \textbf{Curved text and rotation expose a binary capability gap.} The best models correctly identify 88.5--92.3\% of the 26 curved elements, while some models score zero, classifying every element as straight. On text rotation, rotated-only MAE ranges from 18.67\textdegree{} to over 54\textdegree{} across models.
    \item \textbf{Style range localization is achievable but exact recovery is not.} The best models achieve span IoU of up to 0.982 (vs.\ ${\sim}$0.376 for Gemini), but exact match is 0\% across all models and all stratifications, indicating that while models can localize styled spans, they cannot jointly recover the full typographic specification within those spans. For representative failure cases, see Figure~\ref{fig:typo-failures}.
\end{itemize}

\begin{figure}[t]
\centering

\begin{minipage}[c]{0.3\columnwidth}
\centering
\includegraphics[width=\linewidth]{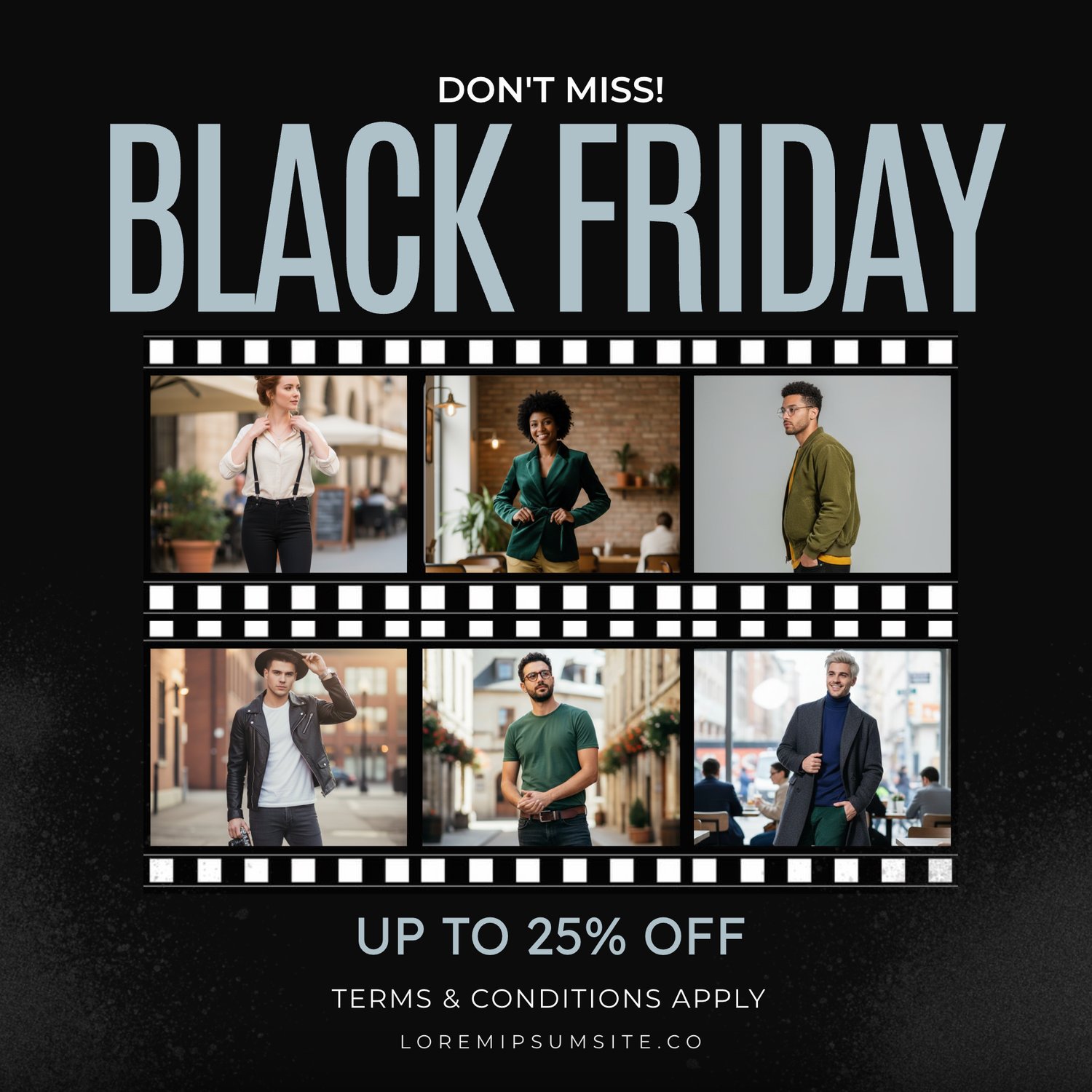}
\end{minipage}
\hspace{0.05\columnwidth}
\begin{minipage}[c]{0.3\columnwidth}
\centering
\includegraphics[width=\linewidth]{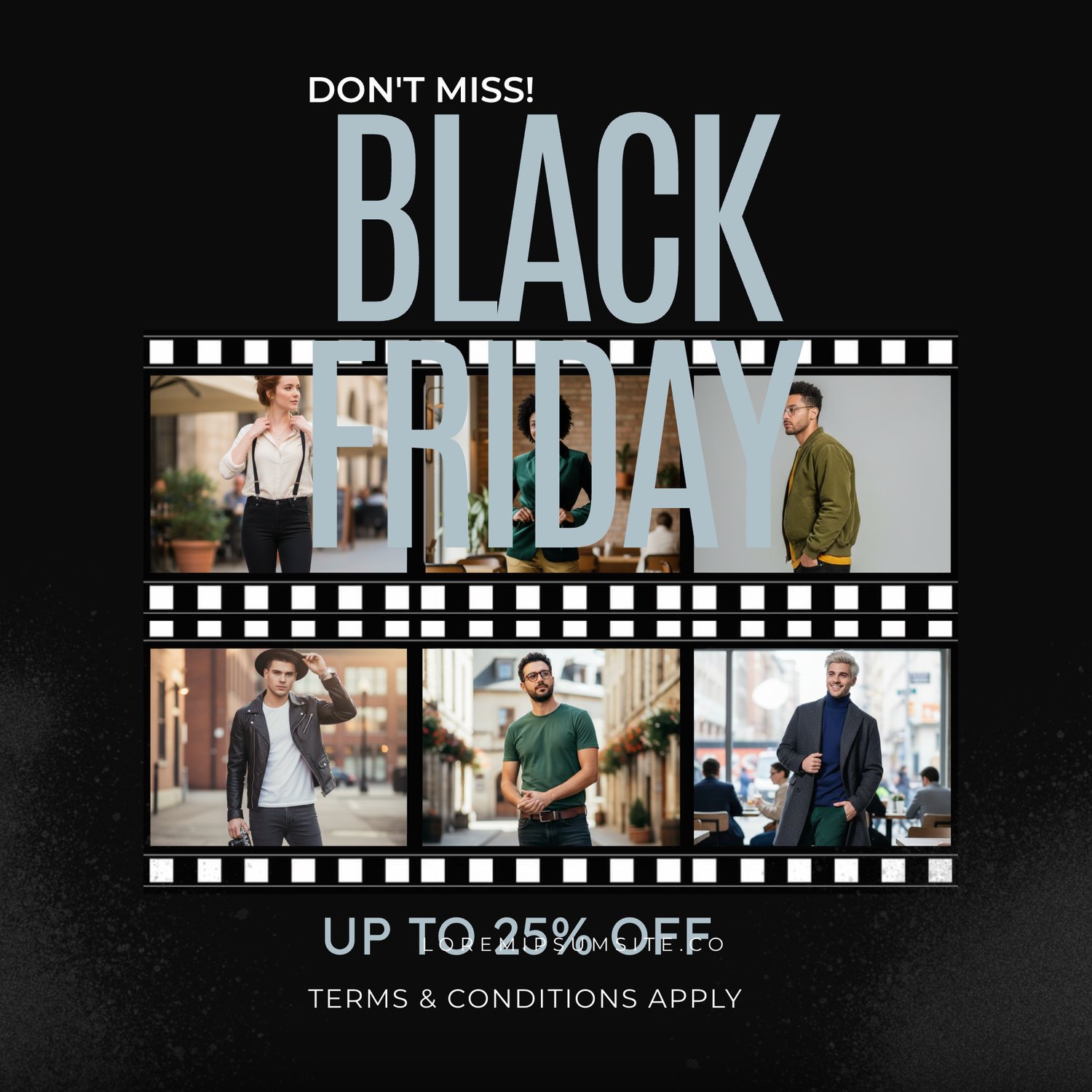}
\end{minipage}

\vspace{2pt}

\begin{minipage}[t]{0.3\columnwidth}
\raggedright\small
\centering
\textbf{(a)} Original layout
\end{minipage}
\hspace{0.05\columnwidth}
\begin{minipage}[t]{0.3\columnwidth}
\raggedright\small
\centering
\textbf{(b)} Layout re-rendered using predicted font size and text position.
\end{minipage}

\caption{Text parameter prediction rendering effect. The model (Gemini-3.1-Pro) predicts an incorrect position and size for the text, causing mis-alignments and overlaps. (only font size and text box position were modified in this example).}
\label{fig:text-param-failure}
\end{figure}  
\FloatBarrier

\subsection{Styled Text Generation}
\label{sec:styled-text-gen}

Styled text generation evaluates a model’s ability to translate structured typographic constraints into visually accurate text rendering. In addition to producing the correct string, the model must faithfully reproduce the full typographic specification, including font family, size, color, alignment, and spacing. We evaluate two settings, with both restricted to 1:1 aspect-ratio samples built from Google open-source fonts to ensure evaluator reliability:
\begin{itemize}
    \item \textbf{Styled text element generation:} the model generates an isolated text patch from a target string and typography specification, without layout context.
    \item \textbf{Styled text rendering to layout:} the model restores missing text into a masked layout image while preserving all surrounding unmasked pixels, with the masked layout provided as additional input.
\end{itemize}
Each model receives the target text content and a structured typography specification (font family, size, color, alignment, spacing); in the layout setting it additionally receives the masked layout image. Table~\ref{tab:styled-text-gen} presents the results. The prompt templates used in this task are provided in Appendix~\ref{app:prompts}. In the layout setting, the editable text region is provided as a tight polygonal mask computed from the convex hull algorithm.

\paragraph{Results.}
Table~\ref{tab:styled-text-gen} reports quantitative results for both element-level generation and layout-level rendering, while Figure~\ref{fig:styled_text_generation} illustrates representative failures in the practical layout-editing setting. The central challenge is not simply rendering plausible styled text, but restoring it within the prescribed masked region without altering surrounding content. Across models, this requirement remains unmet: layout-level spatial fidelity is limited, and qualitative examples show that generated text often extends beyond the editable region or modifies nearby design elements. The main observations are as follows:
\begin{itemize}
    \item \textbf{Mask-constrained text insertion remains unreliable.} In the layout setting, even the best model achieves only IoU\,=\,0.580 and F1\,=\,0.685, indicating limited agreement with the ground-truth text region. Figure~\ref{fig:styled_text_generation} further illustrates that edits frequently overflow outside the input masks and modify the nearby contents. This suggests that current models do not reliably confine text generation to the intended region.
    \item \textbf{Fine-grained typography recovery remains uneven even in the easier element setting.} Gemini-3.1-Flash-Img achieves the best font family accuracy, alignment accuracy, and color MAE in the element setting, indicating stronger adherence to the requested typographic specification than GPT-Image-1.5. Nevertheless, font family accuracy remains low overall, indicating that faithful recovery of detailed typographic attributes remains unsolved.
\end{itemize}

\begin{table*}[h]
    \centering
    \caption{Styled text generation results.
Each model is evaluated on element-level generation (\emph{element}) and layout-level rendering (\emph{layout}).
Font and Align denote accuracy; Size and Color denote MAE.
IoU/F1, LPIPS, and SSIM are reported only for the layout task, where a full rendered target with spatial reference is available. Best result per metric and task in \textbf{bold}.}
    \label{tab:styled-text-gen}
    \small
    \setlength{\tabcolsep}{4.5pt}
    \begin{tabular}{@{}ll cc c ccc cc@{}}
        \toprule
        & & \multicolumn{2}{c}{\textbf{Spatial Accuracy}}
        & \multicolumn{1}{c}{\textbf{Font}}
        & \multicolumn{3}{c}{\textbf{Style Fidelity}}
        & \multicolumn{2}{c}{\textbf{Render}} \\
        \cmidrule(lr){3-4} \cmidrule(lr){5-5} \cmidrule(lr){6-8} \cmidrule(lr){9-10}
        \textbf{Model} & \textbf{Task}
          & IoU$\uparrow$ & F1$\uparrow$
          & Font$\uparrow$
          & Align$\uparrow$ & Size$\downarrow$ & Color$\downarrow$
          & LPIPS$\downarrow$ & SSIM$\uparrow$ \\
        \midrule
        \multirow{2}{*}{GPT-Image-1.5}
          & layout  & 0.523 & 0.658 & 0.190 & \textbf{0.690} & 36.50 & 16.75 & 0.227 & 0.704 \\
          & element & - & - & 0.100 & 0.660 & \textbf{104.2} & 16.04 & - & - \\
        \midrule
        \multirow{2}{*}{Gemini-3.1-Flash-Img}
          & layout  & \textbf{0.580} & \textbf{0.685} & \textbf{0.468} & 0.670 & \textbf{27.59} & \textbf{16.00} & \textbf{0.056} & \textbf{0.923} \\
          & element & - & - & \textbf{0.353} & \textbf{0.696} & 188.7 & \textbf{15.66} & - & - \\
        \bottomrule
    \end{tabular}
\end{table*}

{
\setlength{\tabcolsep}{3pt}
\renewcommand{\arraystretch}{1.0}

\begin{figure*}[!t]
\centering

\newcommand{\tpromptbox}[1]{%
\begin{tcolorbox}[
    width=\linewidth,
    colback=gray!5,
    colframe=gray!45,
    boxrule=0.3pt,
    arc=1.5pt,
    left=3pt,right=3pt,top=3pt,bottom=3pt
]
{\fontsize{5.0}{6.0}\selectfont #1}
\end{tcolorbox}
}

\newcommand{\tpitem}[2]{%
\textbf{#1}~#2\par\vspace{0.8pt}
}

\newcommand{\tpspec}[1]{%
{\fontsize{4.5}{6.0}\selectfont
\textbf{Typography/style specification :}\par
#1\par}
}

\newcommand{\timg}[1]{%
\includegraphics[width=\linewidth]{#1}%
}

\begin{tabular}{
    >{\raggedright\arraybackslash}p{0.24\textwidth}
    >{\centering\arraybackslash}p{0.17\textwidth}
    >{\centering\arraybackslash}p{0.17\textwidth}
    >{\centering\arraybackslash}p{0.17\textwidth}
    >{\centering\arraybackslash}p{0.17\textwidth}
}

\makecell[c]{\textbf{Text Prompt}} &
\makecell[c]{\textbf{Masked Input}} &
\makecell[c]{\textbf{GT}} &
\makecell[c]{\textbf{Gemini-3.1}\\\textbf{Flash Image}} &
\makecell[c]{\textbf{GPT-Image 1.5}} \\[4pt]

\tpromptbox{
\tpitem{Instruction:}{Restore missing text}
\tpitem{Target text:}{``Setting SMART goals for personal and professional growth Creating action plans to achieve objectives Overcoming obstacles and setbacks''}
\tpspec{
font=Montserrat; size=26.5px; \\
weight=400; align=left; color=\\rgb(1,1,1);
lineHeight=37.4px; \\letterSpacing=0.04em; box=(left 325,\\ top 302, width 393, height 331)
}
}
&
\timg{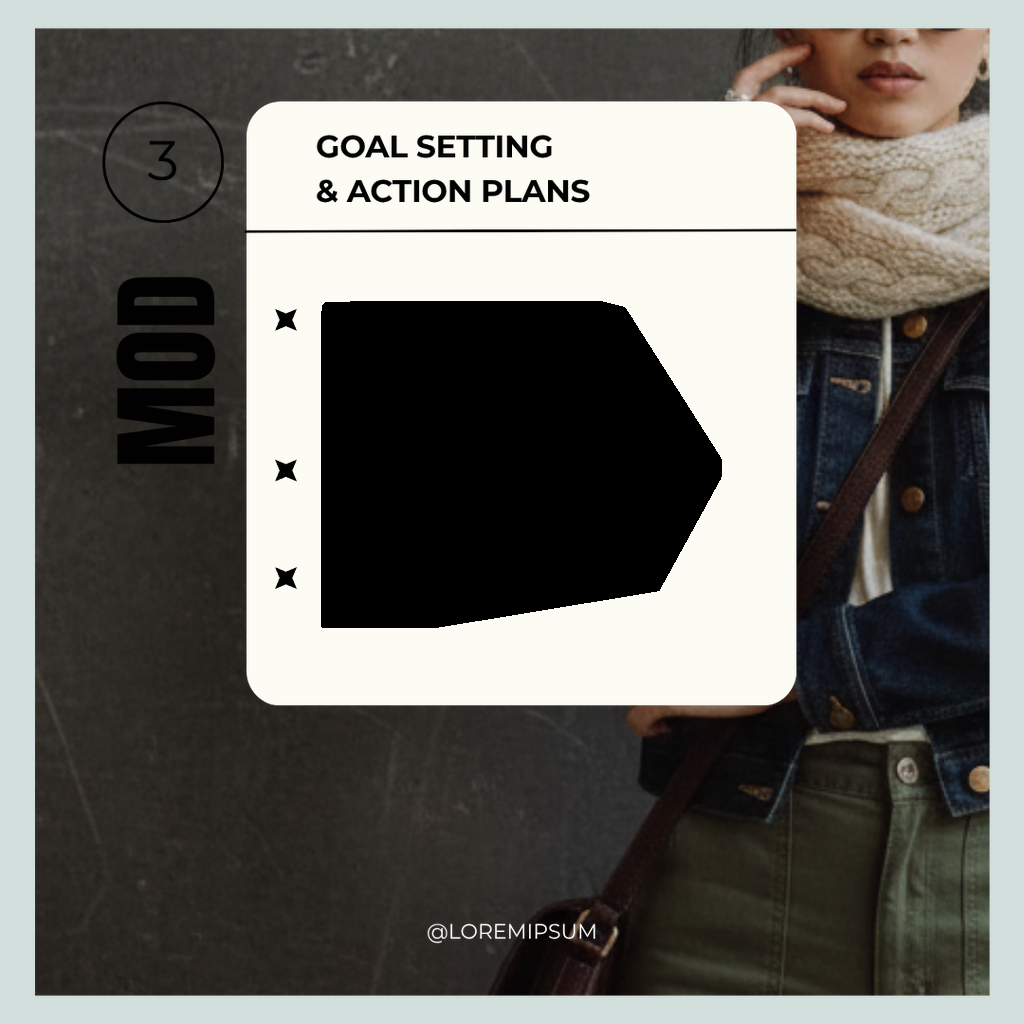}
&
\timg{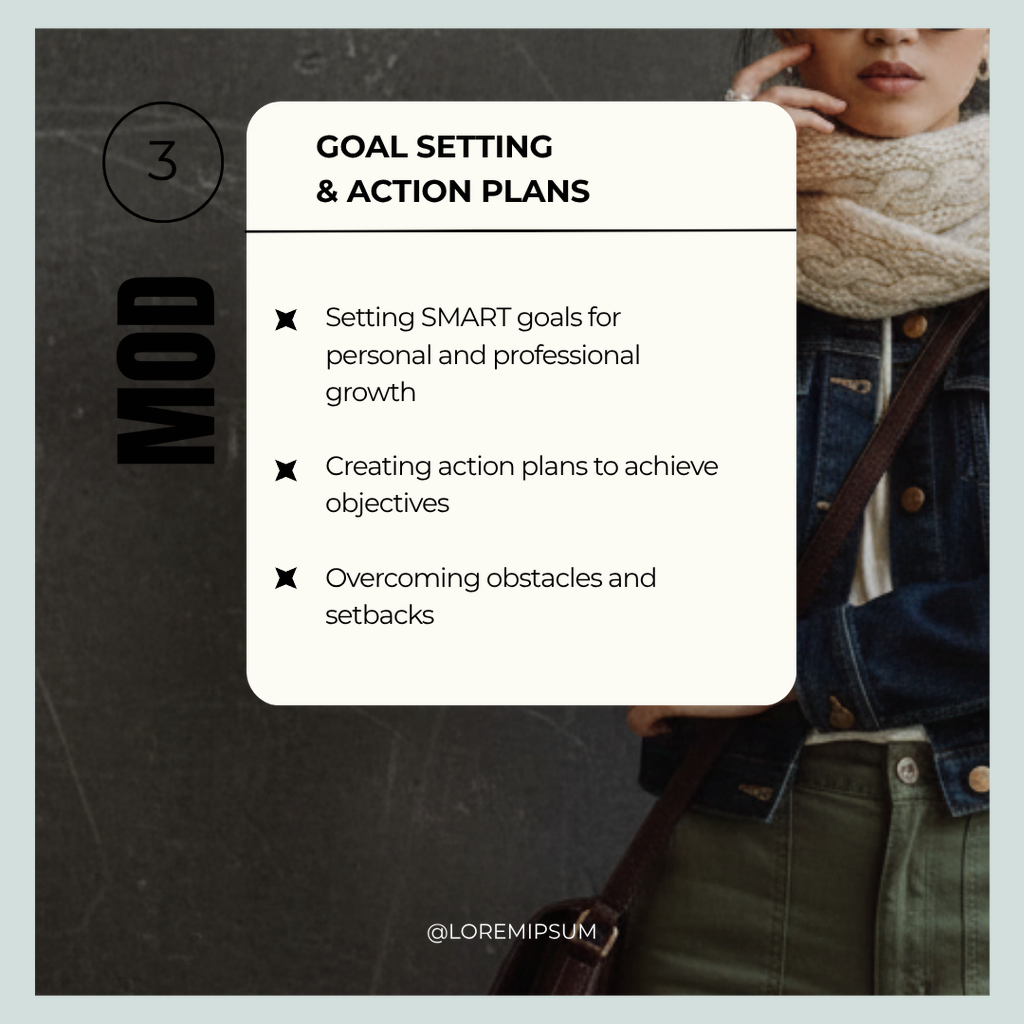}
&
\timg{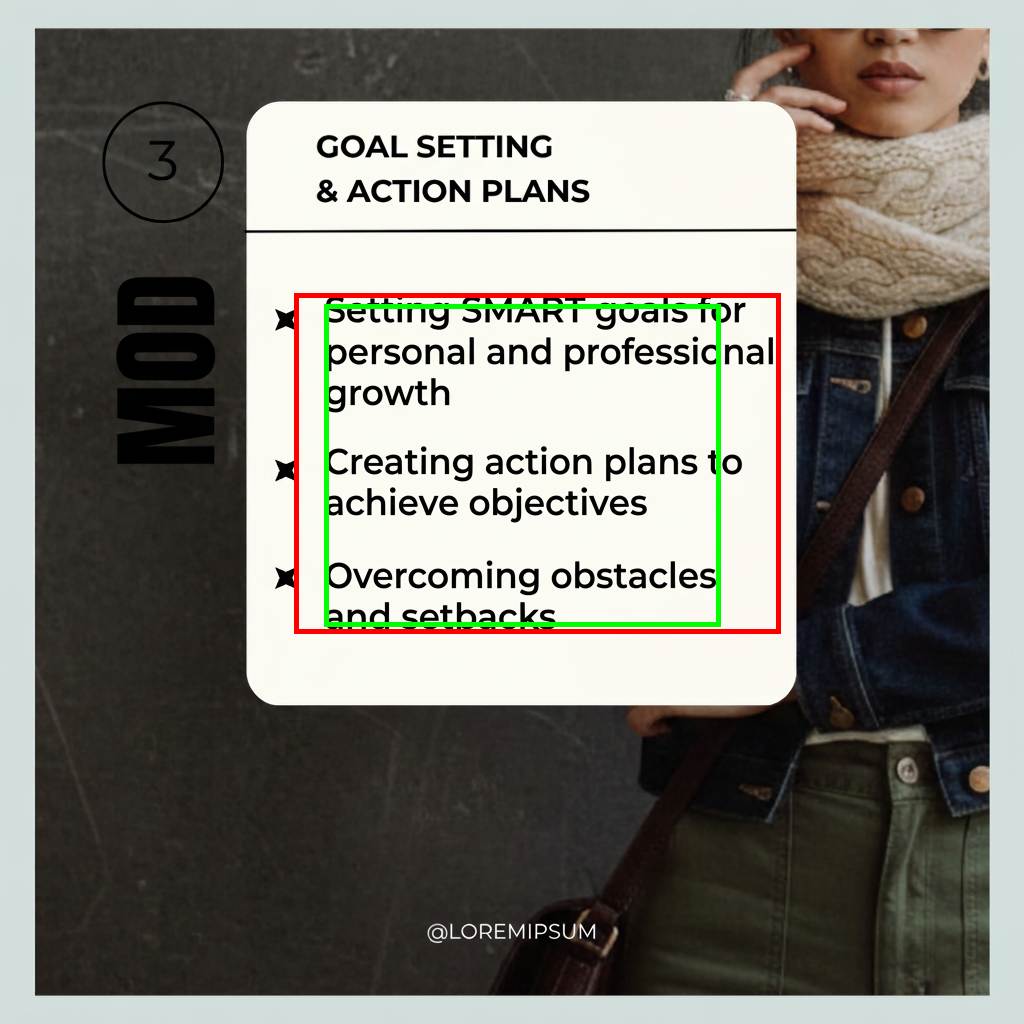}
&
\timg{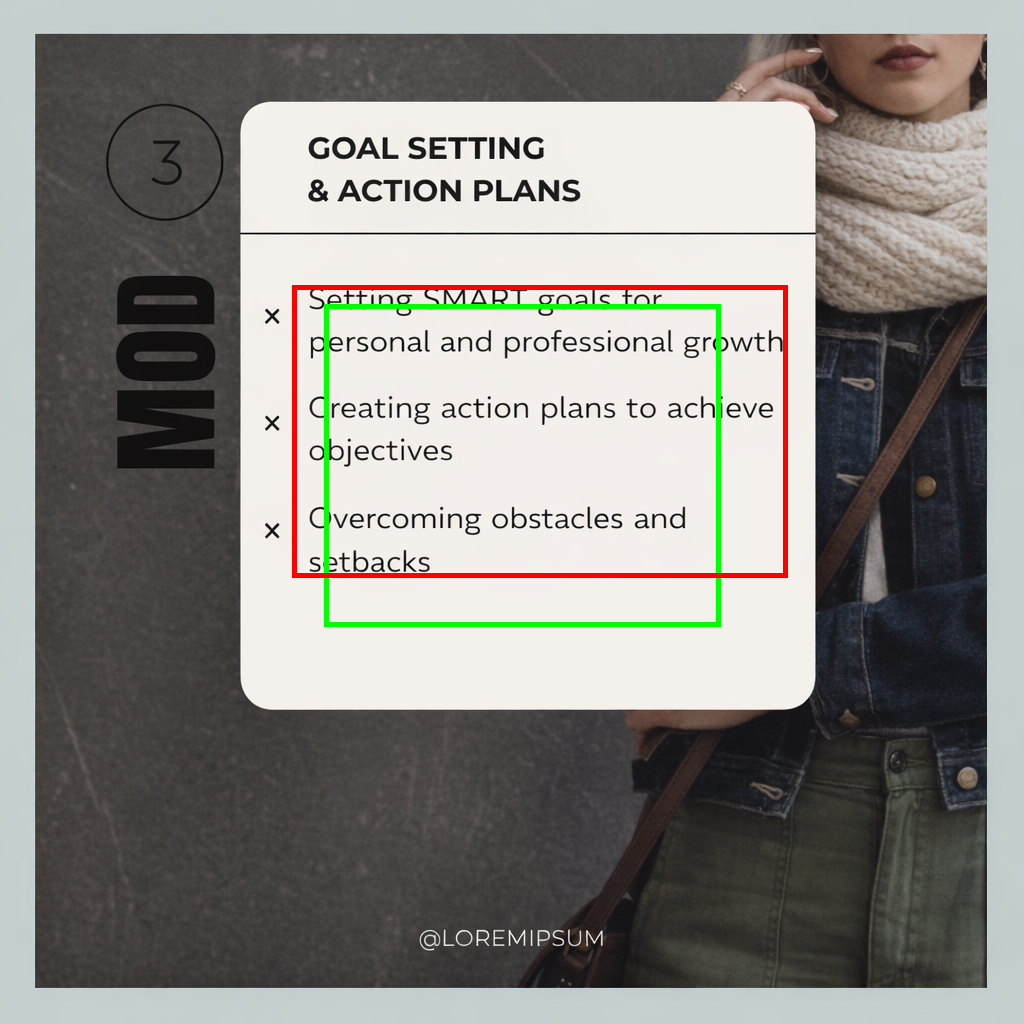}
\\[6pt]

\tpromptbox{
\tpitem{Instruction:}{Restore missing text}
\tpitem{Target text:}{``JOBS''}
\tpspec{
font=Anton; size=169.7px; \\
weight=400;align=left;\\
color=rgb(1,1,1); \\
lineHeight=238.1px; \\letterSpacing=0em;\\
box=(left 462, top 272, width 476, height 204)
}
}
&
\timg{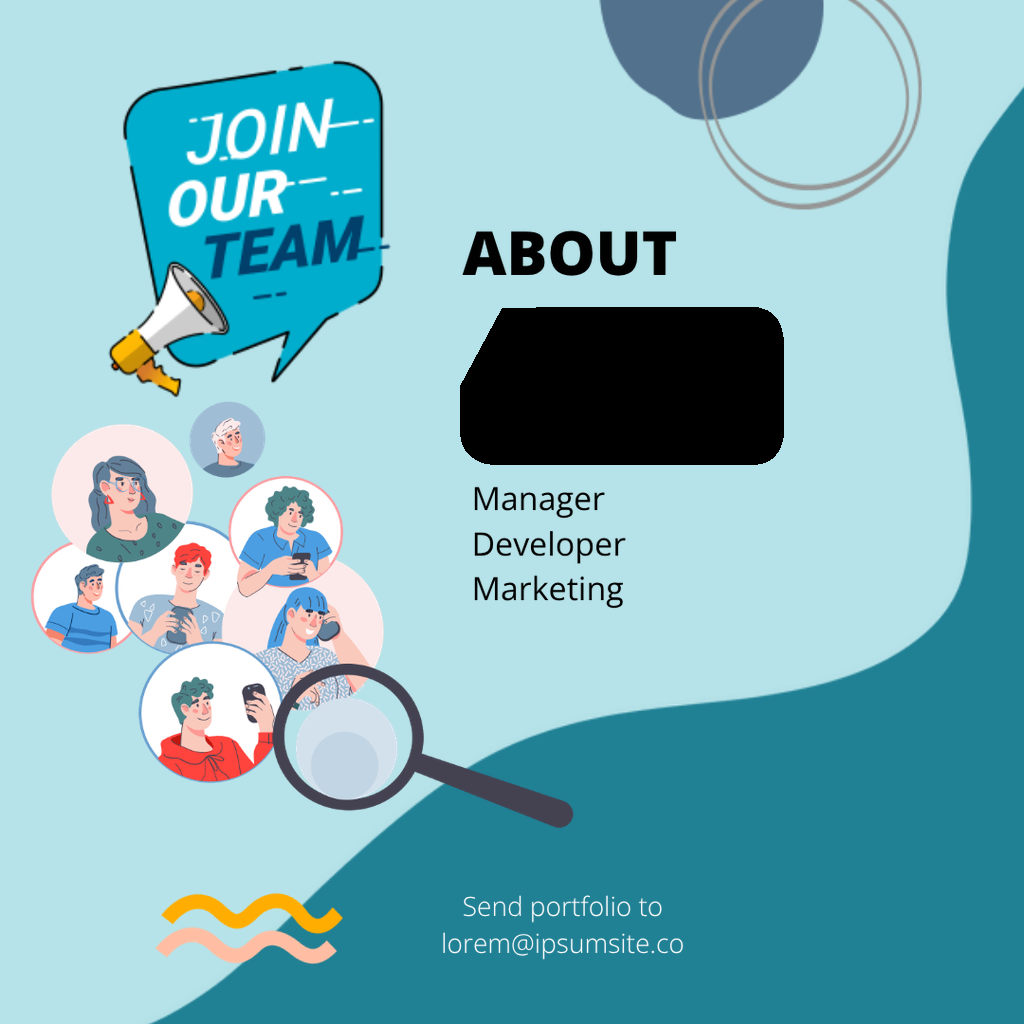}
&
\timg{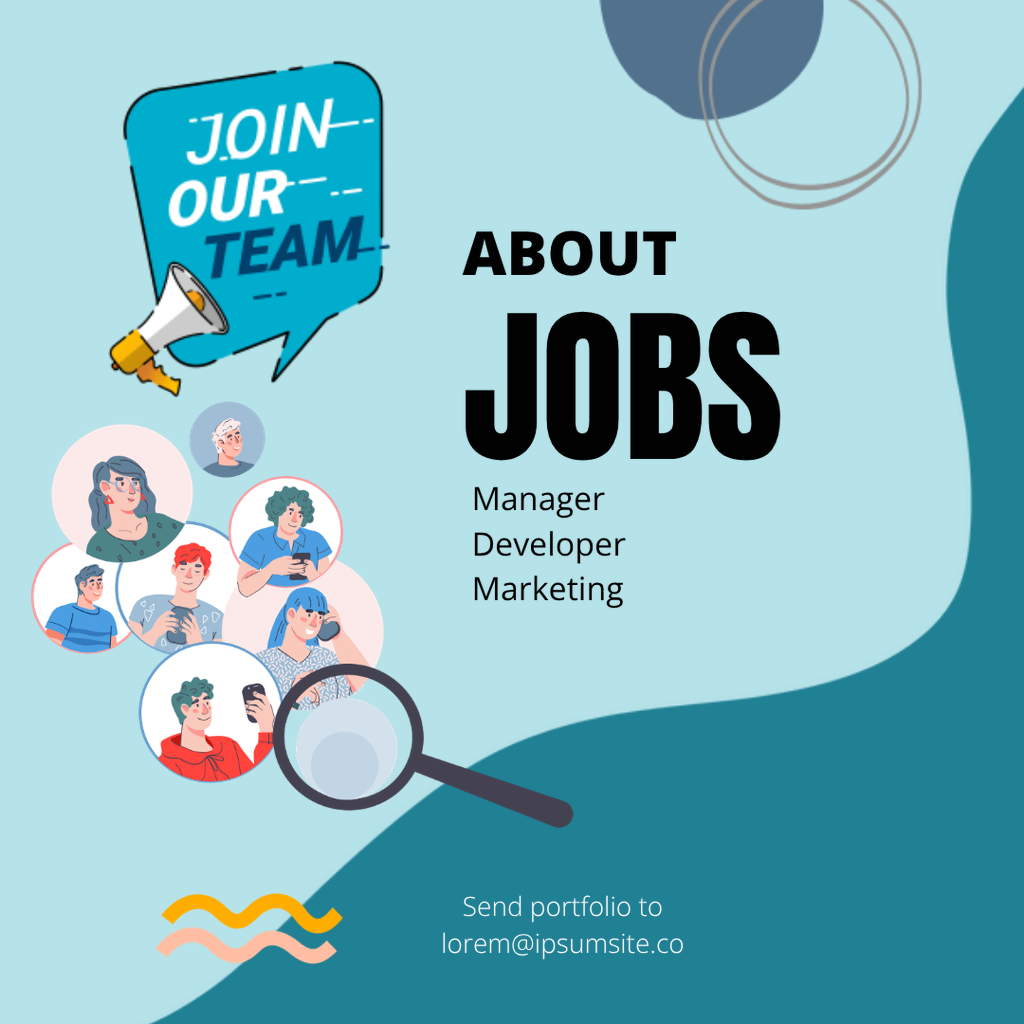}
&
\timg{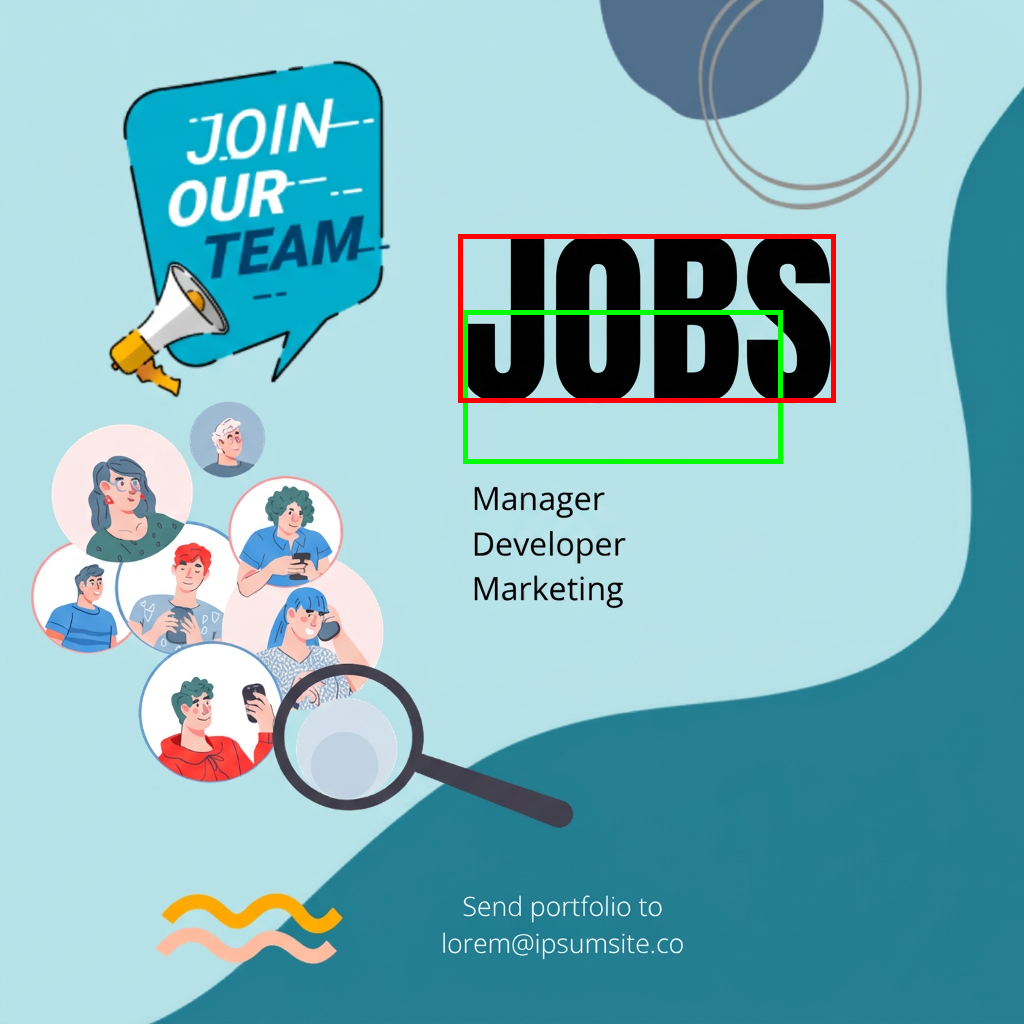}
&
\timg{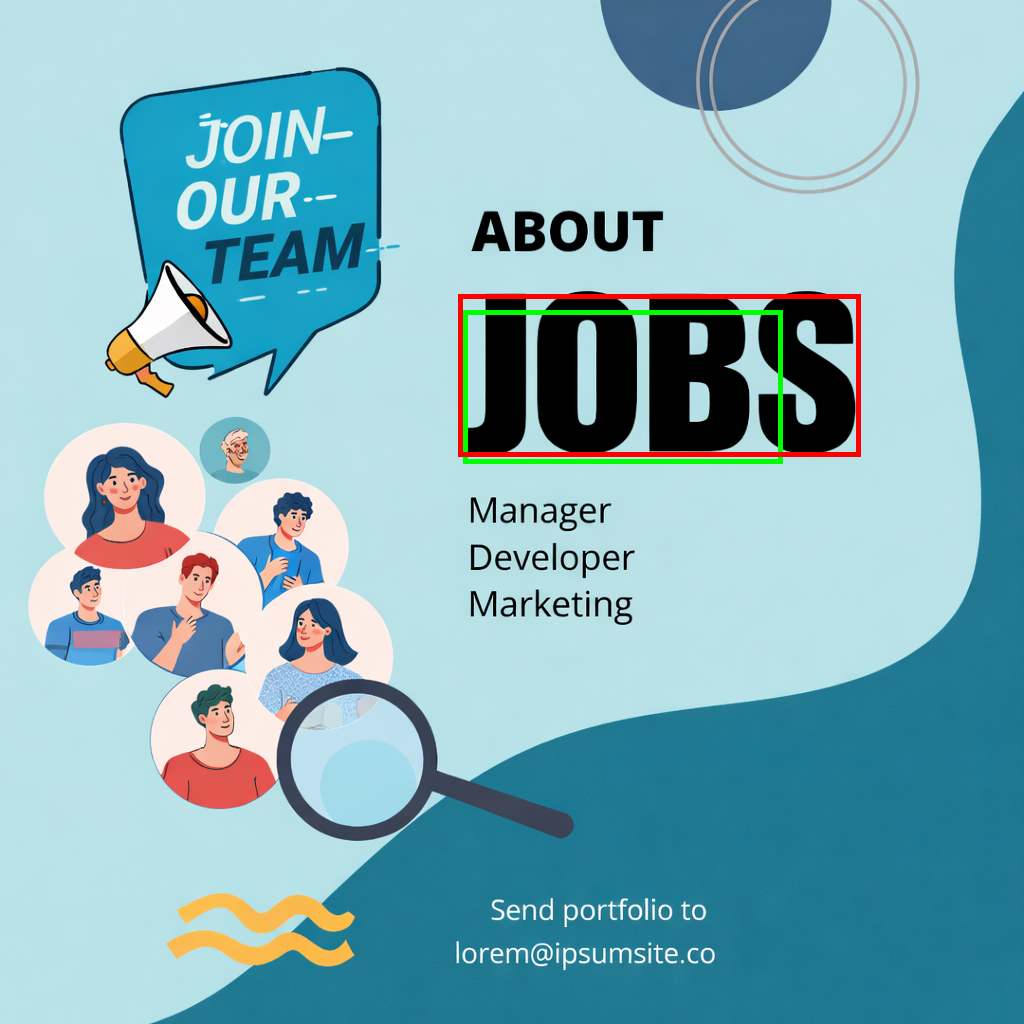}
\\

\end{tabular}

\caption{\textbf{Layout-level styled text generation failures.}
Each example shows the target text and typography specification, the masked input layout, the ground-truth render, and outputs from Gemini-3.1 Flash Image and GPT-Image 1.5.
Both models generate plausible text, but often fail to respect the intended text region: the restored content can extend beyond the masked area or deviate substantially from the prescribed extent.
In the first example, both models overshoot the target region when reconstructing a multi-line paragraph; in the second, both render a readable headline, yet neither matches the intended placement and size within the editable area.
These examples show that the main bottleneck is not merely rendering stylized text, but confining the edit to the prescribed mask while preserving surrounding content.
GT bbox is shown in \textcolor[HTML]{00FF00}{green}, and the predicted text bbox is shown in \textcolor[HTML]{FF0000}{red}.}
\label{fig:styled_text_generation}
\end{figure*}
}


\subsection{Text Removal \& Background Regeneration}
\label{sec:text-removal}

Text removal evaluates whether a model can support practical design editing by erasing typography without leaving ghosting artefacts and faithfully reconstructing the hidden background structure so that the canvas remains reusable for subsequent editing. This capability is essential for template adaptation, multilingual translation, and iterative copy revision~\cite{jia2024cole,cheng2025graphist,lin2024ladeco,zhang2025creatiposter,qu2025igd,hirsch2026lica,chen2025accordion}. Each model receives a source design image together with a binary mask indicating the text regions (white\,=\,editable, black\,=\,preserved) and a unified instruction to remove all text and reconstruct the background naturally. We evaluate on 100 randomly sampled templates from the LICA dataset (Section~\ref{para:dataset}), filtered to exclude low-resolution images (max side $<$\,1{,}024\,px) and extreme aspect ratios ($\geq$\,2.0). Since each model supports a different set of aspect ratios, inputs are resized to model-compatible resolutions before generation and resized back to the original dimensions for evaluation. Table~\ref{tab:text-removal} presents the results.

\paragraph{Results.}
The results reveal two key patterns across the text removal task:
\begin{itemize}
    \item \textbf{Text erasure is effectively solved, but background reconstruction is not.} Both models achieve high text removal rates ($>$\,94\%), confirming that the text itself is effectively erased. However, they diverge sharply on reconstruction fidelity, with PSNR ranging from 16.64 to 30.22 (${\sim}$13.6\,dB difference) and SSIM from 0.743 to 0.944, indicating that faithfully reconstructing the occluded background remains a significant challenge.
    \item \textbf{Higher removal rate does not imply better reconstruction.} The model with a marginally higher raw removal rate (95.3\% vs.\ 94.5\%) produces substantially worse background reconstruction (LPIPS\,=\,0.331 vs.\ 0.088; FID\,=\,115.75 vs.\ 54.05), suggesting it over-generates or hallucinates content in the masked regions rather than faithfully reconstructing what was occluded.
\end{itemize}

\begin{table}[h]
    \centering
    \caption{Text removal and background regeneration results.
    Best result per metric in \textbf{bold}.}
    \label{tab:text-removal}
    \small
    \setlength{\tabcolsep}{4pt}
    \begin{tabular}{@{}l c cccc cc@{}}
        \toprule
        & \multicolumn{1}{c}{\textbf{Text Removal}}
        & \multicolumn{4}{c}{\textbf{Perceptual Quality}}
        & \multicolumn{2}{c}{\textbf{Reconstruction}} \\
        \cmidrule(lr){2-2} \cmidrule(lr){3-6} \cmidrule(lr){7-8}
        \textbf{Model}
          & Remove
          & FID$\downarrow$ & LPIPS$\downarrow$ & CLIP$\uparrow$ & DINO$\uparrow$
          & PSNR$\uparrow$ & SSIM$\uparrow$ \\
        \midrule
        Gemini-3.1-Flash-Img & 0.945 & \textbf{54.05} & \textbf{0.088} & \textbf{0.238} & \textbf{0.906} & \textbf{30.22} & \textbf{0.944} \\
        GPT-Image-1.5 & \textbf{0.953} & 115.75 & 0.331 & 0.236 & 0.820 & 16.64 & 0.743 \\
        \bottomrule
    \end{tabular}
\end{table}

\begin{figure*}[t]
    \centering

    \newlength{\labW}
    \newlength{\colW}
    \setlength{\labW}{0.035\textwidth}
    \setlength{\colW}{0.228\textwidth}

    \newlength{\imgHA} 
    \newlength{\imgHB} 
    \newlength{\imgHC} 
    \setlength{\imgHA}{0.152\textwidth}
    \setlength{\imgHB}{0.228\textwidth}
    \setlength{\imgHC}{0.342\textwidth}

    \newcommand{\imgfit}[2]{%
        \includegraphics[width=\colW,height=#2,keepaspectratio]{#1}%
    }

    \newcommand{\fourimgsA}[5]{%
        \noindent
        \begin{minipage}[c][\imgHA][c]{\labW}\centering \textbf{#1} \end{minipage}\hfill
        \begin{minipage}[c][\imgHA][c]{\colW}\centering \imgfit{#2}{\imgHA} \end{minipage}\hfill
        \begin{minipage}[c][\imgHA][c]{\colW}\centering \imgfit{#3}{\imgHA} \end{minipage}\hfill
        \begin{minipage}[c][\imgHA][c]{\colW}\centering \imgfit{#4}{\imgHA} \end{minipage}\hfill
        \begin{minipage}[c][\imgHA][c]{\colW}\centering \imgfit{#5}{\imgHA} \end{minipage}
    }

    \newcommand{\fourimgsB}[5]{%
        \noindent
        \begin{minipage}[c][\imgHB][c]{\labW}\centering \textbf{#1} \end{minipage}\hfill
        \begin{minipage}[c][\imgHB][c]{\colW}\centering \imgfit{#2}{\imgHB} \end{minipage}\hfill
        \begin{minipage}[c][\imgHB][c]{\colW}\centering \imgfit{#3}{\imgHB} \end{minipage}\hfill
        \begin{minipage}[c][\imgHB][c]{\colW}\centering \imgfit{#4}{\imgHB} \end{minipage}\hfill
        \begin{minipage}[c][\imgHB][c]{\colW}\centering \imgfit{#5}{\imgHB} \end{minipage}
    }

    \newcommand{\fourimgsC}[5]{%
        \noindent
        \begin{minipage}[c][\imgHC][c]{\labW}\centering \textbf{#1} \end{minipage}\hfill
        \begin{minipage}[c][\imgHC][c]{\colW}\centering \imgfit{#2}{\imgHC} \end{minipage}\hfill
        \begin{minipage}[c][\imgHC][c]{\colW}\centering \imgfit{#3}{\imgHC} \end{minipage}\hfill
        \begin{minipage}[c][\imgHC][c]{\colW}\centering \imgfit{#4}{\imgHC} \end{minipage}\hfill
        \begin{minipage}[c][\imgHC][c]{\colW}\centering \imgfit{#5}{\imgHC} \end{minipage}
    }

    \begin{minipage}[c]{\labW}\centering \end{minipage}\hfill
    \begin{minipage}[c]{\colW}\centering \textbf{Input Image} \end{minipage}\hfill
    \begin{minipage}[c]{\colW}\centering \textbf{Input Mask} \end{minipage}\hfill
    \begin{minipage}[c]{\colW}\centering \textbf{Gemini-3.1}\\\textbf{Flash Image} \end{minipage}\hfill
    \begin{minipage}[c]{\colW}\centering \textbf{GPT-Image 1.5} \end{minipage}

    \vspace{6pt}

    \fourimgsA{(a)}
    {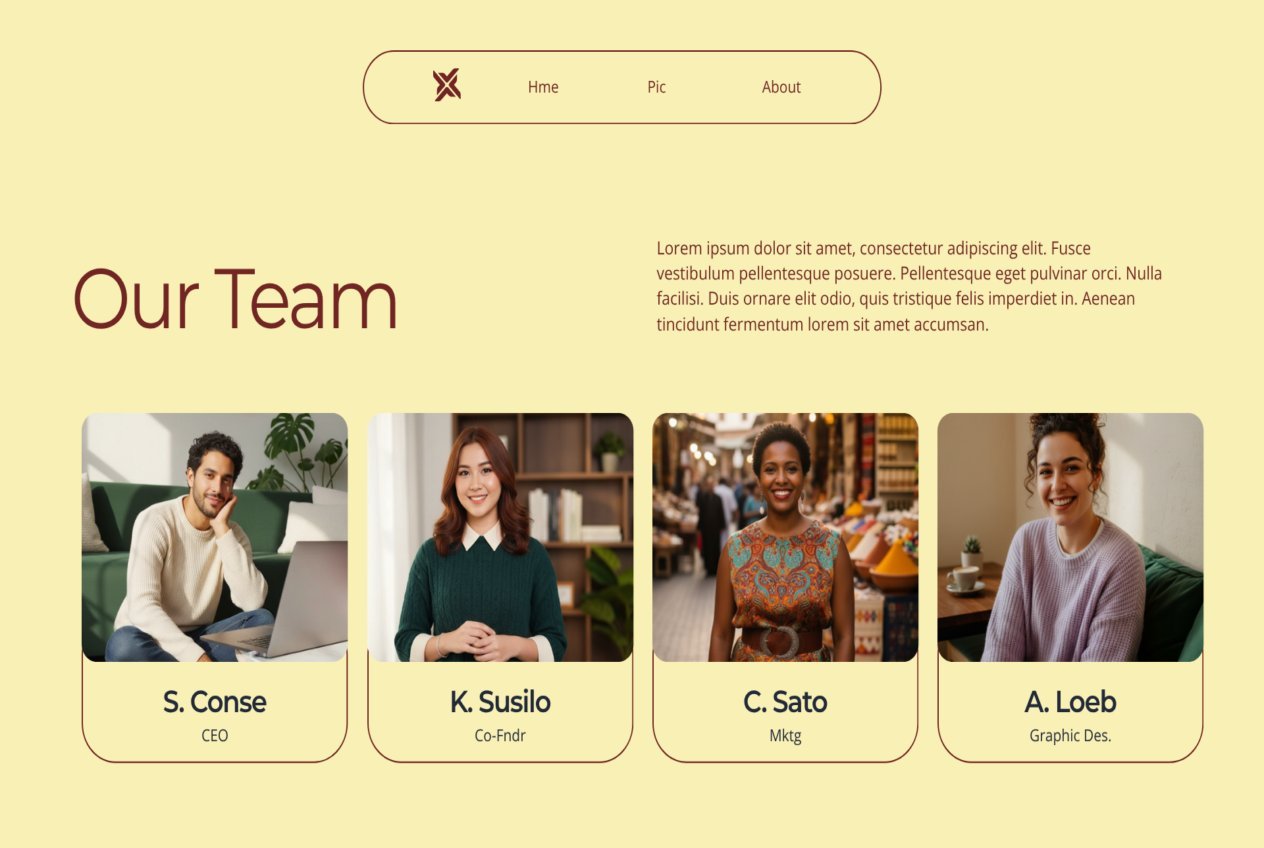}
    {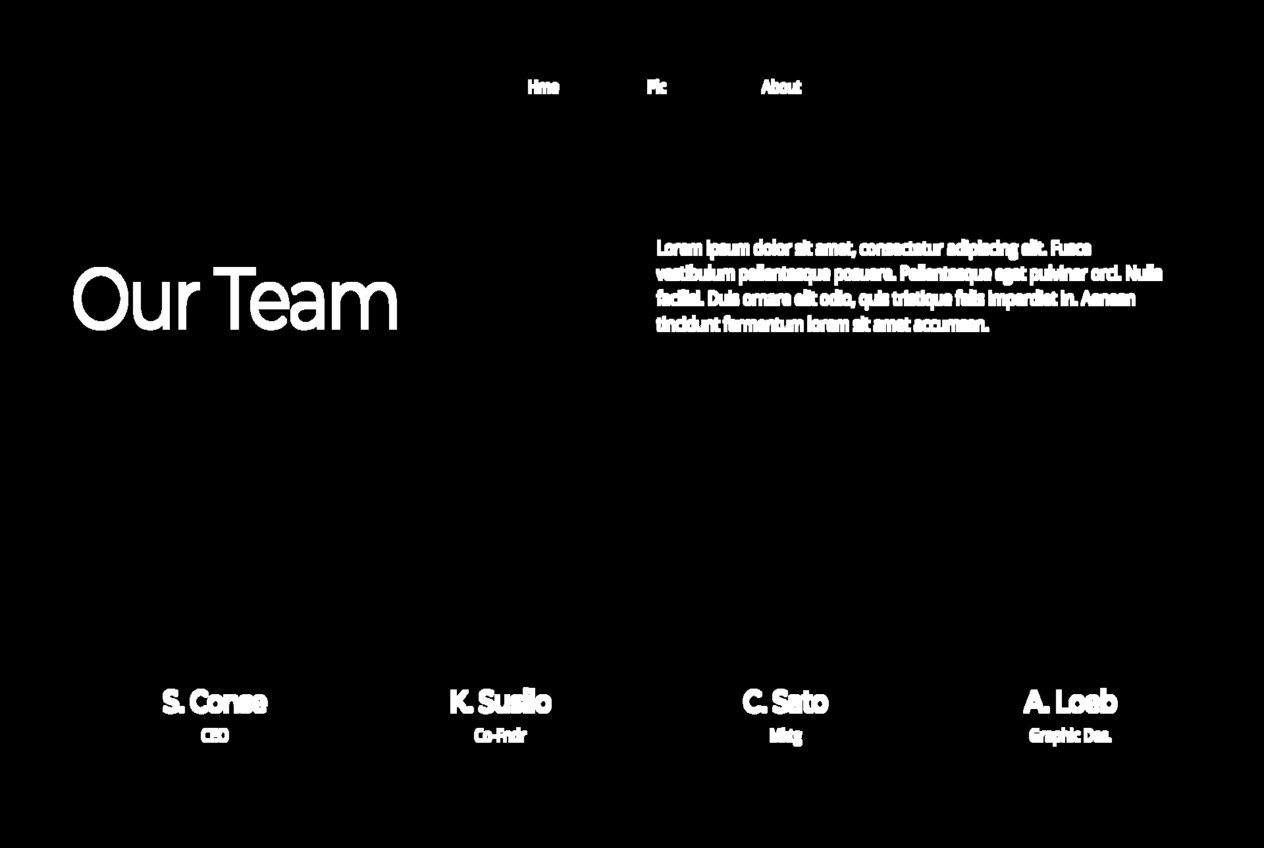}
    {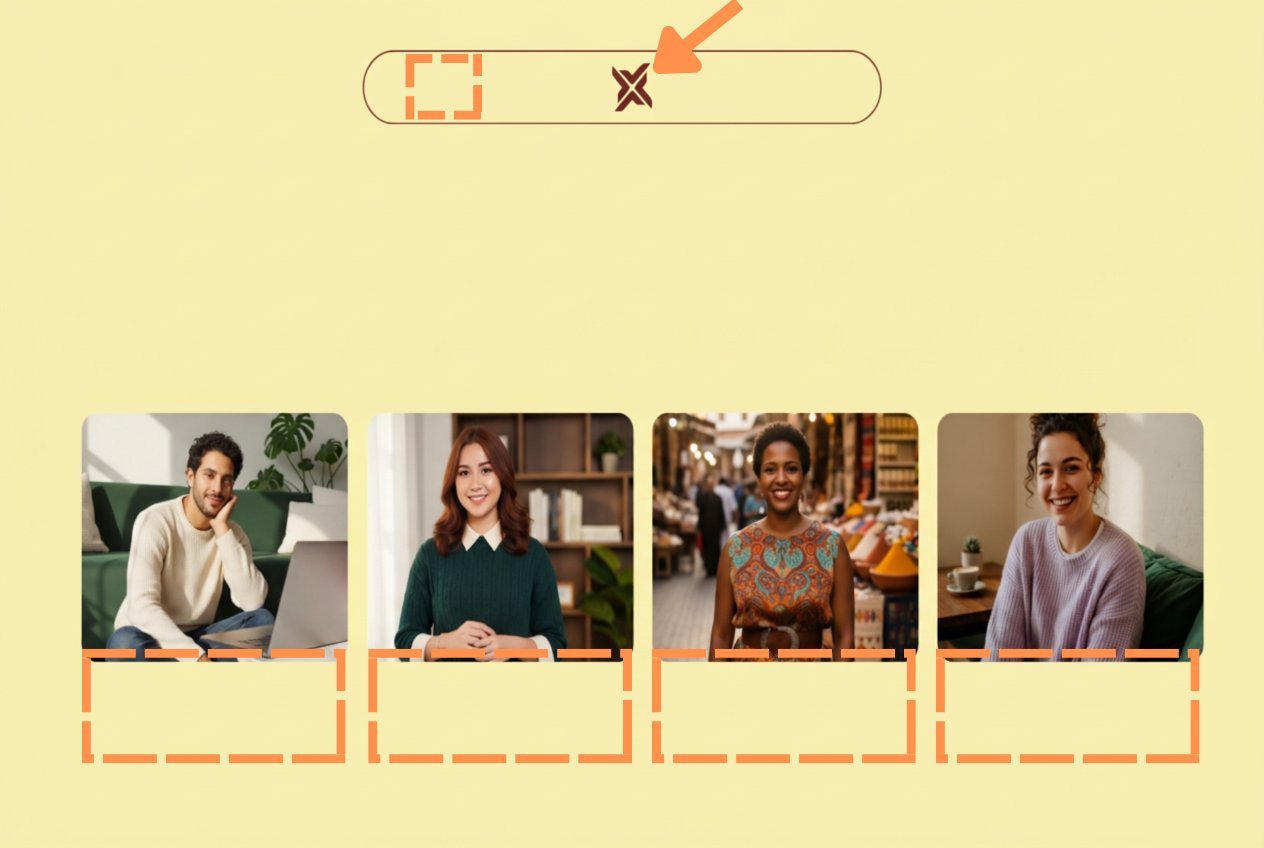}
    {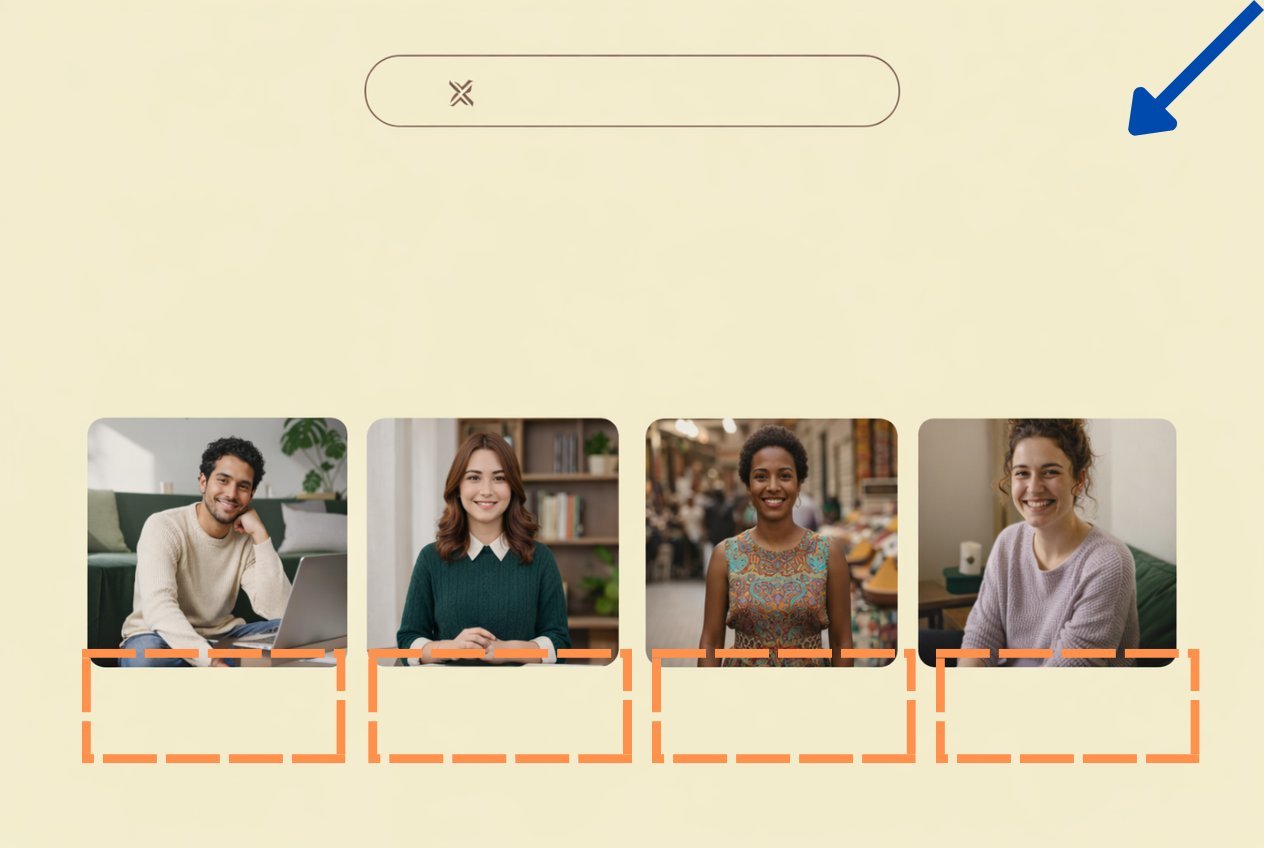}

    \vspace{4pt}

    \fourimgsB{(b)}
    {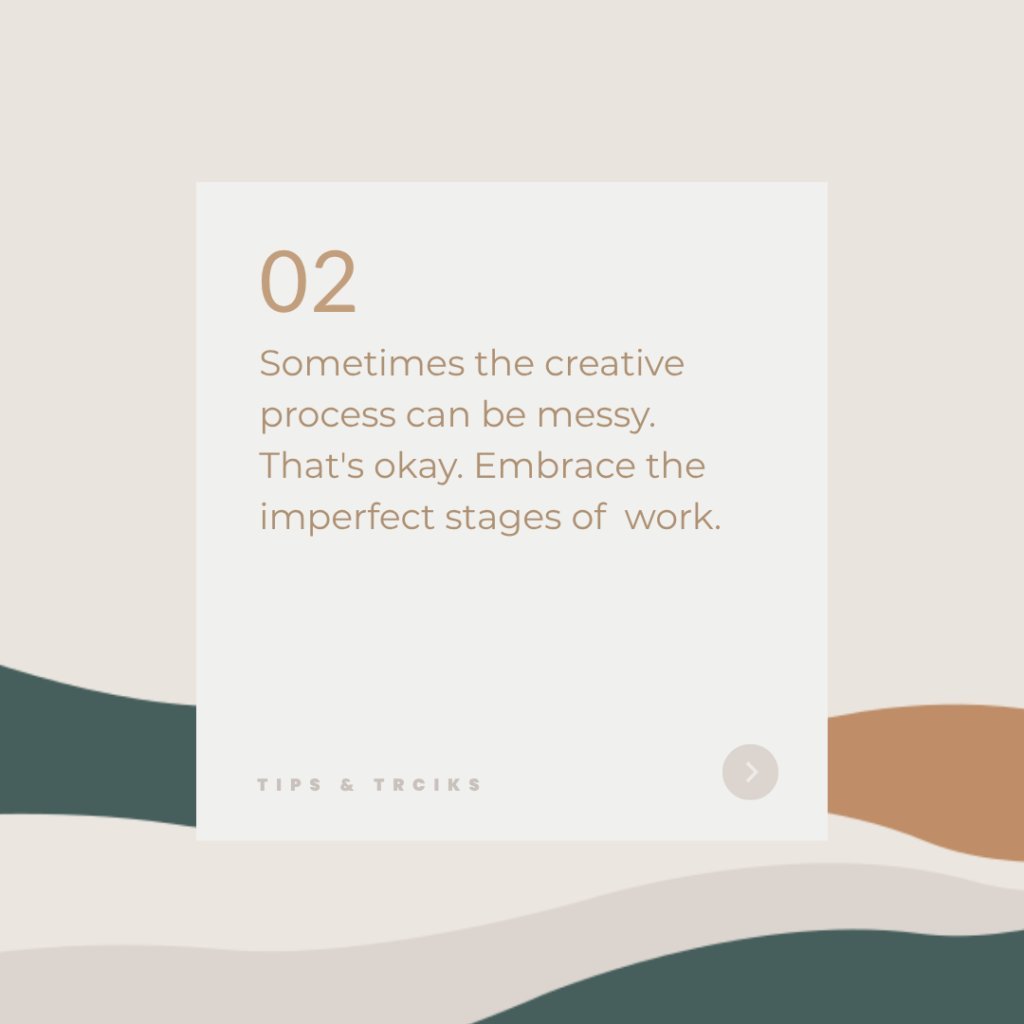}
    {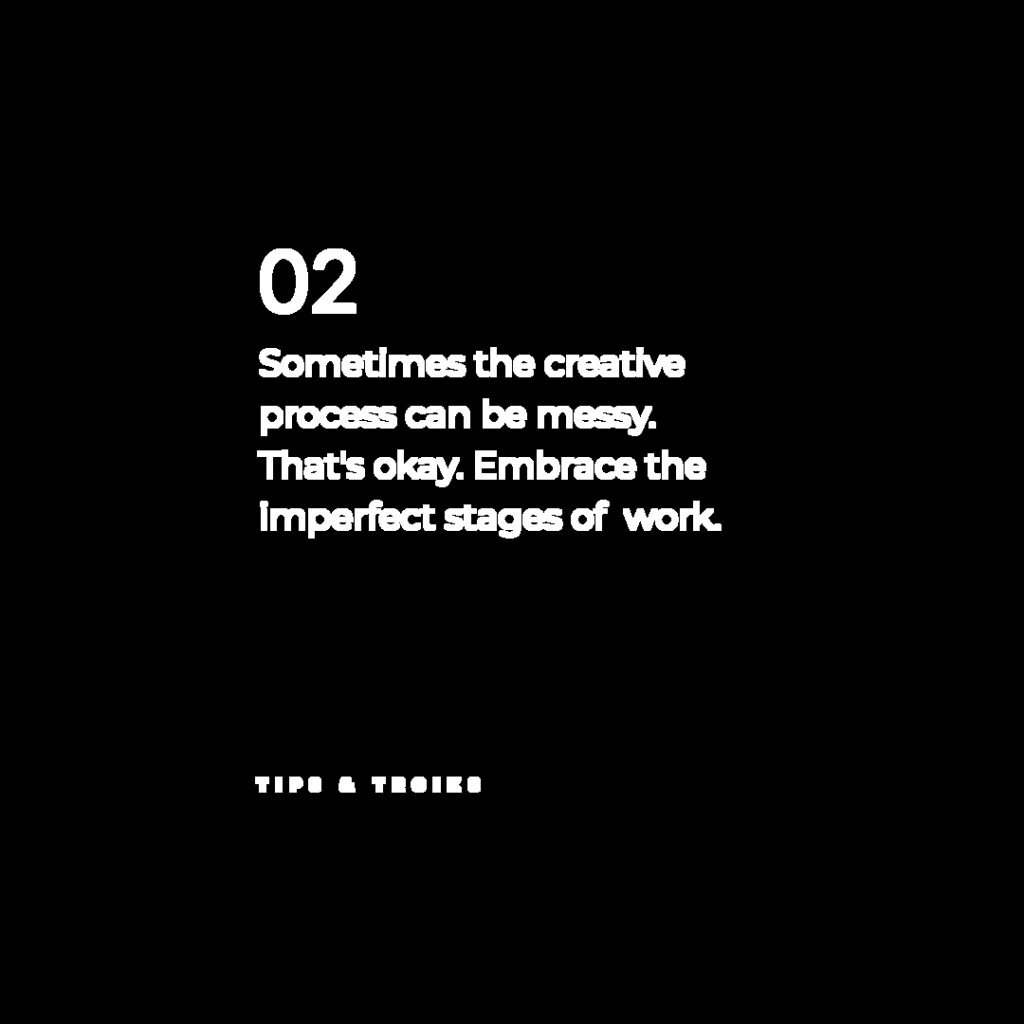}
    {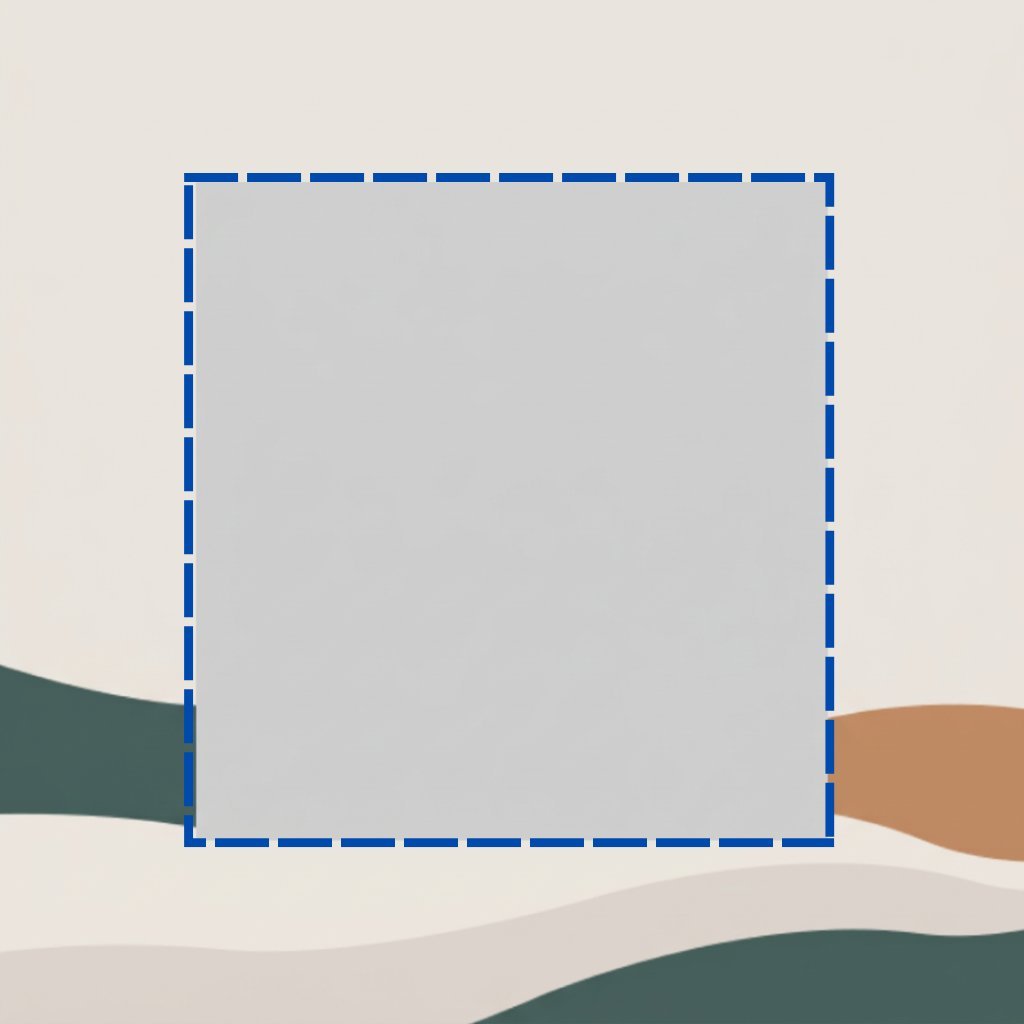}
    {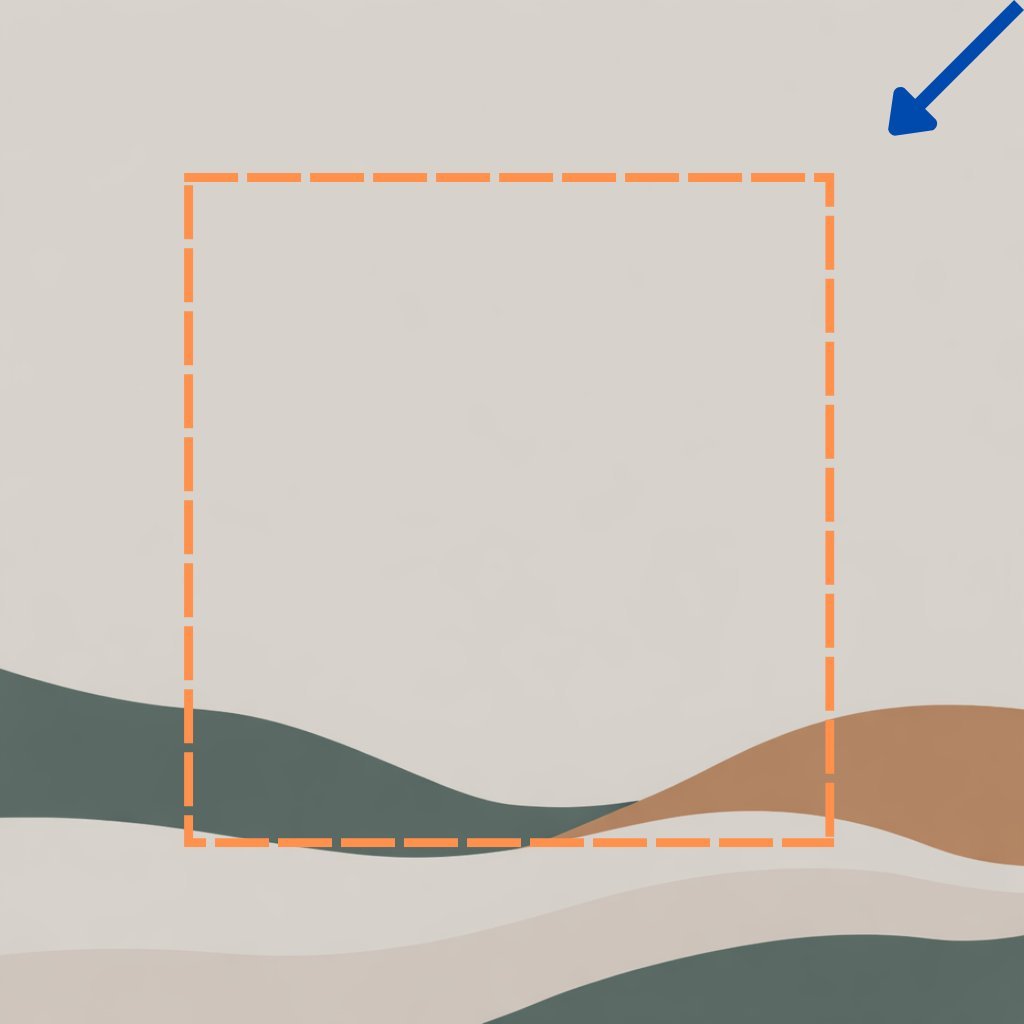}

    \vspace{-24pt}

    \fourimgsC{(c)}
    {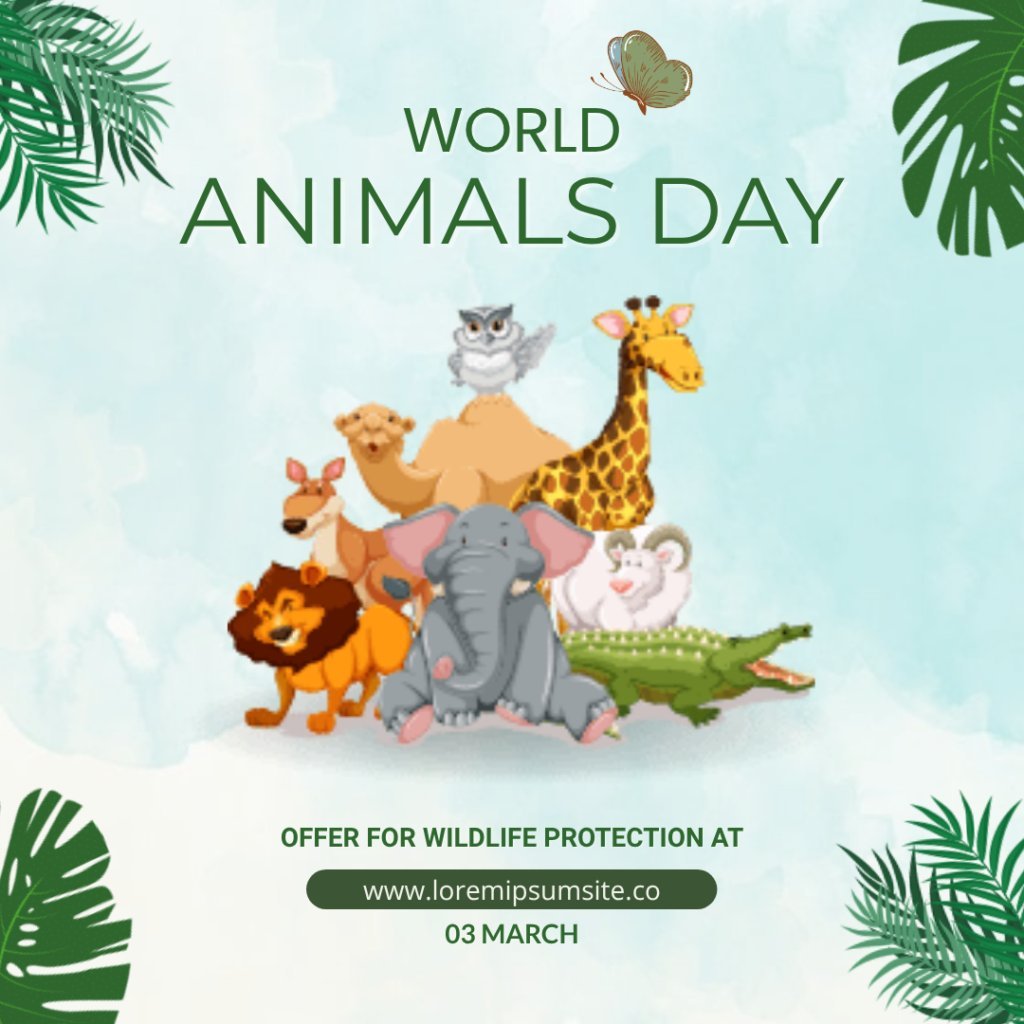}
    {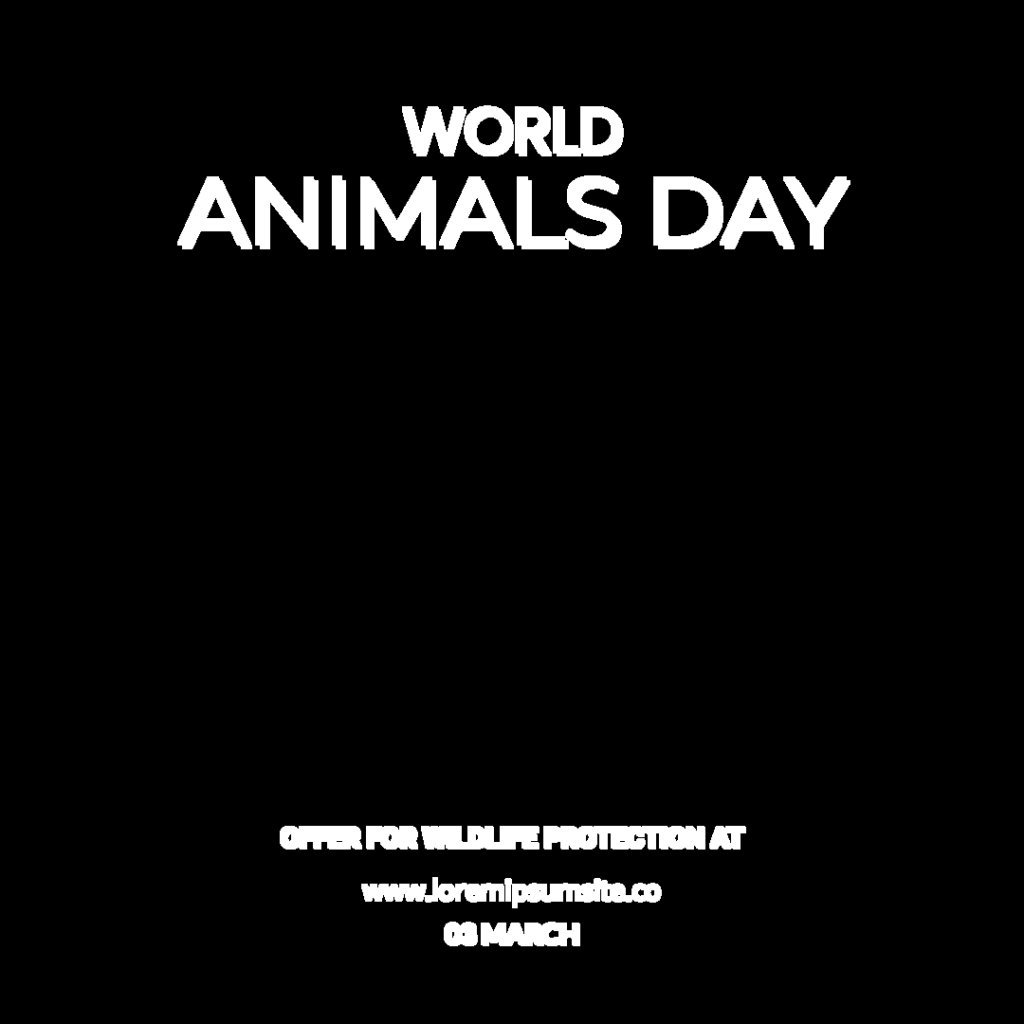}
    {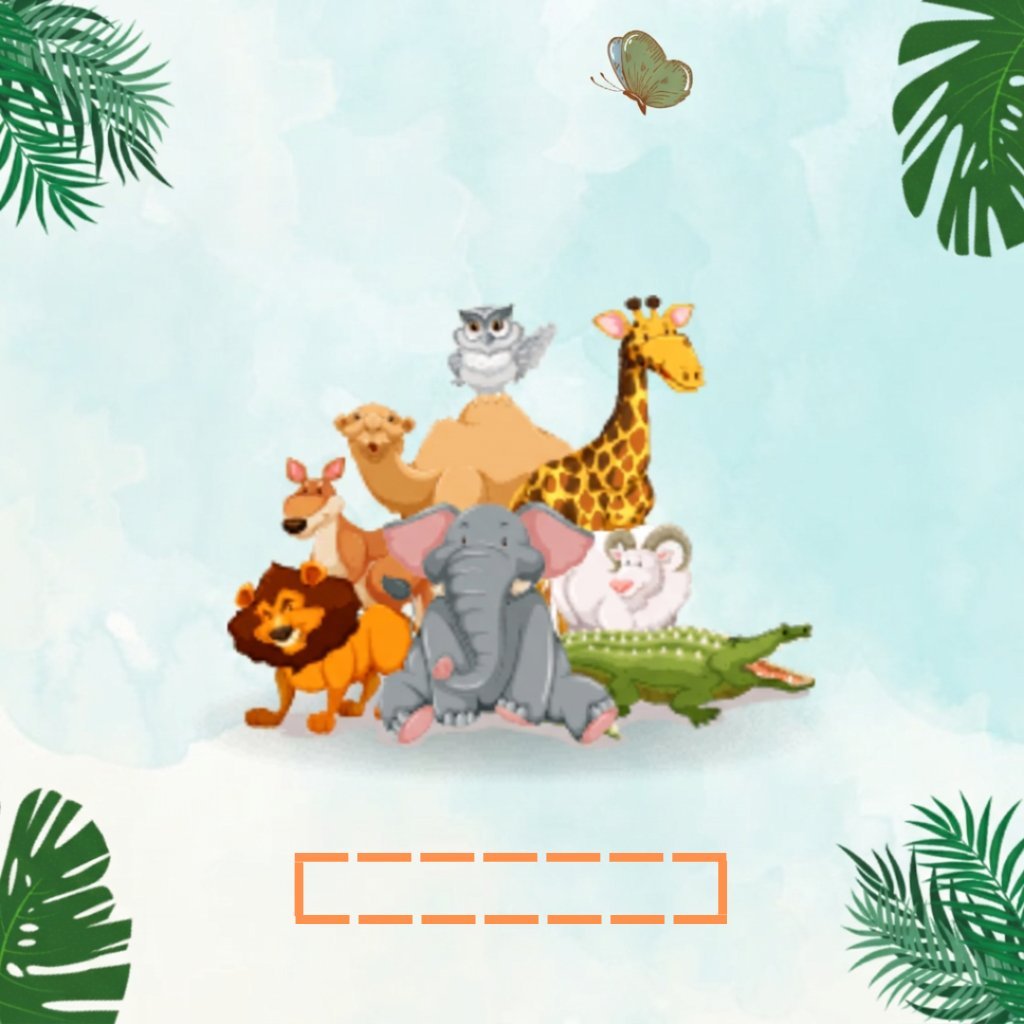}
    {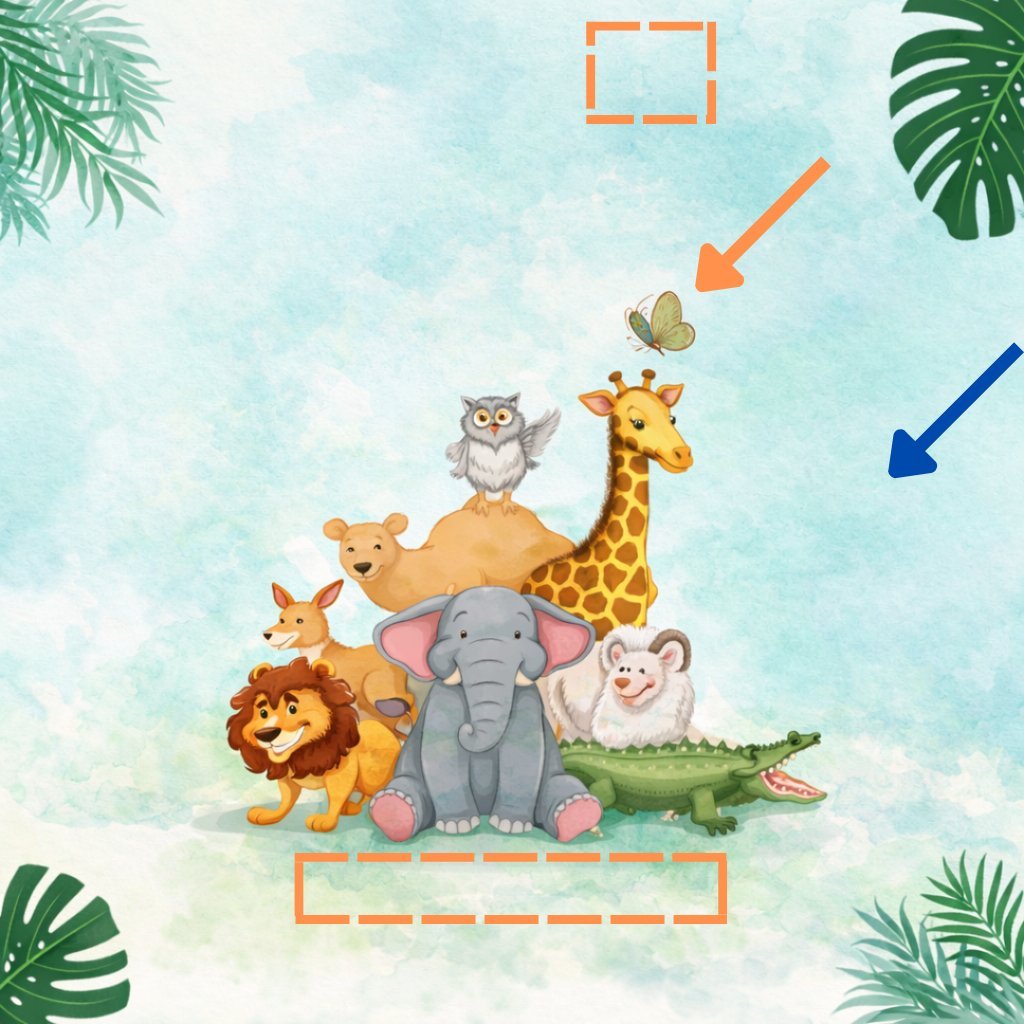}

    \vspace{-24pt}
    \caption{\textbf{Text erasing across aspect ratios.}
Given an input image and a text mask, the models shall remove the masked text and inpaint the removed regions.
Both models fail to confine edits to the masked region: GPT-Image 1.5 alters background color and tone, while both models further modify non-target elements (e.g., removing or recoloring the box in (b)).
These results show that design-aware erasing requires strict preservation of surrounding structure and style.
\textcolor[HTML]{FF914D}{Orange} marks structural changes; \textcolor[HTML]{004AAD}{blue} marks unintended color shifts.}
    \label{fig:text_eraser_qualitative}
\end{figure*}


\begin{table}[h]
\centering
\small
\setlength{\tabcolsep}{5pt}
\caption{Summary of key findings across typography tasks.}
\begin{tabular}{>{\raggedright\arraybackslash}p{2.8cm}>{\raggedright\arraybackslash}p{3.5cm}>{\raggedright\arraybackslash}p{2.5cm}>{\raggedright\arraybackslash}p{2cm}>{\raggedright\arraybackslash}p{2cm}}
\toprule
\textbf{Task Group} & \textbf{Key Finding} & \textbf{Best Performance} & \textbf{Best Model} & \textbf{Status} \\
\midrule
Font Recognition & Weak across the board; correct predictions concentrate on high-frequency typefaces while the vast majority of 167 families score zero F1 & 23.7\% top-1 acc, Macro-F1 $\ll$ acc & GPT-5.4 & Unsolved \\
\midrule
Text Color Prediction & Dramatic variation across models; weakest models produce colors perceptually unrelated to ground truth ($\Delta$E $>$ 52) & $\Delta$E\,=\,5.57, 63.3\% within $\Delta$E\,$<$\,5 & GPT-5.4 & Unsolved \\
\midrule
Typographic Properties (Size, Weight, Align, Spacing) & Font size and line height show the largest gaps across models (up to $4\times$); letter spacing is easiest for all models & Font size MAE\,=\,8.97\,px, Align\,=\,86.1\% & GPT-5.4 (most), Claude (align) & Partially solved \\
\midrule
Curved Text \& Rotation & Binary capability gap; some models correctly detect 88.5--92.3\% of curved elements while others score zero; rotated-only MAE ranges from 18.67° to 54° & 92.3\% curved detection, 18.67° rot. MAE & Claude (curved), GPT-5.4 (rotation) & Partially solved \\
\midrule
Style Range Detection & Span localization is achievable but exact recovery of full typographic specification is not; exact match is 0\% across all models & Span IoU\,=\,0.982, EM\,=\,0\% & Claude-Opus-4.6 & Partially solved \\
\midrule
Styled Text Generation & Layout-level text insertion remains unreliable: models frequently spill beyond the mask and alter nearby content, while fine-grained typography recovery remains limited & IoU\,=\,0.580 (layout), Font Acc.\,=\,46.8\% (layout) & Gemini-3.1-Flash-Img & Partially solved \\
\midrule
Text Removal \& Regeneration & Text removal is effectively solved but background reconstruction is not; higher removal rate does not imply better reconstruction & PSNR\,=\,30.22, SSIM\,=\,0.944, LPIPS\,=\,0.088 & Gemini-3.1-Flash-Image & Partially solved \\
\bottomrule
\end{tabular}
\label{tab:typography_findings}
\end{table}

\clearpage
 
\section{Infographics Tasks}
\label{sec:svg}
\FloatBarrier

\subsection{SVG Understanding \& Editing}
\label{sec:svg-understanding}

SVG is the native format for vector design assets such as icons, illustrations, decorative strokes, and background frames. Understanding and manipulating SVG code is distinct from raster-image reasoning, since it requires parsing structured markup, reasoning about geometric primitives, and producing syntactically valid code. We evaluate five tasks on 300 SVG assets from the LICA dataset (Section ~\ref{para:dataset}), each annotated with a complexity label (\emph{easy}, \emph{medium}, \emph{hard}) based on structural features following the stratification approach of SVGenius~\cite{chen2025svgenius}. 

All tasks use code-only input with no rendered images, and span two capability modes: \emph{understanding} and \emph{editing}. Understanding tasks are perceptual and semantic multiple-choice Q\&As generated by Claude Code agent (Opus-4.6) from the rendered SVG image, with four options and a verified answer. Editing tasks includes bug fixing, code optimization, and style editing, which are programmatically annotated and require the model to parse and reason about an existing SVG before perform the editing; since modifying SVG code presupposes understanding it, we treat editing as an extension of comprehension rather than generation. Bug fixing uses easy (${\sim}$5 errors), medium (${\sim}$7 with misspelled attributes), and hard (${\sim}$7 with path-data corruption) variants; code optimization is evaluated against ground truth generated by SVGO\,v4 (reference compression ratio 62.6\%); and style editing applies 1--3 combined operations from a predefined vocabulary of fill, stroke, opacity, viewBox, transform, and
  color inversion commands.
Table~\ref{tab:svg-understand-tasks} summarizes the five understanding and editing tasks.
Complexity labels are assigned by a rule-based scorer over structural features (path count, element depth, attribute diversity).

\begin{table}[b]
\centering
\small
\setlength{\tabcolsep}{5pt}
\caption{SVG understanding and editing task definitions. All tasks share the same 300 source SVGs; editing tasks generate multiple difficulty variants per SVG.}
\begin{tabular}{>{\raggedright\arraybackslash}p{2.2cm}>{\raggedright\arraybackslash}p{5.5cm}>{\raggedright\arraybackslash}p{1.2cm}>{\raggedright\arraybackslash}p{4.5cm}}
\toprule
\textbf{Task} & \textbf{Description} & \textbf{Samples} & \textbf{Metrics} \\
\midrule
Perceptual Q\&A & Given SVG source code only (no rendered image), answer a multiple-choice question (A--D) about visual properties: dominant color, shape count, element presence, or aspect ratio. & 300 & Accuracy \\
\midrule
Semantic Q\&A & Same format as perceptual Q\&A, but the question targets meaning: what object the graphic depicts, its application context, or design category. & 300 & Accuracy \\
\midrule
Bug Fixing & Repair a deliberately corrupted SVG. Easy (${\sim}$5 errors: missing tags/quotes), Medium (${\sim}$7: misspelled attributes, garbage chars), Hard (${\sim}$7: path-data corruption). & 900 & Repair Similarity \\
\midrule
Code Optimization & Produce a smaller SVG that renders identically to the original. Reference compression ratio from SVGO\,v4 is 62.6\%. & 300 & Compression Ratio \\
\midrule
Style Editing & Apply a natural-language edit command (e.g.\ color change + rotation). Easy (1 op), Medium (2 ops), Hard (3 ops). & 900 & Edit Distance \\
\bottomrule
\end{tabular}
\label{tab:svg-understand-tasks}
\end{table}

\begin{figure}[htbp]
    \centering
    \includegraphics[width=0.75\textwidth]{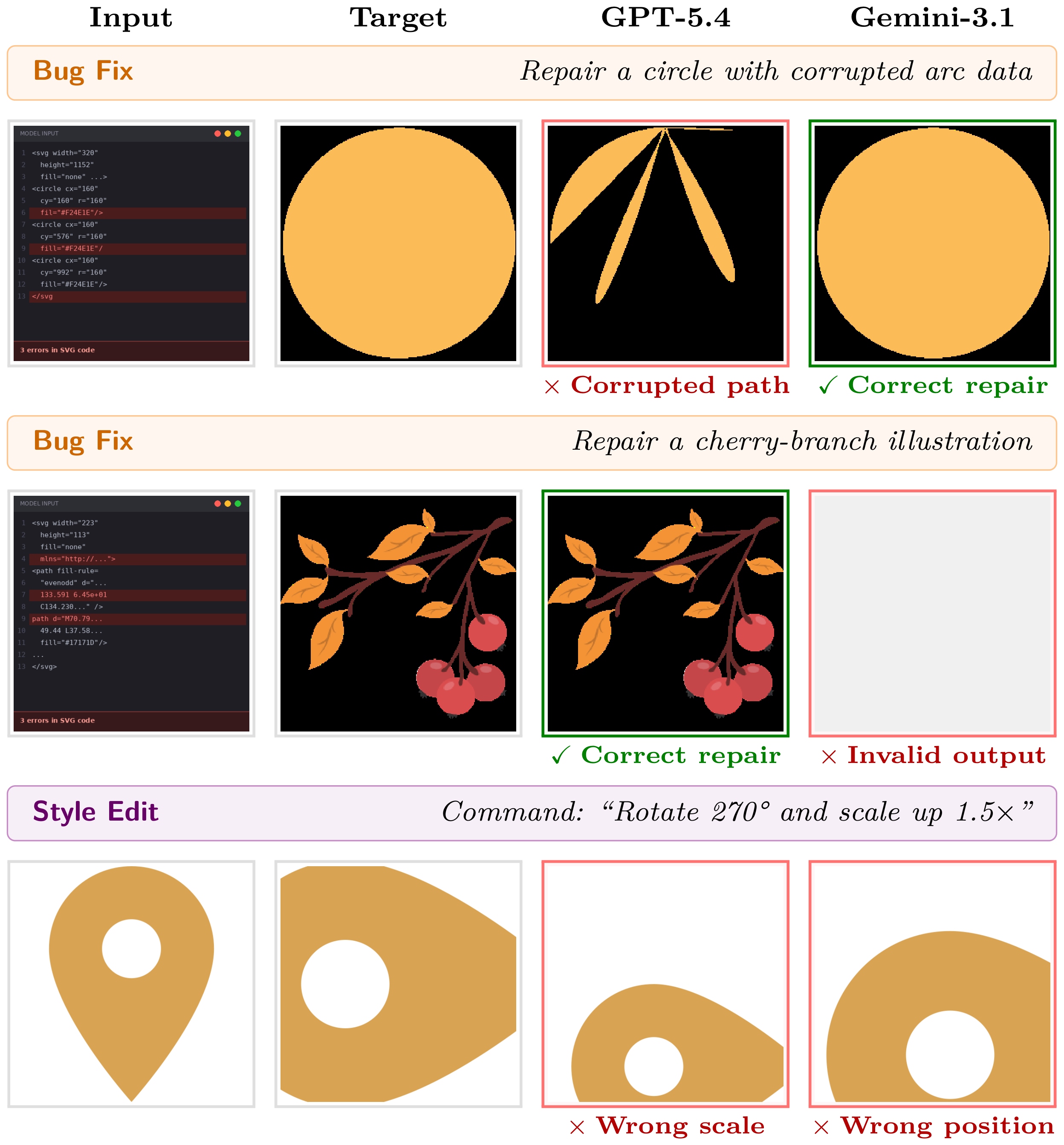}
    
    \caption{SVG editing failures.
    \textbf{Bug Fix} (rows~1--2): GPT-5.4 corrupts the circle into a fan shape while Gemini repairs it correctly (row~1), yet GPT-5.4 succeeds where Gemini produces invalid output (row~2).
    \textbf{Style Edit} (row~3): both models produce incorrect scale or displaced content on the multi-step geometric transform.}
\label{fig:svg-edit-failures}
\end{figure}

\paragraph{Results.}
Table~\ref{tab:svg-results} presents the results with 
all metrics defined in Section~\ref{sec:metrics}. The results reveal three key patterns across the SVG understanding and editing tasks:
\begin{itemize}
    \item \textbf{Perceptual reasoning about SVG code is harder than semantic interpretation.} The gap between perceptual Q\&A (87.0\% best) and semantic Q\&A (93.7\% best) confirms empirically that inferring visual properties directly from SVG markup is harder than understanding what an SVG depicts.
    \item \textbf{Understanding and editing strengths are complementary across models.} The best model achieves repair similarity of 0.932 and leads on style editing (EditD\,=\,0.183), while the other leads on semantic Q\&A (93.7\%) and code optimization (CompR\,=\,0.870), suggesting that SVG tasks draw on different underlying capabilities: structured-code proficiency aids repair and editing, while perceptual grounding aids semantic interpretation and compression.
    \item \textbf{Multi-operation geometric transforms remain challenging for all models.} Failure cases on style editing reveal that combined rotation and scaling commands produce incorrect angles, positions, and displaced content across both models, indicating that composing multiple geometric operations in SVG space is largely unsolved. For representative failure cases, see Figure~\ref{fig:svg-edit-failures}.
\end{itemize}

\begin{table}[h]
\centering
\caption{SVG understanding and editing results (code-only input).
Best result per metric in \textbf{bold}.
RepSim\,=\,repair similarity ($\uparrow$); CompR\,=\,compression ratio ($\downarrow$); EditD\,=\,edit distance ($\downarrow$).
Gemini-3.1$^\dagger$ selects the best-performing Gemini-3.1 variant per task (Pro or Flash-Lite; see Appendix).}
\label{tab:svg-results}
\small
\begin{tabular}{@{}l cc ccc@{}}
\toprule
& \multicolumn{2}{c}{\textbf{Understanding}}
& \multicolumn{3}{c}{\textbf{Editing}} \\
\cmidrule(lr){2-3} \cmidrule(lr){4-6}
\textbf{Model}
  & \shortstack{Perceptual Q\&A \\ Acc$\uparrow$} & \shortstack{Semantic Q\&A \\ Acc$\uparrow$}
  & \shortstack{Bug Fixing \\ RepSim$\uparrow$} & \shortstack{Code Optim. \\ CompR$\downarrow$} & \shortstack{Style Edit \\ EditD$\downarrow$} \\
\midrule
GPT-5.4               & 0.847          & \textbf{0.937} & 0.793          & \textbf{0.870} & 0.242 \\
Gemini-3.1$^\dagger$  & \textbf{0.870} & 0.900          & \textbf{0.932} & 0.884            & \textbf{0.183}            \\
\bottomrule
\end{tabular}
\end{table}



\subsection{Infographics Generation}
\label{sec:svg-generation}

Generating valid SVG from a text description, reference image, or both is a code-generation task with visual considerations: the model shall produce syntactically correct vector markup whose rendering matches a target graphic.
We evaluate three tasks on 300 SVG assets from the LICA dataset (Section~\ref{para:dataset}), distinguished by input modality:
Text-to-SVG (description only),
Image-to-SVG (rendered PNG only), and
Text+Image-to-SVG (both).
Generated SVGs are rendered at $256{\times}256$ pixels for evaluation.
To further investigate SVG generation across different element types, we curated 50 additional samples for each of the following categories: \texttt{chart} (charts and data visualizations), \texttt{stroke} (decorative strokes and borders), and \texttt{container} (frames and backgrounds).

We additionally evaluate generation of Lottie animations using a curated subset of 50 animations (${\le}$10 layers, ${\le}$50\,KB, mean duration 4.4\,s), extending vector generation to the temporal domain. Generated Lottie files are rendered using the official LottieFiles player to ensure evaluation reflects real-world playback fidelity. Table~\ref{tab:svg-gen-tasks} summarizes the five generation tasks.

\begin{table}[htbp]
\centering
\small
\setlength{\tabcolsep}{5pt}
\caption{SVG and Lottie generation task definitions. All SVG tasks share the same 300 assets; Lottie tasks use a curated 50-animation subset.}
\begin{tabular}{>{\raggedright\arraybackslash}p{2.2cm}>{\raggedright\arraybackslash}p{5.5cm}>{\raggedright\arraybackslash}p{1.2cm}>{\raggedright\arraybackslash}p{4.5cm}}
\toprule
\textbf{Task} & \textbf{Description} & \textbf{Samples} & \textbf{Metrics} \\
\midrule
Text-to-SVG & Generate SVG code from a natural-language description only. Tests semantic-to-code translation without visual reference. & 300 & Valid, SSIM, LPIPS, MSE, Complexity \\
\midrule
Image-to-SVG & Generate SVG code from a rendered PNG of the target. Tests visual-to-code translation by inferring vector structure from raster input. & 300 & Valid, SSIM, LPIPS, MSE, Complexity \\
\midrule
Text+Image-to-SVG & Generate SVG code from both description and rendered PNG. Provides the richest input, serving as an upper bound. & 300 & Valid, SSIM, LPIPS, MSE, Complexity \\
\midrule
Text-to-Lottie & Generate Lottie JSON from a natural-language description of the animation. & 50 & Valid, FrameSSIM, FrameMSE, StructSim \\
\midrule
Text+Image-to-Lottie & Generate Lottie JSON from both a description and a rendered keyframe PNG. & 50 & Valid, FrameSSIM, FrameMSE, StructSim \\
\bottomrule
\end{tabular}
\label{tab:svg-gen-tasks}
\end{table}

\begin{figure*}[htbp]
    \centering
    \includegraphics[width=0.8\textwidth]{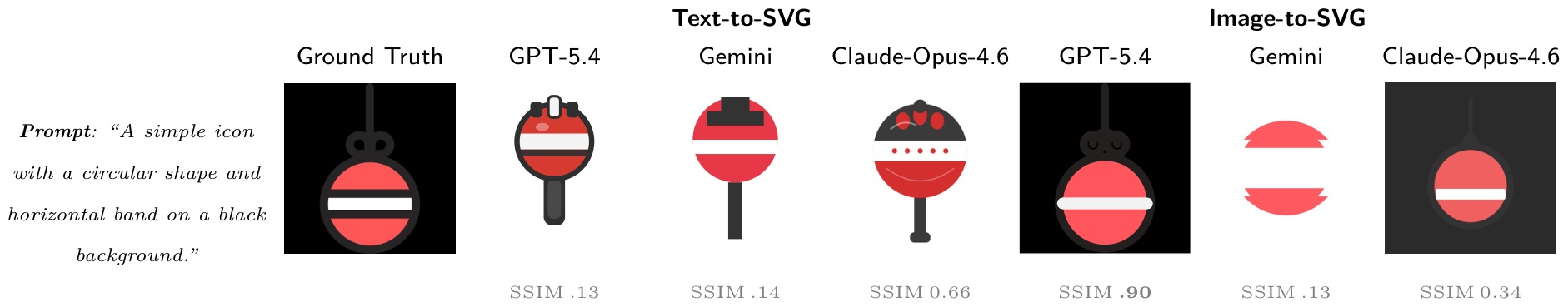}
    \caption{SVG generation examples. From left to right: the text prompt, ground truth, and six model outputs for text-to-SVG and image-to-SVG from GPT-5.4, Gemini-3.1 and Claude-Opus-4.6. All models fail to follow the instruction of a black background in the text-to-SVG generation, and introduce artifacts in the image-to-SVG task.}
    \label{fig:svg-gen-examples}
\end{figure*}

\begin{figure}[htbp]
    \centering
    \includegraphics[width=0.9\textwidth]{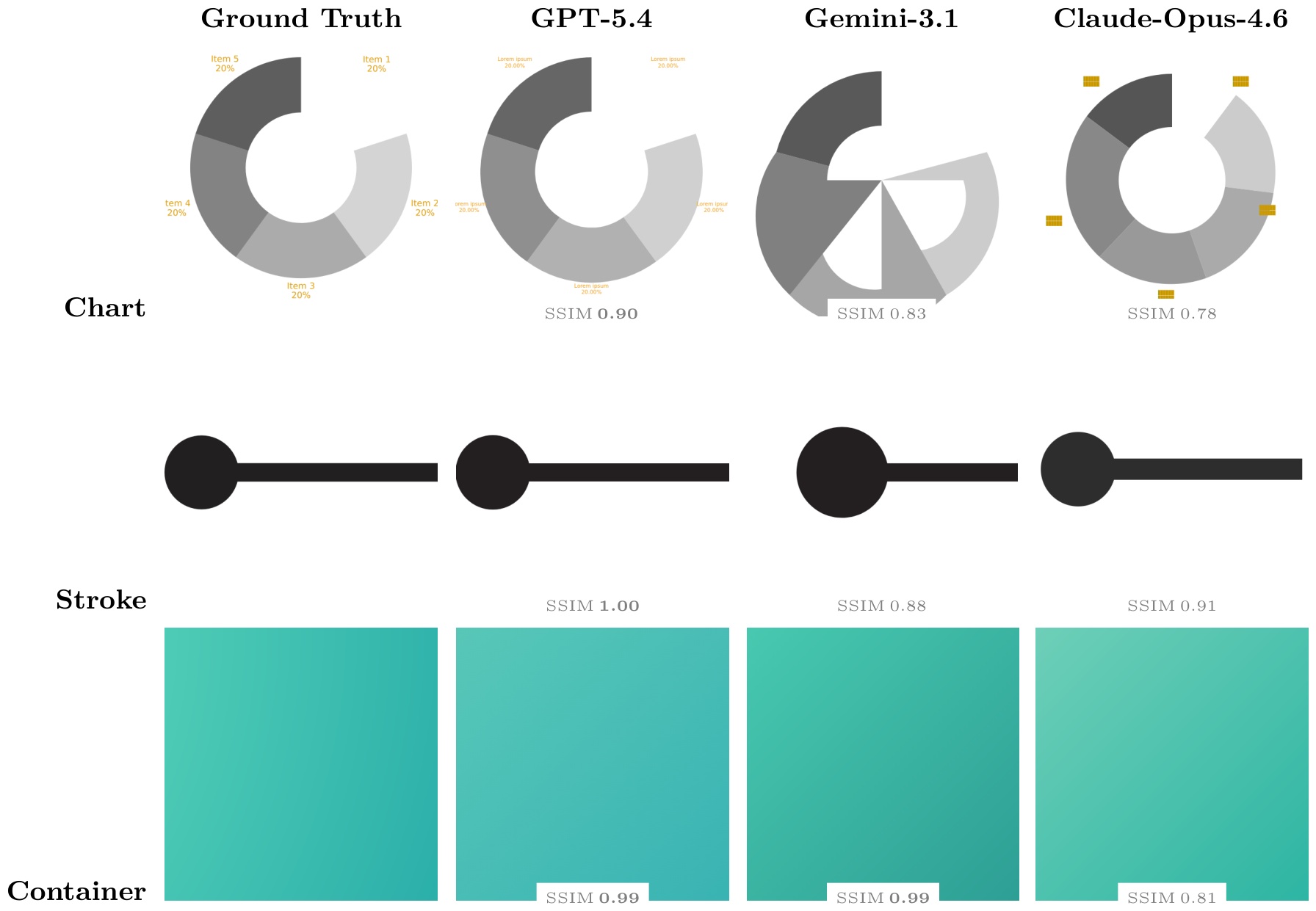}
    \caption{Qualitative image-to-SVG examples across SVG types. For the Chart element, GPT-5.4 achieves the closest reproduction (SSIM 0.90), whereas Gemini omits the central hole and labels (0.83) and Claude modifies the endpoints gap  and labels (0.78). GPT-5.4 accurately reconstructs the stroke, while Claude-Opus-4.6 exaggerates its width and Gemini-3.1 reduces it. All model produce incorrect gradient values for the container type.}
    \label{fig:svg-gen-by-type-qual}
\end{figure}

\begin{figure}[htbp]
    \centering
    \includegraphics[width=0.85\textwidth]{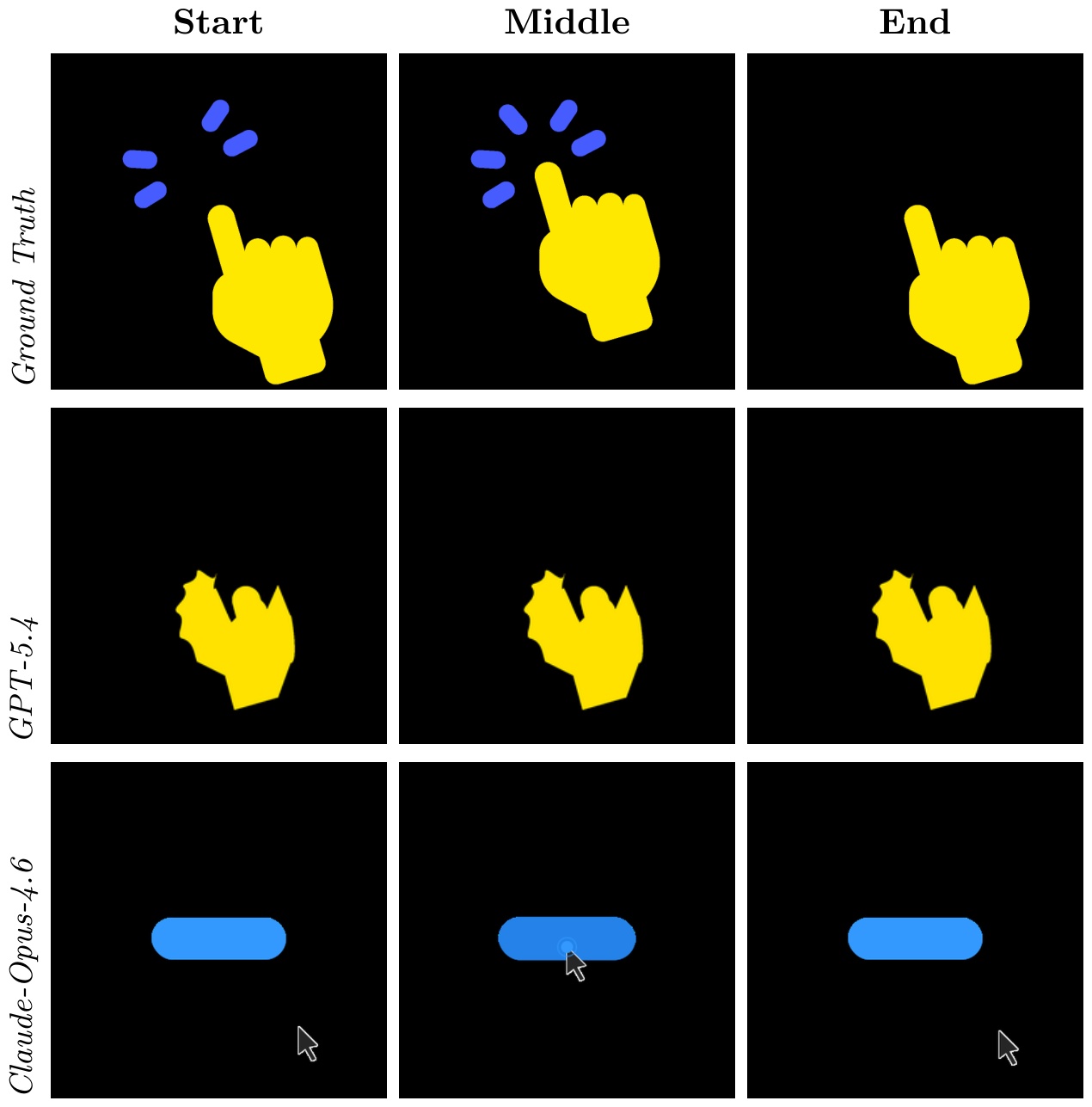}
    \caption{Lottie generation failure (animated CTA click cursor).
\textit{Top}: ground-truth keyframes showing a pointing hand with animated blue sparks.
\textit{Middle}: GPT-5.4 produces a static, distorted hand with no spark layer and no animation.
\textit{Bottom}: Claude-Opus-4.6 miss the hand completely, and produce a moving cursor instead to represent the click animation instead.
Gemini-3.1's visual is excluded since output is invalid. All models fail to reproduce the multi-layer animated structure.}
\label{fig:lottie-failure-example}
\end{figure}

\paragraph{Results.} For qualitative examples, see Figures~\ref{fig:svg-gen-examples}, \ref{fig:svg-gen-by-type-qual}, and~\ref{fig:lottie-failure-example}.
 
The results across static SVG and Lottie generation reveal three key patterns:
\begin{itemize}
    \item \textbf{SVG generation quality is strongly modality-dependent.} Image conditioning dramatically improves fidelity: the best model achieves 0.918 SSIM on image-to-SVG versus 0.733 on text-only (Table~\ref{tab:svg-gen-results}). GPT-5.4 and Claude-Opus-4.6 achieves near perfect validity (successful render) across all modalities, while Gemini-3.1's validity drops to 81\% (image-only)/84\% (text+image) on image-conditioned tasks as output truncation becomes more frequent on complex SVGs.
    \item \textbf{Lottie animation generation is feasible but far from solved.} Even the best result (FrameSSIM\,=\,0.598) is far below static SVG performance (Table~\ref{tab:lottie-gen-results}). Multimodal input provides large gains for GPT-5.4 but degrades Gemini's output quality, dropping its validity from 100\% to 66\%.
\end{itemize}

\begin{table}[htbp]
\centering
\caption{SVG generation results across input modalities.
Best result per metric in \textbf{bold}.
Cmplx\,=\,weighted complexity.
Gemini-3.1$^\dagger$ selects the best-performing Gemini-3.1 variant per task.}
\label{tab:svg-gen-results}
\small
\begin{tabular}{@{}ll ccccc@{}}
\toprule
\textbf{Input} & \textbf{Model}
  & Valid$\uparrow$ & SSIM$\uparrow$
  & LPIPS$\downarrow$ & MSE$\downarrow$
  & Cmplx \\
\midrule
\multirow{3}{*}{Text}
  & Claude-Opus-4.6      & 0.970          & 0.724          & 0.450          & 0.118          & 27.4 \\
  & GPT-5.4              & \textbf{1.000} & \textbf{0.733} & 0.498          & \textbf{0.058} & 25.3 \\
  & Gemini-3.1$^\dagger$ & 0.977          & 0.723          & \textbf{0.448} & 0.127          & 18.4 \\
\midrule
\multirow{3}{*}{Image}
  & Claude-Opus-4.6      & \textbf{1.000} & 0.744          & 0.315          & 0.109          & 17.3 \\
  & GPT-5.4              & \textbf{1.000} & \textbf{0.918} & \textbf{0.098} & \textbf{0.005} & 21.8 \\
  & Gemini-3.1$^\dagger$ & 0.840          & 0.695          & 0.348          & 0.220          & 23.4 \\
\midrule
\multirow{3}{*}{Text+Image}
  & Claude-Opus-4.6      & 0.987          & 0.741          & 0.334          & 0.116          & 16.3 \\
  & GPT-5.4              & \textbf{1.000} & \textbf{0.870} & \textbf{0.197} & \textbf{0.017} & 22.7 \\
  & Gemini-3.1$^\dagger$ & 0.813          & 0.691          & 0.354          & 0.235          & 19.4 \\
\bottomrule
\end{tabular}
\end{table}
 
\begin{table}[htbp]
\centering
\caption{SVG generation results stratified by SVG type (50 samples each).
Best result per metric within each input modality in \textbf{bold}.}
\label{tab:svg-gen-by-svgtype}
\small
\setlength{\tabcolsep}{4pt}
\begin{tabular}{@{}ll l cccc@{}}
\toprule
\textbf{Input} & \textbf{SVG Type} & \textbf{Model}
  & Valid$\uparrow$ & SSIM$\uparrow$
  & LPIPS$\downarrow$ & MSE$\downarrow$ \\
\midrule
\multirow{9}{*}{\rotatebox{90}{Text}}
  & \multirow{3}{*}{chart}
    & Claude-Opus-4.6 & 0.98          & 0.687          & \textbf{0.526} & 0.084 \\
    & & GPT-5.4    & \textbf{1.00} & \textbf{0.757} & 0.551 & \textbf{0.078} \\
  & & Gemini-3.1 & \textbf{1.00} & 0.733          & 0.555          & 0.094 \\
  \cmidrule{2-7}
  & \multirow{3}{*}{stroke}
    & Claude-Opus-4.6 & \textbf{1.00} & \textbf{0.690} & 0.537          & \textbf{0.190} \\
    & & GPT-5.4    & \textbf{1.00} & 0.689 & \textbf{0.501} & 0.217 \\
  & & Gemini-3.1 & 0.980          & 0.643          & 0.551          & 0.258 \\
  \cmidrule{2-7}
  & \multirow{3}{*}{container}
    & Claude-Opus-4.6 & \textbf{1.00} & \textbf{0.685} & 0.499          & \textbf{0.175} \\
    & & GPT-5.4    & \textbf{1.00} & 0.673 & 0.512          & 0.193 \\
  & & Gemini-3.1 & \textbf{1.00} & 0.671          & \textbf{0.496} & 0.194 \\
\midrule
\multirow{9}{*}{\rotatebox{90}{Image}}
  & \multirow{3}{*}{chart}
    & Claude-Opus-4.6 & \textbf{1.00} & 0.731          & 0.399          & 0.049 \\
    & & GPT-5.4    & \textbf{1.00} & \textbf{0.833} & \textbf{0.322} & \textbf{0.039} \\
  & & Gemini-3.1 & \textbf{1.00} & 0.821          & 0.367          & 0.045 \\
  \cmidrule{2-7}
  & \multirow{3}{*}{stroke}
    & Claude-Opus-4.6 & \textbf{1.00} & 0.771          & 0.228          & 0.202 \\
    & & GPT-5.4    & \textbf{1.00} & 0.792          & 0.200          & 0.197 \\
  & & Gemini-3.1 & \textbf{1.00} & \textbf{0.837} & \textbf{0.185} & \textbf{0.146} \\
  \cmidrule{2-7}
  & \multirow{3}{*}{container}
    & Claude-Opus-4.6 & \textbf{1.00} & 0.818          & 0.260          & 0.083 \\
    & & GPT-5.4    & \textbf{1.00} & \textbf{0.921} & \textbf{0.136} & \textbf{0.041} \\
  & & Gemini-3.1 & \textbf{1.00} & 0.838          & 0.241          & 0.079 \\
\midrule
\multirow{9}{*}{\rotatebox{90}{Text+Image}}
  & \multirow{3}{*}{chart}
    & Claude-Opus-4.6 & 0.98          & 0.713          & 0.416          & 0.068 \\
    & & GPT-5.4    & \textbf{1.00} & 0.820 & 0.376          & \textbf{0.044} \\
  & & Gemini-3.1 & 0.920          & \textbf{0.824}          & \textbf{0.351} & 0.045 \\
  \cmidrule{2-7}
  & \multirow{3}{*}{stroke}
    & Claude-Opus-4.6 & \textbf{1.00} & 0.686          & 0.397          & 0.239 \\
    & & GPT-5.4    & \textbf{1.00} & 0.738          & 0.324          & 0.236 \\
  & & Gemini-3.1 & \textbf{1.00} & \textbf{0.819} & \textbf{0.218} & \textbf{0.153} \\
  \cmidrule{2-7}
  & \multirow{3}{*}{container}
    & Claude-Opus-4.6 & \textbf{1.00} & 0.797          & 0.294          & 0.095 \\
    & & GPT-5.4    & \textbf{1.00} & \textbf{0.861}          & 0.226          & 0.085 \\
  & & Gemini-3.1 & \textbf{1.00} & 0.857          & \textbf{0.196} & \textbf{0.071} \\
\bottomrule
\end{tabular}
\end{table}
 
\begin{table}[htbp]
\centering
\caption{Lottie animation generation results (50 samples each).
Best result per metric in \textbf{bold}.
FrameSSIM and FrameMSE are averaged over five rendered keyframes.
StructSim measures layer, type and dimension similarity.}
\label{tab:lottie-gen-results}
\small
\begin{tabular}{@{}ll ccccc@{}}
\toprule
\textbf{Input} & \textbf{Model}
  & Valid$\uparrow$ & FrameSSIM$\uparrow$
  & FrameMSE$\downarrow$ & StructSim$\uparrow$
  & Length \\
\midrule
\multirow{3}{*}{Text}
  & Claude-Opus-4.6 & 0.600          & 0.324          & 0.596          & 0.372          & 12{,}443 \\
  & GPT-5.4        & 0.980          & 0.242          & 0.666          & \textbf{0.612} & 4{,}933 \\
  & Gemini-3.1-pro & \textbf{1.000} & \textbf{0.362} & \textbf{0.507} & 0.597          & 4{,}795 \\
\midrule
\multirow{3}{*}{Text+Image}
  & Claude-Opus-4.6 & 0.820          & 0.486          & 0.430          & 0.613          & 8{,}650 \\
  & GPT-5.4        & \textbf{0.960} & \textbf{0.598} & \textbf{0.292} & \textbf{0.734} & 5{,}172 \\
  & Gemini-3.1-pro & 0.660          & 0.381          & 0.565          & 0.426          & 11{,}602 \\
\bottomrule
\end{tabular}
\end{table}

\begin{table}[htbp]
\centering
\small
\setlength{\tabcolsep}{5pt}
\caption{Summary of key findings across SVG \& vector tasks.}
\begin{tabular}{>{\raggedright\arraybackslash}p{2.5cm}>{\raggedright\arraybackslash}p{4cm}>{\raggedright\arraybackslash}p{2.5cm}>{\raggedright\arraybackslash}p{2cm}>{\raggedright\arraybackslash}p{2cm}}
\toprule
\textbf{Task Group} & \textbf{Key Finding} & \textbf{Best Performance} & \textbf{Best Model} & \textbf{Status} \\
\midrule
Perceptual Q\&A & Semantic questions are easier than perceptual ones across all models, suggesting that inferring \emph{what} an SVG depicts from code is easier than reasoning about precise visual properties & 87.0\%  & Gemini-3.1 & Partially solved \\
\midrule
Semantic Q\&A & High accuracy from code alone & 93.7\% accuracy & GPT-5.4 & Partially solved \\
\midrule
Bug Fixing & precise SVG code repair remains difficult even for frontier models as there is a notable gap from perfect repair & 0.932 Repair Similarity & Gemini-3.1 & Partially solved \\
\midrule
Code Optimization & both frontier models lags behind the SVGO baseline (0.626)  & 0.870 compression ratio & GPT-5.4 &  Partially solved \\
\midrule
Style Editing & GPT produces minimal, targeted edits closest to the reference; other models introduce extraneous changes & 0.183 edit distance & Gemini-3.1 & Partially solved \\
\midrule
Text-to-SVG Generation & Both models produce valid SVGs at high rates; text-only input yields schematic interpretations rather than pixel-faithful reproductions & SSIM\,=\,0.733, 100\% validity & GPT-5.4 & Unsolved \\
\midrule
Image-to-SVG Generation & Image conditioning dramatically improves fidelity; GPT-5.4 achieves near-faithful reproduction while Gemini's validity drops due to output truncation & SSIM\,=\,0.918, LPIPS\,=\,0.098 & GPT-5.4 & Partially solved \\
\midrule
Text+Image-to-SVG Generation & Multimodal input (text+image) generally under-performs image-only input, suggesting that current models do not effectively integrate textual and visual signals for vector code generation & SSIM\,=\,0.870, 100\% validity & GPT-5.4 & Partially solved \\
\midrule
Text-to-Lottie & Gemini achieves perfect validity and best frame similarity from text alone; structural similarity is moderate for both models & F-SSIM\,=\,0.362, 100\% valid & Gemini-3.1-pro & Unsolved \\
\midrule
Text+Image-to-Lottie & Image input provides large gains for GPT-5.4 but degrades Gemini's validity from 100\% to 66\%; animated structure remains largely unreproduced & F-SSIM\,=\,0.598, StructSim\,=\,0.734 & GPT-5.4 & Unsolved \\
\bottomrule
\end{tabular}
\label{tab:svg_findings}
\end{table}

\clearpage
 
\section{Template \& Design Semantics}
\label{sec:semantics}
\FloatBarrier

This section evaluates models on higher-level design semantics: classifying templates by purpose, predicting designer intent, and understanding and generating template variants. Table~\ref{tab:semantics_tasks} lists the task definitions and dataset statistics for this domain.

\begin{table}[h]
\centering
\small
\setlength{\tabcolsep}{5pt}
\caption{Template \& design semantics task definitions and dataset statistics.}
\begin{tabular}{>{\raggedright\arraybackslash}p{2.5cm}>{\raggedright\arraybackslash}p{5cm}>{\raggedright\arraybackslash}p{1.2cm}>{\raggedright\arraybackslash}p{2cm}>{\raggedright\arraybackslash}p{2.5cm}}
\toprule
\textbf{Task} & \textbf{Description} & \textbf{Samples} & \textbf{Classes} & \textbf{Metrics} \\
\midrule
Category Classification & Predict the parent category and sub-category of a design template from a rendered image. Tested under open-vocabulary and label-constrained prompting. & 989 & 29 parent, 858 sub & Top-1/5 Acc, Macro-F1 \\
\midrule
User Intent Prediction & Given a rendered design, describe the purpose, audience, and desired outcome in a single sentence. Free-text generation task. & 989 & Free-text & BERTScore, Cosine Sim. \\
\midrule
Template Variant Understanding & Recognize that designs originate from the same template despite differing content. Three sub-tasks: pairwise matching, ranking, and clustering. & 1{,}000 / 200 / 50 & Binary / Ranked / Clustered & Acc, MRR, MAP, ARI, AMI \\
\midrule
Template Variant Generation & Produce design layouts respecting template conventions. two sub-tasks: style completion and color transfer. & 140 / 200  & — & Validity, font/color match, SSIM, LPIPS \\
\bottomrule
\end{tabular}
\label{tab:semantics_tasks}
\end{table}

\subsection{Semantics Understanding: Category Classification}
\label{sec:category-classification}

Category classification predicts the intended use of a visual design template from a rendered image alone. Each template is annotated with a two-level hierarchy: a parent category, denoting a broad template type (e.g.\ ``business card'', ``Instagram post'') and a sub-category describing the visual aesthetic or theme (e.g.\ ``professional'', ``skincare''). We test under \emph{open-vocabulary} and \emph{label-constrained} prompting (see Appendix Table~\ref{tab:prompts-category-und}).

\begin{figure}[t]
\centering
\begin{subfigure}{0.49\textwidth}
  \centering
  \includegraphics[width=\linewidth]{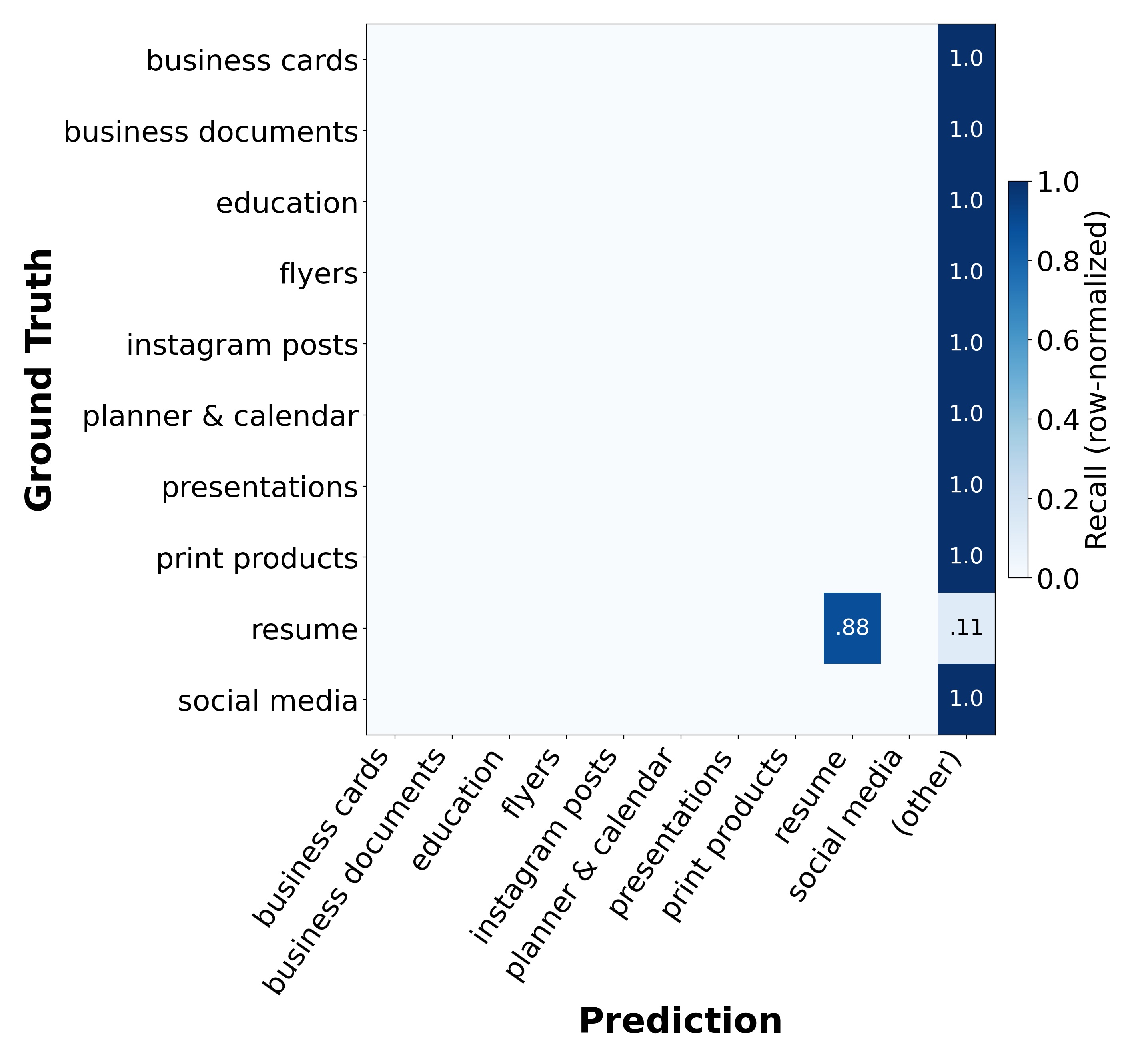}
  \caption{Open-vocabulary prompting}
  \label{fig:test1}
\end{subfigure}
\hfill
\begin{subfigure}{0.49\textwidth}
  \centering
  \includegraphics[width=\linewidth]{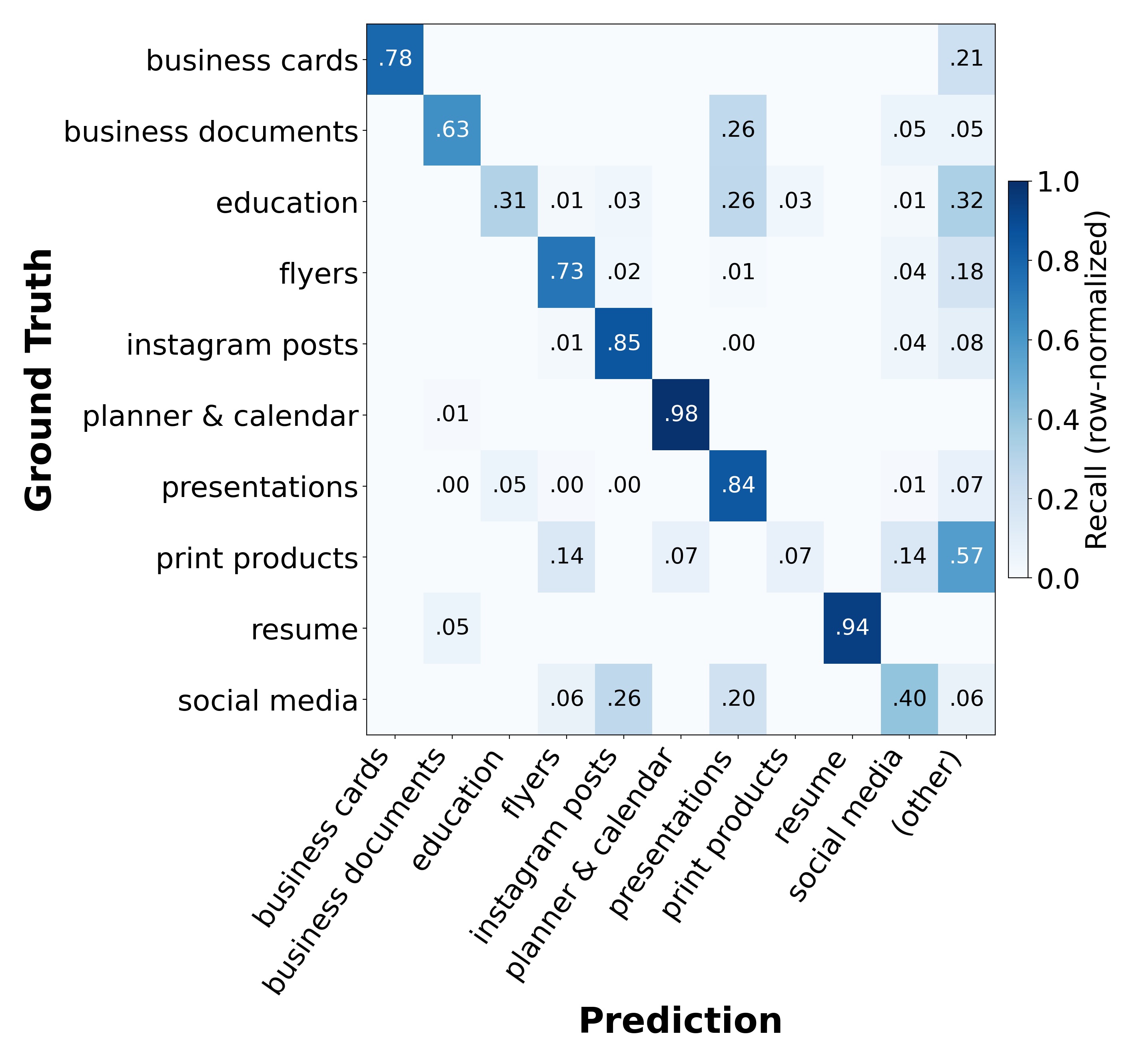}
  \caption{Constrained prompting}
  \label{fig:test2}
\end{subfigure}
\caption{Confusion matrices for top-10 highest recurring parent category classification on Claude Opus~4.6 (best overall model). \textbf{(a)}~Under open-vocabulary prompting, nearly all predictions fall into the ``other'' bin, illustrating pervasive label aliasing. \textbf{(b)}~Providing the closed label set recovers strong diagonal structure; well-delineated categories such as planner \& calendar (.98) and resume (.94) are near-perfectly recognized, while visually overlapping
categories such as education (.31) and social media (.40) remain challenging.}
\label{fig:category-classification}
\end{figure}

\paragraph{Results.}
Table~\ref{tab:category_classification} presents the results for both prompting strategies with all metrics defined in Section~\ref{sec:metrics}. Figure~\ref{fig:category-classification} illustrates the confusion matrices of the best performing model on label-constrained prompting.

The results reveal three key patterns:
\begin{itemize}
    \item \textbf{Label constraining yields +30--62\,pp gains on parent accuracy.} Under open-vocabulary prompting, models frequently produce semantically reasonable but lexically mismatched labels (e.g. ``social media graphic" instead of ``Instagram post"), which are scored as incorrect. In practice, this also exposes a usability challenge: a user has no clear way to anticipate which label variants the model will recognize versus treat as mismatches, making effective prompting non-trivial. Providing the closed label set recovers strong diagonal structure in the confusion matrix (Figure~\ref{fig:category-classification}b), with well-delineated categories.
    \item \textbf{Model rankings shift between prompting conditions.} Gemini~3.1 leads open-vocabulary (45.5\%), suggesting stronger internalized design vocabulary, while Claude Opus leads label-constrained (78.7\% Top-1). GPT~5.4's open-vocabulary deficit is primarily a naming problem (91.0\% Top-5 under constraining).
    \item \textbf{Sub-category classification remains an open challenge}, with the best Top-1 reaching only 10.13\% even under label-constrained prompting.
\end{itemize}

\begin{table}[h]
    \centering
    \caption{Parent ($N{=}989$) and sub-category ($N{=}858$) classification accuracy and macro-F1 across prompting conditions. Claude Opus 4.6 leads label-constrained prompting, while Gemini 3.1 leads open-vocabulary prompting suggesting a stronger internalized design vocabulary.}
    \begin{tabular}{c|c|ccc|ccc}
        \toprule
        \multirow{2}{*}{\textbf{Model}} & \multirow{2}{*}{\textbf{Prompt}} & \multicolumn{3}{c}{\textbf{Parent Category}} & \multicolumn{3}{|c}{\textbf{Sub-Category}} \\[3pt]
        & & Top-1 Acc. & Top-5 Acc. & Macro-F1 & Top-1 Acc. & Top-5 Acc. & Macro-F1 \\
        \midrule
        \midrule
        \multirow{2}{*}{Gemini-3.1} & open-vocab. & \textbf{45.50} & \textbf{64.00} & \textbf{0.339} & \textbf{03.72} & \textbf{19.58} & \textbf{0.049} \\
        & label-const. & 76.23 & 91.50 & 0.474 & \textbf{10.13} & 36.48 & \textbf{0.131} \\
        \midrule
        \multirow{2}{*}{GPT-5.4} & open-vocab. & 11.42 & 21.84 & 0.200 & 02.79 & 17.24 & 0.021 \\
        & label-const. & 55.81 & 91.00 & 0.372 & 09.90 & \textbf{39.39} & 0.108 \\
        \midrule
        \multirow{2}{*}{Opus-4.6} & open-vocab. & 17.08 & 41.05 & 0.312 & 1.39 & 15.38 & 0.009 \\
        & label-const. & \textbf{78.66} & \textbf{94.13} & \textbf{0.506} & \textbf{10.13} & 36.94 & 0.126 \\
        \bottomrule
    \end{tabular}
    \label{tab:category_classification}
\end{table}

\footnotetext{We omit Gemini-3.1-Pro from this evaluation. Its extended thinking budget produces verbose, chain-of-thought-laden responses that significantly inflate output length, depressing both BERTScore and embedding similarity (54.10 and 58.97, respectively). We therefore report Gemini-3.1-Flash-Lite as the representative Gemini variant.}        
\subsection{Semantics Understanding: User Intent Prediction}
\label{sec:intent-prediction}

User intent prediction is a free-text generation task: given only a rendered image, the model shall describe the purpose, audience, and desired outcome that motivated the design. We evaluate on 989 templates, each paired with a human-authored single-sentence intent description, and report BERTScore~(F1) and Llama embedding cosine similarity (Section~\ref{sec:metrics}).

\begin{table}[h] 
    \centering 
    \caption{Semantic similarity between model-generated and human-annotated user intents for design templates.\protect\footnotemark}
    \begin{tabular}{c|cc} 
        \toprule 
        \textbf{Model} & \textbf{BERTScore (F1) $\uparrow$} & \textbf{Llama Cosine Similarity $\uparrow$} \\ 
        \midrule 
        Gemini-3.1-Flash-Lite & \textbf{89.55} & \textbf{92.71} \\
        GPT-5.4 & 89.03 & 91.40 \\ 
        Opus-4.6 & 88.46 & 92.29 \\ 
        \bottomrule 
    \end{tabular} 
    \label{tab:user_intent} 
\end{table}


\paragraph{Results.}
Table~\ref{tab:user_intent} presents the results.
\begin{itemize}
    \item \textbf{All models converge within ${\sim}$1\,pp} (BERTScore 88.5--89.6, cosine similarity 91.4--92.7), suggesting that single-sentence intent prediction is near ceiling at this annotation granularity.
    \item \textbf{High scores reflect semantic overlap, not exact alignment}: BERTScore and cosine similarity reward topical proximity, so a prediction that captures the right domain (e.g.\ ``food promotion'') scores well even if it misidentifies the specific goal (e.g.\ ``brand awareness'' vs.\ ``limited-time offer'').
    \item \textbf{Intent is inherently underdetermined from a single image}: multiple plausible intents can explain the same design, which compresses the score distribution and makes it difficult to distinguish genuinely superior reasoning from surface-level pattern matching.
    \item \textbf{Richer intent specifications}: multi-sentence descriptions decomposing audience, platform, and emotional response would likely be needed to reveal meaningful differentiation.
\end{itemize}
\FloatBarrier

\subsection{Template Variant Understanding}
\label{sec:template-variant-understanding}

\begin{figure*}[t]
    \centering
    \includegraphics[width=\textwidth]{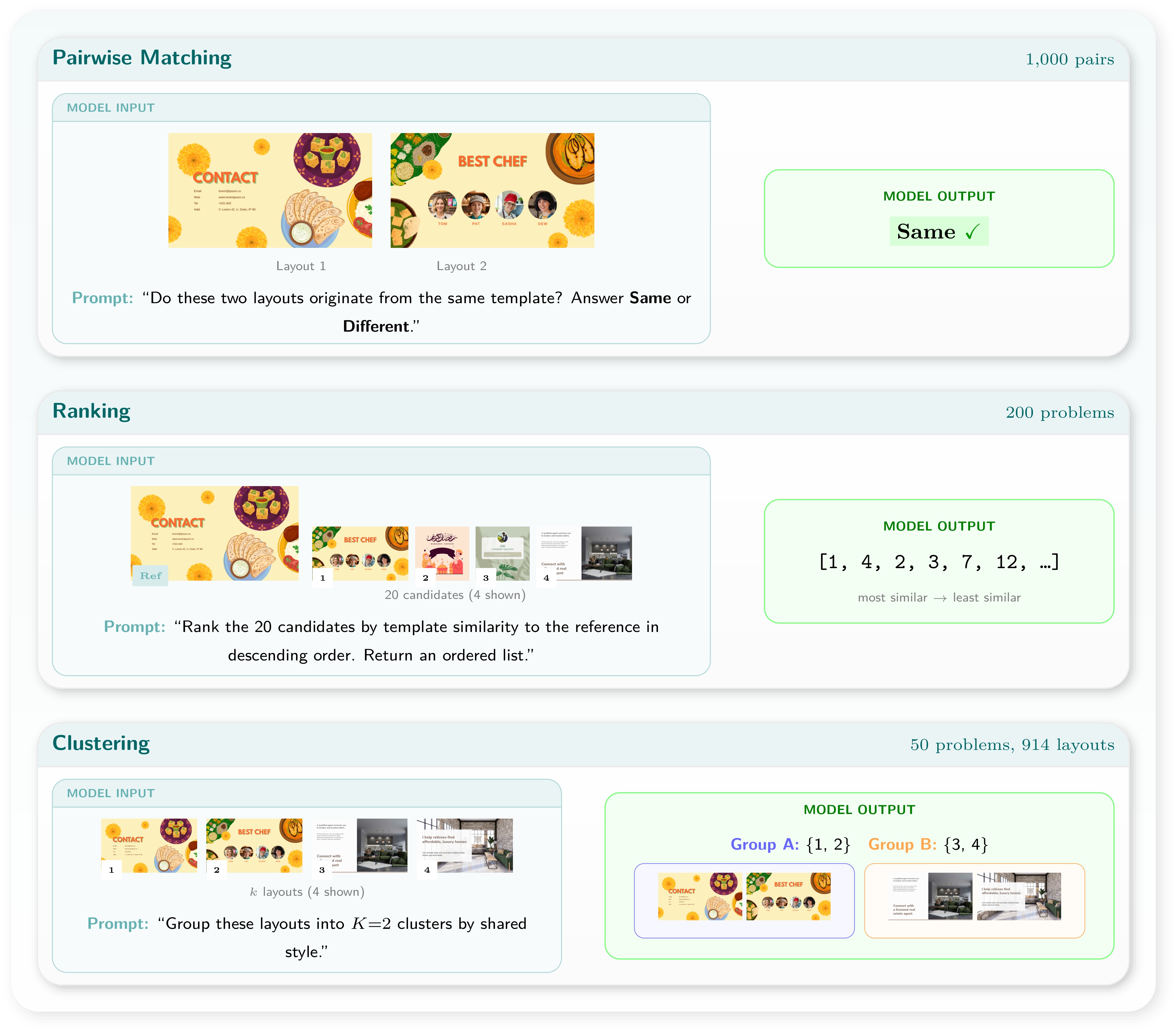}
    \caption{Template variant understanding tasks, see Appendix~\ref{app:prompts} for the exact input prompts. All tasks receive rendered layout screenshots and a text prompt.
\textbf{Pairwise Matching}: binary classification of template identity.
\textbf{Ranking}: rank 20 candidates by similarity to a reference.
\textbf{Clustering}: partition $k$ layouts into template-identity groups.}
    \label{fig:template-und-tasks}
\end{figure*}

Template variant understanding evaluates whether models can recognize that multiple designs originate from the same underlying template despite differing content. We define three progressively harder tasks: pairwise matching, ranking, and clustering, on sibling layouts from the dataset. Figure~\ref{fig:template-und-tasks} illustrates the task structure.

\begin{figure*}[t]
    \centering
    \includegraphics[width=\textwidth]{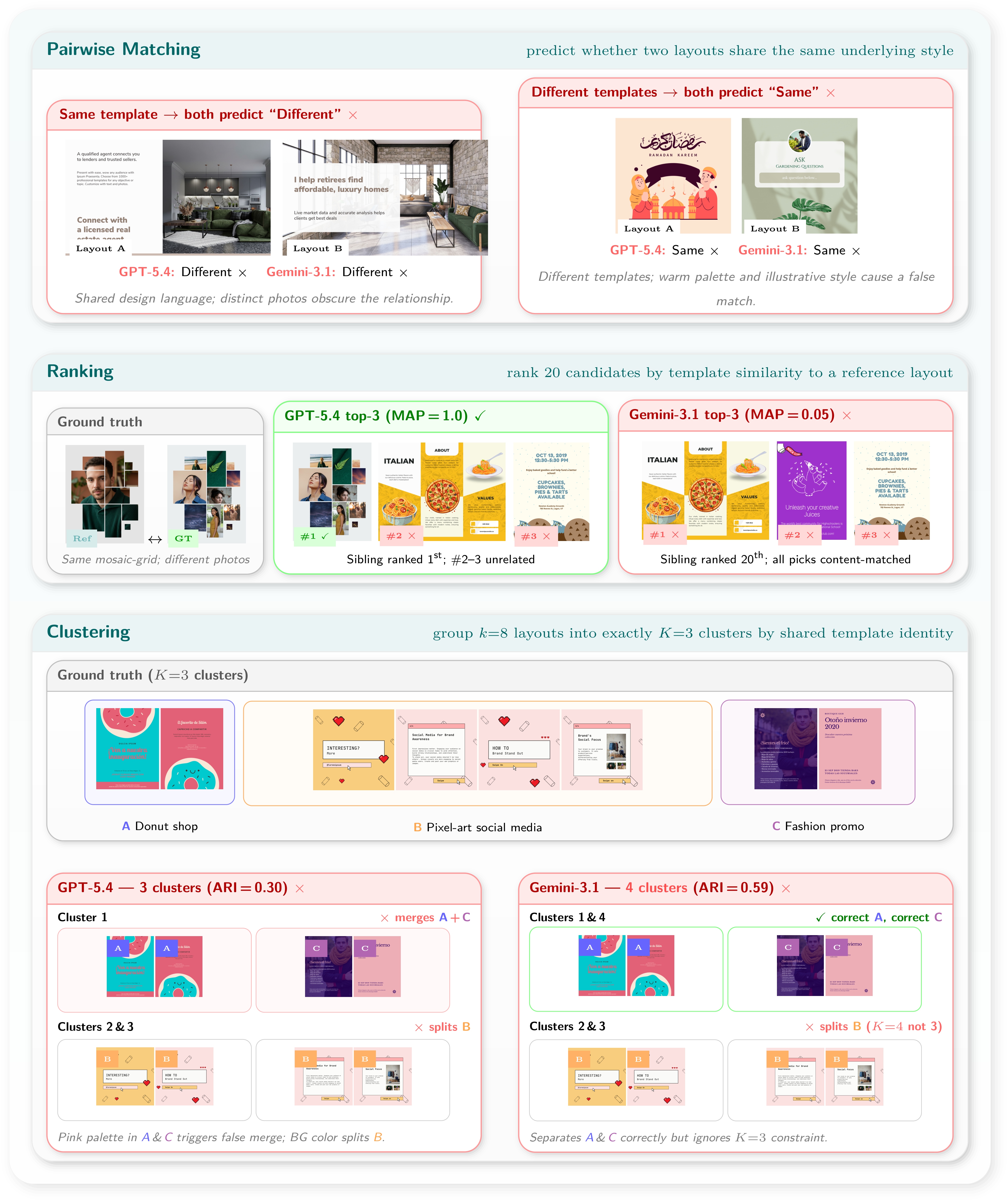}
    \caption{Template variant understanding failure cases.
    \textbf{Pairwise matching:} both models misclassify a same-template pair as ``different'' (distinct photos) and a different-template pair as ``same'' (shared warm palette).
    \textbf{Ranking:} GPT-5.4 identifies the correct mosaic-grid sibling despite different content; Gemini-3.1 ranks by surface content and places the sibling 20\textsuperscript{th}.
    \textbf{Clustering:} colored tags show ground-truth membership.
    GPT-5.4 merges \textcolor{blue!60}{\bfseries A}+\textcolor{violet!60}{\bfseries C} (similar pink palettes);
    Gemini-3.1 separates them but violates the $K{=}3$ constraint, producing 4 clusters by splitting \textcolor{orange!70}{\bfseries B} on background color.}
    \label{fig:template-understanding-failures}
\end{figure*}

\paragraph{Results.}
Table~\ref{tab:template-understanding} presents LLM and non-LLM baseline results (CLIP, DINOv2, LPIPS, color EMD, font Jaccard, and structural features) across all three tasks. Figure~\ref{fig:template-understanding-failures} illustrates failure modes across all three tasks: models confuse shared design language with content similarity in matching, rank by surface appearance rather than structural identity in ranking, and either merge visually similar clusters or violate cardinality constraints in clustering.

\begin{table*}[h]
    \centering
    \caption{Template variant understanding results (all metrics in \%), across
    pairwise Matching ($n{=}1{,}000$ pairs),
    ranking ($n{=}200$ queries),
    and clustering ($n{=}50$ problems, 914 layouts; ARI and AMI are chance-corrected).
    Best per metric in \textbf{bold}.
    Gemini-3.1$^\dagger$: best-performing variant per task.}
    \label{tab:template-understanding}
    \small
    \setlength{\tabcolsep}{3.5pt}
    \begin{tabular}{@{}l ccc cccc cccc@{}}
        \toprule
        & \multicolumn{3}{c}{\textbf{Matching}}
        & \multicolumn{4}{c}{\textbf{Ranking}}
        & \multicolumn{4}{c}{\textbf{Clustering}} \\
        \cmidrule(lr){2-4} \cmidrule(lr){5-8} \cmidrule(lr){9-12}
        \textbf{Model / Baseline}
          & Acc$\uparrow$ & F1$\uparrow$ & AUC$\uparrow$
          & MRR$\uparrow$ & MAP$\uparrow$ & nDCG@10$\uparrow$ & Rec@10$\uparrow$
          & ARI$\uparrow$ & AMI$\uparrow$ & V-m.$\uparrow$ & FMI$\uparrow$ \\
        \midrule
        \multicolumn{12}{@{}l}{\emph{LLM models (text+image input)}} \\
        \midrule
        GPT-5.4
          & 62.60 & 40.26 & 62.60
          & \textbf{99.35} & \textbf{99.84} & \textbf{99.55} & \textbf{89.39}
          & 85.94 & 89.40 & 94.86 & 88.44 \\
        Gemini-3.1$^\dagger$
          & 96.70 & 96.59 & 96.70
          & 97.00 & 89.00 & 93.51 & 82.40
          & \textbf{93.69} & \textbf{94.64} & 96.40 & 90.37 \\
        \midrule
        \multicolumn{12}{@{}l}{\emph{Non-LLM baselines}} \\
        \midrule
        Font Jaccard
          & \textbf{97.7} & \textbf{97.7} & \textbf{98.8}
          & 96.3 & 95.6 & 96.1 & 87.2
          & 92.8 & 94.6 & \textbf{97.0} & \textbf{94.0} \\
        LPIPS
          & 78.2 & 76.5 & 84.2
          & 92.6 & 85.5 & 88.5 & 83.3
          & 52.2 & 60.9 & 76.6 & 63.2 \\
        DINOv2
          & 71.9 & 72.0 & 78.6
          & 89.4 & 80.9 & 83.4 & 77.7
          & 36.3 & 47.9 & 68.0 & 52.2 \\
        color EMD
          & 76.7 & 77.4 & 82.8
          & 89.4 & 82.6 & 85.1 & 80.1
          & 44.5 & 53.9 & 72.2 & 58.1 \\
        Elem.\ count
          & 74.1 & 73.2 & 81.4
          & 83.3 & 76.9 & 79.6 & 76.5
          & 33.5 & 42.5 & 65.3 & 48.9 \\
        CLIP cosine
          & 50.0 & 66.7 & 50.0
          & 54.7 & 46.2 & 46.8 & 49.3
          & $-$1.7 & $-$1.7 & 34.8 & 28.7 \\
        \bottomrule
    \end{tabular}
\end{table*}


\begin{itemize}
    \item \textbf{Font Jaccard baseline matches or exceeds both LLMs} on all three tasks (97.7\% matching, 96.3\% MRR, 92.8\% ARI), indicating that font-family metadata captures most of the template identity signal in this corpus — yet LLMs, despite having access to both image and text inputs, fail to leverage this signal, suggesting they do not reliably attend to typographic consistency as a grouping cue.
\item \textbf{Matching and clustering are dissociated from ranking}: a model can achieve near-perfect ranking (MRR 99.4\%) while failing at binary match decisions (62.6\% accuracy), suggesting these sub-tasks probe fundamentally different similarity reasoning abilities.
\item \textbf{Binary match decisions are subject to strong prediction bias}: one model predicts ``different'' for 874 of 1{,}000 pairs (precision 1.0, recall 0.25), revealing that pairwise classification collapses under conservative decision thresholds even when relative ordering is well-calibrated.
\end{itemize}


\FloatBarrier

\subsection{Template Variant Generation}
\label{sec:template-variant-gen}

\begin{figure*}[htbp]
    \centering
    \includegraphics[width=\textwidth]{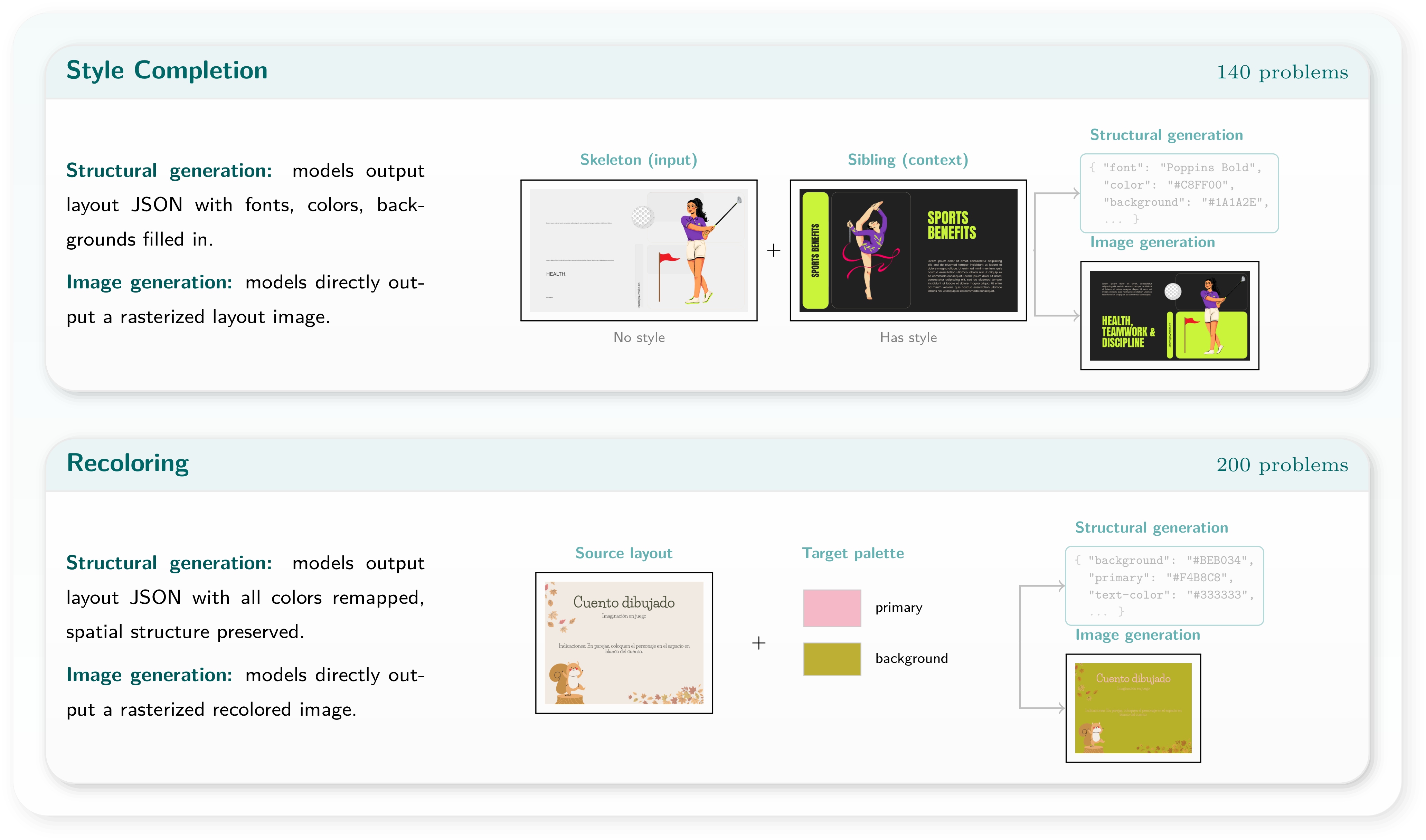}
    \caption{Template variant generation tasks, and see Appendix~\ref{app:prompts} for the exact input prompts. In \emph{structural generation}, the model receives reference layout JSON file(s) and their rendered image(s) and shall generate a complete layout specification in JSON format. In \emph{image generation}, the model receives rendered reference image(s) and directly produces a rasterized layout.
    \textbf{Style Completion:} infer style properties from styled siblings from the same template.
    \textbf{Recoloring:} recolor to match a target palette from a different template family (shown as color swatches). 
    }
    \label{fig:template-gen-task-overview}
\end{figure*}

The generation tasks evaluate whether models can produce design layouts that adhere to template conventions. We consider two output modalities. In \emph{structural generation}, the model receives reference layout JSON file(s) and their rendered image(s) from LICA (Section~\ref{sec:overview}) and shall generate a complete layout specification in JSON format, enabling explicit control over components and attributes. In \emph{image generation}, the model instead receives rendered reference image(s) and directly produces a rasterized layout.
Metrics span three tiers: JSON-level validity and adherence (Tier~1), aesthetic quality (Tier~2), and rendered-image similarity (Tier~3).  Figure~\ref{fig:template-gen-task-overview} illustrates the two structural-mode tasks across these modalities.

\begin{figure*}[t]
    \centering
    \includegraphics[width=\textwidth]{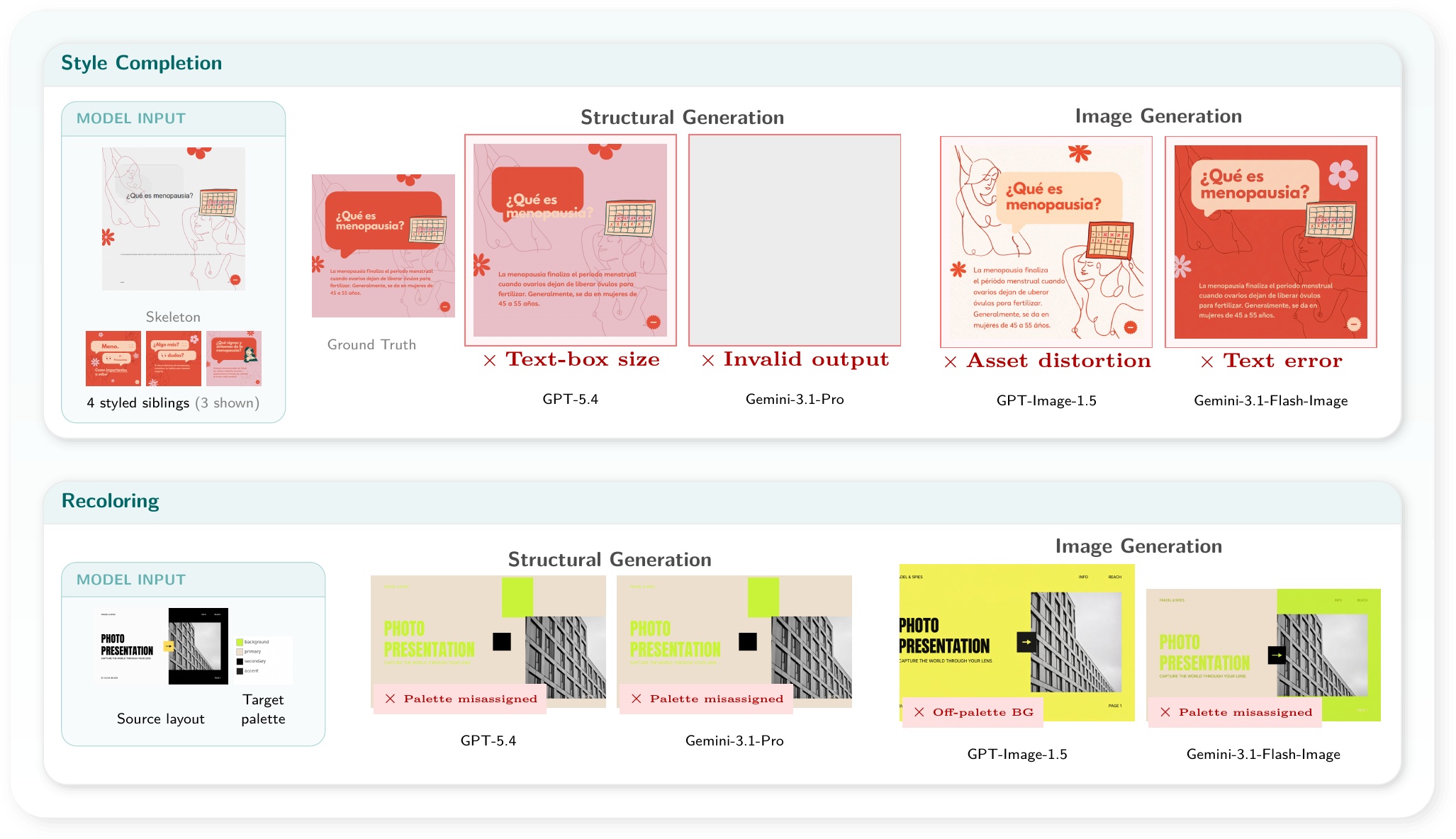}
    \caption{Template generation failure cases, and see Appendix~\ref{app:prompts} for the exact input prompts.
    \textbf{Style Completion}: GPT-5.4 fails to infer the proper text box size, while Gemini-3.1-Pro fails to generate valid JSON file;both image generation models fail to be text or content faithful.
    \textbf{Recoloring}: All models fail to correctly assign the colors based on the palette, which shows that models lack the understanding of color palette in design. The size differences reflect fixed output resolutions of image generation APIs.}
    \label{fig:template-gen-qualitative}
\end{figure*}

\paragraph{Results.}
Tables~\ref{tab:template4}--\ref{tab:template-image-metrics} present results.
Figure~\ref{fig:template-gen-qualitative} shows qualitative examples.

\begin{figure}[h]
    \centering
    \includegraphics[width=0.9\textwidth]{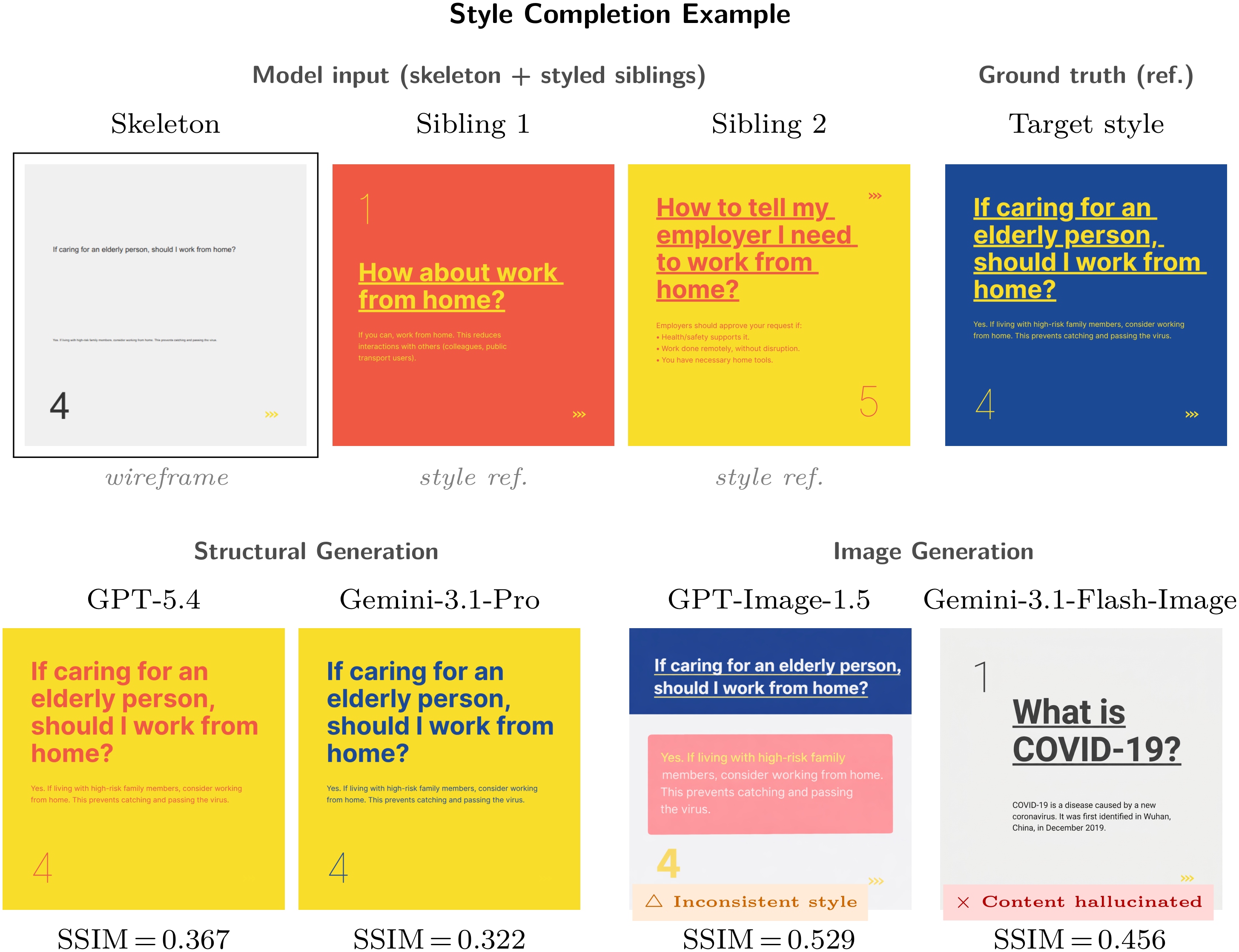}
    \caption{\textbf{Failure of image similarity metrics to reflect stylistic validity.} See Appendix~\ref{app:prompts} for the exact input prompts. In this style completion example, both structural generation predictions are visually coherent and preserve the sibling style language, but they assign a different palette from the reference. Although the outputs are style consistent, pixel-level similarity remains low (SSIM 0.367 and 0.322) in contrast to image generation results where GPT-Image-1.5 (0.529) preserves the content but generates inconsistent style and Gemini-3.1-Flash-Image (0.456) hallucinates the content, showing that image metrics can undervalue stylistically plausible alternatives. See Appendix~\ref{app:prompts} for the exact input prompts.}
    \label{fig:template-metric-limitation}
\end{figure}

\begin{itemize}
    \item \textbf{Validity is the primary differentiator.} GPT-5.4 achieves at least 94\% JSON validity across all tasks; Gemini-3.1-Pro achieves at most 83\%, limiting the sample on which its style metrics are computed.
    \item \textbf{Pixel-level metrics are insufficient for transformation tasks.} Figure~\ref{fig:template-metric-limitation} shows a style completion example where both structural generation predictions receive low SSIM (0.367 and 0.322) despite remaining stylistically plausible, indicating that image similarity metrics can penalize palette role swaps and do not capture style consistency; 
\end{itemize}

\begin{table}[htbp]
    \centering
    \caption{Style Completion ($n{=}140$).
    \textbf{Font}: per-component font-family match.
    \textbf{color}: per-component color match.
    \textbf{FS}: font-size MAE (px, $\downarrow$).
    \textbf{Op.}: opacity MAE ($\downarrow$).
    \textbf{BG}: background $\Delta E$ ($\downarrow$).}
    \label{tab:template4}
    \small
    \setlength{\tabcolsep}{3.5pt}
    \begin{tabular}{@{}l c cc cc c@{}}
        \toprule
        \textbf{Model}
          & Valid$\uparrow$ & Font$\uparrow$ & color$\uparrow$
          & FS$\downarrow$ & Op.$\downarrow$ & BG$\downarrow$ \\
        \midrule
        GPT-5.4
          & \textbf{0.943} & 0.961 & 0.771
          & \textbf{1.83} & \textbf{0.010} & 10.56 \\
        Gemini-3.1-Pro
          & 0.550 & \textbf{0.968} & \textbf{0.782}
          & 2.47 & 0.023 & \textbf{8.42} \\
        \bottomrule
    \end{tabular}
\end{table}

\begin{table}[htbp]
    \centering
    \caption{Recoloring ($n{=}200$).
    \textbf{Pos.}: position fidelity.
    \textbf{Area}: area fidelity.
    \textbf{Pal.}: palette adherence ($\Delta E{<}10$).
    \textbf{Cov.}: palette coverage.
    \textbf{BG}: background $\Delta E$ ($\downarrow$).}
    \label{tab:template5}
    \small
    \setlength{\tabcolsep}{3.5pt}
    \begin{tabular}{@{}l c cc cc c@{}}
        \toprule
        \textbf{Model}
          & Valid$\uparrow$ & Pos.$\uparrow$ & Area$\uparrow$
          & Pal.$\uparrow$ & Cov.$\uparrow$ & BG$\downarrow$ \\
        \midrule
        GPT-5.4
          & \textbf{0.965} & \textbf{1.000} & \textbf{1.000}
          & \textbf{0.946} & \textbf{0.776} & \textbf{14.48} \\
        Gemini-3.1-Pro
          & 0.635 & \textbf{1.000} & \textbf{1.00}
          & 0.929 & 0.690 & 17.76 \\
        \bottomrule
    \end{tabular}
\end{table}


\begin{table}[h]
    \centering
    \caption{Tier~2 aesthetic metrics (text+image input). All values in $[0,1]$.}
    \label{tab:template-tier2}
    \small
    \begin{tabular}{@{}ll ccc@{}}
        \toprule
        \textbf{Model} & \textbf{Task}
          & Harmony$\uparrow$ & Contrast$\uparrow$ & Hierarchy$\uparrow$ \\
        \midrule
        \multirow{2}{*}{GPT-5.4}
          & Style Completion & 0.888 & 0.703 & 1.000 \\
          & Recoloring & 0.904 & 0.573 & 1.000 \\
        \midrule
        \multirow{2}{*}{Gemini-3.1-Pro}
          & Style Completion & 0.848 & \textbf{0.768} & 1.000 \\
          & Recoloring & 0.877 & 0.562 & 1.000 \\
        \bottomrule
    \end{tabular}
\end{table}

\begin{table}[h]
    \centering
    \caption{Tier~3 image-level metrics.
    For Style Completion, reference is the ground-truth styled layout. For Recoloring, reference is the source layout before transformation (structural preservation).
    \emph{Structural}: model outputs JSON, rendered for comparison. \emph{Image}: model directly generates a rasterized layout image.$N$: valid, renderable predictions.}
    \label{tab:template-image-metrics}
    \small
    \setlength{\tabcolsep}{4pt}
    \begin{tabular}{@{}ll r ccc@{}}
        \toprule
        \textbf{Model} & \textbf{Task} & $N$
          & SSIM$\uparrow$ & PSNR$\uparrow$ & LPIPS$\downarrow$ \\
        \midrule
        \multicolumn{6}{@{}l}{\emph{Style Completion (ref.\,= ground-truth styled layout)}} \\
        \midrule
        \multicolumn{6}{@{}l}{\quad\textit{Structural generation}} \\
        GPT-5.4       &  & 132 & 0.796 & 16.64 & 0.238 \\
        Gemini-3.1-Pro&  & 77  & \textbf{0.810} & \textbf{17.98} & \textbf{0.210} \\
        \multicolumn{6}{@{}l}{\quad\textit{Image generation}} \\
        GPT-Image-1.5       &  & 140 & 0.617 & 10.62 & 0.579 \\
        Gemini-3.1-Flash-Img&  & 138 & \textbf{0.649} & \textbf{11.35} & \textbf{0.488} \\
        \midrule
        \multicolumn{6}{@{}l}{\emph{Recoloring (ref.\,= source layout, structural preservation)}} \\
        \midrule
        \multicolumn{6}{@{}l}{\quad\textit{Structural generation}} \\
        GPT-5.4       &  & 193 & 0.644 & 10.34 & 0.381 \\
        Gemini-3.1-Pro&  & 127 & \textbf{0.652} & \textbf{10.98} & \textbf{0.357} \\
        \multicolumn{6}{@{}l}{\quad\textit{Image generation}} \\
        GPT-Image-1.5       &  & 200 & 0.519 & 8.06 & 0.575 \\
        Gemini-3.1-Flash-Img&  & 193 & \textbf{0.662} & \textbf{10.23} & \textbf{0.340} \\
        \bottomrule
    \end{tabular}
\end{table}

\begin{table}[h]
\centering
\small
\setlength{\tabcolsep}{5pt}
\caption{Summary of key findings across template \& design semantics tasks.}
\begin{tabular}{>{\raggedright\arraybackslash}p{2.8cm}>{\raggedright\arraybackslash}p{3.5cm}>{\raggedright\arraybackslash}p{2.5cm}>{\raggedright\arraybackslash}p{2cm}>{\raggedright\arraybackslash}p{2cm}}
\toprule
\textbf{Task} & \textbf{Key Finding} & \textbf{Best Performance} & \textbf{Best Model} & \textbf{Status} \\
\midrule
Category Classification & Label constraining yields +30--62\,pp gains; open-vocab difficulty is mostly label aliasing, not perceptual failure & 78.7\% Top-1 (constrained) & Claude-Opus-4.6 & Unsolved \\
\midrule
User Intent Prediction & All models converge within ${\sim}$1\,pp; near ceiling at current annotation granularity & BERTScore 89.55, Cos.\,Sim.\ 92.71 & Gemini Flash Lite & Partially solved \\
\midrule
Pairwise Matching & Font Jaccard baseline matches or exceeds LLMs; simple font metadata suffices for binary identity decisions & 97.7\% acc (baseline), 96.7\% (LLM) & Font Jaccard / Gemini & Mostly solved \\
\midrule
Ranking & GPT achieves near-perfect MRR despite binary-decision bias; ranking by similarity is tractable & MRR 99.4\%, MAP 99.8\% & GPT-5.4 & Mostly solved \\
\midrule
Clustering & Gemini leads but non-LLM baselines remain competitive; cardinality violations are common & ARI 93.7\%, AMI 94.6\% & Gemini-3.1 & Partially solved \\
\midrule
Style Completion & Validity is the primary differentiator (94.3\% GPT vs.\ 55\% Gemini); font match is high when output is valid & 96.1\% font match, 94.3\% valid & GPT-5.4 & Partially solved \\
\midrule
Recoloring & Structural fidelity is high (position and area ${\sim}$100\%) but palette coverage lags at 77.6\% & 96.5\% valid, 94.6\% palette & GPT-5.4 & Partially solved \\
\bottomrule
\end{tabular}
\label{tab:semantics_findings}
\end{table}

\clearpage
 
\section{Animation \& Temporal Tasks}
\label{sec:animation}
\FloatBarrier

This section evaluates models on animation understanding and generation tasks: temporal ordering, motion classification, property extraction, and animated video synthesis. Table~\ref{tab:animation_tasks} lists the task definitions and dataset statistics for this domain. All understanding tasks draw from 100 animated compositions. Gemini receives the full video as input; GPT and Claude receive uniformly sampled keyframe sequences.

\begin{table}[h]
\centering
\small
\setlength{\tabcolsep}{5pt}
\caption{Animation \& temporal task definitions and dataset statistics.}
\begin{tabular}{>{\raggedright\arraybackslash}p{2.5cm}>{\raggedright\arraybackslash}p{5cm}>{\raggedright\arraybackslash}p{1.2cm}>{\raggedright\arraybackslash}p{2cm}>{\raggedright\arraybackslash}p{2.5cm}}
\toprule
\textbf{Task} & \textbf{Description} & \textbf{Samples} & \textbf{Classes} & \textbf{Metrics} \\
\midrule
Keyframe Ordering & Recover the chronological sequence of four shuffled keyframes from an animated composition. & 100 & Permutation of 4 keyframes & Exact Match, Kendall's $\tau$, Pairwise Acc \\
\midrule
Motion Type Classification & Identify the entrance animation effect (from 32 canonical types) applied to each element. Tested under open-vocab and constrained prompting. & 100 & 32 motion classes & Component Acc (all/single/multi), Count MAE \\
\midrule
Animation Property Extraction & Three sub-tasks: video-level duration, component-level duration, and component-level start-time prediction. & 100 & Continuous (seconds) & MAE, tolerance rates, Count MAE \\
\midrule
Animation Parameter Generation & Produce correct animations given a static layout and explicit per-component animation specifications. & 10 & Animated video & Motion Acc, Duration MAE, Direction Acc \\
\midrule
Motion Trajectory Generation & Synthesize a video depicting a specified motion primitive given a static layout image and component metadata. & 10 & Animated video & Motion Acc, LPIPS, SSIM \\
\midrule
Short-Form Video Layout & Produce a complete animated marketing video from a text brief alone, without visual input. & 10 & Animated video & Human eval (4 criteria) \\
\bottomrule
\end{tabular}
\label{tab:animation_tasks}
\end{table}

\subsection{Temporal Understanding: Keyframe Ordering}
\label{sec:keyframe-ordering}

Keyframe ordering evaluates temporal reasoning over design animations: given four shuffled keyframes from each of 100 videos, the model must recover the correct chronological sequence. Understanding temporal order is fundamental to animation comprehension, a model that cannot infer narrative or motion progression from visual snapshots cannot be a reliable collaborator in animated design workflows. We report exact match, Kendall's~$\tau$, pairwise accuracy, and first-frame accuracy (random baseline: 4.2\%, 0.0, 50.0\%, 25.0\%).


\paragraph{Results.}
Table~\ref{tab:keyframe-ordering} presents the results.

\begin{table}[h]
    \centering
    \caption{Keyframe ordering: models are given four shuffled keyframes from each of 100 animated compositions and must predict the correct chronological sequence.}
    \begin{tabular}{l|cccc}
        \toprule
        \textbf{Model}
            & \textbf{Exact Match $\uparrow$}
            & \textbf{Kendall's $\tau$ $\uparrow$}
            & \textbf{Pairwise Acc.\ $\uparrow$}
            & \textbf{First-Frame Acc.\ $\uparrow$} \\
        \midrule
        \textit{Random baseline}
            & \textit{4.2\%} & \textit{0.000} & \textit{50.0\%} & \textit{25.0\%} \\
        \midrule
        Gemini-3.1-Pro   & 14.00 & 17.66 & 58.83 & 67.00 \\
        GPT-5.4          & \textbf{16.00} & \textbf{22.33} & \textbf{61.16} & \textbf{80.00} \\
        Claude-Opus-4.6   & 15.00 & 17.99 & 59.00 & 77.00 \\
        \bottomrule
    \end{tabular}
    \label{tab:keyframe-ordering}
\end{table}

\begin{itemize}
    \item \textbf{All models beat the random baseline but remain weak}: exact match ranges from 14--16\%, with pairwise accuracy only ${\sim}$10\,pp above chance (58.8--61.2\%), indicating this as an unsolved task. 
    \item \textbf{First-frame identification is substantially easier than full ordering} (67--80\% vs.\ 14--16\% exact match), suggesting models detect the initial state but struggle with fine-grained temporal progression.
\end{itemize}

\subsection{Animation Understanding: Motion Type Classification}
\label{sec:motion-type}

\begin{figure}[htbp]
    \centering
    \includegraphics[width=0.6\textwidth]{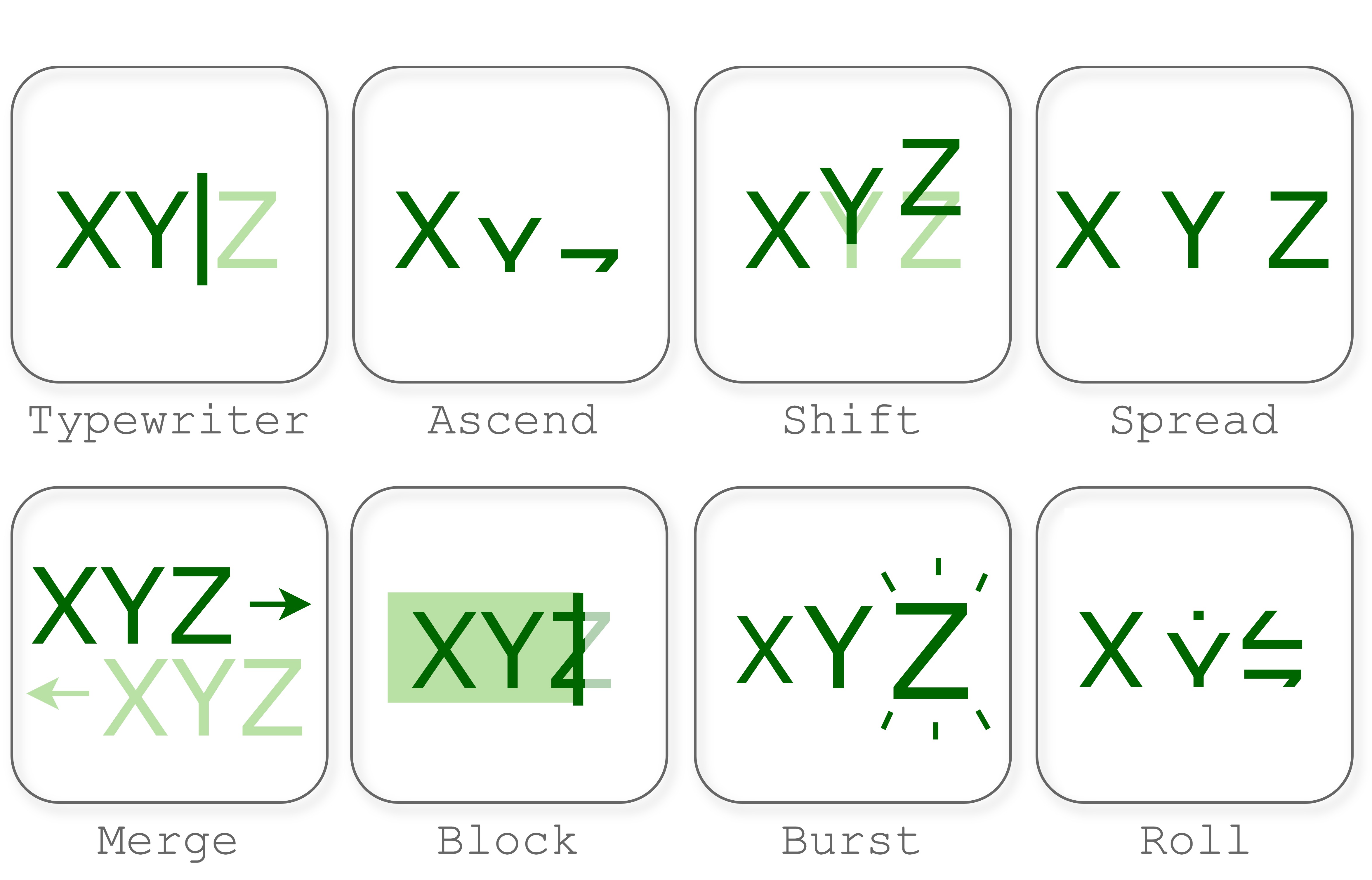}
    \caption{Example motion types. A full list of the 32 motion types are given in the Appendix Table~\ref{tab:prompts-temporal-und}).}
    \label{fig:motion-types}
\end{figure}

Motion type classification evaluates whether models can identify the entrance animation effect from 32 canonical types, e.g.\ \texttt{rise}, \texttt{fade}, \texttt{pop}, \texttt{tumble}) applied to individual elements in 100 LICA animated compositions (Section~\ref{para:dataset}). Compositions contain 1--20 animated elements; we test under open-vocabulary and label-constrained prompting. Gemini receives full video; GPT and Claude receive keyframe sequences. See Figure~\ref{fig:motion-types} as an example for top motion types in the LICA dataset. 

\paragraph{Results.}
Table~\ref{tab:motion_type} presents the results. We omit qualitative examples for this task as model outputs are too inaccurate to yield meaningful visual analysis. Key findings show:
\begin{itemize}
    \item \textbf{Motion type classification is largely unsolved}: all models achieve below 13\% accuracy, with most scoring 0\% on single-component scenes.
    \item \textbf{Single-component accuracy is near-zero}: five of six model/prompt combinations score 0\% on single-component scenes (the exception being Claude constrained at 13.3\%). Because these scenes contain only a single animated element and require no decomposition, scene segmentation alone cannot explain the failure. This points to a more fundamental inability to map perceived motion patterns. This is likely because UI motion-graphics primitives (e.g.\ \texttt{tumble}, \texttt{pop}, \texttt{stomp}) are underrepresented in pretraining data relative to natural-video motion, leaving models without a reliable basis for distinguishing them.
    \item \textbf{Constrained prompting does not reliably help}: providing the 32-label vocabulary improves Claude marginally on single-component scenes but degrades Gemini substantially, suggesting that models are not close to solving the task and simply guessing differently when the label set is supplied.
\end{itemize}

\begin{table}[h]
\centering
\caption{Motion type classification on each animated element. Results are stratified by scene complexity: Single (1 animated element) vs. Multi ($\geq$2). Gemini receives the full animation video; GPT and Claude receive uniformly sampled keyframes.}
\begin{tabular}{c|c|c|ccc|c}
\toprule
\textbf{Model} & \textbf{Input} & \textbf{Prompt}
  & Acc.\,(All) $\uparrow$ & Acc.\,(Single) $\uparrow$ & Acc.\,(Multi) $\uparrow$
  & Count MAE $\downarrow$ \\
\midrule\midrule
\multirow{2}{*}{Gemini-3.1} & \multirow{2}{*}{Video}
  & open        & 7.12  & 0.00  & 8.37  & \textbf{3.60} \\
  & & constrained & 3.00  & 0.00  & 3.52 & 5.79 \\
\midrule
\multirow{2}{*}{GPT-5.4} & \multirow{2}{*}{Keyframes}
  & open        & 8.95 & 0.00 & 10.53 & 3.90 \\
  & & constrained & 8.72 & 0.00 & 10.26 & 4.10 \\
\midrule
\multirow{2}{*}{Claude-Opus-4.6} & \multirow{2}{*}{Keyframes}
  & open          & 10.60 & 0.00  & 12.47 & 3.92 \\
  & & constrained & 10.67 & \textbf{13.33} & 10.20 & 4.22 \\
\bottomrule
\end{tabular}
\label{tab:motion_type}
\end{table}

\subsection{Animation Understanding: Animation Property Extraction}
\label{sec:animation-properties}

Animation property extraction evaluates a model's ability to perceive and quantify temporal characteristics of animated design compositions across three sub-tasks of increasing difficulty on 100 LICA compositions (Section~\ref{para:dataset}): video-level duration prediction, component-level duration prediction, and component-level start-time prediction. Here, a component refers to an individual design element in the layout (e.g., text/image) that is animated independently. The latter two tasks therefore require identifying and tracking each element over time (implicit scene decomposition) as well as fine-grained temporal estimation. The latter two require implicit scene segmentation and fine-grained temporal estimation. Gemini receives full video; GPT and Claude receive keyframe sequences.

\begin{figure}[h]
    \centering
    \includegraphics[width=\textwidth]{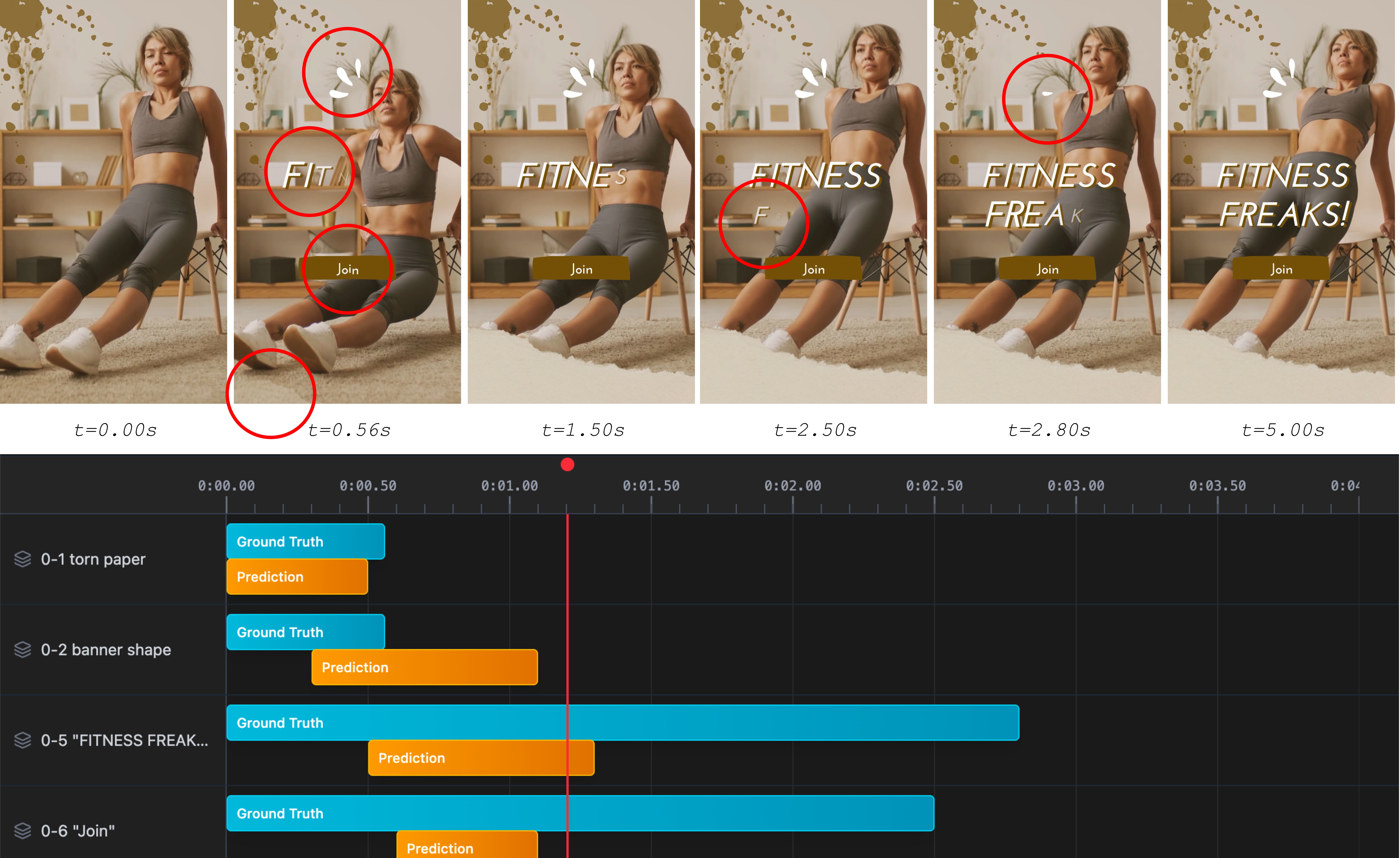}
    \caption{Ground-truth versus predicted entrance windows on one sample. Each bar spans from the component's entrance start time to the end of its entrance animation (start + duration). Claude-Opus-4.6 correctly identifies the two short rise animations (0-1, 0-2) within ~0.25s but substantially underestimates the longer burst and bounce durations (2.80 → 0.80s and 2.50 → 0.30s). The model also hallucinates staggered entrance times (0.0–0.8s offsets) when all four components enter simultaneously at t = 0. Duration MAE: 1.12s; start time MAE: 0.40s.}
    \label{fig:animations-parameters-extraction}
\end{figure}

\paragraph{Results.}
Tables~\ref{tab:video_duration} and~\ref{tab:component_start_time} present the full results, stratified by scene complexity (single-component vs.\ multi-component). All metrics are defined in Section~\ref{sec:metrics}. Figure~\ref{fig:animations-parameters-extraction} illustrates an example of the entrance window predictions on Claude Opus 4.6. 

\begin{table}[h]
\centering
\caption{Video-level and Component-level duration prediction, stratified by scene complexity. $\leq$1\ scores report the percentage of predictions within the given tolerance. MAE is in seconds. Count MAE measures the absolute error in predicting the number of animated components.}
\begin{tabular}{c|c|c|ccc|c}
\toprule
  & \textbf{Model} & \textbf{Input} & \multicolumn{4}{c}{}  \\
\midrule
\midrule
\multicolumn{3}{l|}{\textit{Video-level}} & MAE $\downarrow$ & $\leq$1\,s $\uparrow$ & $\leq$2\,s $\uparrow$ & \\
\midrule
 \multirow{4}{*}{All} 
  & Gemini 3.1 & Video     & 6.21  & \textbf{25.00} & \textbf{46.00} & \\
  & GPT-5.4    & Keyframes & 6.44  & 9.00  & 28.00 & \\
  & Claude-Opus-4.6   & Keyframes & 6.28  & 14.00 & 25.00 & \\
\midrule\midrule
\multicolumn{3}{l|}{\textit{Component-level}} & MAE $\downarrow$ & $\leq$0.1\,s $\uparrow$ & $\leq$0.25\,s $\uparrow$
  & Count MAE $\downarrow$ \\
\midrule
\multirow{4}{*}{All} 
& Gemini-3.1 & Video     & 0.62 & 32.72 & 40.88 & 5.98 \\
& GPT-5.4    & Keyframes & 0.56 & 26.67 & \textbf{59.82} & \textbf{3.58}   \\
& Claude-Opus-4.6   & Keyframes & \textbf{0.51} & \textbf{43.29} & 58.46 & 3.91   \\
\midrule
\multirow{4}{*}{Single-component} 
& Gemini-3.1 & Video     & 0.25 & 46.66 & 53.33 & 3.40 \\
& GPT-5.4    & Keyframes & 0.14 & 46.66 & \textbf{86.66} & 1.20 \\
& Claude-Opus-4.6   & Keyframes & \textbf{0.12} & \textbf{73.33} & 80.00 & 3.91 \\
\midrule
\multirow{4}{*}{Multi-component} 
& Gemini-3.1 & Video     & 0.68 & 30.26 & 38.68 & 6.43 \\
& GPT-5.4    & Keyframes & 0.63 & 23.14 & \textbf{55.08} & \textbf{4.00} \\
& Claude-Opus-4.6   & Keyframes & \textbf{0.58} & \textbf{37.99} & 54.65 & 4.40 \\
\bottomrule
\end{tabular}
\label{tab:video_duration}
\end{table}

\begin{table}[h]
\centering
\caption{Component-level start-time offset prediction, stratified by scene complexity. MAE is in seconds. The model predicts when each element's entrance animation begins relative to the start of the video. Count MAE measures the absolute error in predicting the number of animated components.}
\begin{tabular}{c|c|c|ccc|c}
\toprule
& \textbf{Model} & \textbf{Input}
  & MAE $\downarrow$ & $\leq$0.5\,s $\uparrow$ & $\leq$1.0\,s $\uparrow$
  & Count MAE $\downarrow$ \\
\midrule\midrule
\multirow{4}{*}{All}
& Gemini-3.1 & Video     & \textbf{1.41}  & 64.95 & \textbf{82.14} & 5.82 \\
& GPT-5.4    & Keyframes & 1.44  & 61.19 & 80.31 & 4.07 \\
& Claude-Opus-4.6   & Keyframes & 1.98  & 48.61 & 63.37 & \textbf{3.74} \\
\midrule
\multirow{4}{*}{Single-component}
& Gemini-3.1 & Video     & 0.15  & 93.33 & 93.33 & 4.2 \\
& GPT-5.4    & Keyframes & \textbf{0.00} & \textbf{100} & \textbf{100} & 2.93 \\
& Claude-Opus-4.6   & Keyframes & 0.03  & \textbf{100} & \textbf{100} & 2.26 \\
\midrule
\multirow{4}{*}{Multi-component}
& Gemini-3.1 & Video     & \textbf{1.63}  & 59.95 & \textbf{80.16} & 6.10 \\
& GPT-5.4    & Keyframes & 1.69  & 54.35  & 76.83 & 4.27 \\
& Claude-Opus-4.6   & Keyframes & 2.33  & 39.54  & 56.91 & \textbf{4.00} \\
\bottomrule
\end{tabular}
\label{tab:component_start_time}
\end{table}




\begin{itemize}
    \item \textbf{Video-level duration is poorly estimated across the board}: all models exhibit MAE $>$\,6\,s, and the majority of predictions are off by more than two seconds, demonstrating that models lack a reliable internal clock for overall composition length.
    \item \textbf{Scene decomposition is the primary bottleneck}: single-to-multi performance gap is large and consistent across all sub-tasks, and models with lower component Count MAE estimate timing more accurately.
\end{itemize}

\clearpage

\subsection{Animation Generation: Parameter Generation}
\label{sec:animation-parameter-generation}

Animation parameter generation evaluates whether video generation models can produce correct animations when given a static design layout and explicit per-component animation specifications. The model receives the last frame of the ground-truth video as a static layout reference along with a text prompt that enumerates, for each component, the animation type (e.g.\ fade, rise, pan, tumble, flicker, rotate), duration, speed, direction, and animate phase (entrance, continuous, or both). Components are identified only by an integer index and their element type (\textsc{image}, \textsc{group}, or \textsc{text}); the model must resolve which visual region each index refers to using the layout image alone.

\begin{figure}[htbp]
    \centering
    \includegraphics[width=\textwidth]{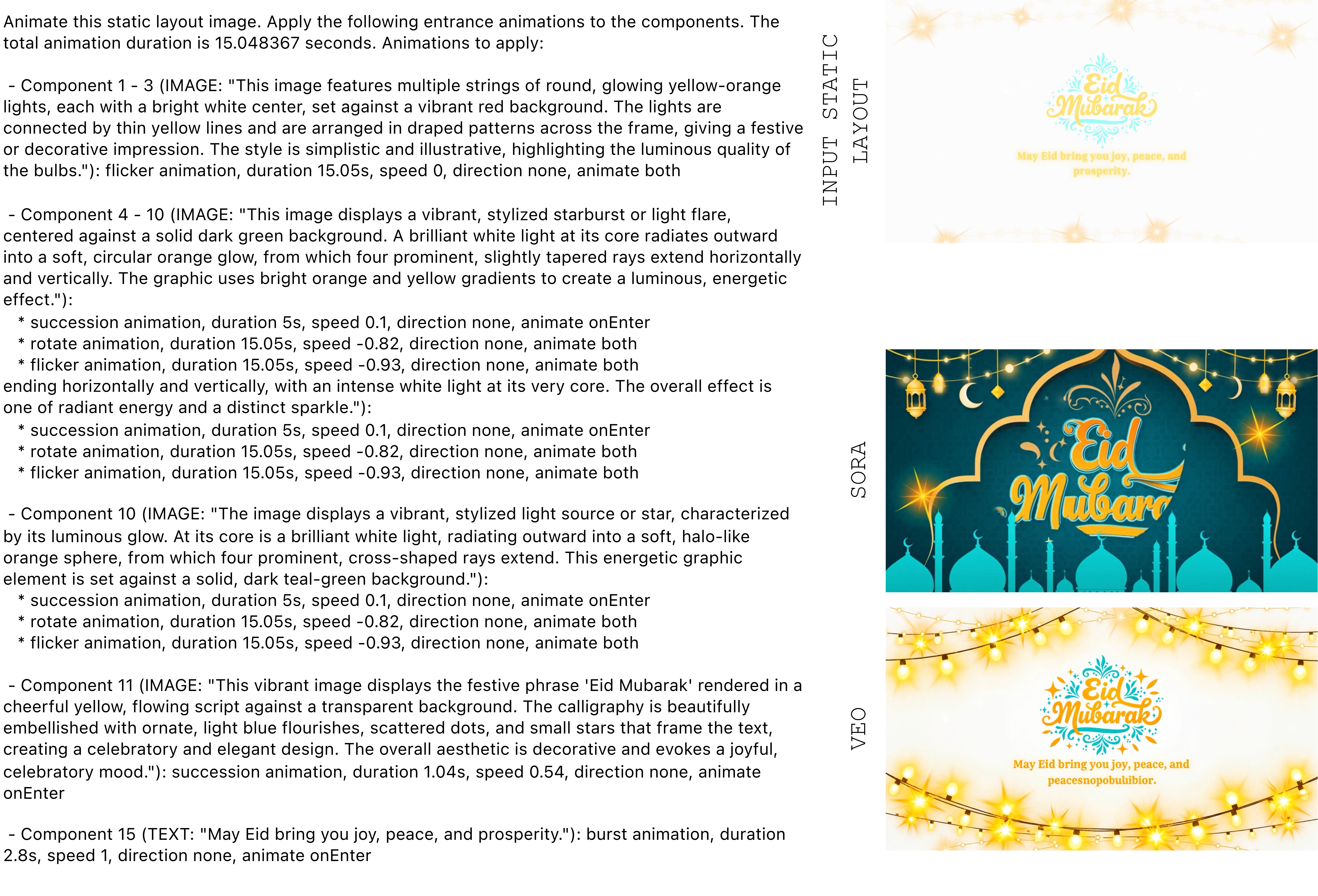}
    \caption{Example outputs from Sora and Veo given a static design layout with explicit per-component animation instructions. The static input is the final frame of the ground-truth video, which fades to white in this example. Sora more faithfully follows the prompt specifications (e.g., correctly matching the described primary color), while Veo adheres more strongly to the input image, preserving its appearance and color scheme. In this example, both models fail to generate the correct number of components for groups 1–3 and 4–10. While the motion direction is generally correct, the motion type is incorrect for all components.}
    \label{fig:anim-param-gen}
\end{figure}

\paragraph{Results.}
Qualitative results on 10 samples comparing Sora and Veo are provided in the supplementary HTML\footnote{\url{https://lica.world/video-generation-benchmarks}\label{fn:url}}. Perceptual similarity scores are given in Table~\ref{tab:anim-param-gen} and Figure~\ref{fig:anim-param-gen} illustrates an example of the prompt and input static layout used in this task. 
Key findings:

\begin{itemize}
    \item \textbf{Grounding is the core failure}: given only a static composite frame and a textual list of component indices, models have no reliable way to identify which image region corresponds to ``Component~$k$.'' They must simultaneously parse the layout, segment it, assign consistent indices, and apply distinct animation parameters to each segment. Precisely controlling attributes such as speed, easing, and magnitude proves substantially harder in this setting.
    \item \textbf{Outputs diverge systematically from specifications}: in practice, models either apply a single dominant motion globally, animate the wrong elements, or hallucinate motion unrelated to the prompt.
    \item \textbf{Full video input does not resolve the problem}: faithful animation parameter control would require richer conditioning, such as per-component image crops or masks, so that the model can unambiguously associate each instruction with its target region.
    \item \textbf{Automatic metrics are limited}: frame-level similarity measures such as FID and LPIPS are uninformative when the layout itself is distorted, and per-component motion metrics presuppose correct component isolation, precisely the capability that is lacking. We report perceptual similarity (LPIPS, SSIM, PSNR) between the input static layout and generated frames to verify that models preserve the original composition, but these scores do not capture whether the correct animations were applied to the correct components. We therefore supplement with qualitative results and leave fine-grained quantitative evaluation to future work with component-level human evaluation.
\end{itemize}

\begin{table*}[h]
\centering
\caption{Perceptual similarity measured between the input static layout and the first and last frames of the generated video. High similarity does not imply correct execution of the prompt: qualitative results show that both models fail to produce the correct motion types and component counts.}
\label{tab:anim-param-gen}
\small
\begin{tabular}{lll|c|c}
\toprule
\textbf{Evaluation Dimension} & \textbf{Method} & \textbf{Metric} & \textbf{Sora} & \textbf{Veo} \\
\midrule
\multirow{4}{*}{Perceptual Similarity}
  & Static $\rightarrow$ First Frame & SSIM $\uparrow$    & 48.51 & \textbf{77.24} \\
    & Static $\rightarrow$ Last Frame & SSIM $\uparrow$   & 44.05 & \textbf{46.48} \\
  & Static $\rightarrow$ First Frame & LPIPS $\downarrow$ & 60.81 & \textbf{20.12} \\
  & Static $\rightarrow$ Last Frame  & LPIPS $\downarrow$ & \textbf{47.86} & 50.35 \\
\bottomrule
\end{tabular}
\end{table*}

        \subsection{Animation Generation: Motion Trajectory Generation}
\label{sec:motion-trajectory-generation}

Motion trajectory generation evaluates whether video generation models can synthesize a specific motion primitive (e.g.\ \textit{wipe}, \textit{fade}, \textit{rise}) given a static layout image and component metadata. The model receives the final resting state of a design, a motion-type label from the 32 LICA primitives, and component specifications (index, type, direction, speed, duration), and must produce keyframes or a video depicting the target transition. We measure motion type accuracy, mean LPIPS (transition smoothness), LPIPS variance (motion evenness), and mean SSIM.

\paragraph{Results.} Qualitative results on 10 samples comparing Sora and Veo are provided in the supplementary
HTML\footref{fn:url}.

\begin{table}[h]
    \centering
    \caption{Human evaluated motion trajectory generation accuracy by motion type. Each sample tests whether the generated video depicts the requested motion primitive. Veo-2.0 and Sora-2 are evaluated on 10 samples spanning 7 of the 32 LICA motion primitives.}
    \begin{tabular}{c|c|c|c}
        \toprule
        \multirow{2}{*}{\textbf{Motion Type}} & \multirow{2}{*}{\textbf{No. Samples}} & \textbf{Veo-2.0} & \textbf{Sora-2} \\
         & & Motion Accuracy $\uparrow$ & Motion Accuracy $\uparrow$ \\
        \midrule
        \midrule
        flicker & 1 & 100.00 & 0.00 \\
        blur & 1 & 0.00 & 0.00 \\
        baseline & 1 & 100.00 & 0.00 \\
        rise & 3 & 66.67 & 33.33 \\
        tumble & 1 & 0.00 & 0.00 \\
        pan & 2 & 100.00 & 50.00 \\
        fade & 1 & 0.00 & 100.00 \\
        \midrule
        \textit{Aggregate} & \textit{10} & \textit{60\%} & \textit{30\%} \\
        \bottomrule
    \end{tabular}
    \label{tab:motion-trajectory-generation}
\end{table}

\begin{itemize}
    \item \textbf{Fine-grained parameter control remains elusive}: even when a model successfully reproduces a motion type, precisely controlling attributes such as speed, easing, and magnitude proves substantially harder, indicating that coarse motion generation does not imply fine-grained controllability.
    \item \textbf{Performance is highly motion-dependent}: both models succeed on \texttt{pan} but fail on \texttt{tumble} and \texttt{blur}, suggesting that spatial trajectory primitives are easier to reproduce than visual-effect primitives.
\end{itemize}
\subsection{Animation Generation: Short-Form Video Layout Generation}
\label{sec:short-form-video-gen}

Short-form video layout generation evaluates the model's ability to produce a complete animated marketing video from a text brief alone, without any visual input. The model shall autonomously design the layout, author text, select a color theme, and animate the composition in 9{:}16 format. We evaluate on 10 marketing briefs spanning diverse industries via human evaluation (binary pass/fail per metric, $N{=}10$).


\paragraph{Results.}
Table~\ref{tab:short-form-aggregate} presents the human evaluation results. Qualitative results on 10 samples comparing Sora and Veo are provided in the supplementary
HTML\footref{fn:url}.

\begin{table}[h]
\centering
\caption{Human evaluation results for short-form video layout generation. Each model receives only a text marketing brief and must generate a complete 9{:}16 video from scratch. Scores report average correctness percentages over $N{=}10$ samples. For Text Accuracy and Text Readability, we average the per-sample ratios of correct or readable text instances. Spatial Layout Accuracy is computed as the average proportion of correctly matched positioned elements, and Background Correctness is reported as pass rate.}
\label{tab:short-form-aggregate}
\begin{tabular}{l|c|c}
\toprule
\multirow{2}{*}{\textbf{Metric}} & \multicolumn{2}{c}{\textbf{Accuracy $\uparrow$}} \\
 & Veo-3.1 & Sora-2 \\
\midrule
\quad Text Accuracy           & 68.69 & \textbf{78.77} \\
\quad Text Readability        & 81.59 & \textbf{94.13} \\
\quad Spatial Layout Accuracy & 80.00 & 80.00 \\
\quad Background Correctness  & 80.00 & 80.00 \\
\bottomrule
\end{tabular}
\end{table}

\begin{itemize}
    \item \textbf{High-level accuracy falls short at finer granularity.} While video generation models can adhere to prompts at a high level, performance degrades substantially when evaluation is decomposed into finer-grained criteria such as component-level positioning or text quality.
    \item \textbf{Text quality is a critical differentiator.} Sora-2 renders accurate text in 79\% of samples, compared to 69\% for Veo-3.1. Even in terms of readability, neither model achieves perfect scores, limiting their usability.
\end{itemize}

\begin{figure}[htbp]
    \centering
    \includegraphics[width=\textwidth]{figures/animations_qualitative/short-form-video.jpg}
    \caption{Qualitative example of single-layout short-form video generation on Sora-2 and Veo-3.1. Each model receives only a text marketing brief describing a vertical fashion announcement and must design, lay out, and animate a complete 9:16 short-form video from scratch (no image input). The ground truth (top) shows components entering sequentially over ~4 seconds: brand name, headline with star accents, tagline box, product photo, and social handle, all on a textured marble background. Veo-3.1 (bottom left) reproduces the layout hierarchy and marble background but does not render the video in the requested 9:16 portrait, generates incorrect text, produces the wrong number of star accents flanking the headline, and exhibits visual style inconsistency by the third frame as the layout shifts to a different color scheme. Sora-2 (bottom right) correctly produces a 9:16 portrait video and legible text, but collapses the staggered reveal into a near-simultaneous appearance of all elements, losing the sequential entrance timing specified in the brief. Neither faithfully reproduce the fine-grained compositional structure: Veo-3.1 fails on aspect ratio and animation style, while Sora-2 fails on temporal sequencing and component-level entrance control.}
    \label{fig:short-form-video}
\end{figure}

While the present evaluation focuses on single-layout short-form videos, real-world marketing content frequently comprises multi-layout compositions in which distinct scenes or slides transition sequentially within a single video. Evaluating multi-layout generation introduces additional challenges beyond those observed here, including inter-layout coherence, consistent branding across scenes, transition quality, and correct allocation of content to the appropriate layout. We leave systematic evaluation of multi-layout short-form video generation to future work.

\begin{table}[h]
\centering
\small
\setlength{\tabcolsep}{5pt}
\caption{Summary of key findings across animation \& temporal tasks.}
\begin{tabular}{>{\raggedright\arraybackslash}p{2.8cm}>{\raggedright\arraybackslash}p{3.5cm}>{\raggedright\arraybackslash}p{2.5cm}>{\raggedright\arraybackslash}p{2cm}>{\raggedright\arraybackslash}p{2cm}}
\toprule
\textbf{Task} & \textbf{Key Finding} & \textbf{Best Performance} & \textbf{Best Model} & \textbf{Status} \\
\midrule
Keyframe Ordering & Models beat the random baseline but remain weak; first-frame identification (${\sim}$80\%) is much easier than full ordering (${\sim}$16\%) & 16\% exact match, 80\% first-frame & GPT-5.4 & Unsolved \\
\midrule
Motion Type Classification & All models below 13\% accuracy; full video input does not help over keyframes, suggesting scene decomposition is the bottleneck & 13.3\% single acc (constrained) & Claude-Opus-4.6 & Unsolved \\
\midrule
Video-level Duration & Coarsely estimable; Gemini benefits from full video input but all models show ${\sim}$6\,s MAE & $\leq$2\,s: 46\%, MAE\,=\,6.21\,s & Gemini-3.1 & Unsolved \\
\midrule
Component-level Duration & Single-component is partially tractable (MAE\,=\,0.12\,s) but multi-component degrades sharply (MAE\,=\,0.58\,s) & MAE\,=\,0.51\,s, $\leq$0.1\,s: 43.3\% & Claude-Opus-4.6 & Unsolved \\
\midrule
Component-level Start Time & Single-component is trivially solved (start\,=\,0); multi-component start-time estimation is the hardest temporal sub-task & $\leq$1.0\,s: 82.1\%, MAE\,=\,1.41\,s & Gemini-3.1 & Partially solved \\
\midrule
Animation Parameter Gen. & Grounding is the core failure; models cannot map component indices to spatial regions, leading to globally applied or hallucinated motion & Qualitative only & --- & Unsolved \\
\midrule
Motion Trajectory Gen. & Veo outperforms Sora on aggregate accuracy; spatial trajectory primitives are easier than visual-effect primitives & 60\% motion accuracy & Veo-2.0 & Unsolved \\
\midrule
Short-Form Video Layout & Both models follow briefs well; text legibility and aspect-ratio compliance are the key differentiators & 80\% text accuracy, 80\% spatial accuracy & Sora-2 & Partially solved \\
\bottomrule
\end{tabular}
\label{tab:animation_findings}
\end{table}

\clearpage


\section{Discussion}
\label{sec:discussion}

\subsection{Task Solvability Landscape}

Across the 49 tasks in GDB, a clear solvability hierarchy emerges. Table~\ref{tab:solvability} organizes all task groups into three tiers based on best-model performance and the nature of remaining gaps.

\begin{table}[h]
\centering
\small
\setlength{\tabcolsep}{4pt}
\caption{Task solvability tiers across GDB. Count refers to the number of individual tasks classified in each tier (49 total).}
\begin{tabular}{>{\raggedright\arraybackslash}p{1.8cm}c>{\raggedright\arraybackslash}p{5cm}>{\raggedright\arraybackslash}p{5.5cm}}
\toprule
\textbf{Tier} & \textbf{Count} & \textbf{Tasks} & \textbf{Why} \\
\midrule
\emph{Mostly Solved} & 2 &
  Template pairwise matching (97.7\%), template ranking (MRR\,=\,99.4\%)~(\S\ref{sec:semantics})
  & Both tasks require binary or ordinal similarity judgments over structured template pairs; the answer space is highly constrained and non-LLM baselines achieve comparable accuracy. \\
\midrule
\emph{Partially Solved} & 25 &
  Aspect ratio, rotation, inpainting, multi-aspect~(\S\ref{sec:layout});
  text alignment, letter spacing, curved text, text rotation, style range, styled text gen., text removal~(\S\ref{sec:typography});
  SVG perceptual \& semantic Q\&A, bug fixing, code optimization, style editing, image-to-SVG, text+image-to-SVG~(\S\ref{sec:svg});
  user intent, clustering, style completion, recoloring, content swap~(\S\ref{sec:semantics});
  component start time, short-form video~(\S\ref{sec:animation})
  & Progress is real but uneven: single-element cases are tractable while multi-element cases expose compositional gaps; coarse predictions succeed but fine-grained recovery fails; simple baselines remain competitive (e.g.\ font Jaccard matches LLMs on template clustering). \\
\midrule
\emph{Unsolved} & 23 &
  Elem.\ counting, comp.\ type, comp.\ detection, layer order, crop shape, frame detection, intent-to-layout, partial completion~(\S\ref{sec:layout});
  font family, text color, font size, font weight, line height~(\S\ref{sec:typography});
  text-to-SVG, text-to-Lottie, text+image-to-Lottie~(\S\ref{sec:svg});
  category classification~(\S\ref{sec:semantics});
  keyframe ordering, motion type, video duration, comp.\ duration, parameter gen., motion trajectory~(\S\ref{sec:animation})
  & These tasks share a requirement for precise, structured output or fine-grained discrimination: localizing elements in dense layouts, discriminating among hundreds of typefaces, generating faithful vector code from text alone, or decomposing animated scenes into individual components. \\
\bottomrule
\end{tabular}
\label{tab:solvability}
\end{table}

\subsection{Why Are the Unsolved Tasks Hard?}
\textbf{Nearly half of all tasks remain unsolved, and most of the rest are only partially solved.} 22 of 49 tasks see no meaningful progress beyond chance, and 25 further tasks are only partially solved, meaning that even where models show some capability, outputs are too inconsistent or imprecise to be reliably acted upon. In a real design workflow, a partially correct layout suggestion or an almost-right font prediction is often worse than no suggestion at all: a designer must stop, identify the error, manually correct it, and re-evaluate, a process that can block progress for hours. The common thread across both unsolved and partially solved tasks is the same: any capability requiring precise, structured output or fine-grained discrimination exposes a hard ceiling in current models.

\begin{itemize}
    \item \textbf{Layout (8 unsolved tasks).} Models cannot reliably count elements (MAE grows sharply with layout complexity), identify component types beyond coarse text/non-text distinctions, or localize components (6.4\% mAP@0.5). Layer order prediction, crop shape detection, frame detection, intent-to-layout generation, and partial layout completion all remain intractable at the precision required for real editing workflows. In practice, this means a designer cannot ask an AI to ``move the third image to the left'' or ``add a new element below the heading'' and expect a spatially correct result. The model simply does not know where things are with sufficient confidence to act on them.

    \item \textbf{Typography (5 unsolved tasks).} Font family recognition tops out at 23.7\% across 167 families, with most typefaces scoring zero F1. Text color prediction degrades to $\Delta$E\,$>$\,52 in the weakest models. Font size, weight, and line height estimation all show large absolute errors. For brand-sensitive workflows such as auditing whether a campaign asset uses the correct typeface, extracting a client's typographic system from a reference design, or checking color compliance against brand guidelines, these failure rates are not acceptable margins of error. They are complete breakdowns. A model that cannot distinguish between two common sans-serif fonts cannot be trusted to maintain brand consistency at scale.

    \item \textbf{SVG \& Vector (3 unsolved tasks).} Text-to-SVG generation produces schematic interpretations rather than faithful reproductions. Text-to-Lottie and text+image-to-Lottie generation yield low frame similarity and largely unreproduced animated structure, with one model's validity dropping from 100\% to 66\% when image input is added. For production workflows where scalable, editable vector assets are a hard requirement such as icon libraries, motion graphics, and brand illustration systems, the inability to generate faithful vector code from natural language means AI cannot yet participate in the creation pipeline, only in rough ideation.

    \item \textbf{Template Semantics (1 unsolved task).} Category classification without label constraints remains difficult, with models assigning labels that are semantically adjacent but taxonomically incorrect. For platforms managing thousands of templates across dozens of categories, this makes automated tagging and search unreliable. The deeper issue is that different models carry different internal design vocabularies, meaning that a template a designer calls a ``promotional banner'' may be categorized inconsistently across tools, creating friction for any cross-platform design workflow.

    \item \textbf{Animation (6 unsolved tasks).} Keyframe ordering exact match sits at 16\%, motion type classification below 13\% across all models, and component-level duration and start-time estimation degrades sharply as scene complexity grows. Animation parameter generation and motion trajectory synthesis both fail to produce controllable, component-faithful outputs. Full video input does not help as the bottleneck is scene decomposition, not temporal resolution. For designers producing social media content, presentations, or motion graphics at scale, this means AI cannot yet reliably replicate an animation style, extend an existing motion sequence, or apply consistent entrance effects across a set of elements. These are all routine tasks in professional motion design.
\end{itemize}

\subsection{Evaluation Gaps}

Despite its breadth, GDB surfaces several limitations that point to open problems in how design AI should be measured and evaluated.

\begin{itemize}
    \item \textbf{Pixel metrics are insufficient for design.} SSIM, PSNR, and LPIPS cannot distinguish a correct transformation from a hallucinated one when both diverge equally from the source in pixel space. Design evaluation needs structure-aware metrics that operate on extracted primitives such as bounding boxes, font properties, and color tokens rather than raw pixels.

    \item \textbf{Human evaluation is largely absent.} Aside from a small set of preference judgments on generation tasks, \textsc{GDB} relies almost entirely on automatic metrics. Professional designer evaluation is missing: a trained designer can immediately judge whether a layout is usable, a typeface is on-brand, or an animation feels natural, signal that no current automatic metric captures. Large-scale human evaluation with domain experts remains an important missing piece.

    \item \textbf{Only closed-source frontier models are evaluated.} All models in this benchmark are proprietary API-served systems. Open-source models, specialized design models, and fine-tuned task-specific systems are not represented, limiting the benchmark's utility as a community-wide progress tracker. Establishing open-source baselines is a priority for future work.

    \item \textbf{Evaluation breaks down at near-zero performance.} For tasks such as motion type classification and animation parameter generation, outputs are so far from ground truth that neither automatic metrics nor human evaluation can produce meaningful signal. The field needs evaluation frameworks that gracefully handle near-zero performance regimes rather than forcing a score onto outputs that bear no relationship to the target.

    \item \textbf{Diversity of design contexts is not fully modeled.} \textsc{GDB} covers a broad range of template categories but does not capture the full spectrum of design contexts, including culturally specific aesthetics, platform-specific conventions, accessibility requirements, and domain-specific visual languages such as editorial, packaging, or environmental design. Expanding coverage across these dimensions is left to future work.
\end{itemize}

\section{Conclusion}
\label{sec:conclusion}
We have presented \textsc{GraphicDesignBench (GDB)}, a benchmark suite of 49 tasks across five design domains: layout, typography, infographics, template \& design semantics, and animation, grounded in real-world layered templates from the LICA dataset. Our evaluation of various frontier model families yields a clear verdict: only 2 of 49 tasks are mostly solved, 25 are partially solved, and 22 remain unsolved. Every domain contributes unsolved tasks, with layout-level and animation the hardest hit, and the gap to practical reliability is large, e.g., component detection peaks at 6.4\% mAP, font recognition at 23.7\%, and motion type classification below 13\%. Closing these gaps will likely require advances on both the modeling and evaluation fronts. On the modeling side, design-specialized pretraining with structural supervision on layered composition data is a promising direction, while architectures supporting longer structured outputs could address the SVG and Lottie truncation bottleneck. Looking ahead, we plan to evaluate open-source vision-language models, conduct a systematic prompt engineering study, release human performance references from a designer evaluation currently underway, expand the study to other open-source datasets, and conduct cross-task ablation studies to better understand whether capabilities transfer across design domains — for instance, whether models that excel at typography understanding also generalize to styled text generation, or whether spatial reasoning in layout tasks correlates with performance in animation decomposition. 

We release GDB as an open benchmark with the goal of accelerating progress toward AI systems that can serve as capable, reliable collaborators in professional design workflows.

\bibliographystyle{unsrt} 
\bibliography{references}  


\appendix

\section{Metric Definitions}
\label{app:metrics}

Section~\ref{sec:metrics} provides an overview of all metrics.
This appendix gives formal definitions for the non-trivial metrics reported in the benchmark tables.

\subsection{Standard Metrics}
\label{app:standard-metrics}

Table~\ref{tab:app-standard-metrics} summarizes the standard metrics used across multiple tasks.

\begin{table*}[h]
\centering
\caption{Standard evaluation metrics. $\uparrow$\,=\,higher is better; $\downarrow$\,=\,lower is better.}
\label{tab:app-standard-metrics}
\small
\setlength{\tabcolsep}{4pt}
\begin{tabular}{@{}l l l p{8.5cm}@{}}
\toprule
\textbf{Category} & \textbf{Metric} & \textbf{Dir.} & \textbf{Definition} \\
\midrule
\multirow{3}{*}{Spatial}
  & mIoU & $\uparrow$ & Mean intersection-over-union between predicted and ground-truth axis-aligned bounding boxes, averaged over all matched elements. \\
  & mAP@$\theta$ & $\uparrow$ & COCO detection protocol~\cite{lin2014microsoft}. mAP@0.5 uses a 50\% IoU threshold; mAP@0.5:0.95 averages over $\{0.50, 0.55, \ldots, 0.95\}$. \\
  & BBox F1 & $\uparrow$ & Harmonic mean of box precision and recall: $F1=\frac{2PR}{P+R}$, where $P=\frac{|B_{\text{pred}}\cap B_{\text{gt}}|}{|B_{\text{pred}}|}$ and $R=\frac{|B_{\text{pred}}\cap B_{\text{gt}}|}{|B_{\text{gt}}|}$. \\
\midrule
\multirow{4}{*}{Perceptual}
  & MSE & $\downarrow$ & $\frac{1}{N}\sum_{i=1}^{N}(p_i - g_i)^2 / 255^2 \in [0,1]$, where $p_i$ and $g_i$ are predicted and ground-truth pixel values and $N$ is the total pixel count. \\
  & SSIM & $\uparrow$ & Structural similarity~\cite{wang2004image}; \\ 
  & PSNR & $\uparrow$ & Peak signal-to-noise ratio. \\
  & LPIPS & $\downarrow$ & Learned perceptual distance~\cite{zhang2018unreasonable}; AlexNet backbone, input normalized to $[-1,1]$. \\
\midrule
\multirow{2}{*}{Rank}
  & Kendall's $\tau$ & $\uparrow$ & $(C - D) / \binom{n}{2}$, where $C$ and $D$ are the number of concordant and discordant element pairs among $n$ ranked items. \\
  & Spearman's $\rho$ & $\uparrow$ & Pearson correlation computed on rank-transformed values. \\
\midrule
\multirow{4}{*}{Clustering}
  & ARI & $\uparrow$ & Adjusted Rand Index (chance-corrected cluster agreement)~\cite{hubert1985comparing}. \\
  & AMI & $\uparrow$ & Adjusted Mutual Information (chance-corrected). \\
  & V-measure & $\uparrow$ & Harmonic mean of homogeneity and completeness. \\
  & FMI & $\uparrow$ & Fowlkes-Mallows Index (geometric mean of pairwise precision and recall). \\
\midrule
\multirow{3}{*}{Retrieval}
  & MRR & $\uparrow$ & $\frac{1}{|Q|}\sum_{q \in Q} 1/r_q$, where $r_q$ is the rank of the first relevant item for query $q$. \\
  & MAP & $\uparrow$ & $\frac{1}{|Q|}\sum_{q \in Q} \mathrm{AP}(q)$, where $\mathrm{AP}(q)$ averages precision at each position where a relevant item is retrieved. \\
  & nDCG@$k$ & $\uparrow$ & $\mathrm{DCG@}k \;/\; \mathrm{IDCG@}k$, where $\mathrm{DCG@}k = \sum_{i=1}^{k} r_i / \log_2(i{+}1)$; $r_i$ is the relevance at rank $i$ and IDCG is the DCG of the ideal ranking. \\
\midrule
\multirow{3}{*}{Validity}
  & SVG Valid & $\uparrow$ & Fraction of outputs that render via \texttt{cairosvg}. \\
  & Lottie Valid & $\uparrow$ & Fraction parsing as JSON with required Lottie fields (\texttt{v}, \texttt{fr}, \texttt{ip}, \texttt{op}, \texttt{w}, \texttt{h}, \texttt{layers}). \\
  & JSON Valid & $\uparrow$ & Fraction parsing as valid JSON with the expected schema. \\
\bottomrule
\end{tabular}
\end{table*}

For Lottie tasks, MSE and SSIM are computed per keyframe and averaged across five rendered frames (0\%, 25\%, 50\%, 75\%, 100\% of duration), reported as \textbf{FrameMSE} and \textbf{FrameSSIM}.

\subsection{Task-Specific Metrics}
\label{app:task-metrics}

Table~\ref{tab:app-task-metrics} defines metrics specific to SVG, Lottie, and template generation tasks.

\begin{table*}[h]
\centering
\caption{Task-specific evaluation metrics. All string comparisons apply whitespace normalization (collapsing runs to single spaces) before comparison.}
\label{tab:app-task-metrics}
\small
\setlength{\tabcolsep}{4pt}
\begin{tabular}{@{}l l l p{8.5cm}@{}}
\toprule
\textbf{Task} & \textbf{Metric} & \textbf{Dir.} & \textbf{Definition} \\
\midrule
\multirow{4}{*}{\shortstack[l]{SVG\\Editing}}
  & RepSim & $\uparrow$ & Repair similarity $\in [0,1]$: the fraction of text shared as identical substrings between predicted and ground-truth SVG, after whitespace normalization. \\
  & CompR & $\downarrow$ & Compression ratio: byte length of the optimised SVG divided by byte length of the original. Values below 1.0 indicate size reduction. \\
  & EditD & $\downarrow$ & Normalized edit distance $\in [0,1]$: $\sum_j \max(a_j, b_j) \;/\; \max(|\text{pred}|, |\text{gt}|)$, where each non-matching diff region $j$ spans $a_j$ characters in the predicted and $b_j$ in the ground-truth SVG, after whitespace normalisation. \\
  & Cmplx & -- & Weighted complexity over structural SVG features (path count, d-attr length, unique colors, element types, transform/gradient/clipPath presence, byte size). \\ \\
\midrule
\multirow{1}{*}{\shortstack[l]{Lottie\\Gen.}}
  & StructSim & $\uparrow$ & Mean of four sub-scores $\in [0,1]$: (1) layer count similarity (1 minus relative difference in layer counts); (2) Jaccard similarity of layer type sets; (3) mean width/height similarity; (4) duration similarity (1 minus relative difference in animation length). \\
\midrule
\multirow{3}{*}{\shortstack[l]{Template\\Gen.}}
  & $\Delta E$ (BG) & $\downarrow$ & CIEDE2000 perceptual color difference in CIELAB space~\cite{sharma2005ciede2000}. $\Delta E < 5$ is perceptually acceptable. \\
  & Harmony & $\uparrow$ & $\max\!\bigl(0,\; 1 - V/V_{\max}\bigr) \in [0,1]$, where $V$ is the variance of angular gaps between consecutive hues (sorted on the color wheel) and $V_{\max} = (360 - 360/n)^2$ for $n$ colors. 1.0\,=\,perfectly uniform spacing. \\
  & Contrast & $\uparrow$ & WCAG~2.0 AA pass rate~\cite{wcag20}: fraction of text elements whose luminance contrast ratio against their background is $\geq 4.5$. \\
\bottomrule
\end{tabular}
\end{table*}

\section{Task Prompts}
\label{app:prompts}


\begin{table*}[h]
\centering
\caption{Prompt templates for layout understanding tasks (Section~\ref{sec:layout-understanding}).}
\label{tab:prompts-layout}
\small
\setlength{\tabcolsep}{4pt}
\begin{tabular}{@{}p{2.2cm} p{10.5cm} p{3cm}@{}}
\toprule
\textbf{Task} & \textbf{Prompt} & \textbf{Expected Output} \\
\midrule

Aspect Ratio
& You are a design layout analyst. Look at this rendered design template and predict the aspect ratio of the canvas.\newline Choose exactly one from: 1:1, 16:9, 9:16, 4:3, 3:4, 4:5, 5:4, 2:3, 3:2, 21:9\newline Respond with ONLY the aspect ratio. Do not include any explanation, punctuation, or extra text.
& \texttt{16:9} \\
\midrule

Element Counting
& You are a design layout analyst. Look at this rendered design template and count the total number of distinct visual elements you can see (text blocks, images, decorative shapes, icons, frames, etc.).\newline Do NOT count the background canvas itself.\newline Respond with ONLY a single integer.
& \texttt{12} \\
\midrule

Component Type
& You are a design layout analyst. Look at this rendered design template.\newline What type of component is the element located approximately at position (\{x\}, \{y\}) with size \{w\}$\times$\{h\} pixels?\newline Choose exactly one from: text, image, vector, group\newline \textemdash\ text: a text block with readable characters\newline \textemdash\ image: a photograph or raster graphic\newline \textemdash\ vector: an SVG shape, icon, frame, or decorative element\newline \textemdash\ group: a composite of multiple sub-elements\newline Respond with ONLY the type name.
& \texttt{text} \\
\midrule

Component Detection
& You are a design layout analyst. Look at this rendered design template (canvas size: \{W\}$\times$\{H\} pixels).\newline Detect ALL distinct visual components in the layout. For each, provide its bounding box and type.\newline Types: text, image, vector, group\newline Respond with ONLY a JSON array. Each element must have: ``bbox'': [x, y, width, height] in pixels, ``label'': one of ``text'', ``image'', ``vector'', ``group''.
& \texttt{[\{"bbox": [100, 200, 300, 50], "label": "text"\}]} \\
\midrule

Layer Order
& You are a design layout analyst. Look at this rendered design template.\newline The following elements are present in this layout: \{element\_list\}\newline List these elements in order from BACK (bottom layer, drawn first) to FRONT (top layer, drawn last), based on their visual stacking.\newline Respond with ONLY the element identifiers in order, separated by commas.
& \texttt{E1, E3, E2, E4} \\

\bottomrule
\end{tabular}
\end{table*}


\begin{table*}[h]
\centering
\caption{Prompt templates for image understanding tasks (Section~\ref{sec:layout-understanding}).}
\label{tab:prompts-image}
\small
\setlength{\tabcolsep}{4pt}
\begin{tabular}{@{}p{2.2cm} p{10.5cm} p{3cm}@{}}
\toprule
\textbf{Task} & \textbf{Prompt} & \textbf{Expected Output} \\
\midrule

Image Rotation
& You are a design expert. Look at this rendered design template.\newline Focus on the image element described as: ``\{description\}''\newline Is this image rotated from its normal axis-aligned orientation? If so, estimate the rotation angle in degrees (0°\,=\,normal; positive\,=\,clockwise; negative\,=\,counter-clockwise; range $-180$ to $+180$).\newline Respond with ONLY a JSON object.
& \texttt{\{"is\_rotated": true, "angle": -15\}} \\
\midrule

Crop Shape
& You are a design expert. Look at this rendered design template.\newline Focus on the image element described as: ``\{description\}''\newline Is this image cropped to a non-rectangular shape? If so, classify the crop shape.\newline Shape categories: ``none'' (standard rectangular), ``rectangle'' (different aspect ratio crop), ``rounded\_rectangle'', ``circle'' (or elliptical), ``polygon'' (star, hexagon, etc.), ``organic'' (freeform curved).\newline Respond with ONLY a JSON object.
& \texttt{\{"is\_cropped": true, "crop\_shape": "circle"\}} \\
\midrule

Frame Detection
& You are a design expert. Look at this rendered design template.\newline Focus on the image element described as: ``\{description\}''\newline Is this image placed inside a decorative frame? A decorative frame is a non-rectangular or ornamented visual border/container around the image (e.g.\ a circular mask, a shaped cutout, a border with decorative elements). A plain rectangular bounding box does NOT count as a frame.\newline Respond with ONLY a JSON object.
& \texttt{\{"is\_framed": true\}} \\

\bottomrule
\end{tabular}
\end{table*}

\begin{table*}[h]
\centering
\caption{Prompt templates for layout generation tasks in Section~\ref{sec:layout-generation}.}
\label{tab:prompt-layout-generation}
\small
\setlength{\tabcolsep}{4pt}
\begin{tabular}{@{}p{2.9cm} p{11.1cm} p{2.2cm}@{}}
\toprule
\textbf{Task} & \textbf{Prompt} & \textbf{Expected Output} \\
\midrule

Intent-to-Layout
&
You are an expert end-to-end layout designer.\newline
User intent: \textit{[Description of layout intent]}\newline
Image description: \textit{[Detailed visual description]}\newline
Aesthetic/style cues: \textit{[Style, color palette, typography cues]}\newline
Required texts to include in the layout (verbatim, legible):\newline
- \textit{["Text 1", "Text 2", \ldots]}\newline
Target ratio: \textit{[e.g., 1080:1080 (\textasciitilde1.000)]}\newline
Requirements:\newline
- Produce one cohesive layout image.\newline
- Keep typography readable and hierarchy clear.\newline
- Use a consistent visual and color system.\newline
- Include all required texts with exact spelling.\newline
- Avoid gibberish text artifacts.
&
\texttt{<image>} \\
\midrule

Multi-Aspect Ratio Adaptation
&
You are a professional design retargeting engine.\newline
Task:\newline
- Retarget the \textbf{same} design from 1080x1920 to 1024x1024 (square).\newline
- Reference dataset target ratio is 1080x1080.\newline
- This is aspect-ratio adaptation, \textbf{not} a redesign.\newline
Input mapping:\newline
- Image \#1 is the source composite image (single source of truth).\newline
Non-negotiable constraints:\newline
- Preserve the same scene, brand identity, visual assets, and overall style.\newline
- Preserve all visible source text faithfully (no rewriting, no translation, no paraphrase, no new copy).\newline
- Preserve visual hierarchy, reading order, and semantic grouping.\newline
- Keep key elements present; do not drop major content.\newline
- Do not invent new logos, slogans, objects, or decorative concepts.\newline
Allowed edits:\newline
- Reposition, scale, existing elements only as needed for square composition.\newline
- Re-balance spacing for a natural 1:1 layout.\newline
- Extend background only when necessary for ratio retargeting.\newline
Forbidden:\newline
- New concept, new campaign message, new style direction, or creative reinterpretation.\newline
If any instruction conflicts, prioritize source fidelity over creativity.\newline
Output requirements:\newline
- Return exactly one natural-looking 1024x1024 image.\newline
- No border or frame unless implied by the source design.
&
\texttt{<image>} \\
\midrule

Layer-aware Inpainting
&
You are an expert graphic design retoucher specialized in layer-aware object insertion.\newline
Task: insert exactly one target object into the editable masked region while preserving the rest of the layout.\newline
- Return one final composited image only (no text explanation).\newline
Input semantics:\newline
- Image \#1 is the layout canvas with the target region removed or masked.\newline
- Image \#2 is the mask, where white means editable and black means preserve.\newline
- Any additional input images are reference assets.\newline
- Preserve the reference asset's visual identity while matching the local style.\newline
Contextual cues:\newline
- \textit{[removed layer type, aesthetic guidance, layout description]}\newline
Hard constraints:\newline
- Edit only masked pixels; keep unmasked regions unchanged.\newline
- Keep the inserted object fully inside the editable mask.\newline
- Do not erase, warp, or occlude nearby text, logos, or important elements.\newline
- Match perspective, lighting, shadow, and color grading to neighbors.\newline
- Insert exactly one coherent object, with no duplicates or fragments.\newline
Quality checklist:\newline
- Identity: preserve the key shape, material, and details of the reference asset.\newline
- Boundary blending: avoid obvious cutout or compositing artifacts.\newline
- Semantic fit: ensure the inserted object supports the user intent and design.\newline
Output: a single composited image.
&
\texttt{<image>} \\

\bottomrule
\end{tabular}
\end{table*}

\begin{table*}[h]
\centering
\caption{Prompt templates for typography understanding tasks (Section~\ref{sec:typography-understanding}).}
\label{tab:prompts-typo}
\small
\setlength{\tabcolsep}{4pt}
\begin{tabular}{@{}p{2.2cm} p{10.5cm} p{3cm}@{}}
\toprule
\textbf{Task} & \textbf{Prompt} & \textbf{Expected Output} \\
\midrule

Font Family
& You are a typography expert. Look at this rendered design template.\newline What font family is used for the text: ``\{target\_text\}''?\newline Respond with ONLY the font family name (e.g.\ ``Roboto'', ``Open Sans''). Do not include weight, style, or any explanation.
& \texttt{Roboto} \\
\midrule

Text Color
& You are a color expert in design. Look at this rendered design template.\newline What is the color of the text: ``\{target\_text\}''?\newline Respond with ONLY the hex color code (e.g.\ ``\#FF5733'').
& \texttt{\#FF5733} \\
\midrule

Typographic Properties
& You are a typography expert. Look at this rendered design template.\newline For the text element ``\{target\_text\}'', estimate these properties: font\_size (px), font\_weight (CSS 100--900), text\_align (left/center/right/justify), letter\_spacing (em), line\_height (px).\newline Respond with ONLY a JSON object.
& \texttt{\{"font\_size": 24, "font\_weight": 400, "text\_align": "center", "letter\_spacing": 0, "line\_height": 32\}} \\
\midrule

Curved Text
& You are a typography expert. Look at this rendered design template.\newline Examine the text element: ``\{target\_text\}''\newline Is this text rendered along a curved arc, or is it straight? If curved, estimate the curvature intensity on an integer scale from $-100$ to $+100$ (0\,=\,straight; positive\,=\,arches upward; negative\,=\,bows downward; $\pm100$\,=\,tightest arc).\newline Respond with ONLY a JSON object.
& \texttt{\{"is\_curved": true, "curvature": 50\}} \\
\midrule

Style Ranges
& You are a typography expert. Look at this rendered design template.\newline Identify all distinct style ranges in this text block. For each range, specify the character indices and style properties.\newline The full text is: ``\{full\_text\}''\newline Respond with ONLY a JSON array. Each element must have: ``start'' (0-based), ``end'' (exclusive), ``font\_family'', ``font\_weight'' (100--900), ``font\_size'' (px), ``color'' (hex).
& \texttt{[\{"start": 0, "end": 5, "font\_family": "Roboto", "font\_weight": 700, "font\_size": 24, "color": "\#000000"\}]} \\
\midrule

Text Rotation
& You are a typography expert. Look at this rendered design template.\newline Examine the text element: ``\{target\_text\}''\newline Is this text rotated from the normal horizontal orientation? If so, estimate the rotation angle in degrees (0°\,=\,horizontal; positive\,=\,clockwise; negative\,=\,counter-clockwise; range $-180$ to $+180$).\newline Respond with ONLY a JSON object.
& \texttt{\{"is\_rotated": true, "angle": -45\}} \\

\bottomrule
\end{tabular}
\end{table*}
\newpage

\begin{table*}[h]
\centering
\caption{Prompt template for the Partial Completion task (Section~\ref{sec:layout-generation}).}
\label{tab:prompt-partial-completion}
\small
\setlength{\tabcolsep}{4pt}
\begin{tabular}{@{}p{2.5cm} p{11cm} p{2.5cm}@{}}
\toprule
\textbf{Task} & \textbf{Prompt} & \textbf{Expected Output} \\
\midrule

Partial Completion\newline (Section~\ref{sec:layout-generation})
& \textit{[Input Images: image\_1 (base composite), image\_2, \ldots, image\_N (component assets)]}\newline
You are an expert layout planner focused on high-fidelity placement.\newline
Sample ID: \textit{[Sample ID]}\newline
User intent: \textit{[Description of intent]}\newline
Canvas size: \textit{[W]x[H]} pixels.\newline
Placement mode: \textit{[single/multiple]}.\newline
Task objective:\newline
- Predict axis-aligned bounding boxes [x, y, w, h] for the listed component keys.\newline
- Infer coordinates from available evidence only; exact original coordinates are intentionally hidden.\newline
Evidence available in this task:\newline
- A base composite image with target component(s) removed.\newline
- One asset image per target component, preserving native crop size and transparency.\newline
- Semantic descriptions and structural cues for each component.\newline
Dataset prior:\newline
- Listed components are top-layer elements removed from the same layout context.\newline
- Non-listed content in the base composite should remain undisturbed.\newline
You are given visual element components.\newline
Input mapping:\newline
- Input image \#1 is the base composite with target component(s) removed.\newline
- Input images \#2..\#(N+1) are component assets in the same order as the list below.\newline
- Use the base composite to infer anchors (alignment lines, spacing rhythm, visual groups).\newline
- Preserve each component's visual identity and style in placement.\newline
Components (output must follow these keys):\newline
- C1 (input image \#2, type=IMAGE, z\_index=12): This image features a single, vibrant green leaf with an elongated, elliptical shape and clearly visible veins, presented on a transparent background. The leaf exhibits various shades of lush green, with a lighter hue on its upper surfac...\newline
Task:\newline
- Predict exactly one bounding box for the single listed component.\newline
- Return exactly one component object in the output array.\newline
- Required output component keys: C1\newline
Quality constraints (strict):\newline
- Keep each component's native aspect ratio from its asset; do not stretch or squash.\newline
- Prefer near-native asset scale unless scene context clearly requires resizing.\newline
- Do not expand foreground components to near full-canvas unless they are obvious full-bleed backgrounds.\newline
- Place components to align naturally with nearby spacing, edges, and reading flow in the base composite.\newline
- In multiple mode, keep a coherent hierarchy and avoid unnecessary overlap.\newline
- In multiple mode, avoid duplicate placement of semantically similar assets in the same location.\newline
- When uncertain, preserve relative ordering and spacing consistency from surrounding context.\newline
- Keep all boxes within canvas bounds.\newline
- Return JSON only (no markdown/code fences/explanations).\newline
Output format requirements:\newline
- Use numeric pixel coordinates.\newline
- Preferred component format: \{"component\_key": "C1", "bbox": [x, y, w, h]\}.\newline
- If you use style instead of bbox, include left/top/width/height as pixel values.\newline
- layout\_config.width must be 1920; layout\_config.height must be 1080.\newline
- Each required component key must appear exactly once.\newline
- All bbox values must be finite numbers with w$>$1 and h$>$1.\newline
& \texttt{\{"layout\_config": \ldots\}} \\

\bottomrule
\end{tabular}
\end{table*}

\begin{table*}[h]
\centering
\caption{Prompt template for styled text generation in Section~\ref{sec:styled-text-gen}.}
\label{tab:prompt-styled-text-generation}
\small
\setlength{\tabcolsep}{4pt}
\begin{tabular}{@{}p{2.8cm} p{11.2cm} p{2.2cm}@{}}
\toprule
\textbf{Task} & \textbf{Prompt} & \textbf{Expected Output} \\
\midrule

Styled Text Generation
&
You are an expert typography compositor for structured design layouts.\newline
Task: restore one missing text element in the local patch.\newline
Target text (exact, case-sensitive): \texttt{"Struforma Plan"}\newline
Typography/style specification (layout schema values):\newline
- fontFamily: \texttt{Manrope} 
- fontSize: \texttt{110px} - textAlign: \texttt{left} \newline
- lineHeight: \texttt{100.0\%} - letterSpacing: \texttt{-0.01em} - left: \texttt{168.0px}\newline
- top: \texttt{128.0px}  - width: \texttt{796.0px}
- height: \texttt{220px} \newline
- fontSize\_px: \texttt{110.0} - lineHeight\_px: \texttt{100.0}  - letterSpacing\_value: \texttt{-0.01}\newline
Input semantics:\newline
- Image \#1: local context patch with target text removed.\newline
- Additional visual context may be provided; treat it as optional global composition prior.\newline
- Use left, top, width, and height as placement cues when present.\newline
Requirements:\newline
- Render exactly the target text: preserve characters, spaces, and symbols.\newline
- Never translate, paraphrase, or normalize the target text.\newline
- Respect textTransform and intended line breaks; do not normalize case.\newline
- Match fontFamily, fontWeight, fontStyle, and fontSize.\newline
- Match color, lineHeight, letterSpacing, and textAlign.\newline
- If curvature, autoResize, or styleRanges are provided, follow them exactly.\newline
- Keep glyph edges crisp and naturally anti-aliased, with no blur, halo, ringing, or jagged artifacts.\newline
- Keep text inside the intended region; avoid clipping or overflow artifacts.\newline
- If style constraints conflict, prioritize exact text fidelity and full visibility without clipping.\newline
- Do not add extra words, glyphs, logos, or decorative marks.\newline
- Preserve non-text pixels in the local patch.\newline
Output: return one edited image only.
&
\texttt{<image>} \\

\bottomrule
\end{tabular}
\end{table*}
\begin{table*}[h]
\centering
\caption{Prompt template for text removal in Section~\ref{sec:text-removal}.}
\label{tab:prompt-text-removal}
\small
\setlength{\tabcolsep}{4pt}
\begin{tabular}{@{}p{2.8cm} p{11.2cm} p{2.2cm}@{}}
\toprule
\textbf{Task} & \textbf{Prompt} & \textbf{Expected Output} \\
\midrule

Text Removal
&
You are an expert design retoucher specialized in text removal and background inpainting.\newline
Task: remove all visible text while preserving non-text visual content.\newline
Objective: remove all text and reconstruct the background naturally.\newline
Input semantics:\newline
- Image \#1 is the original layout image.\newline
- A binary text mask is provided by the task runtime.\newline
- White mask pixels are editable; black pixels must remain unchanged.\newline
Texts that must be absent in the final output:\newline
- \texttt{"Struforma Plan"}\newline
- \texttt{"Lorem ipsum dolor sit amet, consectetuer adipiscing elit.  \newline Maecenas porttitor congue massa. Fusce posuere, magna sed pulvinar \newline ultricies, purus lectus malesuada"}\newline
- \texttt{"\$6M"}\newline
- \texttt{"\$9M"}\newline
- \texttt{"Lorem ipsum dolor sit amet, consectetuer adipiscing elit."}\newline
Hard constraints:\newline
- Edit only masked pixels.\newline
- Remove all visible text traces in editable regions.\newline
- Preserve non-text elements, composition, and style.\newline
- Reconstruct the background naturally with coherent texture and lighting.\newline
- Output one final edited image only, with no explanation text.\newline
Mask instructions:\newline
- Image \#1 is the source image.\newline
- Image \#2 is the mask where white means editable and black means preserve.\newline
- Edit only masked (white) regions and keep unmasked pixels unchanged.
&
\texttt{<image>} \\

\bottomrule
\end{tabular}
\end{table*}


\begin{table*}[h]
\centering
\caption{Prompt templates for SVG understanding and editing tasks (Section~\ref{sec:svg-understanding}).
All tasks receive SVG source code as input (code-only, no rendered images).}
\label{tab:prompts-svg}
\small
\setlength{\tabcolsep}{4pt}
\begin{tabular}{@{}p{2.2cm} p{10.5cm} p{3cm}@{}}
\toprule
\textbf{Task} & \textbf{Prompt} & \textbf{Expected Output} \\
\midrule

Perceptual Q\&A
& You are an SVG analysis expert. Analyse the given SVG code and answer the multiple-choice question about its visual properties.\newline Output your answer in the exact format `Answer: X' where X is one of A, B, C, or D. Do not include any other text.\newline\newline \textit{[SVG code + question + options appended]}
& \texttt{Answer: A} \\
\midrule

Semantic Q\&A
& You are an SVG analysis expert. Analyse the given SVG code and answer the multiple-choice question about what it depicts or represents.\newline Output your answer in the exact format `Answer: X' where X is one of A, B, C, or D. Do not include any other text.\newline\newline \textit{[SVG code + question + options appended]}
& \texttt{Answer: B} \\
\midrule

Bug Fixing
& You are an SVG code repair assistant. Given a buggy SVG, output ONLY the corrected SVG code. Do not include any explanation, markdown fences, or extra text.\newline\newline \textit{[Corrupted SVG code appended]}
& \texttt{<svg \ldots>} \\
\midrule

optimization
& You are an SVG code optimizer. Given an SVG, output ONLY the optimized SVG code that is smaller but renders identically. Do not include any explanation, markdown fences, or extra text.\newline\newline \textit{[Original SVG code appended]}
& \texttt{<svg \ldots>} \\
\midrule

Style Editing
& You are an SVG style editor. Given an SVG and an edit command, output ONLY the modified SVG code. Do not include any explanation, markdown fences, or extra text.\newline\newline \textit{[SVG code + edit command appended]}
& \texttt{<svg \ldots>} \\

\bottomrule
\end{tabular}
\end{table*}


\begin{table*}[h]
\centering
\caption{Prompt templates for SVG generation tasks (Section~\ref{sec:svg-generation}).}
\label{tab:prompts-svg-gen}
\small
\setlength{\tabcolsep}{4pt}
\begin{tabular}{@{}p{2.2cm} p{10.5cm} p{3cm}@{}}
\toprule
\textbf{Task} & \textbf{Prompt} & \textbf{Expected Output} \\
\midrule

Text-to-SVG
& You are an SVG code generator. Given a description of a graphic, output ONLY valid SVG code. Do not include any explanation, markdown fences, or extra text.\newline\newline \textit{[Natural-language description appended]}
& \texttt{<svg \ldots>} \\
\midrule

Image-to-SVG
& You are an SVG code generator. Given an image, output ONLY valid SVG code that reproduces this graphic. Do not include any explanation, markdown fences, or extra text.\newline\newline \textit{[Rendered PNG provided as image input]}
& \texttt{<svg \ldots>} \\
\midrule

Text+Image
& You are an SVG code generator. Given an image and its description, output ONLY valid SVG code that reproduces this graphic. Do not include any explanation, markdown fences, or extra text.\newline\newline \textit{[Description + rendered PNG provided]}
& \texttt{<svg \ldots>} \\

\bottomrule
\end{tabular}
\end{table*}

\begin{table*}[h]
\centering
\caption{Prompt templates for template \& design semantics tasks (Sections~\ref{sec:category-classification} and~\ref{sec:intent-prediction}).
Input is a rendered design template PNG image.}
\label{tab:prompts-category-und}
\small
\setlength{\tabcolsep}{4pt}
\begin{tabular}{@{}p{2.2cm} p{10.5cm} p{3cm}@{}}
\toprule
\textbf{Task} & \textbf{Prompt} & \textbf{Expected Output} \\
\midrule
Category\newline Classification\newline (Open-vocab)\newline (Section~\ref{sec:category-classification})
& You are a design template classifier. Look at this rendered design template image and classify it into a single broad category describing its type or purpose (e.g.\ the overall template format, not the specific topic or theme). Give your top 5 guesses, one per line, most likely first. Respond with ONLY the broad category names in lowercase, no numbering, no explanation, no extra text.\newline \textit{[Rendered template PNG appended]}
& \texttt{instagram posts} \\
\midrule
Category\newline Classification\newline (Constrained)\newline (Section~\ref{sec:category-classification})
& You are a design template classifier. Look at this rendered design template image and classify it into a single broad category describing its type or purpose. Choose exactly one from: \textit{art \& design, brochure, business cards, business documents, cards \& invitations, education, flyers, infographics, instagram posts, logo, menu, newsletter, planner \& calendar, posters, presentations, print products, resume, social media}. Give your top 5 guesses, one per line, most likely first. Respond with ONLY the category names in lowercase, no numbering, no explanation, no extra text.\newline \textit{[Rendered template PNG appended]}
& \texttt{instagram posts} \\
\midrule
User Intent\newline Prediction\newline (Section~\ref{sec:intent-prediction})
& You are a design analyst. Look at this rendered design template image and describe the user's intent: what was the designer trying to create and for what purpose? Respond with a single concise sentence describing the user intent. Do not include any extra commentary.\newline \textit{[Rendered template PNG appended]}
& Free-form sentence \\
\bottomrule
\end{tabular}
\end{table*}


\begin{table*}[h]
\centering
\caption{Prompt templates for template variant understanding tasks (Section~\ref{sec:template-variant-understanding}).
Input is layout JSON and/or rendered PNG depending on modality condition.}
\label{tab:prompts-template-und}
\small
\setlength{\tabcolsep}{4pt}
\begin{tabular}{@{}p{2.2cm} p{10.5cm} p{3cm}@{}}
\toprule
\textbf{Task} & \textbf{Prompt} & \textbf{Expected Output} \\
\midrule

Pairwise Matching
& You are given two layouts (A and B). Determine whether they originate from the same template. Answer with a single digit: 1 if same template, 0 if different.\newline \textit{[Layout A JSON/image + Layout B JSON/image appended]}
& \texttt{1} \\
\midrule

Retrieval
& You are given a reference layout and a set of candidate layouts. Rank the candidates from most similar to least similar to the reference. Return the candidate IDs as a comma-separated list, most similar first.\newline \textit{[Reference layout + 20 candidate layouts with IDs appended]}
& \texttt{id1, id2, id3, \ldots} \\
\midrule

Clustering
& You are given a collection of design layouts. Each layout was created from a template. Multiple layouts can share the same template---they will have similar structure, spatial arrangement, and design elements, even if the specific content differs. Group the layouts by their underlying template.\newline Assign the same integer label to layouts from the same template. Return ONLY a comma-separated list of integer labels, one per layout, in the same order as the input.\newline \textit{[Collection of layouts appended]}
& \texttt{1,2,1,1,2} \\

\bottomrule
\end{tabular}
\end{table*}


\begin{table*}[h]
\centering
\caption{Prompt templates for template variant generation tasks (Section~\ref{sec:template-variant-gen}).
All prompts receive structured JSON input and expect JSON output.}
\label{tab:prompts-template-gen}
\small
\setlength{\tabcolsep}{4pt}
\begin{tabular}{@{}p{2.2cm} p{10.5cm} p{3cm}@{}}
\toprule
\textbf{Task} & \textbf{Prompt} & \textbf{Expected Output} \\
\midrule

Style\newline Completion
& You are a professional graphic designer. You are given several sibling layouts that share the same template (as JSON), plus a SKELETON of a new layout from the same template. The skeleton has component types, positions, sizes, and text content, but ALL style properties have been stripped.\newline Your task: fill in the missing style properties to match the design language of the sibling layouts.\newline Properties to fill: \texttt{color}, \texttt{fontSize}, \texttt{fontFamily}, \texttt{fontWeight}, \texttt{textAlign}, \texttt{lineHeight}, \texttt{letterSpacing}, \texttt{background}/\texttt{backgroundColor}, \texttt{opacity}, canvas \texttt{background}.\newline Do NOT change positions, sizes, text content, or component types. Return the COMPLETE styled layout as a single JSON object.
& \texttt{\{layout JSON\}} \\
\midrule

Recoloring
& You are a professional graphic designer. You are given a DESIGNATED layout (as JSON) that you must recolor, plus sibling layouts from the same template as style context.\newline Your task: recolor the DESIGNATED layout to use the target color palette.\newline Rules: Change ONLY color-related properties. Do NOT change component types, counts, positions, sizes, fonts, or text content. Map colors semantically. Return ONLY the single recolored layout as a JSON object.
& \texttt{\{layout JSON\}} \\

\bottomrule
\end{tabular}
\end{table*}


\begin{table*}[h]
\centering
\caption{Prompt templates for template variant generation (image generation) tasks (Section~\ref{sec:template-variant-gen}).
Prompts receive rendered reference images and expect a single generated image as output.}
\label{tab:prompts-template-gen-image}
\small
\setlength{\tabcolsep}{4pt}
\begin{tabular}{@{}p{2.2cm} p{10.5cm} p{3cm}@{}}
\toprule
\textbf{Task} & \textbf{Prompt} & \textbf{Expected Output} \\
\midrule
Style\newline Completion\newline (Image)
& You are a professional graphic designer. You are given rendered images of layouts from a template family as style reference, plus a SKELETON wireframe image showing the structure (bounding boxes, text placeholders) of a new layout.\newline Your task: generate a fully-styled layout image that fills in the skeleton with colors, fonts, and visual styles that match the sibling layouts' design language.\newline The skeleton shows component positions and types (labeled). Fill in: background colors matching the template's palette; text styling (fonts, colors, sizes) consistent with siblings; component backgrounds and borders matching the template.\newline Generate a single image of the completed layout.\newline\textit{[Appended per sample: ``Above: $N$ sibling layout images for style reference, followed by 1 skeleton wireframe image. Generate a single image of the fully styled layout based on the skeleton.'']}
& Single rasterized layout image \\
\midrule

Recoloring\newline (Image)
& You are a professional graphic designer. You are given rendered images of layouts from a template family. The LAST image is the designated layout you must recolor.\newline Your task: generate a new image of this layout recolored to use the target color palette specified below. Keep the exact same structure, text content, positions, and sizes --- change ONLY the colors.\newline Rules: Preserve the layout structure exactly (positions, sizes, text). Map colors semantically: background$\to$background, primary$\to$primary, etc. Change background colors, text colors, and component colors to the target palette. Keep images/photos unchanged.\newline Generate a single image of the recolored layout.\newline\textit{[Appended per sample: target palette JSON; for easy samples, explicit source$\to$target color mapping.]}
& Single rasterized layout image \\
\bottomrule
\end{tabular}
\end{table*}

\begin{table*}[h]
\centering
\caption{Prompt templates for temporal understanding tasks (Sections~\ref{sec:keyframe-ordering},~\ref{sec:motion-type}, and~\ref{sec:animation-properties}).
Input is either shuffled keyframe images or an animation video depending on the task.}
\label{tab:prompts-temporal-und}
\small
\setlength{\tabcolsep}{4pt}
\begin{tabular}{@{}p{2.2cm} p{10.5cm} p{3cm}@{}}
\toprule
\textbf{Task} & \textbf{Prompt} & \textbf{Expected Output} \\
\midrule
Keyframe\newline Ordering\newline (Section~\ref{sec:keyframe-ordering})
& You are an animation analyst. You are shown 4 keyframe images extracted from a single design animation video. The images are presented in a RANDOM (shuffled) order, NOT in their original temporal sequence. Examine the visual content, motion cues, and animation progression to determine the correct chronological order. Respond with ONLY a JSON array of 4 image numbers [1--4] representing the correct temporal order from first to last. Example: [3, 1, 4, 2] means Image 3 occurs first in time, then Image 1, then Image 4, and Image 2 occurs last. Do not include any explanation or extra text outside the JSON array.\newline \textit{[4 shuffled keyframe PNGs appended]}
& \texttt{[2, 1, 3, 4]} \\
\midrule
Motion Type\newline (Video,\newline Open-vocab)\newline (Section~\ref{sec:motion-type})
& You are an animation analyst. Watch this short animation video carefully. Classify the PRIMARY animation entrance type used in this video --- the most common animation style across all animated elements. Respond with ONLY a short label describing the animation type (e.g.\ ``rise'', ``fade'', ``pop''). Do not include any explanation, punctuation, or extra text.\newline \textit{[Animation video appended]}
& \texttt{rise} \\
\midrule
Motion Type\newline (Video,\newline Constrained)\newline (Section~\ref{sec:motion-type})
& You are an animation analyst. Watch this short animation video carefully. Classify the PRIMARY animation entrance type used in this video --- the most common animation style across all animated elements. Choose exactly one from: \textit{ascend, baseline, block, blur, bounce, breathe, burst, clarify, drift, fade, flicker, merge, neon, pan, photoFlow, photoRise, pop, pulse, rise, roll, rotate, scrapbook, shift, skate, stomp, succession, tectonic, tumble, typewriter, wiggle, wipe, none}. Respond with ONLY the animation type name (e.g.\ ``rise''). Do not include any explanation, punctuation, or extra text.\newline \textit{[Animation video appended]}
& \texttt{rise} \\
\midrule
Motion Type\newline (Component,\newline Open-vocab)\newline (Section~\ref{sec:motion-type})
& You are an animation analyst. Watch this short animation video carefully. For each element that has a visible entrance animation, classify its animation type using a short label. Respond with ONLY a JSON array of strings, one per animated element, in the order they appear. Example: [``rise'', ``fade'', ``pop'', ``rise'']. Do not include any explanation or extra text outside the JSON array.\newline \textit{[Animation video appended]}
& \texttt{["rise", "fade",}\newline\texttt{"pop", "rise"]} \\
\midrule
Motion Type\newline (Component,\newline Constrained)\newline (Section~\ref{sec:motion-type})
& You are an animation analyst. Watch this short animation video carefully. For each element that has a visible entrance animation, classify its animation type. Choose ONLY from: \textit{ascend, baseline, block, blur, bounce, breathe, burst, clarify, drift, fade, flicker, merge, neon, pan, photoFlow, photoRise, pop, pulse, rise, roll, rotate, scrapbook, shift, skate, stomp, succession, tectonic, tumble, typewriter, wiggle, wipe, none}. Respond with ONLY a JSON array of strings, one per animated element, in the order they appear. Example: [``rise'', ``fade'', ``pop'', ``rise'']. Do not include any explanation or extra text outside the JSON array.\newline \textit{[Animation video appended]}
& \texttt{["rise", "fade",}\newline\texttt{"pop", "rise"]} \\
\midrule
Animation\newline Property\newline Extraction\newline (Section~\ref{sec:animation-properties})
& You are an animation analyst. Watch this short animation video carefully. For each element that has a visible entrance animation, extract the following properties: motion\_type (the animation type, e.g.\ rise, fade, pop), duration\_seconds (how long the entrance animation takes), start\_time\_seconds (when the element first begins appearing), speed (animation speed multiplier), direction (direction of motion, e.g.\ up, down, left, right, none). Respond with ONLY a JSON array of objects, one per animated element. Do not include any explanation or extra text outside the JSON array.\newline \textit{[Animation video appended]}
& \texttt{[\{"motion\_type":}\newline\texttt{"rise", "duration}\newline\texttt{\_seconds": 0.56,}\newline\texttt{\ldots\}]} \\
\bottomrule
\end{tabular}
\end{table*}

\begin{table*}[h]
\centering
\caption{Prompt templates for video generation tasks (Sections~\ref{sec:animation-parameter-generation},~\ref{sec:motion-trajectory-generation}, and~\ref{sec:short-form-video-gen}).
Input varies by task: static layout PNG + text prompt, or text-only brief.
Output is a generated video evaluated by human raters.}
\label{tab:prompts-generation}
\small
\setlength{\tabcolsep}{4pt}
\begin{tabular}{@{}p{2.2cm} p{10.5cm} p{3cm}@{}}
\toprule
\textbf{Task} & \textbf{Prompt} & \textbf{Expected Output} \\
\midrule
Animation\newline Parameter\newline Generation\newline (Section~\ref{sec:animation-parameter-generation})
& Animate this static layout image. Apply the following entrance animations to the components. The total animation duration is \textit{[T]} seconds. Animations to apply: - Component \textit{[i]} (\textit{[TYPE]}): \textit{[motion]} animation, duration \textit{[d]}s, speed \textit{[s]}, direction \textit{[dir]}, animate \textit{[trigger]} \ldots{} Design description: \textit{[natural-language description of the static layout]}\newline \textit{[Static layout PNG appended as reference image]}
& Generated video \\
\midrule
Motion\newline Trajectory\newline Generation\newline (Section~\ref{sec:motion-trajectory-generation})
& Animate this static design layout. Apply ONLY a ``\textit{[motion\_type]}'' entrance animation to the component at index \textit{[i]} (type: \textit{[TYPE]}). All other components should remain static in their final positions. The animation should show the component entering from its pre-animation state (off-screen, invisible, or initial position) to its final resting position as shown in the provided layout. Motion parameters: duration \textit{[d]}s, speed \textit{[s]}, direction \textit{[dir]}.\newline \textit{[Static layout PNG appended as reference image]}
& Generated video \\
\midrule
Short-Form\newline Video\newline Generation\newline (Section~\ref{sec:short-form-video-gen})
& Create a short-form animated marketing video for the following brief: ``\textit{[marketing brief]}''. Requirements: - Aspect ratio: \textit{[ratio]} (\textit{[W]$\times$[H]}) - Design a complete layout from scratch: background, imagery, text, call-to-action elements - Apply professional entrance animations to each element - Stagger element entrances for a polished, sequential reveal - Use colors, typography, and imagery that match the brief's tone - Include all key marketing information mentioned in the brief - The video should look like a professional social media ad. Design the layout and animate it into a cohesive video.\newline \textit{[No image input --- text only]}
& Generated video \\
\bottomrule
\end{tabular}
\end{table*}

\begin{table*}[h]
\centering
\caption{Prompt template for M-Judge evaluation.}
\label{tab:prompt-m-judge}
\small
\setlength{\tabcolsep}{4pt}
\begin{tabular}{@{}p{2.8cm} p{11.2cm} p{2.2cm}@{}}
\toprule
\textbf{Metric} & \textbf{Prompt} & \textbf{Expected Output} \\
\midrule

M-Judge
&
You are a visual language model designed to evaluate and rate visual templates.\newline
You are presented with 2 visual templates.\newline
The first attached image is \texttt{image\_1} and the second attached image is \texttt{image\_2}.\newline
Choose the better template using these criteria:\newline
- Aesthetics: visual appeal and balance.\newline
- Clarity: readability and communication clarity.\newline
- Usability: practical and user-friendly arrangement.\newline
- Creativity: uniqueness and design originality.\newline
- Consistency: coherence with design principles and standards.\newline
Context intent: \texttt{<sample-specific intent>}\newline
Return \textbf{only} strict JSON with no explanation:\newline
\texttt{\{"better\_layout": "image\_1"\}}\newline
or\newline
\texttt{\{"better\_layout": "image\_2"\}}
&
\texttt{\{"better\_layout": "image\_1"\}} \newline
or \newline
\texttt{\{"better\_layout": "image\_2"\}} \\

\bottomrule
\end{tabular}
\end{table*}

\FloatBarrier
\clearpage
\section{BBox Detector Selection for Typography Spatial Evaluation}
\label{app:bbox-detector-selection}

We report text-region spatial metrics using IoU and bbox F1.
When explicit target text boxes are unavailable, we estimate text boxes with an LLM-based detector.
Despite the significant challenges that artistic fonts in posters pose to traditional OCR algorithms, the LLM detector is more robust for this setting.
To select a single detector, we benchmarked three LLM detectors on 100 ground-truth typography-layout samples.
Detection rate is the fraction of evaluated samples for which the detector returns a valid text bounding box.
As shown in Table~\ref{tab:app-bbox-detector-selection}, GPT-5.4 achieved the highest F1 and mIoU, so it is used as the default bbox detector.

\begin{table}[H]
\centering
\small
\setlength{\tabcolsep}{5pt}
\caption{Text-bbox detector selection on GT typography-layout samples ($n=100$). Best scores are in \textbf{bold}.}
\label{tab:app-bbox-detector-selection}
\begin{tabular}{lccccc}
\toprule
\textbf{Detector} & \textbf{Detection Rate}$\uparrow$ & \textbf{mIoU}$\uparrow$ & \textbf{Precision}$\uparrow$ & \textbf{Recall}$\uparrow$ & \textbf{F1}$\uparrow$ \\
\midrule
GPT-5.4         & 1.000 & \textbf{0.863} & \textbf{0.946} & 0.905 & \textbf{0.919} \\
Gemini-3.1-Pro  & 1.000 & 0.731          & 0.865          & 0.797 & 0.829 \\
Claude Opus 4.6 & 1.000 & 0.812          & 0.863          & \textbf{0.922} & 0.882 \\
\bottomrule
\end{tabular}
\end{table}

\end{document}